\RequirePackage{fix-cm}
\documentclass[10pt,twocolumn,letterpaper]{article}
\usepackage[pagenumbers]{cvpr}
\input{preamble}
\makeatletter
\let\mainaddtocurcol\@addtocurcol
\usepackage{flafter}
\let\suppaddtocurcol\@addtocurcol
\let\@addtocurcol\mainaddtocurcol
\makeatother
\usepackage{placeins,longtable}
\usepackage{tikz}
\usetikzlibrary{arrows.meta,positioning,fit,backgrounds,decorations.pathreplacing}
\definecolor{cpfaOriginalRed}{rgb}{.929,.110,.141}%
\definecolor{cpfaOriginalGreen}{rgb}{0,.631,.294}%
\definecolor{cpfaOriginalBlue}{rgb}{.169,.400,.886}%
\definecolor{cpfaOriginalInk}{rgb}{.137,.122,.125}%
\pgfkeys{/cpfa/.is family,/cpfa,
  width/.store in=\cpfaWidth,
  red/.store in=\cpfaRed,
  green/.store in=\cpfaGreen,
  blue/.store in=\cpfaBlue,
  ink/.store in=\cpfaInk,
  stripe spacing/.store in=\cpfaStripeSpacing,
  stripe count/.code={\pgfmathsetmacro{\cpfaStripeSpacing}{28.346/(#1)}},
  stripe width/.store in=\cpfaStripeWidth,
  grid width/.store in=\cpfaGridWidth,
  border width/.store in=\cpfaBorderWidth
}%
\newcommand{\cpfaSensorPixel}[4]{%
  \begin{scope}[shift={(#1,#2)}]
    \clip (0,0) rectangle (28.346,28.346);
    \begin{scope}[rotate=#3]
      \foreach \stripe in {-\cpfaStripeLimit,...,\cpfaStripeLimit} {
        \draw[\cpfaInk,line width=\cpfaStripeWidth,line cap=butt]
          (-60,{(#4/2.706228+\stripe)*\cpfaStripeSpacing}) --
          (60,{(#4/2.706228+\stripe)*\cpfaStripeSpacing});
      }
    \end{scope}
  \end{scope}%
}%
\newcommand{\cpfaSensorSetup}[1]{%
  \pgfkeys{/cpfa/.cd,width=40mm,
    red=cpfaOriginalRed,green=cpfaOriginalGreen,blue=cpfaOriginalBlue,
    ink=cpfaOriginalInk,stripe spacing=2.706228,
    stripe width=.5412456bp,grid width=.5bp,border width=\cpfaGridWidth,#1}%
  \pgfmathtruncatemacro{\cpfaStripeLimit}{ceil(57/\cpfaStripeSpacing)+1}%
  \pgfmathsetlengthmacro{\cpfaHalfBorderWidth}{.5*(\cpfaBorderWidth)}%
}%
\newcommand{\CPFASensor}[1][]{%
\begingroup%
\cpfaSensorSetup{#1}%
\resizebox{\cpfaWidth}{!}{%
\begin{tikzpicture}[x=1bp,y=1bp]
  \path[use as bounding box] (0,0) rectangle (113.386,113.386);
  \clip (0,0) rectangle (113.386,113.386);
  \fill[\cpfaRed] (0,56.693) rectangle (56.693,113.386);
  \fill[\cpfaGreen] (56.693,56.693) rectangle (113.386,113.386);
  \fill[\cpfaGreen] (0,0) rectangle (56.693,56.693);
  \fill[\cpfaBlue] (56.693,0) rectangle (113.386,56.693);
  \cpfaSensorPixel{0}{85.039}{90}{.848800}
  \cpfaSensorPixel{28.346}{85.039}{45}{1.507464}
  \cpfaSensorPixel{0}{56.693}{135}{.407486}
  \cpfaSensorPixel{28.347}{56.693}{0}{2.138080}
  \cpfaSensorPixel{56.693}{85.039}{90}{.851800}
  \cpfaSensorPixel{85.039}{85.039}{45}{1.509585}
  \cpfaSensorPixel{56.693}{56.693}{135}{.402536}
  \cpfaSensorPixel{85.039}{56.693}{0}{2.138080}
  \cpfaSensorPixel{0}{28.346}{90}{.848800}
  \cpfaSensorPixel{28.347}{28.346}{45}{1.510292}
  \cpfaSensorPixel{0}{0}{135}{.412436}
  \cpfaSensorPixel{28.346}{0}{0}{2.141080}
  \cpfaSensorPixel{56.693}{28.346}{90}{.851800}
  \cpfaSensorPixel{85.039}{28.346}{45}{1.511706}
  \cpfaSensorPixel{56.693}{0}{135}{.407486}
  \cpfaSensorPixel{85.039}{0}{0}{2.141080}
  \foreach \edge in {28.346,56.693,85.039} {
    \draw[\cpfaInk,line width=\cpfaGridWidth] (\edge,0) -- (\edge,113.386);
    \draw[\cpfaInk,line width=\cpfaGridWidth] (0,\edge) -- (113.386,\edge);
  }
  \draw[\cpfaInk,line width=\cpfaBorderWidth,line join=miter]
    ([xshift=\cpfaHalfBorderWidth,yshift=\cpfaHalfBorderWidth]0,0)
    rectangle
    ([xshift=-\cpfaHalfBorderWidth,yshift=-\cpfaHalfBorderWidth]113.386,113.386);
\end{tikzpicture}%
}%
\endgroup%
}%
\newcommand{\CPFAPolarizer}[2][]{%
\begingroup%
\cpfaSensorSetup{width=3mm,#1}%
\ifnum#2=0 \def\cpfaPhase{2.138080}%
\else\ifnum#2=45 \def\cpfaPhase{1.507464}%
\else\ifnum#2=90 \def\cpfaPhase{.848800}%
\else\ifnum#2=135 \def\cpfaPhase{.407486}%
\else\PackageError{cpfa}{Unsupported polarizer angle #2}{Use 0, 45, 90, or 135.}%
  \def\cpfaPhase{0}%
\fi\fi\fi\fi%
\resizebox{\cpfaWidth}{!}{%
\begin{tikzpicture}[x=1bp,y=1bp]
  \path[use as bounding box] (0,0) rectangle (28.346,28.346);
  \clip (0,0) rectangle (28.346,28.346);
  \fill[white] (0,0) rectangle (28.346,28.346);
  \cpfaSensorPixel{0}{0}{#2}{\cpfaPhase}
  \draw[\cpfaInk,line width=\cpfaBorderWidth,line join=miter]
    ([xshift=\cpfaHalfBorderWidth,yshift=\cpfaHalfBorderWidth]0,0)
    rectangle
    ([xshift=-\cpfaHalfBorderWidth,yshift=-\cpfaHalfBorderWidth]28.346,28.346);
\end{tikzpicture}%
}%
\endgroup%
}%

\definecolor{cpfaChannelRed}{RGB}{240,104,108}
\definecolor{cpfaChannelGreen}{RGB}{104,213,140}
\definecolor{cpfaChannelBlue}{RGB}{56,136,218}

\newcommand{\CPFAStokesPlateEdges}[2]{%
  \coordinate (#1-back-nw) at ([shift={(.26,.26)}]#1.north west);
  \coordinate (#1-back-ne) at ([shift={(.26,.26)}]#1.north east);
  \coordinate (#1-back-se) at ([shift={(.26,.26)}]#1.south east);
  \fill[#2!80] (#1.north west) -- (#1-back-nw)
    -- (#1-back-ne) -- (#1.north east) -- cycle;
  \fill[#2] (#1.north east) -- (#1-back-ne)
    -- (#1-back-se) -- (#1.south east) -- cycle;
  \draw[#2!70!black,line width=.25pt,line join=round,line cap=butt]
    (#1.south west) -- (#1.south east) -- (#1-back-se)
    -- (#1-back-ne) -- (#1-back-nw) -- (#1.north west) -- cycle
    (#1.north west) -- (#1.north east) -- (#1.south east)
    (#1.north east) -- (#1-back-ne);
}

\newcommand{\CPFAStokesTriplet}[4]{%
  \begin{scope}[shift={(#2)},
    stokes sheet/.style={inner sep=0pt,outer sep=0pt}]
    \node[stokes sheet,fill=cpfaChannelBlue!65,
      minimum width=#3,minimum height=\dimexpr#3*5/6\relax]
      (#1-blue) at (1.1,1.1) {};
    \CPFAStokesPlateEdges{#1-blue}{cpfaChannelBlue}
    \node[stokes sheet,fill=cpfaChannelGreen!65,
      minimum width=#3,minimum height=\dimexpr#3*5/6\relax]
      (#1-green) at (.55,.55) {};
    \CPFAStokesPlateEdges{#1-green}{cpfaChannelGreen}
    \node[stokes sheet] (#1) at (0,0)
      {\csname ArxivImage:figures/cpfa-intro-panels/s#4.png\endcsname{width=#3}};
    \CPFAStokesPlateEdges{#1}{cpfaChannelRed}
  \end{scope}%
}

\definecolor{linkblue}{rgb}{0.21,0.49,0.74}
\usepackage[pagebackref,breaklinks,colorlinks,allcolors=linkblue]{hyperref}
\renewcommand*{\HyperDestNameFilter}[1]{main-#1}
\usepackage{bookmark}
\usepackage[capitalize]{cleveref}
\crefname{figure}{Figure}{Figures}
\Crefname{figure}{Figure}{Figures}
\hypersetup{pdftitle={HDR-Polar1K: An HDR Stokes Image Dataset for Learning Color-Polarization Reconstruction},pdfauthor={Ryota Maeda, Yonghee Oh, Yunseong Moon, Seung-Hwan Baek}}
\title{HDR-Polar1K: An HDR Stokes Image Dataset\\for Learning Color-Polarization Reconstruction}
\author{Ryota Maeda\qquad Yonghee Oh\qquad Yunseong Moon\qquad Seung-Hwan Baek\\[4pt]POSTECH}
\makeatletter
\let\SuppMakeTitle\maketitle
\let\SuppInnerTitle\@maketitle
\def\sharedtitle{HDR-Polar1K: An HDR Stokes Image Dataset\\for Learning Color-Polarization Reconstruction}
\apptocmd{\@maketitle}{\input{figures/main_scene_teaser}}{}{\PackageError{arxiv}{Could not insert teaser}{}}
\makeatother

\begin{document}
\bookmark[level=0,page=1,view={XYZ 72 756 null}]{Main Paper}
\makeatletter
\let\OriginalAddToContents\addtocontents
\def\addtocontents#1#2{\ifstrequal{#1}{toc}{}{\OriginalAddToContents{#1}{#2}}}
\makeatother
\maketitle
\begin{abstract}
Color-polarization filter array (CPFA) cameras require joint demosaicing and denoising to recover dense color Stokes images from RAW measurements. Existing training datasets are limited by acquisition artifacts and insufficient coverage of controlled optical scenes. We introduce HDR-Polar1K, a dataset of 1,008 HDR color Stokes images spanning everyday and controlled scenes. Repeated, exposure-bracketed CPFA capture and CPFA-aware downsampling produce images free of saturation and nonphysical Stokes vectors. These HDR images enable brightness variation and calibrated sensor simulation for training under diverse sensor conditions. Training on HDR-Polar1K improves reconstruction across multiple networks. We also introduce Raw2Stokes, which produces physically valid Stokes images with optional sensor conditioning. Trained on HDR-Polar1K using our calibrated sensor simulation, Raw2Stokes enables practical color-polarization reconstruction from full-resolution real RAW captures.

\end{abstract}

\section{Introduction}
\label{sec:introduction}

\begin{figure}[t]
  \centering
  \input{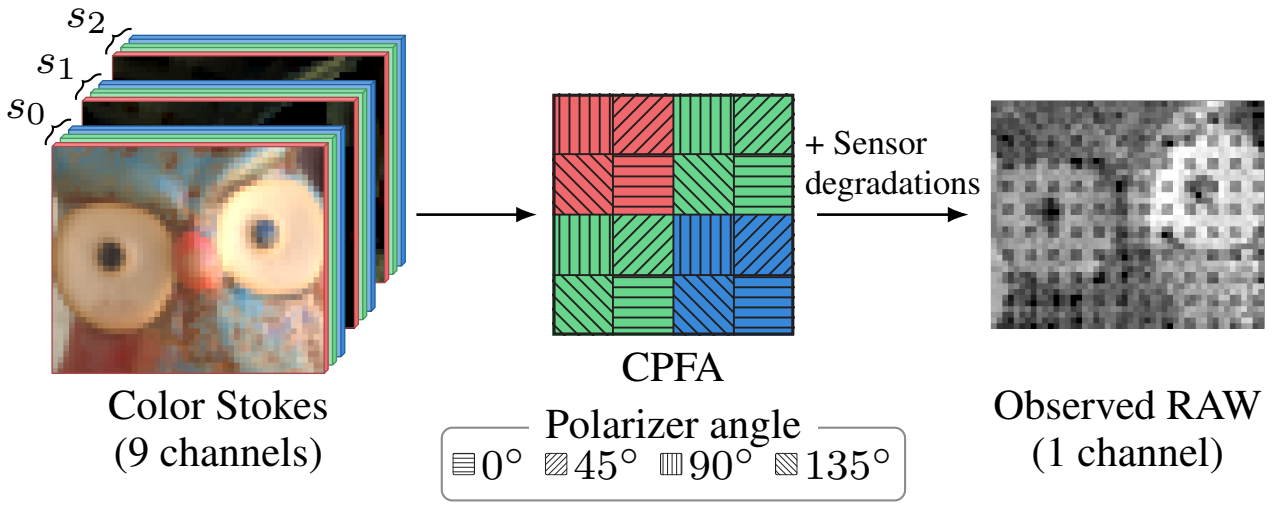}%
  \caption{\textbf{CPFA image formation.} Each sensor pixel measures a single color channel and polarization orientation, yielding a RAW mosaic from which the color Stokes image needs to be reconstructed. Real RAW measurements are additionally affected by sensor degradations.}
  \label{fig:cpfa-capture}
\end{figure}

Recent advances in color-polarization filter array (CPFA) sensors, such as the Sony IMX250MYR, have made color-polarization imaging more accessible. These sensors integrate spatially varying color filters and linear polarizers, with each pixel sampling a specific color channel and polarization orientation~\cite{maruyama2018backilluminated}, as shown in \cref{fig:cpfa-capture}. This design enables single-shot capture in a compact form factor, facilitating applications that exploit polarization cues, including reflection separation~\cite{lei2020polarized,yao2025polarfree}, shape reconstruction~\cite{cui2017polarimetric}, material analysis~\cite{liang2022multimodal}, and inverse rendering~\cite{li2024neisf}.

Realizing these benefits requires reconstructing a color Stokes image from a mosaiced CPFA RAW measurement affected by sensor noise and other degradations. A color Stokes image represents linear polarization using three Stokes parameters for each RGB channel, yielding nine components per pixel. Recovering these quantities from real RAW measurements requires joint color-polarization demosaicing and denoising (JDD). Learning-based approaches address these coupled tasks by learning image priors~\cite{zhang2026cpddnet}. Their performance depends on both network design and the quality and diversity of the Stokes images used for supervision.

We identify limitations of existing datasets. First, their Stokes images often contain artifacts caused by saturation, spatial misalignment from sequential capture with a rotating polarizer~\cite{bigue2023production,dumoulin2024impact}, and acquisition-induced color inconsistencies~\cite{wen2021sparse}. Second, existing datasets primarily cover everyday scenes, whereas polarization imaging is also used for controlled optical measurements involving active polarized illumination~\cite{hwang2022sparse,choi2026snapshot} and photoelastic imaging~\cite{dave2025nest,maeda2025event}. The illumination and material conditions in these measurements are not well represented in existing training data.

To address existing dataset limitations, we introduce \textbf{HDR-Polar1K}, a dataset of 1,008 HDR color Stokes images spanning everyday and controlled scenes. Controlled scenes include colored and polarized illumination, structured light, and polarizing or photoelastic objects. Across both categories, we intentionally capture rich spatial variations in color and polarization to provide informative supervision for patch-based training. We use a fixed CPFA camera to avoid rotation-induced spatial and color inconsistencies. Repeated captures, exposure bracketing, and CPFA-aware downsampling yield low-noise HDR Stokes images at reduced spatial resolution. The resulting images contain no saturated pixels, and all Stokes vectors satisfy the physical constraints in \cref{eq:physical-stokes-validity}.

To train networks with realistically simulated sensor degradations, we develop a reusable calibrated forward model of the Sony IMX250MYR sensor. The model simulates CPFA measurements affected by noise, dark signal, hot pixels, clipping, and quantization across different sensor gain, exposure time, and temperature settings. By uniformly scaling the HDR Stokes vectors before simulation, we generate training pairs of degraded RAW mosaics and corresponding clean Stokes images at varying scene brightness levels.

Alongside the dataset, we introduce \textbf{Raw2Stokes}, a simple but effective network for joint color-polarization demosaicing and denoising. Its output parameterization enforces the physical constraints of Stokes vectors, guaranteeing physically valid predictions and improving reconstruction accuracy. It optionally incorporates sensor metadata, allowing the same trained model to operate with or without this information.

We evaluate HDR-Polar1K against existing training datasets across multiple reconstruction networks for joint demosaicing and denoising. The results show that training on HDR-Polar1K improves reconstruction accuracy across all evaluated networks, with Raw2Stokes achieving the best performance. We further apply Raw2Stokes trained on HDR-Polar1K directly to real RAW captures across diverse scenes and acquisition conditions. These results demonstrate that the dataset and network together support practical color-polarization reconstruction from real CPFA measurements. We will release the dataset, trained models, and calibrated sensor model.

\section{Related Work}
\label{sec:related-work}

\paragraph{Color-polarization image datasets.}
Existing datasets provide dense color-polarization images for developing and evaluating color-polarization reconstruction methods~\cite{wen2019cpdnet,morimatsu2020eari,qiu2021linear,wen2021sparse,zhou2025pidsr,rahman2025polarization,luo2026updem,wu2026deep}. These datasets commonly use division-of-time acquisition with an RGB camera and a rotating polarizer. Ideally, this approach samples the same scene location at each polarizer orientation, allowing direct Stokes estimation at every pixel. In practice, spatial misalignment~\cite{bigue2023production,dumoulin2024impact}, acquisition-induced color inconsistencies~\cite{wen2021sparse}, and saturation can corrupt the resulting Stokes images. CPFA cameras have also been used to construct datasets for 3D reconstruction~\cite{han2024nersp}, glass segmentation~\cite{mei2022glass}, and reflection removal~\cite{yao2025polarfree}, with task-specific ground truth. These datasets contain scenes tailored to their target tasks and do not provide HDR Stokes images suitable for training color-polarization reconstruction. HDR-Polar1K provides high-quality HDR Stokes images across diverse everyday and controlled scenes.

\paragraph{RAW-to-Stokes reconstruction.}
Polarization demosaicing recovers dense polarization information from mosaiced measurements, with color-polarization demosaicing methods additionally recovering color. Existing approaches include interpolation-based methods~\cite{gao2011bilinear,morimatsu2020eari,morimatsu2021igri,lu2024hybrid,su2025universal}, optimization-based methods~\cite{qiu2021linear,wen2021sparse,luo2024nlcsr,dumoulin2025lmmse}, and neural networks~\cite{wen2019cpdnet,nguyen2022tcpdnet,zheng2024scfnet,zhou2025pidsr,wang2026swincpd,luo2026updem,wu2026deep,li2026pugdiff}. Complementary work addresses polarization image denoising~\cite{zhang2017pca,abubakar2018bm3d,li2020polarimetricdenoising,liang2022bm3d,liu2023pol2pol,liu2023pm3dcnn}, low-light enhancement~\cite{zhou2023lowlight,yang2026zeropie}, defective-pixel correction~\cite{ratliff2007dead}, and dynamic-range enhancement through polarization redundancy~\cite{lapray2020exploiting}. Joint demosaicing and denoising has also been studied for polarization mosaics~\cite{li2023joint,zhang2026cpddnet}. However, existing polarization demosaicing and denoising networks generally do not guarantee physically valid Stokes outputs. Raw2Stokes addresses this limitation by enforcing these constraints through its output parameterization.

\section{Background on Stokes Vectors}
\label{sec:polarization-representation}

For each RGB channel, the linear Stokes vector \(\mathbf{s}=[s_0,s_1,s_2]^{\mathsf T}\) describes total intensity through \(s_0\) and linear polarization through \(s_1\) and \(s_2\). The three color channels together yield nine Stokes components per pixel.
We omit the circular component \(s_3\), which conventional CPFA sensors cannot measure. A physically valid linear Stokes vector satisfies
\begin{equation}
  s_0\geq0,\qquad s_1^2+s_2^2\leq s_0^2.
  \label{eq:physical-stokes-validity}
\end{equation}

We visualize linear polarization using the angle of linear polarization (AoLP) and degree of linear polarization (DoLP):
\begin{equation}
  \psi=\frac12\operatorname{atan2}(s_2,s_1), \qquad
  \rho=\frac{\sqrt{s_1^2+s_2^2}}{s_0}.
  \label{eq:dolp-aolp}
\end{equation}

CPFA sensors sample intensities through linear polarizers oriented at \(\theta\in\{0^\circ,45^\circ,90^\circ,135^\circ\}\).
For a given color channel and scene location, the ideal intensity measured through a linear polarizer at orientation \(\theta\) is
\begin{equation}
  I_\theta=\frac12\left(s_0+s_1\cos2\theta+s_2\sin2\theta\right).
  \label{eq:linear-analyzer-model}
\end{equation}

Given four polarizer measurements for the same color channel and scene location, the Stokes vector can be estimated as
\begin{equation}
  \begin{aligned}
    s_0 &= \frac{I_0+I_{45}+I_{90}+I_{135}}{2}, \\
    s_1 &= I_0-I_{90}, \qquad s_2=I_{45}-I_{135}.
  \end{aligned}
  \label{eq:four-analyzer-stokes}
\end{equation}

Because a CPFA samples different color--polarizer combinations at different sensor locations, as shown in \cref{fig:cpfa-capture}, dense Stokes reconstruction requires recovering the missing measurements.

\begin{figure}[t]
  \centering
  \begingroup
\newlength{\referenceunit}
\setlength{\referenceunit}{0.9807mm}
\def\native{figures/main-reference-construction-ball-panels}

\newcommand{\cimg}[2]{\csname ArxivImage:figures/main-reference-construction-ball-panels/#1.png\endcsname{width=#2\referenceunit}}
\newcommand{\ctriplet}[3]{%
 \begin{scope}[shift={(#2)},sheet/.style={inner sep=0pt,outer sep=0pt}]
 \node[sheet,fill=cpfaChannelBlue!65,minimum width=13.72980mm,minimum height=11.44477mm] (#1-blue) at (1.1,1.1) {};
 \CPFAStokesPlateEdges{#1-blue}{cpfaChannelBlue}
 \node[sheet,fill=cpfaChannelGreen!65,minimum width=13.72980mm,minimum height=11.44477mm] (#1-green) at (.55,.55) {};
 \CPFAStokesPlateEdges{#1-green}{cpfaChannelGreen}
 \node[sheet] (#1) at (0,0) {\cimg{s#3}{14}};
 \CPFAStokesPlateEdges{#1}{cpfaChannelRed}
 \end{scope}}
\begin{tikzpicture}[x=0.98070mm,y=0.98070mm,font=\fontsize{7}{8}\selectfont,
 image/.style={inner sep=0pt,outer sep=0pt,draw=black!65,line width=.3pt},
 flow/.style={-{Latex[length=1.22588mm]},line width=.45pt},
 small/.style={font=\fontsize{7}{8}\selectfont,inner sep=0pt,align=center},
 roi/.style={draw=teal!75!black,line width=.5pt},
 link/.style={draw=teal!75!black,line width=.4pt,densely dotted}]
 \foreach \i/\y in {0/30,2/18,5/6}{
  \node[image] (back\i) at (10.5,\y+2.5) {\cimg{raw-\i}{9}};
  \node[image] (middle\i) at (8.6,\y+.6) {\cimg{raw-\i}{9}};
  \node[image] (raw\i) at (8,\y) {\cimg{raw-\i}{9}};
  \foreach \d in {.55,.95,1.35}{\fill ([xshift=\d \referenceunit,yshift=\d \referenceunit]middle\i.north west) circle[radius=0.10788mm];}
  \coordinate (stack-center\i) at (9.25,\y+1.25);
  \draw[line width=.45pt] ([xshift=0.68649mm]back\i.east |- stack-center\i) -- (19,\y+1.25);
 }
 \foreach \y in {25.25,13.25}{
  \foreach \d in {-.45,0,.45}{\fill (8.6,\y+\d) circle[radius=0.10788mm];}
 }
 \draw[line width=.45pt] (19,7.25) -- (19,31.25);
 \draw[decorate,decoration={brace,amplitude=0.58842mm},line width=.4pt] (2.7,2.25)--(2.7,36.25)
  node[midway,rotate=90,anchor=south,yshift=0.98070mm,small] {6 exposures};
 \node[small,anchor=west] at (15.4,34.4) {32 frames / exposure};
 \node[image] (hdr) at (41,19.25) {\cimg{hdr}{14}};
 \draw[flow] (19,19.25) -- (hdr.west) node[midway,above=0.98070mm,small] {Average} node[midway,below=0.98070mm,small] {HDR fusion};
 \ctriplet{s2}{76,25.25}{2}
 \ctriplet{s1}{74,22.25}{1}
 \ctriplet{s0}{72,19.25}{0}
 \foreach \n/\k in {s2/2,s1/1,s0/0}{
  \draw[decorate,decoration={brace,amplitude=0.44132mm},line width=.35pt]
  ([xshift=-0.39228mm,yshift=0.09807mm]\n.north west) -- ([xshift=-0.09807mm,yshift=0.39228mm]\n-blue-back-nw)
  node[midway,anchor=south east,inner sep=0pt,xshift=-0.39228mm,yshift=0.29421mm,font=\fontsize{7}{8}\selectfont] {$s_\k$};
 }
 \draw[flow] (hdr.east) -- (s0.west) node[midway,above=0.98070mm,small] {CPFA-aware} node[midway,below=0.98070mm,small] {downsampling};
 \node[small,anchor=north] at (9.25,1.1) {RAW frames};
 \node[small,anchor=north] at (41,12.35) {Full-resolution\\HDR mosaic\\$2448\times2048$};
 \node[small,anchor=north] at (74,12.35) {Low-resolution\\HDR color Stokes\\$153\times128$};
\end{tikzpicture}%
\endgroup%
  \caption{\textbf{Color Stokes image construction.} Repeated RAW frames are averaged at each exposure and merged into an HDR RAW image. Downsampling produces color Stokes images.}
  \label{fig:main-reference-construction}
\end{figure}
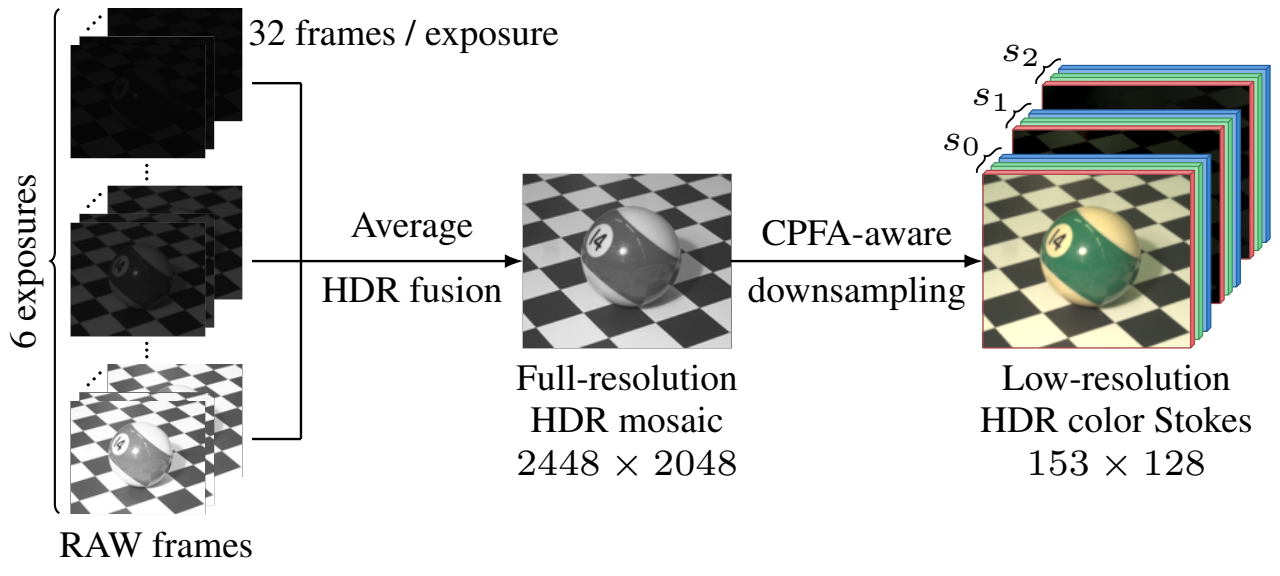

\begin{figure}[!t]
  \centering
  \input{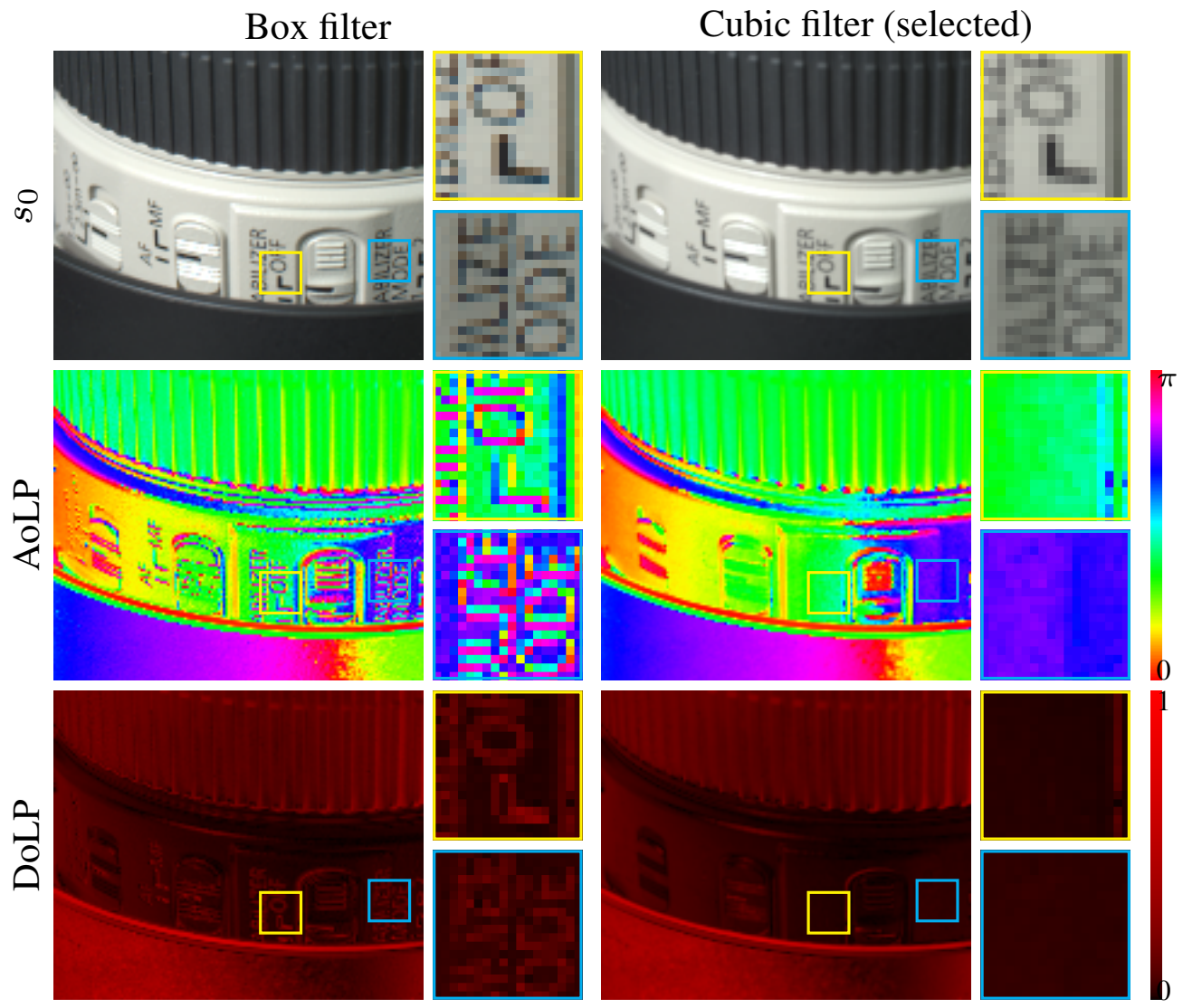}
  \caption{\textbf{Filter comparison for CPFA-aware downsampling.} The box filter produces false color and polarization (AoLP and DoLP) near intensity boundaries. The selected cubic filter matches low-order spatial responses across color--polarizer channels for CPFA, reducing these artifacts.}
  \label{fig:main-filter-comparison}
\end{figure}

\begin{table*}[t]
  \centering
  \small
  \definecolor{DatasetGreen}{HTML}{E8F3E8}
  \definecolor{DatasetRed}{HTML}{F9E4E4}
  \setlength{\tabcolsep}{2.4pt}
  \renewcommand{\arraystretch}{1.12}
  \caption{\textbf{Comparison of public color-polarization image datasets for demosaicing.} HDR-Polar1K provides HDR Stokes images without image shifts, color inconsistencies, saturated pixels, or nonphysical Stokes vectors. Although HDR-Polar1K has lower image resolution, it provides accurate Stokes images for patch-based JDD training, where the accuracy of Stokes values takes priority over image resolution. For acquisition consistency, checks (\(\checkmark\)) and crosses (\(\times\)) indicate the absence and presence of issues, respectively. Reported zeros are exact.}
  \label{tab:supp-dataset-saturation}
  \begin{tabular*}{\textwidth}{@{\extracolsep{\fill}}llrcc>{\centering\arraybackslash}p{14mm}>{\centering\arraybackslash}p{14mm}rr@{}}
    \toprule
    & \multicolumn{4}{c}{Properties} & \multicolumn{2}{c}{Acquisition consistency} & \multicolumn{2}{c}{Invalid samples} \\
    \cmidrule(lr){2-5}\cmidrule(lr){6-7}\cmidrule(l){8-9}
    Dataset & Capture method & Scenes & Resolution & HDR & Spatial & Color & Saturation \(\downarrow\) & Nonphysical \(\downarrow\) \\
    \midrule
    Wen et al. (2019)~\cite{wen2019cpdnet} & RGB + rotating polarizer & 105 & \(1456\mathord{\times}1088\) & \cellcolor{DatasetRed}\(\times\) & \cellcolor{DatasetRed}\(\times\) & \cellcolor{DatasetRed}\(\times\) & 1.23\% & 0.00111\% \\
    Morimatsu et al. (2020)~\cite{morimatsu2020eari} & RGB + rotating polarizer & 40 & \(1024\mathord{\times}768\) & \cellcolor{DatasetRed}\(\times\) & \cellcolor{DatasetRed}\(\times\) & \cellcolor{DatasetGreen}\(\checkmark\) & 0.52\% & 0.00002\% \\
    Qiu et al. (2021)~\cite{qiu2021linear} & RGB + rotating polarizer & 40 & \(1024\mathord{\times}1024\) & \cellcolor{DatasetRed}\(\times\) & \cellcolor{DatasetGreen}\(\checkmark\) & \cellcolor{DatasetGreen}\(\checkmark\) & 0.30\% & \textbf{0.00000\%} \\
    Wen et al. (2021)~\cite{wen2021sparse} & RGB + rotating polarizer & 50 & \(720\mathord{\times}540\) & \cellcolor{DatasetRed}\(\times\) & \cellcolor{DatasetRed}\(\times\) & \cellcolor{DatasetGreen}\(\checkmark\) & 0.21\% & 0.22179\% \\
    Zhou et al. (2025)~\cite{zhou2025pidsr} & RGB + rotating polarizer & 138 & \(2448\mathord{\times}2048\) & \cellcolor{DatasetRed}\(\times\) & \cellcolor{DatasetRed}\(\times\) & \cellcolor{DatasetGreen}\(\checkmark\) & 1.27\% & 0.00329\% \\
    Rahman et al. (2025)~\cite{rahman2025polarization} & RGB + rotating polarizer & 40 & \(1024\mathord{\times}768\) & \cellcolor{DatasetRed}\(\times\) & \cellcolor{DatasetRed}\(\times\) & \cellcolor{DatasetGreen}\(\checkmark\) & 2.44\% & 0.00014\% \\
    Luo et al. (2026)~\cite{luo2026updem} & RGB + rotating polarizer & 80 & Multiple\textsuperscript{a} & \cellcolor{DatasetRed}\(\times\) & \cellcolor{DatasetRed}\(\times\) & \cellcolor{DatasetGreen}\(\checkmark\) & 5.34\% & 0.07134\% \\
    Wu et al. (2026)~\cite{wu2026deep} & RGB + rotating polarizer & 177 & Multiple\textsuperscript{a} & \cellcolor{DatasetRed}\(\times\) & \cellcolor{DatasetRed}\(\times\) & \cellcolor{DatasetGreen}\(\checkmark\) & 1.64\% & 0.02137\% \\
    \midrule
    \textbf{HDR-Polar1K (ours)} & CPFA + downsampling & 1,008 & \(153\mathord{\times}128\) & \cellcolor{DatasetGreen}\(\checkmark\) & \cellcolor{DatasetGreen}\(\checkmark\) & \cellcolor{DatasetGreen}\(\checkmark\) & \textbf{0.00\%} & \textbf{0.00000\%} \\
    \bottomrule
  \end{tabular*}
  \par\smallskip
  {\footnotesize\raggedright\textsuperscript{a}The datasets of Luo et al. and Wu et al. contain images at multiple resolutions, ranging from \(1024\mathord{\times}1024\) to \(2448\mathord{\times}2048\).\par}
\end{table*}

\input{figures/main_dataset_issues}

\section{HDR-Polar1K Dataset}
\label{sec:dataset}

HDR-Polar1K provides HDR color Stokes images of 1,008 everyday and controlled scenes.

\subsection{Scene Content}

\paragraph{Everyday and controlled scenes.}
The dataset comprises 782 everyday scenes and 226 controlled scenes. Everyday scenes include indoor and outdoor captures of common objects and surfaces under everyday lighting conditions, covering diverse materials, colors, shapes, and textures. Controlled scenes include setups with polarized illumination~\cite{hwang2022sparse}, colored illumination~\cite{choi2026snapshot}, structured light~\cite{baek2021transport}, and photoelastic materials~\cite{dave2025nest,maeda2025event}, as well as polarizers and displays. These setups capture strongly polarized regions, color-dependent polarization, projected patterns, and photoelastic fringes.

\paragraph{Rich color-polarization details.}
We deliberately include spatial detail in both color and polarization throughout each scene to provide informative supervision, as shown in \cref{fig:dataset-scene-teaser}. This strategy is inspired by prior work that uses challenging patches to improve color demosaicing~\cite{gharbi2016deep,tedla2025examining}.

\subsection{Color Stokes Image Construction}

We construct HDR color Stokes images using a CPFA camera, as shown in \cref{fig:main-reference-construction}.

First, we acquire repeated, exposure-bracketed RAW images using a fixed camera equipped with a Sony IMX250MYR CPFA sensor. Temporal averaging suppresses noise~\cite{abdelhamed2018sidd,liu2023mobilehdr}, while exposure bracketing retains unsaturated measurements across a wide dynamic range. Specifically, we record \(32\) frames at each of six exposure times spaced \(2\)~EV apart, using the minimum gain of \(0\)~dB. We average the frames at each exposure in linear RAW space, subtract the calibrated dark signal, and merge the bracketed exposures into an HDR RAW image~\cite{debevec1997recovering}.

Second, we construct color Stokes images from the merged HDR RAW mosaic through CPFA-aware downsampling. At each output location, we aggregate nearby RAW measurements separately for each color--polarizer channel, forming a virtual pixel with four polarizer intensities per color. Because these channels sample different positions, simple averaging can produce false polarization near texture boundaries, even when the light is unpolarized. Inspired by the construction of RGB images through downsampling~\cite{khashabi2014joint}, we find that an existing cubic filter~\cite{mitchell1988reconstruction} has effective spatial sampling kernels with matching first and second moments across CPFA color--polarizer channels. This makes the channels respond consistently to low-order spatial intensity variations, reducing spurious polarization near boundaries. \Cref{fig:main-filter-comparison} provides a qualitative comparison with box filtering, showing reduced false color and polarization. We downsample the \(2448\mathord{\times}2048\) HDR RAW mosaic to \(153\mathord{\times}128\) images and convert the aggregated polarization measurements to Stokes images using \cref{eq:four-analyzer-stokes}. These images retain sufficient spatial extent for our \(64\mathord{\times}64\) training patches, while dense color-polarization detail provides informative supervision throughout each scene. The moment analysis and filter selection are detailed in the supplementary material.

\subsection{Comparison with Existing Datasets}
\label{sec:dataset-comparison}

\Cref{tab:supp-dataset-saturation} compares HDR-Polar1K with existing datasets in terms of capture method, scene count, image resolution, acquisition consistency, and invalid samples. Our comparison identifies spatial and color inconsistencies, as well as saturation, in existing datasets. First, division-of-time capture with an RGB camera ideally samples the same scene location at every polarizer orientation, but rotating the polarizer can shift images slightly. The resulting pixel-level intensity differences can appear as spurious polarization in the estimated Stokes images~\cite{bigue2023production,dumoulin2024impact}. \Cref{fig:main-dataset-issues}(a) illustrates this displacement in an example from Zhou et al.~\cite{zhou2025pidsr}. Second, acquisition-induced color inconsistencies can distort Stokes estimates, as reported by Wen et al.~\cite{wen2021sparse} for the earlier Wen et al.~\cite{wen2019cpdnet} dataset. Third, clipped intensities can corrupt polarization estimates near highlights, as shown in \cref{fig:main-dataset-issues}(b).

In contrast, as reported in \cref{tab:supp-dataset-saturation}, HDR-Polar1K contains no saturated pixels, and every Stokes vector satisfies the physical constraints in \cref{eq:physical-stokes-validity}.

\section{Calibrated Sensor Simulation}
\label{sec:sensor-simulation}

We calibrate a forward model of the Sony IMX250MYR sensor to synthesize realistic CPFA RAW measurements across varying sensor gain, exposure time, and temperature settings. The model accounts for Poisson--Gaussian noise~\cite{foi2008poisson} and pixel-dependent dark signals, including hot pixels. We use it to generate CPFA RAW mosaics from brightness-scaled Stokes images under sampled sensor conditions, followed by quantization and clipping. Each simulated mosaic is paired with the corresponding clean, brightness-scaled Stokes image. Calibration and simulation details are provided in the supplementary material.

\section{Raw2Stokes Network}
\label{sec:raw2stokes}

\begin{figure}[t]
  \centering
  \input{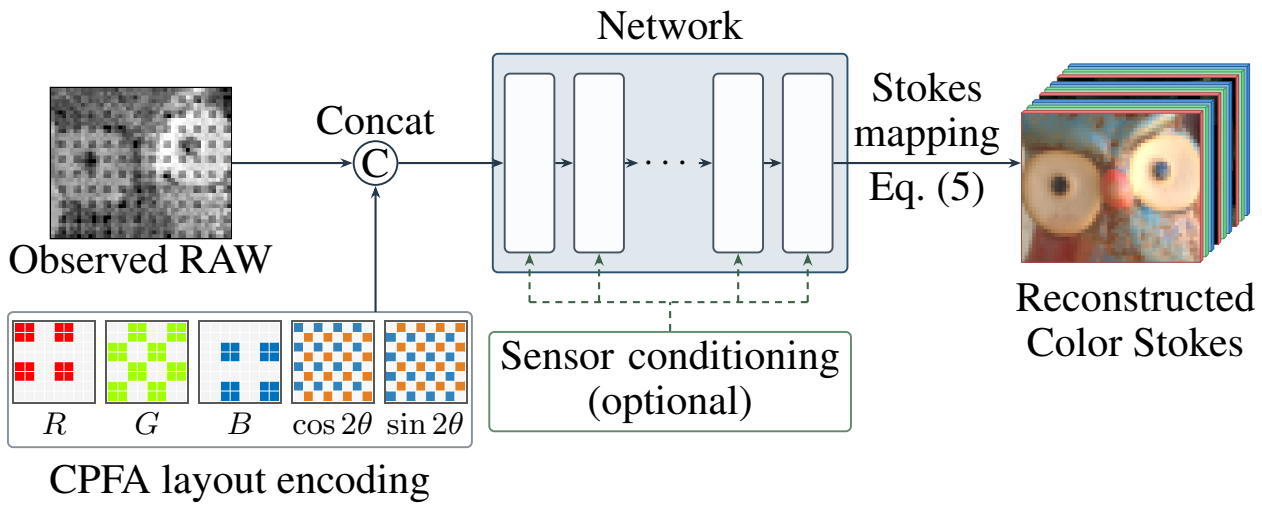}
  \caption{\textbf{Overview of the Raw2Stokes network.} The observed RAW mosaic is concatenated with CPFA encoding maps and processed by the network. A constrained output mapping produces physically valid color Stokes images. Optional metadata (e.g., gain, exposure time, and temperature) conditions the network.}
  \label{fig:main-raw2stokes-network}
\end{figure}

\begin{figure*}[t]
  \centering
  \input{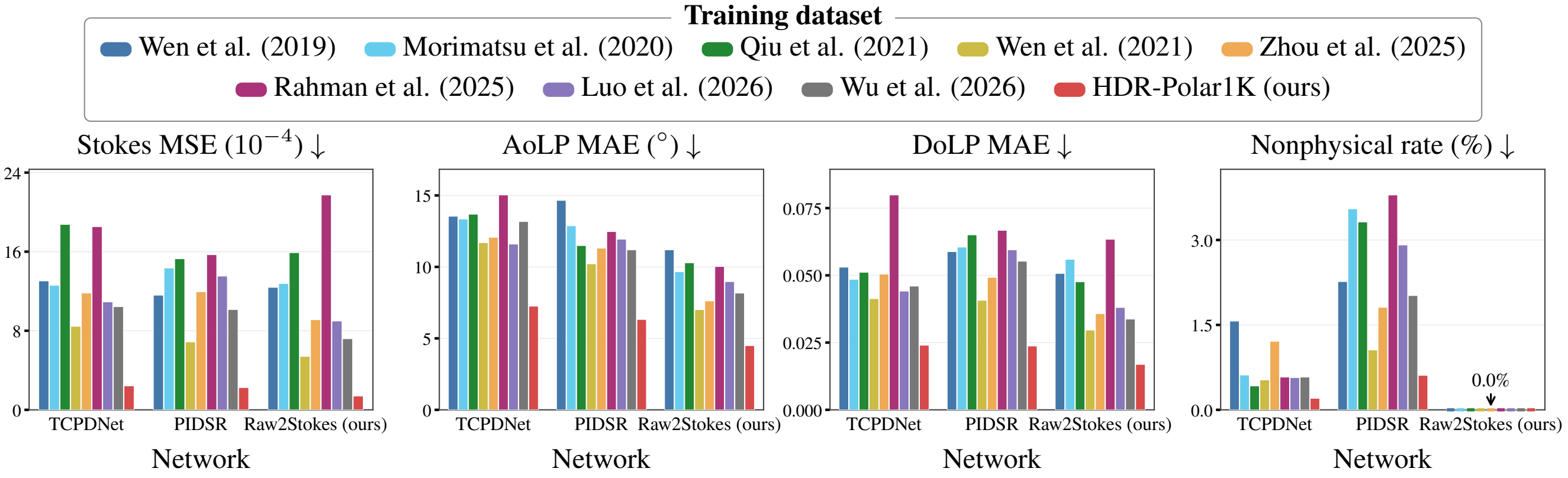}
  \caption{\textbf{Quantitative comparison of training datasets.} We compare nine training datasets using TCPDNet, PIDSR, and Raw2Stokes on 100 HDR-Polar1K test scenes under low-noise conditions. For Stokes MSE, we exclude the five scenes at each tail of each model's ranking and average the remaining 90 scenes because a few scenes produce outliers for models trained on existing datasets. All other metrics use the mean over all 100 scenes. Raw2Stokes trained on HDR-Polar1K achieves the lowest reconstruction errors.}
  \label{tab:training-data-comparison}
\end{figure*}

We introduce Raw2Stokes as a simple but effective baseline for joint color-polarization demosaicing and denoising.

\paragraph{Network architecture and training.}
The network shown in \cref{fig:main-raw2stokes-network} takes a RAW CPFA mosaic and its CPFA sensor layout as inputs. We encode the RGB filter identity at each pixel with a three-dimensional one-hot vector and the polarizer orientation \(\theta\) as \((\cos2\theta,\sin2\theta)\), inspired by the pattern-encoding of Tedla et al.~\cite{tedla2025examining}. These five encoding channels accompany the RAW image as spatial input maps to an RCAN-based backbone~\cite{zhang2018rcan}. The network outputs a color Stokes image and is trained by minimizing a Charbonnier loss~\cite{charbonnier1997deterministic} between predicted and target Stokes images.

\paragraph{Physically valid Stokes output.}
The output head predicts nine unconstrained values per pixel, comprising a triplet \((z_0,z_1,z_2)\) for each RGB channel. We map each triplet to a linear Stokes vector as
\begin{equation}
  \begin{aligned}
    \widehat{s}_0 &= \operatorname{softplus}(z_0),\\
    (\widehat{s}_1,\widehat{s}_2) &= \widehat{s}_0\frac{(z_1,z_2)}{\sqrt{1+z_1^2+z_2^2}}.
  \end{aligned}
  \label{eq:main-stokes-output}
\end{equation}
This parameterization guarantees the physical constraints of the Stokes vector (\cref{eq:physical-stokes-validity}).

\paragraph{Optional sensor conditioning.}
The network additionally accepts gain, exposure time, temperature, the quantization step, and white balance, allowing it to adapt reconstruction to the sensor conditions. Since this metadata is not always available, Raw2Stokes supports both blind and non-blind operation depending on metadata availability. We encode each metadata value together with its own binary availability indicator and use the encoded metadata to modulate the network's feature maps through feature-wise linear modulation~\cite{perez2018film}. During training, we randomly withhold individual metadata fields, including cases with no auxiliary information, so that the same trained model can handle complete, partial, or missing metadata.

\section{Experiments}
\label{sec:experiments}

We evaluate HDR-Polar1K as training data for joint color-polarization demosaicing and denoising (JDD). We first compare training datasets across multiple reconstruction networks on HDR-Polar1K test scenes, then examine reconstruction from full-resolution real sensor measurements.

\subsection{Effect of the training dataset and network}
\label{sec:hardness-results}

\paragraph{Datasets and networks.}
We compare HDR-Polar1K with the eight existing datasets listed in \cref{tab:supp-dataset-saturation} using three networks: TCPDNet~\cite{nguyen2022tcpdnet}, PIDSR~\cite{zhou2025pidsr}, and Raw2Stokes. For PIDSR, we omit the super-resolution stage and train only its demosaicing network. We divide HDR-Polar1K into 858 training, 50 validation, and 100 test scenes.

\paragraph{Training settings.}
We train all networks on all datasets using the same sensor simulation procedure, random \(64\mathord{\times}64\) patch sampling, batch size of 16, learning rate and its schedule, and number of training iterations. Sensor degradations are simulated across varied brightness levels and sensor conditions as described in \cref{sec:sensor-simulation}. TCPDNet and PIDSR retain their original loss functions. We select the network weights with the lowest validation loss, computed using the training loss function on the validation split of the corresponding dataset.

\paragraph{Evaluation.}
We evaluate on HDR-Polar1K because acquisition artifacts and saturation in existing ground-truth images can distort the evaluation, as discussed in \cref{sec:dataset-comparison}. All networks receive the same simulated low-noise RAW inputs. For a fair comparison, Raw2Stokes is evaluated in the blind setting.

\paragraph{Metrics.}
We report mean squared error (MSE) for Stokes images, mean absolute error (MAE) for DoLP, mean absolute angular error for AoLP, and the percentage of Stokes vectors violating the physical constraints in \cref{eq:physical-stokes-validity}. We exclude low-intensity regions, where DoLP and AoLP are unstable, and additionally exclude weakly polarized regions, where AoLP is unreliable. Predicted DoLP is clipped to \([0,1]\) before computing its MAE. We summarize scene-level Stokes MSE with a trimmed mean, removing 5\% from each tail to limit the influence of extreme errors. Other metrics use the mean over all 100 test scenes. Detailed training and evaluation procedures and untrimmed results are provided in the supplementary material.

\paragraph{Quantitative results.}
\Cref{tab:training-data-comparison} compares the nine training datasets, showing that HDR-Polar1K improves reconstruction accuracy across all tested networks. When trained on HDR-Polar1K, Raw2Stokes achieves better reconstruction accuracy than the compared networks. Regardless of the training dataset, Raw2Stokes achieves a zero nonphysical rate through the output parameterization in \cref{eq:main-stokes-output}. An ablation in the supplementary material shows that this parameterization also improves reconstruction accuracy over unconstrained Stokes prediction.

\begin{figure*}[t]
  \centering
  \input{figures/training-data-results-grid}%
  \caption{\textbf{Visual comparison of RAW-to-Stokes reconstruction across training datasets.} Training datasets are compared using the same Raw2Stokes architecture on our test scenes (\(153\mathord{\times}128\)) under low-noise conditions. Bilinear interpolation~\cite{maeda2019polanalyser} is included as a baseline. Numbers at the bottom right of each image report full-image \(s_0\) MSE in units of \(10^{-4}\), AoLP MAE in degrees, and DoLP MAE.}
  \label{fig:training-data-results}
\end{figure*}

\paragraph{Qualitative results.}
\Cref{fig:training-data-results} compares training datasets using the same Raw2Stokes architecture. In (b), even Bilinear interpolation performs better than networks trained on Morimatsu et al.~\cite{morimatsu2020eari} or Zhou et al.~\cite{zhou2025pidsr} in some regions. This suggests that networks struggle with color-polarization patterns poorly represented in their training data. In contrast, training on HDR-Polar1K produces more accurate reconstructions in both scenes.

\subsection{Reconstruction from real sensor measurements}
\label{sec:real-raw-results}

\begin{figure*}[p]
  \centering
  \PaperSRGBRasters
  \input{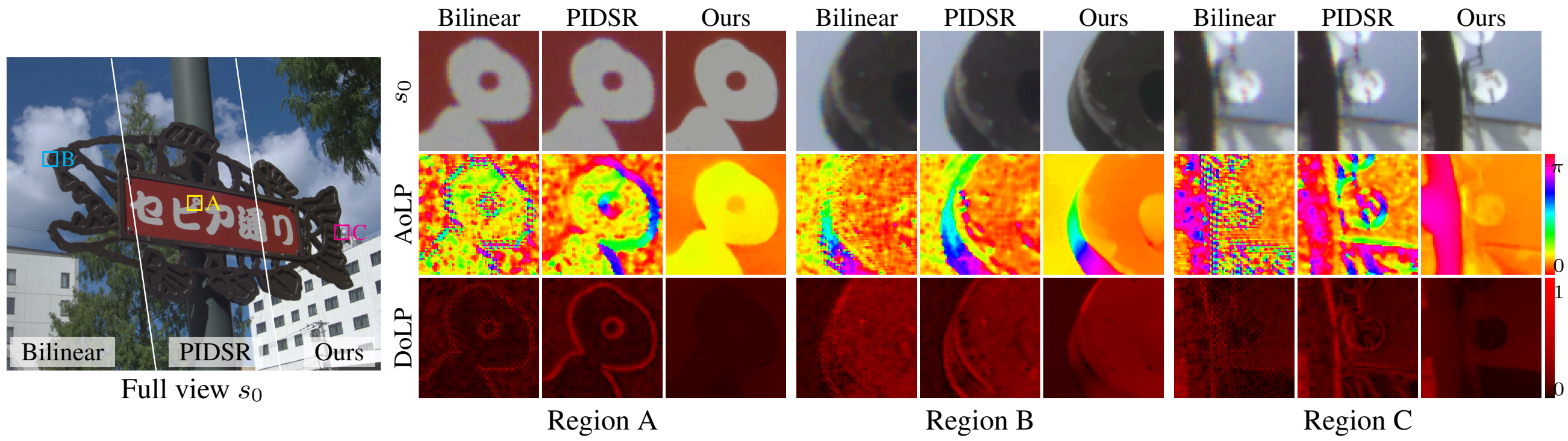}
  \par\nointerlineskip
  \vspace{1mm}
  {\small (a) Low-noise daytime scene (gain: \(0\,\mathrm{dB}\), exposure time: \(1/1000\,\mathrm{s}\))\par}
  \vspace{0.5mm}
  \input{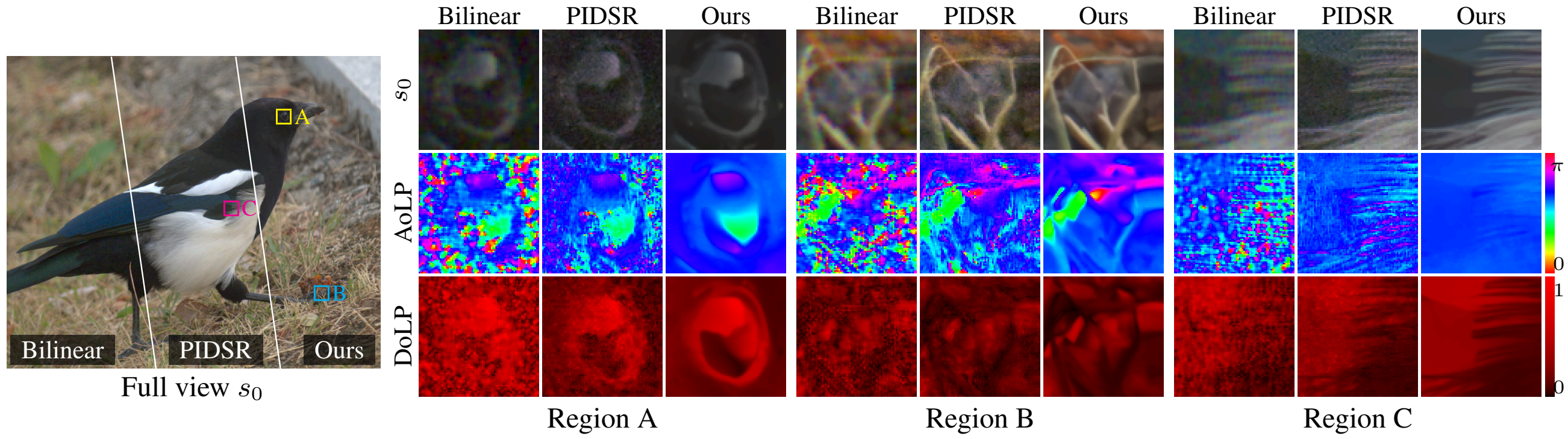}
  \par\nointerlineskip
  \vspace{1mm}
  {\small (b) Moderate-noise evening scene (gain: \(26\,\mathrm{dB}\), exposure time: \(1/770\,\mathrm{s}\))\par}
  \vspace{0.5mm}
  \input{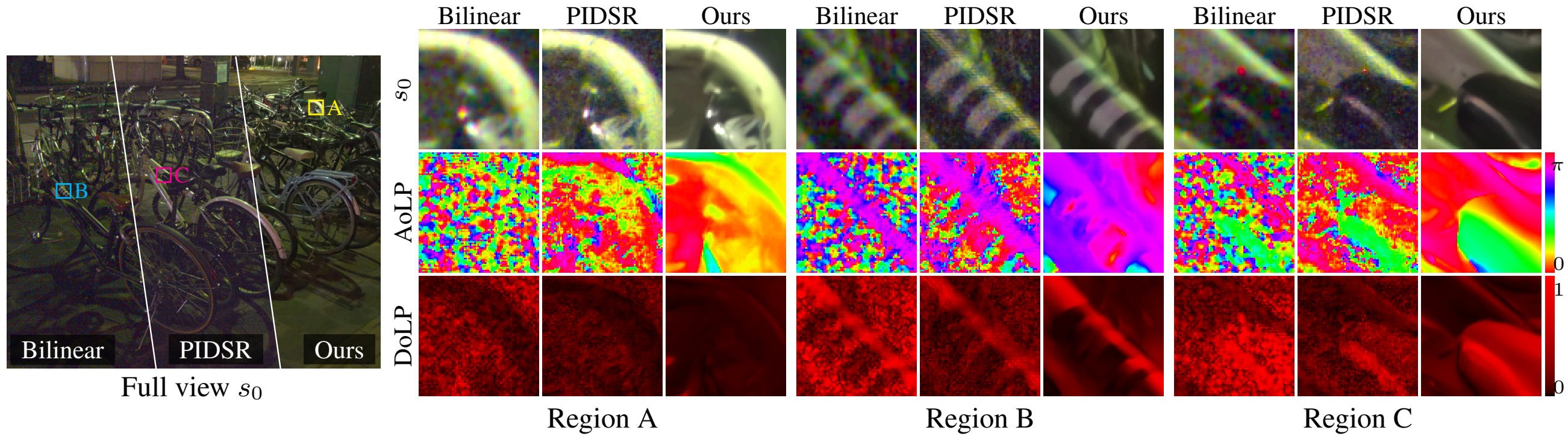}
  \par\nointerlineskip
  \vspace{1mm}
  {\small (c) High-noise nighttime scene (gain: \(31\,\mathrm{dB}\), exposure time: \(1/20\,\mathrm{s}\))\par}
  \vspace{0.5mm}
  \input{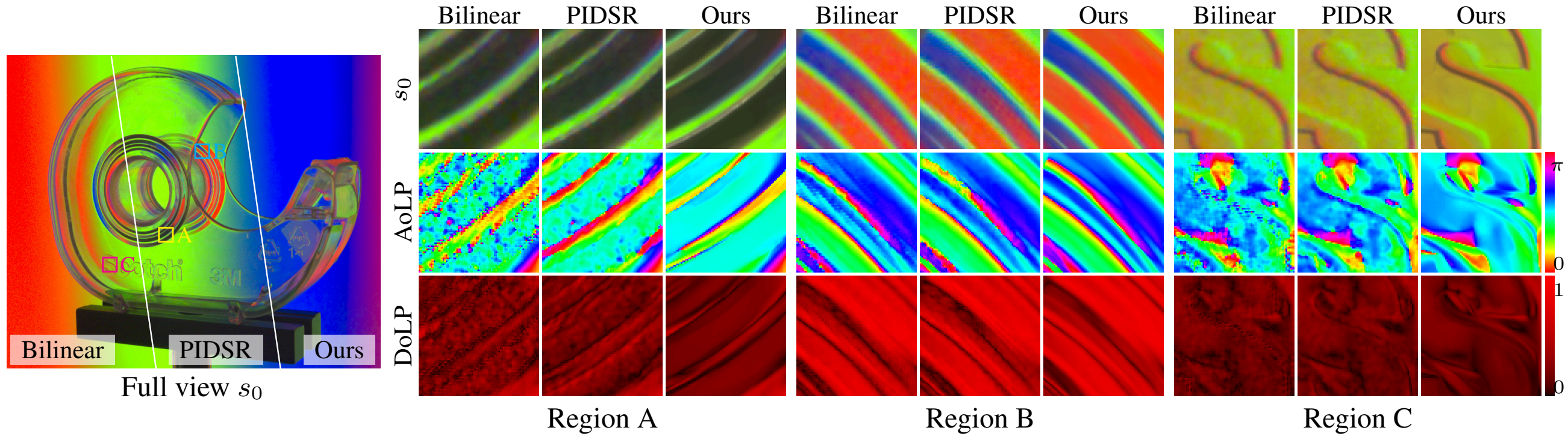}
  \par\nointerlineskip
  \vspace{1mm}
  {\small (d) Photoelastic scene (gain: \(12\,\mathrm{dB}\), exposure time: \(1/130\,\mathrm{s}\))\par}
  \caption{\textbf{RAW-to-Stokes reconstruction of full-resolution real captures.} We compare Raw2Stokes with Bilinear interpolation~\cite{maeda2019polanalyser} and PIDSR~\cite{zhou2025pidsr}. All images were captured with a Sony IMX250MYR sensor at \(2448\mathord{\times}2048\). These results demonstrate the applicability of a single Raw2Stokes model to real RAW reconstruction across different scenes and acquisition conditions.}
  \label{fig:main-real-jdd}
\end{figure*}

\paragraph{Setup.}
We further apply Raw2Stokes trained on HDR-Polar1K directly to full-resolution (\(2448\mathord{\times}2048\)) RAW measurements captured with a Sony IMX250MYR color-polarization sensor in the non-blind setting. \Cref{fig:main-real-jdd} presents qualitative comparisons with Bilinear interpolation~\cite{maeda2019polanalyser} and PIDSR~\cite{zhou2025pidsr} (official pretrained weights) under varying noise conditions and in a photoelastic scene. All methods process each full frame without spatial tiling.

\paragraph{Results.}
Raw2Stokes suppresses noise while preserving image structures and polarization details, even in high-gain captures where noise is more pronounced. The same model reconstructs both everyday scenes and the photoelastic scene across different acquisition conditions. Full-frame reconstructions of these captures and additional examples are provided in the supplementary material.

\section{Discussion}
\label{sec:discussion}

Our Stokes image construction involves a trade-off between image fidelity and spatial resolution. Although substantial downsampling improves the quality of the Stokes images, residual errors may occur in scenes with very high spatial frequencies. More aggressive downsampling can further suppress these errors, but reduces the number of pixels per image, leaving fewer crop locations and limiting spatial context. Capturing more scenes could compensate for fewer crop locations, but the limited spatial context would remain an issue. A higher-resolution CPFA sensor could alleviate this constraint by retaining larger Stokes images under stronger downsampling. Obtaining a large, diverse collection of accurate, high-resolution color Stokes images remains an open challenge.

\section{Conclusion}
\label{sec:conclusion}

We presented HDR-Polar1K, a dataset of 1,008 HDR color Stokes images spanning diverse everyday and controlled scenes. Repeated, exposure-bracketed CPFA captures and CPFA-aware downsampling yield low-noise, unsaturated, and physically valid Stokes images. Using these images and a calibrated sensor model, we synthesize training pairs across varied scene brightness and sensor conditions for joint demosaicing and denoising. Training on HDR-Polar1K improves all evaluated networks on our test set. We also introduced Raw2Stokes, which guarantees physically valid Stokes outputs and achieves the best accuracy. Its results on full-resolution real RAW captures demonstrate the practical applicability of HDR-Polar1K and Raw2Stokes to color-polarization reconstruction.

\section*{LLM Usage Statement}
We used ChatGPT for manuscript clarity and grammar and OpenAI Codex for coding and debugging. All generated content was reviewed and verified by the authors, who take full responsibility for the work.

\clearpage
\onecolumn
\raggedbottom\fussy
\makeatletter
\let\@addtocurcol\suppaddtocurcol
\let\addtocontents\OriginalAddToContents
\renewcommand*{\HyperDestNameFilter}[1]{supp-#1}
\setcounter{section}{0}\setcounter{subsection}{0}\setcounter{figure}{0}\setcounter{table}{0}\setcounter{equation}{0}\setcounter{footnote}{0}\setcounter{page}{1}
\setlength{\parindent}{1pc}
\let\@maketitle\SuppInnerTitle
\def\@title{\sharedtitle}\def\@author{Supplementary Material}\let\@thanks\@empty
\let\maketitle\SuppMakeTitle
\let\thebibliography\BRorg@thebibliography
\let\bibitem\BRorg@bibitem
\let\Hy@backout\@gobble
\patchcmd{\l@section}{1.5em}{2.2em}{}{}
\renewcommand*\l@subsection{\@dottedtocline{2}{1.5em}{3.0em}}
\makeatother
\crefname{section}{Sec.}{Secs.}
\Crefname{section}{Section}{Sections}
\Crefname{table}{Table}{Tables}
\crefname{table}{Tab.}{Tabs.}
\hypersetup{linkcolor=black}
\makeatletter
\let\SuppDefaultPNG\Gread@png
\let\SuppSetSRGB\PaperSRGBRasters
\renewcommand{\PaperSRGBRasters}{%
  \SuppSetSRGB
  \let\SuppTaggedPNG\Gread@png
  \def\Gread@png##1{%
    \edef\SuppPNGType{\pdffiledump offset 25 length 1 {##1}}%
    \ifnum\SuppPNGType=2 \let\SuppReadPNG\SuppTaggedPNG
    \else\ifnum\SuppPNGType=6 \let\SuppReadPNG\SuppTaggedPNG
    \else\let\SuppReadPNG\SuppDefaultPNG\fi\fi
    \SuppReadPNG{##1}}}
\makeatother

\newcommand{\suppScoredPanel}[3]{\tikz{%
  \node[inner sep=0pt,outer sep=0pt] (panel) {#1};
  \path (panel.south west) -- (panel.south east) coordinate[pos=#2] (corner);
  \node[overlay,anchor=south east,fill=black,fill opacity=.65,text=white,text opacity=1,inner xsep=.6mm,inner ysep=.5mm,outer sep=0pt] at (corner) {\fontsize{7}{8}\selectfont \(#3\)};}}

\PaperSRGBRasters
\bookmark[level=0,page=\the\numexpr\ReadonlyShipoutCounter+1\relax,view={XYZ 72 756 null}]{Supplementary Material}
\pagestyle{plain}
\maketitle
\NewCommandCopy\OrigCite\cite
\makeatletter
\DeclareRobustCommand{\cite}[1]{\begingroup\let\citelist\@empty\@for\next:=#1\do{\ifx\citelist\@empty\edef\citelist{S-\next}\else\edef\citelist{\citelist,S-\next}\fi}\edef\temp{\endgroup\noexpand\OrigCite{\citelist}}\temp}
\makeatother
\leavevmode\hspace*{\parindent}\ignorespaces This supplement first describes HDR-Polar1K acquisition, CPFA-aware downsampling, and image quality comparisons across datasets in \cref{sec:supp-hdr-reference-construction,sec:supp-cpfa-cubic,sec:supp-dataset-comparison}. We then present calibrated sensor simulation, the experimental setup, and Raw2Stokes implementation in \cref{sec:supp-raw-simulation,sec:supp-implementation-training}. Next, we compare training datasets in \cref{sec:benchmark}, evaluate existing reconstruction methods in \cref{sec:supp-additional-comparisons}, and analyze Raw2Stokes design choices in \cref{sec:supp-model-analyses}. We conclude with full-resolution real-capture reconstruction in \cref{sec:supp-reconstruction-full-resolution} and downstream applications in \cref{sec:supp-downstream-applications}.

\paragraph{Visualization of color Stokes vector images.}
Throughout the paper, we visualize color Stokes vector images in terms of RGB intensity (\(s_0\)), angle of linear polarization (AoLP), and degree of linear polarization (DoLP). For RGB \(s_0\) images from our camera, we apply white balance (WB) followed by a color correction matrix (CCM) in linear RGB. We estimate a separate CCM for each lighting condition using ColorChecker images captured outdoors (direct sunlight and shade on a clear day) and indoors (fluorescent lighting). We use C5~\cite{afifi2021crosscamera} to estimate the illuminant and select a calibrated WB and CCM for each scene. We manually adjust WB when the automatic setting leaves a noticeable color cast. For each comparison, we apply the same WB and CCM to all results for a given scene. Quantitative evaluation of \(s_0\) uses the values before either correction. For external datasets, we retain the original color processing. When a single AoLP or DoLP image is shown for a color scene, we compute it from the luminance-weighted components \(s_{k,Y}=0.2126s_{k,R}+0.7152s_{k,G}+0.0722s_{k,B}\) for \(k=0,1,2\). In AoLP maps, gray indicates pixels with undefined orientation, and in DoLP maps, gray indicates values outside \([0,1]\).

\tableofcontents

\clearpage

\section{HDR-Polar1K acquisition and color Stokes image construction}
\label{sec:supp-hdr-reference-construction}

\subsection{Capture settings and controlled scene setups}
\label{sec:supp-hdr-polar1k-scenes}

\paragraph{Imaging hardware.}
We use a tripod-mounted Blackfly S USB3 camera (FLIR, BFS-U3-51S5PC-C) equipped with a color-polarization sensor (Sony, IMX250MYR). The camera is fitted with a 35~mm F/1.85 lens (Edmund Optics, \#33-306), providing a 35-mm-format-equivalent focal length of approximately 137~mm and a narrow, telephoto field of view. To increase the depth of field, we stop down the lens as far as the available light allows.

  \newsavebox{\captureSettingsTable}
  \sbox{\captureSettingsTable}{\small\setlength{\tabcolsep}{5pt}\renewcommand{\arraystretch}{1.12}%
  \begin{tabular}{@{}ll@{}}
    \toprule
    Spinnaker node & Value \\
    \midrule
    \texttt{Width}, \texttt{Height} & 2448, 2048 \\
    \texttt{AdcBitDepth} & \texttt{Bit12} \\
    \texttt{PixelFormat} & \texttt{BayerRG12p} \\
    \texttt{ExposureAuto}, \texttt{GainAuto} & \texttt{Off} \\
    \texttt{BlackLevel} & 0\% \\
    \texttt{BlackLevelClampingEnable} & \texttt{False} \\
    \texttt{IspEnable} & \texttt{False} \\
    \texttt{GammaEnable} & \texttt{False} \\
    \texttt{LUTEnable} & \texttt{False} \\
    \texttt{DefectCorrectStaticEnable} & \texttt{False} \\
    \texttt{BalanceWhiteAuto} & \texttt{Off} \\
    \texttt{BalanceRatio} (\texttt{Red}, \texttt{Blue}) & 1, 1 \\
    \bottomrule
  \end{tabular}}%
\begin{wraptable}{r}{\wd\captureSettingsTable}
  \vspace{-6pt}
  \centering
  \small
  \setlength{\tabcolsep}{5pt}
  \renewcommand{\arraystretch}{1.12}

  \captionsetup{width=\wd\captureSettingsTable}
  \caption{\textbf{RAW frame acquisition settings.} Node names follow the FLIR Spinnaker SDK.}
  \label{tab:supp-camera-settings}
  \usebox{\captureSettingsTable}\par
  \vspace{-6pt}
\end{wraptable}

\paragraph{Acquisition settings.}
To capture low-noise RAW frames, we configure the camera with a sensor gain of 0~dB, the maximum ADC bit depth of 12 bits, and in-camera processing disabled, as detailed in Table~\ref{tab:supp-camera-settings}. For each scene, we capture 32 frames at each of six exposure levels separated by 2~EV, yielding 192 frames in total. A complete sequence takes approximately 5 seconds to 1 minute, depending on scene brightness. We restrict acquisition to scenes that remain static throughout the sequence.

\paragraph{RAW storage and metadata.}
The camera outputs 12-bit pixel values in packed \texttt{BayerRG12p} format, which we store losslessly. For each frame, we also record the actual exposure time, gain, and camera settings as metadata.

\begin{figure*}[tp]
  \centering
  \newcommand{\panelHeight}{.140\linewidth}
  \newcommand{\setupPhoto}[2][\panelHeight]{\csname ArxivImage:figures/controlled-scene-setups/#2.jpg\endcsname{height=#1}}
  \newcommand{\annotatedSetup}[2]{%
    \begin{tikzpicture}[baseline=(photo.south),font=\fontsize{8}{9}\selectfont]
      \node[inner sep=0pt,outer sep=0pt,anchor=south west] (photo) at (0,0) {\setupPhoto{#1}};
      \begin{scope}[shift={(photo.south west)},x={(photo.south east)},y={(photo.north west)},overlay,
          callout/.style={-{Latex[length=1.1mm]},line width=.4pt,shorten >=.65mm},
          every node/.style={inner sep=.3mm}]
        #2
      \end{scope}
    \end{tikzpicture}}
  \small
  \begin{tabular}{@{}r@{\hspace{1.5mm}}c@{\hspace{0.8mm}}c@{\hspace{0.8mm}}c@{}}
    \annotatedSetup{display-illumination}{
      \node[white] (label) at (.35,.072) {Display};
      \draw[callout,white] (label.north) -- (.385,.222);
      \node[white] (subject) at (.69,.115) {Objects};
      \draw[callout,white] (subject.north) -- (.605,.366);
    } &
    \includegraphics[height=\panelHeight]{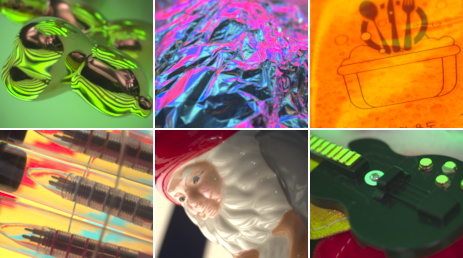} &
    \includegraphics[height=\panelHeight]{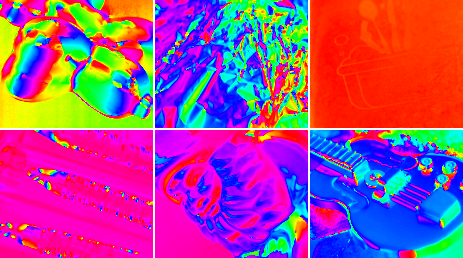} &
    \includegraphics[height=\panelHeight]{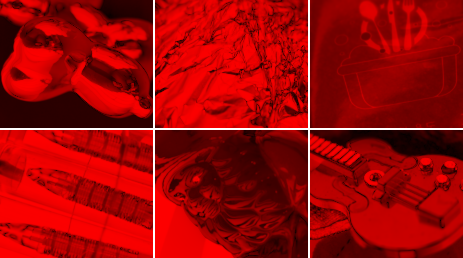} \\
    \multicolumn{1}{@{}c@{\hspace{1.5mm}}}{Scene setup} & $s_0$ & AoLP & DoLP \\
    \multicolumn{4}{c}{(a) Display-illuminated scene (50 scenes)} \\[2mm]
    \annotatedSetup{projector}{
      \node[white,anchor=east] (label) at (.99,.79) {Projector};
      \draw[callout,white] (label.south) -- (.79,.55);
      \node[white,align=center] (pattern) at (.27,.16) {Projected\\structured light};
      \draw[callout,white] (pattern.north) -- (.28,.388);
    } &
    \includegraphics[height=\panelHeight]{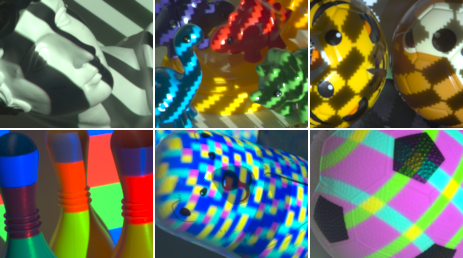} &
    \includegraphics[height=\panelHeight]{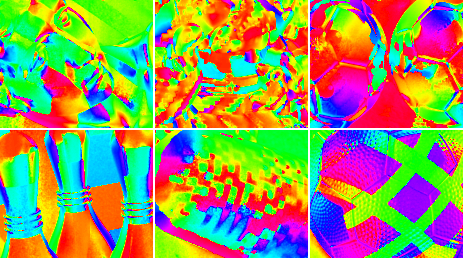} &
    \includegraphics[height=\panelHeight]{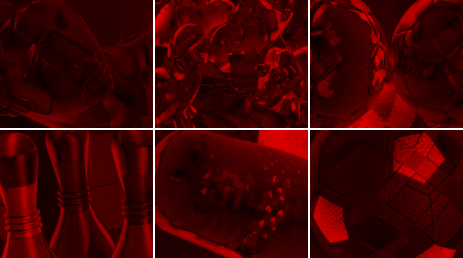} \\
    \multicolumn{1}{@{}c@{\hspace{1.5mm}}}{Scene setup} & $s_0$ & AoLP & DoLP \\
    \multicolumn{4}{c}{(b) Projector scene (52 scenes)} \\[2mm]
    \annotatedSetup{pink-filtered-light}{
      \node[white,align=center] (filterlabel) at (.22,.85) {Color filter\\or polarizer};
      \draw[callout,white] (filterlabel.south east) -- (.436,.64);
      \node[white] (objects) at (.18,.48) {Objects};
      \draw[callout,white] (objects.south) -- (.23,.325);
      \node[white,align=center] (led) at (.82,.488) {LED light\\source};
      \draw[callout,white] (led.north) -- (.81,.675);
      \node[anchor=south east,inner sep=0pt] (filters) at (.718,.098) {\setupPhoto[.034\linewidth]{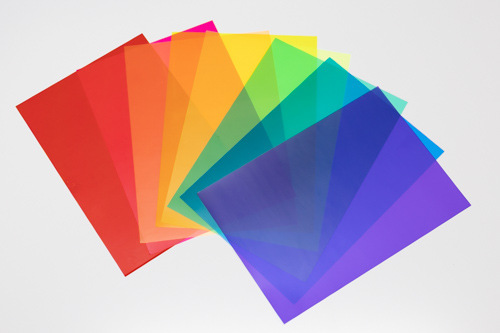}};
      \node[anchor=south east,inner sep=0pt] (sheet) at (.99,.098) {\setupPhoto[.034\linewidth]{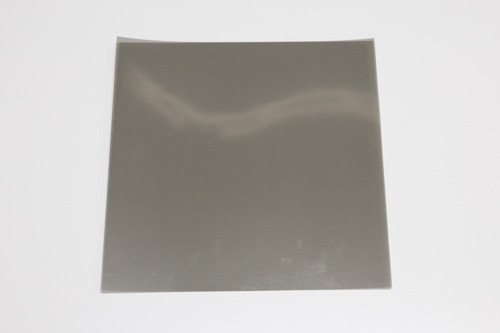}};
      \node[white,anchor=north,font=\fontsize{6.5}{7.5}\selectfont] at ([yshift=.2mm]filters.south) {Color filters};
      \node[white,anchor=north,font=\fontsize{6.5}{7.5}\selectfont] at ([yshift=.2mm]sheet.south) {Polarizer};
    } &
    \includegraphics[height=\panelHeight]{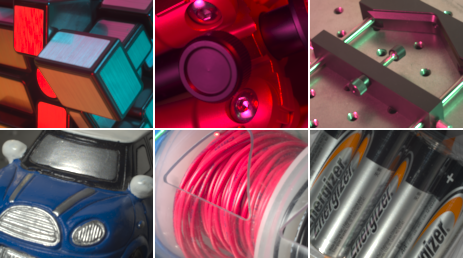} &
    \includegraphics[height=\panelHeight]{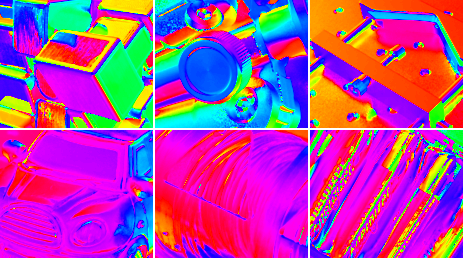} &
    \includegraphics[height=\panelHeight]{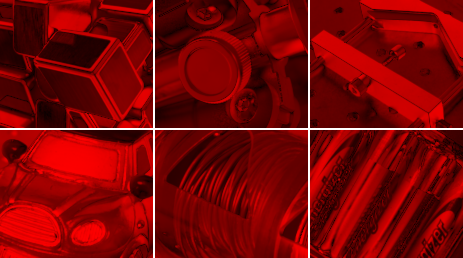} \\
    \multicolumn{1}{@{}c@{\hspace{1.5mm}}}{Scene setup} & $s_0$ & AoLP & DoLP \\
    \multicolumn{4}{c}{(c) Colored or polarized illumination (53 scenes)} \\[2mm]
    \begin{tikzpicture}[baseline=(photo.south),font=\fontsize{8}{9}\selectfont]
      \node[inner sep=0pt,outer sep=0pt,anchor=south west] (photo) at (0,0) {\setupPhoto{photoelastic-setup}};
      \begin{scope}[shift={(photo.south west)},x={(photo.south east)},y={(photo.north west)},overlay,
          callout/.style={-{Latex[length=1.1mm]},line width=.4pt,shorten >=.65mm}]
        \node[inner sep=.3mm] (displaylabel) at (.125,.195) {Display};
        \draw[callout] (displaylabel.north) -- (.155,.405);
        \node[align=center,inner sep=.3mm] (objectlabel) at (.375,.785) {Photoelastic object};
        \draw[callout] (objectlabel.south) -- (.326,.598);
        \node[anchor=south east,inner sep=0pt,outer sep=0pt] (detail) at (1,.11) {\includegraphics[width=.0927\linewidth]{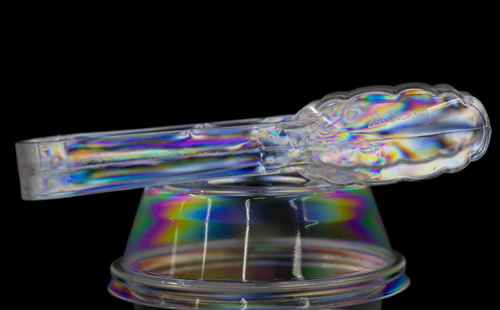}};
        \draw[black,line width=.6pt] (detail.south west) rectangle (detail.north east);
        \node[anchor=north east,inner sep=0pt,font=\fontsize{7.5}{8.5}\selectfont] at ([yshift=-.55mm]detail.south east) {Photo with a polarizer};
      \end{scope}
    \end{tikzpicture} &
    \includegraphics[height=\panelHeight]{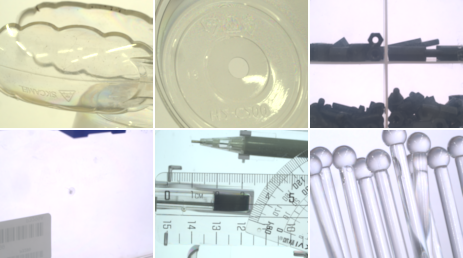} &
    \includegraphics[height=\panelHeight]{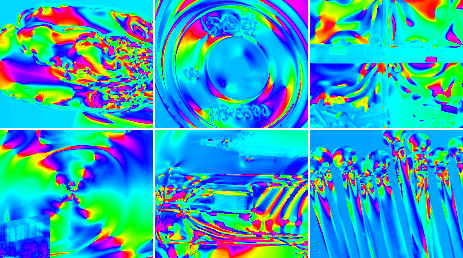} &
    \includegraphics[height=\panelHeight]{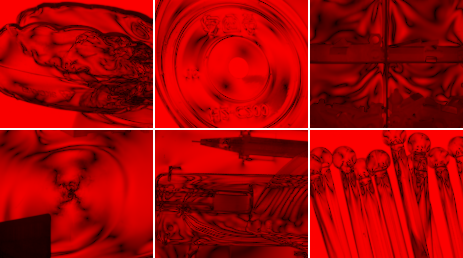} \\
    \multicolumn{1}{@{}c@{\hspace{1.5mm}}}{Scene setup} & $s_0$ & AoLP & DoLP \\
    \multicolumn{4}{c}{(d) Photoelastic scene (50 scenes)} \\[2mm]
    \setupPhoto{display} &
    \includegraphics[height=\panelHeight]{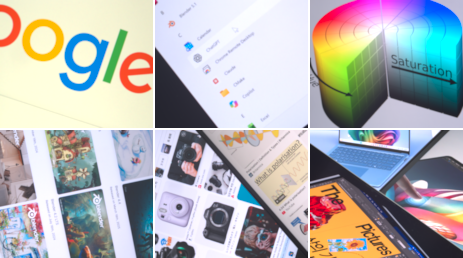} &
    \includegraphics[height=\panelHeight]{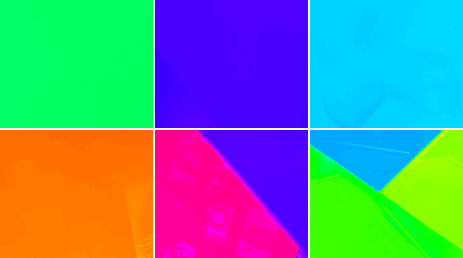} &
    \includegraphics[height=\panelHeight]{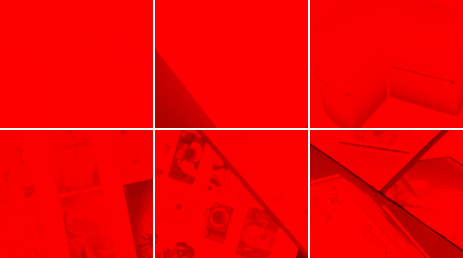} \\
    \multicolumn{1}{@{}c@{\hspace{1.5mm}}}{Scene setup} & $s_0$ & AoLP & DoLP \\
    \multicolumn{4}{c}{(e) Display scene (16 scenes)} \\[2mm]
    \annotatedSetup{polarizer}{
      \node[anchor=north west] at (.05,.975) {Polarizing film strips};
    } &
    \includegraphics[height=\panelHeight]{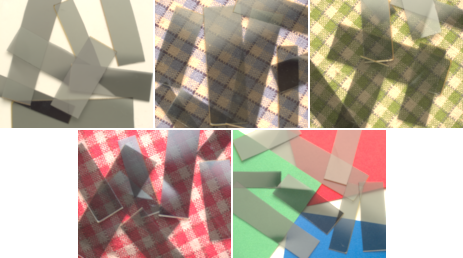} &
    \includegraphics[height=\panelHeight]{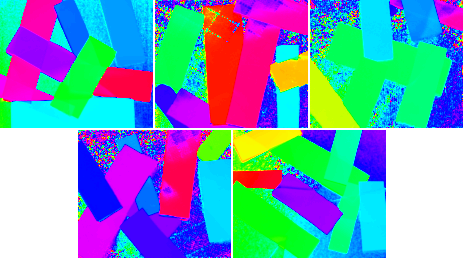} &
    \includegraphics[height=\panelHeight]{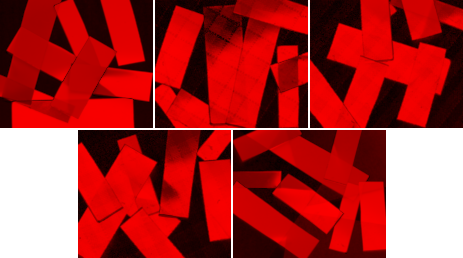} \\
    \multicolumn{1}{@{}c@{\hspace{1.5mm}}}{Scene setup} & $s_0$ &
    AoLP\rlap{\hspace{2mm}{\scriptsize $0$\,\raisebox{0.4ex}{\includegraphics[width=11mm,height=1mm]{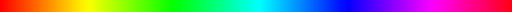}}\,$\pi$}} &
    DoLP\rlap{\hspace{2mm}{\scriptsize $0$\,\raisebox{0.4ex}{\includegraphics[width=11mm,height=1mm]{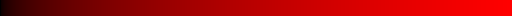}}\,$1$}} \\
    \multicolumn{4}{c}{(f) Polarizer scene (5 scenes)}
  \end{tabular}
  \caption{\textbf{Controlled scene examples.} Each row shows an acquisition setup and the corresponding $s_0$, AoLP, and DoLP images for representative scenes.}
  \label{fig:supp-controlled-scenes}
\end{figure*}

\paragraph{Controlled scene acquisition.}
We capture 226 controlled scenes using six specialized acquisition setups. These setups are designed to capture polarization patterns rarely observed in everyday scenes, including those arising in applications such as active imaging with polarized illumination~\cite{hwang2022sparse,choi2026snapshot} and photoelastic imaging~\cite{dave2025nest,maeda2025event}. \Cref{fig:supp-controlled-scenes} shows the six setups and representative examples.

Setups (a) and (b) use structured illumination from an LCD~\cite{choi2026snapshot} and a projector~\cite{baek2021transport}, respectively, where the LCD provides colored, polarized illumination. In both setups, we display sinusoidal fringes, binary Gray codes, checkerboards, line grids, and dot arrays at varying scales, orientations, and colors. In setup (c), we place a color filter or polarizer in front of an LED source to control the color or polarization of the incident light~\cite{hwang2022sparse}. For photoelastic imaging (d), we illuminate transparent objects with polarized light from a display so that stress-induced birefringence produces color-dependent polarization fringes~\cite{dave2025nest,maeda2025event}. In setup (e), we image the display directly to capture the polarization of its emitted light, and in setup (f), we arrange strips of polarizing film at different orientations to create spatial variations in the polarization angle.

\FloatBarrier
\subsection{HDR color Stokes image construction}
\label{sec:supp-stokes-construction}

\begin{figure}[htbp]
  \centering
  \input{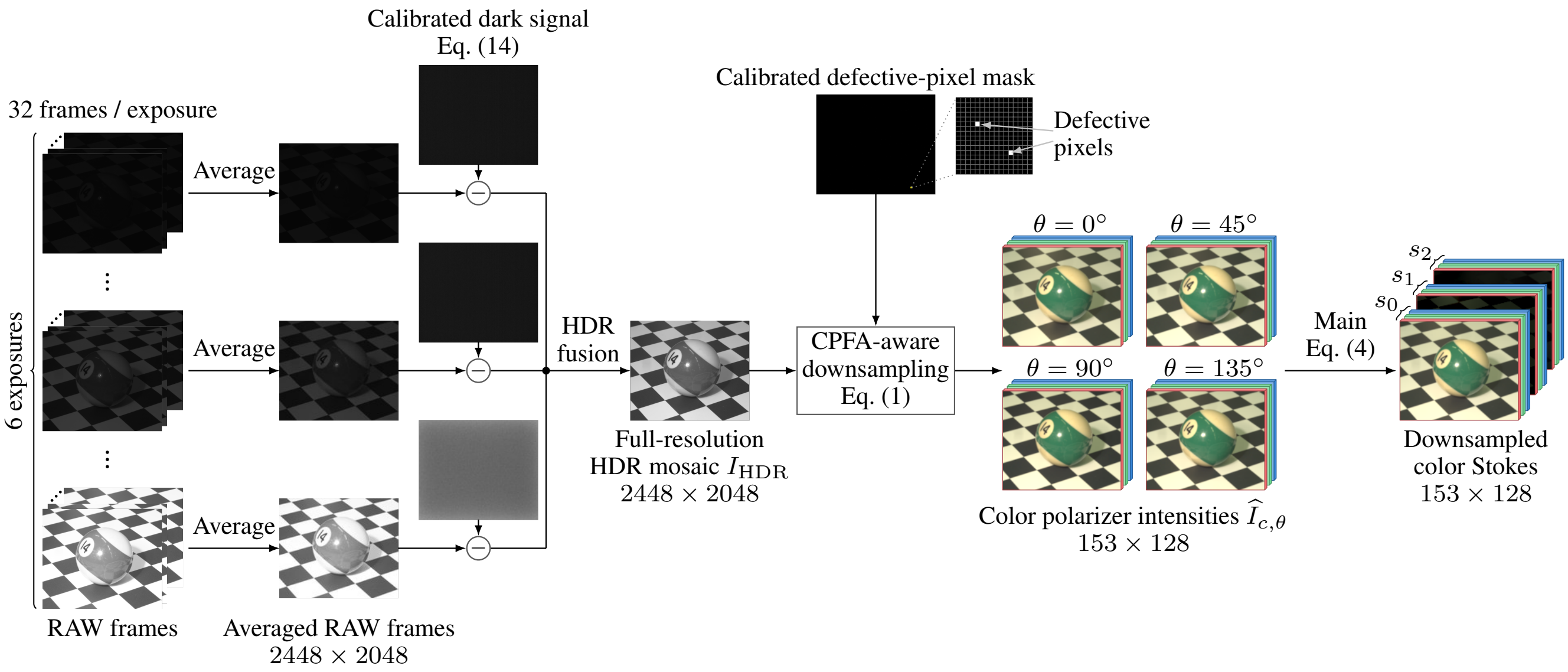}%
  \captionsetup{skip=6pt}%
  \caption{\textbf{HDR color Stokes image construction pipeline.} Repeated RAW frames are averaged at each exposure, dark-subtracted, and fused into a full-resolution HDR mosaic. CPFA-aware downsampling excludes unreliable pixels and yields a color intensity image for each of the four polarizer orientations. These four images are then converted to color Stokes images.}
  \label{fig:supp-dataset-construction}
\end{figure}

For each scene, we construct an HDR color Stokes image from the exposure-bracketed RAW frames acquired in \cref{sec:supp-hdr-polar1k-scenes}. As illustrated in \cref{fig:supp-dataset-construction}, the pipeline consists of two stages: frame averaging and HDR fusion of the bracketed RAW frames into a full-resolution mosaic, followed by CPFA-aware downsampling and conversion to color Stokes vectors. Below, we formulate both stages along with hot-pixel exclusion. The design and selection of the downsampling filter are detailed in \cref{sec:supp-cpfa-cubic}.

\paragraph{Frame averaging and HDR fusion.}
We average repeated RAW frames at each exposure and merge the bracketed exposures using intensity-based weights that favor well-exposed mid-range values~\cite{debevec1997recovering}. Specifically, we take the pixel-wise mean at each exposure time, normalize by the white level, and compute a parabolic weight that peaks at mid-range and drops to zero near the dark and saturated extremes. We then subtract the calibrated dark signal from \cref{sec:supp-sensor-calibration}, divide by the exposure time, and take the weighted average across exposures to obtain the full-resolution HDR mosaic.

\paragraph{Downsampling.}
Instead of demosaicing the full-resolution RAW image first, we compute four polarizer intensities per color channel directly at each output pixel \(\mathbf u\) on a grid downsampled by a factor of \(r\) from the fused HDR mosaic \(I_{\mathrm{HDR}}[\mathbf n]\):
\begin{equation}
\widehat I_{c,\theta}[\mathbf u]
=\sum_{\mathbf n\in\mathcal D_{c,\theta}} w_{c,\theta}[\mathbf n]\,I_{\mathrm{HDR}}[\mathbf n],
\label{eq:supp-raw-to-pgt-intensity}
\end{equation}
where \(\mathcal D_{c,\theta}\) is the set of valid RAW pixels measuring color \(c\) and polarizer orientation \(\theta\), and \(w_{c,\theta}[\mathbf n]\) are the normalized sampling weights (\(\sum_{\mathbf n\in\mathcal D_{c,\theta}}w_{c,\theta}[\mathbf n]=1\)) obtained by evaluating a shared separable spatial kernel centered at \(\mathbf u\). For each green-polarizer channel, contributions from both green sublattices are summed before normalization. Finally, the resulting 12 channel intensities at each output pixel are converted into one linear Stokes vector per color. \Cref{sec:supp-cpfa-cubic} analyzes how we design the downsampling weights \(w_{c,\theta}[\mathbf n]\) and select \(r\) to account for the CPFA layout.

\begin{figure}[htbp]
  \centering
  \input{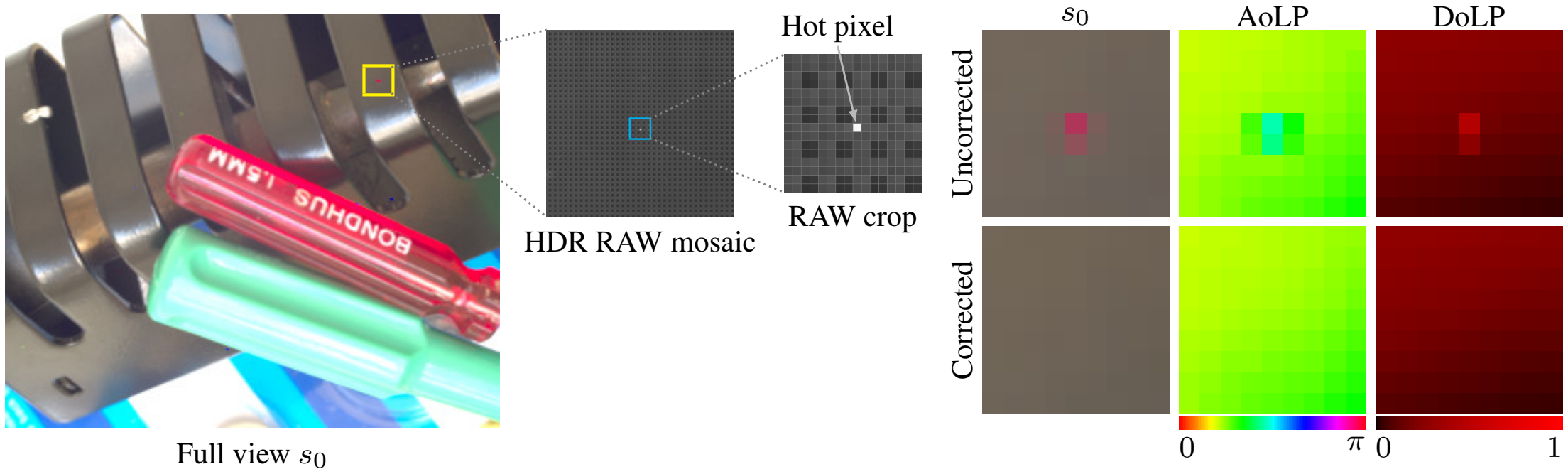}%
  \captionsetup{skip=2pt}%
  \caption{\textbf{Effect of dark subtraction and hot-pixel exclusion on HDR Stokes images.} A persistent hot pixel in the HDR RAW mosaic (top row) contaminates the uncorrected \(s_0\), AoLP, and DoLP outputs after downsampling, whereas dark subtraction and pixel exclusion (bottom row) remove the artifact.}
  \label{fig:supp-reference-correction}
\end{figure}

\paragraph{Hot-pixel exclusion.}
Hot pixels produce persistently bright values due to excessive dark signal. Unlike random noise, they survive temporal averaging~\cite{abdelhamed2018sidd} and cause localized artifacts in the downsampled Stokes images. While per-pixel dark subtraction corrects most of these pixels, some still leave noticeable residuals. To remove them without unnecessarily discarding usable measurements, we construct a defective-pixel mask for each scene from the dark-frame calibration in \cref{sec:supp-sensor-calibration}. Specifically, pixels whose dark signal cannot be reliably calibrated are always masked out, whereas other high-dark-signal pixels are masked only when they remain abnormally bright relative to same-channel neighbors. Using this mask, we remove masked pixels from the valid pixel sets \(\mathcal D_{c,\theta}\) in \cref{eq:supp-raw-to-pgt-intensity}, which sets their weights to zero and automatically renormalizes the remaining weights in each channel. As shown in \cref{fig:supp-reference-correction}, dark subtraction and hot-pixel masking remove artifacts that would otherwise distort \(s_0\), AoLP, and DoLP.

\section{CPFA-aware downsampling}
\label{sec:supp-cpfa-cubic}

Downsampling RAW mosaic images constructs demosaicing-free ground-truth images without learned or heuristic interpolation. For Bayer RGB sensors, Khashabi et al.~\cite{khashabi2014joint} showed that designing the downsampling filter to align first-order sampling moments across color channels prevents spatial phase shifts and color artifacts.

Extending this approach to a CPFA introduces a distinct problem that first-order moment alignment alone does not resolve. A CPFA samples 12 color-polarizer channels at different spatial offsets within each \(4\mathord{\times}4\) tile, as shown in \cref{fig:supp-downsampling-tile}(a), and linear Stokes components depend on intensity differences across polarizer orientations rather than intensities alone. Consequently, any mismatch in the first- or second-order sampling moments across channels converts unpolarized intensity gradients and curvature into spurious polarization.

In this section, we formulate CPFA-aware downsampling through second-order sampling-moment matching across all 12 channels and show that the Mitchell--Netravali cubic filter family satisfies these conditions analytically. We derive the moment conditions and cubic filter in \cref{sec:supp-cpfa-moments,sec:supp-cpfa-cubic-filter}, generalize the analysis via a frequency-domain Stokes transfer matrix in \cref{sec:supp-stokes-transfer}, and present quantitative analysis, parameter selection, and real-image validation in \cref{sec:supp-filter-evaluation,sec:supp-filter-selection,sec:supp-cpfa-real-comparison}.

\subsection{Moment-matching conditions for CPFA downsampling}
\label{sec:supp-cpfa-moments}

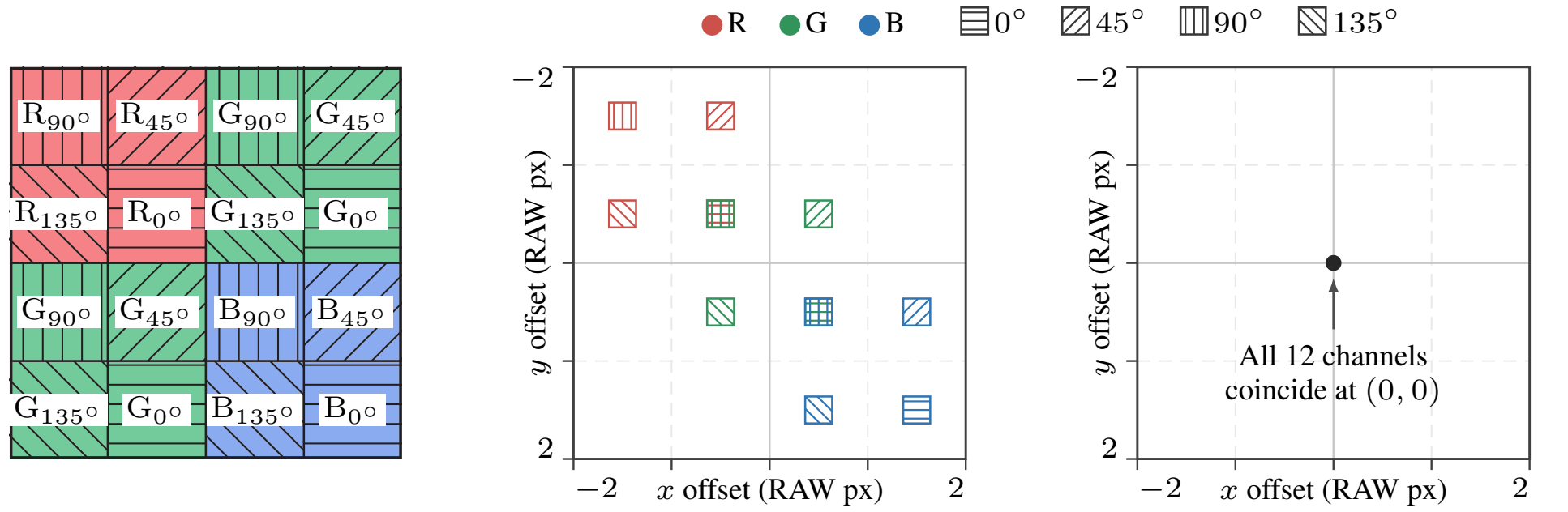
\begin{figure}[!ht]
  \centering
  \begingroup
\definecolor{ctrRed}{HTML}{CC514B}
\definecolor{ctrGreen}{HTML}{32935B}
\definecolor{ctrBlue}{HTML}{3076B4}
\newcommand{\PolMarker}[5]{%
  \begin{scope}[shift={(#1,#2)}]
    \draw[#4,line width=.48pt] ({-#3/2},{-#3/2}) rectangle ({#3/2},{#3/2});
    \begin{scope}
      \clip ({-#3/2},{-#3/2}) rectangle ({#3/2},{#3/2});
      \begin{scope}[rotate=#5]
        \foreach \yoff in {-0.32,0,0.32} {
          \draw[#4,line width=.42pt] ({-#3},{\yoff*#3}) -- ({#3},{\yoff*#3});
        }
      \end{scope}
    \end{scope}
  \end{scope}%
}
\begin{tikzpicture}[x=1mm,y=1mm,font=\fontsize{8}{9}\selectfont,
  cell label/.style={inner xsep=.3mm,inner ysep=.2mm,outer sep=0pt,fill=white}]

  \node[anchor=south,inner sep=0pt,font=\fontsize{7}{8}\selectfont] at (85,34.5) {%
    \tikz[baseline=-0.55ex]{\fill[ctrRed] (0,0) circle[radius=0.85mm];}\,\text{R}\hspace{2.6mm}%
    \tikz[baseline=-0.55ex]{\fill[ctrGreen] (0,0) circle[radius=0.85mm];}\,\text{G}\hspace{2.6mm}%
    \tikz[baseline=-0.55ex]{\fill[ctrBlue] (0,0) circle[radius=0.85mm];}\,\text{B}\hspace{4.5mm}%
    \tikz[x=1mm,y=1mm,baseline=-0.95mm]{\PolMarker{0}{0}{2.2}{black!78}{0}}\,\(0^\circ\)\hspace{2.6mm}%
    \tikz[x=1mm,y=1mm,baseline=-0.95mm]{\PolMarker{0}{0}{2.2}{black!78}{45}}\,\(45^\circ\)\hspace{2.6mm}%
    \tikz[x=1mm,y=1mm,baseline=-0.95mm]{\PolMarker{0}{0}{2.2}{black!78}{90}}\,\(90^\circ\)\hspace{2.6mm}%
    \tikz[x=1mm,y=1mm,baseline=-0.95mm]{\PolMarker{0}{0}{2.2}{black!78}{135}}\,\(135^\circ\)%
  };

  \node[anchor=south west,inner sep=0pt,outer sep=0pt] at (0,0)
    {\CPFASensor[width=32mm,stripe count=5,stripe width=.48bp,
      grid width=.6bp,border width=.8bp,
      red=cpfaOriginalRed!55,green=cpfaOriginalGreen!55,blue=cpfaOriginalBlue!55]};
  \foreach \row/\col/\band/\angle in {
    0/0/R/90, 0/1/R/45, 0/2/G/90, 0/3/G/45,
    1/0/R/135,1/1/R/0,  1/2/G/135,1/3/G/0,
    2/0/G/90, 2/1/G/45, 2/2/B/90, 2/3/B/45,
    3/0/G/135,3/1/G/0,  3/2/B/135,3/3/B/0} {
    \node[cell label,font=\fontsize{7}{8}\selectfont] at ({8*\col+4},{28-8*\row})
      {\(\mathrm{\band}_{\angle^\circ}\)};
  }
  \node[anchor=north,align=center] at (16,-5.8) {(a) \(4\mathord{\times}4\) CPFA tile};

  \begin{scope}[shift={(46,0)}]
    \draw[black!9,line width=.35pt,densely dashed] (0,8) -- (32,8) (0,24) -- (32,24) (8,0) -- (8,32) (24,0) -- (24,32);
    \draw[black!22,line width=.45pt] (0,16) -- (32,16) (16,0) -- (16,32);
    \draw[black!75,line width=.55pt] (0,0) rectangle (32,32);
    \foreach \tx in {0,8,16,24,32} {\draw[black!75,line width=.45pt] (\tx,0) -- (\tx,-0.9);}
    \foreach \ty in {0,8,16,24,32} {\draw[black!75,line width=.45pt] (0,\ty) -- (-0.9,\ty);}
    \node[anchor=north west,inner sep=0pt,font=\fontsize{7}{8}\selectfont] at (0,-1.5) {\(-2\)};
    \node[anchor=north east,inner sep=0pt,font=\fontsize{7}{8}\selectfont] at (32,-1.5) {\(2\)};
    \node[anchor=north,inner sep=0pt,font=\fontsize{7}{8}\selectfont] at (16,-1.5) {\(x\) offset (RAW px)};
    \node[anchor=north east,inner sep=0pt,font=\fontsize{7}{8}\selectfont] at (-1.5,32) {\(-2\)};
    \node[anchor=south east,inner sep=0pt,font=\fontsize{7}{8}\selectfont] at (-1.5,0) {\(2\)};
    \node[rotate=90,anchor=south,inner sep=0pt,font=\fontsize{7}{8}\selectfont] at (-1.7,16) {\(y\) offset (RAW px)};
    \PolMarker{4}{28}{2.2}{ctrRed}{90}
    \PolMarker{12}{28}{2.2}{ctrRed}{45}
    \PolMarker{4}{20}{2.2}{ctrRed}{135}
    \PolMarker{12}{20}{2.2}{ctrRed}{0}
    \PolMarker{12}{20}{2.2}{ctrGreen}{90}
    \PolMarker{20}{20}{2.2}{ctrGreen}{45}
    \PolMarker{12}{12}{2.2}{ctrGreen}{135}
    \PolMarker{20}{12}{2.2}{ctrGreen}{0}
    \PolMarker{20}{12}{2.2}{ctrBlue}{90}
    \PolMarker{28}{12}{2.2}{ctrBlue}{45}
    \PolMarker{20}{4}{2.2}{ctrBlue}{135}
    \PolMarker{28}{4}{2.2}{ctrBlue}{0}
    \node[anchor=north,align=center] at (16,-5.8) {(b) Box-filter sampling centers\\(center error: 2.12 RAW px)};
  \end{scope}

  \begin{scope}[shift={(92,0)}]
    \draw[black!9,line width=.35pt,densely dashed] (0,8) -- (32,8) (0,24) -- (32,24) (8,0) -- (8,32) (24,0) -- (24,32);
    \draw[black!22,line width=.45pt] (0,16) -- (32,16) (16,0) -- (16,32);
    \draw[black!75,line width=.55pt] (0,0) rectangle (32,32);
    \foreach \tx in {0,8,16,24,32} {\draw[black!75,line width=.45pt] (\tx,0) -- (\tx,-0.9);}
    \foreach \ty in {0,8,16,24,32} {\draw[black!75,line width=.45pt] (0,\ty) -- (-0.9,\ty);}
    \node[anchor=north west,inner sep=0pt,font=\fontsize{7}{8}\selectfont] at (0,-1.5) {\(-2\)};
    \node[anchor=north east,inner sep=0pt,font=\fontsize{7}{8}\selectfont] at (32,-1.5) {\(2\)};
    \node[anchor=north,inner sep=0pt,font=\fontsize{7}{8}\selectfont] at (16,-1.5) {\(x\) offset (RAW px)};
    \node[anchor=north east,inner sep=0pt,font=\fontsize{7}{8}\selectfont] at (-1.5,32) {\(-2\)};
    \node[anchor=south east,inner sep=0pt,font=\fontsize{7}{8}\selectfont] at (-1.5,0) {\(2\)};
    \node[rotate=90,anchor=south,inner sep=0pt,font=\fontsize{7}{8}\selectfont] at (-1.7,16) {\(y\) offset (RAW px)};
    \fill[black!85] (16,16) circle[radius=0.65mm];
    \node[align=center,font=\fontsize{7}{8}\selectfont] (coincide) at (16,6.8) {All 12 channels\\coincide at \((0,0)\)};
    \draw[-{Latex[length=1.2mm]},black!70,line width=.5pt] (coincide.north) -- (16,14.8);
    \node[anchor=north,align=center] at (16,-5.8) {(c) Cubic-filter sampling centers\\(center error: 0 RAW px)};
  \end{scope}
\end{tikzpicture}%
\endgroup
  \caption{\textbf{CPFA layout and channel-wise sampling-center alignment.} (a) Sampling positions in one \(4\mathord{\times}4\) CPFA tile. For each polarizer orientation, red and blue occur once and green occurs twice. (b, c) Weighted sampling-center offsets of all 12 channels at downsampling factor \(r=16\), computed from \(w_{c,\theta}[\mathbf n]\). Box averaging (b) scatters the channel centers by up to 2.12 RAW pixels, mirroring the tile layout in (a), whereas any Mitchell--Netravali cubic filter with \(B+2C=1\) (c) aligns all 12 centers at \((0,0)\).}
  \label{fig:supp-downsampling-tile}
\end{figure}

When downsampling averages each channel uniformly over each \(r\mathord{\times}r\) block (Box averaging), the repeating tile offsets in \cref{fig:supp-downsampling-tile}(a) shift each channel's effective sampling center away from the output pixel center in \cref{fig:supp-downsampling-tile}(b). Consequently, on an unpolarized intensity gradient, subtracting two orthogonal polarizer channels whose effective centers differ produces a nonzero difference and therefore spurious polarization.

To analyze this leakage systematically, let \(\mathbf n=(n_y,n_x)\) denote RAW pixel coordinates and let \(\tilde{\mathbf u}=(\tilde u_y,\tilde u_x)\) denote the center of the \(r\mathord{\times}r\) RAW block corresponding to output pixel \(\mathbf u\). Consider an unpolarized one-dimensional intensity profile \(f(\delta)\) varying along the horizontal axis, where \(\delta[\mathbf n]=(n_x-\tilde u_x)/r\) is the horizontal offset from the output pixel center in output-pixel units. The same analysis applies vertically with \(\delta[\mathbf n]=(n_y-\tilde u_y)/r\). Substituting \(I_{\mathrm{HDR}}[\mathbf n]=f(\delta[\mathbf n])\) into \cref{eq:supp-raw-to-pgt-intensity} and expanding \(f\) in a Taylor series around \(\delta=0\) gives
\begin{equation}
\widehat I_{c,\theta}[\mathbf u]=\sum_{\mathbf n\in\mathcal D_{c,\theta}} w_{c,\theta}[\mathbf n]\,f(\delta[\mathbf n])=m_{c,\theta}^{(0)}f(0)+m_{c,\theta}^{(1)}f'(0)+\frac12 m_{c,\theta}^{(2)}f''(0)+\cdots,
\label{eq:supp-taylor-intensity}
\end{equation}
where the \(k\)-th sampling moment of channel \((c,\theta)\) is defined as
\begin{equation}
m_{c,\theta}^{(k)}=\sum_{\mathbf n\in\mathcal D_{c,\theta}} w_{c,\theta}[\mathbf n]\,\delta[\mathbf n]^{\,k}.
\label{eq:supp-sampling-moment}
\end{equation}

Each linear polarization component is an intensity difference between two orthogonal polarizer channels of the same color. Since both channels have unit total weight, subtracting them cancels the constant term \(f(0)\) and leaves
\begin{equation}
\begin{aligned}
\widehat s_{1,c}[\mathbf u]&=\widehat I_{c,0^\circ}[\mathbf u]-\widehat I_{c,90^\circ}[\mathbf u]\\
&=\bigl(m_{c,0^\circ}^{(1)}-m_{c,90^\circ}^{(1)}\bigr)f'(0)+\frac12\bigl(m_{c,0^\circ}^{(2)}-m_{c,90^\circ}^{(2)}\bigr)f''(0)+\cdots,
\end{aligned}
\label{eq:supp-taylor-difference}
\end{equation}
and analogously for \(\widehat s_{2,c}[\mathbf u]=\widehat I_{c,45^\circ}[\mathbf u]-\widehat I_{c,135^\circ}[\mathbf u]\). Thus, any first-moment mismatch between orthogonal channels leaks the first-order intensity derivative \(f'(0)\) into \(s_1\), while any second-moment mismatch leaks the second-order derivative \(f''(0)\).

To eliminate this low-order leakage across all 12 channels, every channel must satisfy the same moment conditions through second order along both axes:
\begin{equation}
m_{c,\theta}^{(0)}=1,\qquad m_{c,\theta}^{(1)}=0,\qquad m_{c,\theta}^{(2)}=\bar m_2,
\label{eq:supp-moment-conditions}
\end{equation}
where the common second moment \(\bar m_2\) is independent of \(c\) and \(\theta\). Under \cref{eq:supp-moment-conditions}, both the \(f'(0)\) and \(f''(0)\) leakage terms in \cref{eq:supp-taylor-difference} vanish: \(m_{c,\theta}^{(0)}=1\) preserves constant intensity, \(m_{c,\theta}^{(1)}=0\) aligns all sampling centers with the output pixel center in \cref{fig:supp-downsampling-tile}(c), and a channel-independent \(m_{c,\theta}^{(2)}=\bar m_2\) cancels the curvature term. Matching \cref{eq:supp-moment-conditions} across colors \(c\) as well as polarizer orientations \(\theta\) also prevents inter-color spatial shifts and chromatic blur mismatches.

With a separable kernel on the regular CPFA lattice, satisfying \cref{eq:supp-moment-conditions} along each axis automatically aligns all 2D moments through second order across all 12 channels and both green sublattices. The mixed cross-moment also vanishes since it factorizes into the product of the zero first moments along the two axes. For polarized scenes, \cref{eq:supp-moment-conditions} further ensures that \(s_0\), \(s_1\), and \(s_2\) undergo identical low-order spatial filtering without cross-component leakage or relative spatial shift, as formalized in the frequency domain in \cref{sec:supp-stokes-transfer}.

\subsection{Cubic filter satisfying the moment conditions}
\label{sec:supp-cpfa-cubic-filter}

The Mitchell--Netravali cubic filter family~\cite{mitchell1988reconstruction} satisfies \cref{eq:supp-moment-conditions} analytically whenever the downsampling factor \(r\) is a positive integer multiple of four. We construct the sampling weights \(w_{c,\theta}[\mathbf n]\) from a separable product of a piecewise cubic one-dimensional kernel \(\phi_{B,C}(\delta)\) with shape parameters \((B,C)\) and compact support \([-2,2]\) in output-pixel units:
\begin{equation}
\begin{aligned}
w_{c,\theta}[\mathbf n]&\propto\phi_{B,C}\!\left(\frac{n_y-\tilde u_y}{r}\right)\phi_{B,C}\!\left(\frac{n_x-\tilde u_x}{r}\right),\qquad \mathbf n\in\mathcal D_{c,\theta},\\
\phi_{B,C}(\delta)&=\frac16
\begin{cases}
(12-9B-6C)|\delta|^3+(-18+12B+6C)|\delta|^2+(6-2B),&0\leq|\delta|<1,\\
(-B-6C)|\delta|^3+(6B+30C)|\delta|^2+(-12B-48C)|\delta|+(8B+24C),&1\leq|\delta|<2,\\
0,&\text{otherwise},
\end{cases}
\end{aligned}
\label{eq:supp-mn-kernel}
\end{equation}
normalized to unit sum as in \cref{eq:supp-raw-to-pgt-intensity}. This separable construction yields a nominal \(4r\mathord{\times}4r\) RAW support.

For interior pixels without missing samples on the regular CPFA lattice with \(B+2C=1\) and any downsampling factor \(r\in\{4,8,12,\ldots\}\), every channel satisfies \cref{eq:supp-moment-conditions} with \(\bar m_2=B/3\):
\begin{equation}
m_{c,\theta}^{(0)}=1,\qquad m_{c,\theta}^{(1)}=0,\qquad m_{c,\theta}^{(2)}=\frac{B}{3}.
\label{eq:supp-cubic-moments}
\end{equation}
Each CPFA sublattice repeats every 4 RAW pixels along each axis, so scaling by \(1/r\) splits its samples into \(r/4\) interleaved lattices, each spaced by one output pixel (\(r\) RAW pixels). On every shifted unit lattice, the \(B+2C=1\) kernel has identical unnormalized weight sums regardless of tile position and satisfies \cref{eq:supp-cubic-moments}, so summing over these lattices and normalizing preserves \cref{eq:supp-cubic-moments} across all 12 channels and both green sublattices. The residual third-order moment scales with \(1-B\), decreases with \(r\), and vanishes at \(B=1\).

\subsection{Frequency-domain Stokes transfer matrix}
\label{sec:supp-stokes-transfer}

While \cref{sec:supp-cpfa-moments} analyzes unpolarized inputs to motivate the moment-matching conditions, evaluating candidate filters across arbitrary polarized scenes and all spatial frequencies requires a general linear model. In the frequency domain, for a single input frequency \(\boldsymbol\omega\) in each color channel \(c\), downsampling maps the input Stokes spectrum \(\mathbf s_c(\boldsymbol\omega)=[s_{0,c}(\boldsymbol\omega),s_{1,c}(\boldsymbol\omega),s_{2,c}(\boldsymbol\omega)]^\top\) to the output Stokes spectrum \(\widehat{\mathbf s}_c(\boldsymbol\omega)\) through a Stokes transfer matrix \(\mathbf T_c(\boldsymbol\omega)\in\mathbb C^{3\mathord{\times}3}\),
\begin{equation}
\widehat{\mathbf s}_c(\boldsymbol\omega)=\mathbf T_c(\boldsymbol\omega)\mathbf s_c(\boldsymbol\omega),
\label{eq:supp-stokes-io}
\end{equation}
where \(\boldsymbol\omega\in\mathbb R^2\) is the angular spatial frequency in radians per output pixel. Ideally, \(\mathbf T_c(\boldsymbol\omega)\) should be a diagonal matrix whose diagonal entries are identical across all three Stokes components and all three color channels, so that every component undergoes the same spatial filtering without cross-component leakage.

To express \(\mathbf T_c(\boldsymbol\omega)\) in terms of the actual downsampling weights \(w_{c,\theta}[\mathbf n]\) in \cref{eq:supp-raw-to-pgt-intensity}, we first take the complex frequency response of each channel's weights relative to the output pixel center \(\tilde{\mathbf u}\),
\begin{equation}
W_{c,\theta}(\boldsymbol\omega)=\sum_{\mathbf n\in\mathcal D_{c,\theta}}w_{c,\theta}[\mathbf n]\,e^{\mathrm{i}\boldsymbol\omega^\top(\mathbf n-\tilde{\mathbf u})/r}.
\label{eq:supp-color-stokes-transfer}
\end{equation}
Stacking Eq.~(3) of the main paper for \(\theta\in\{0^\circ,45^\circ,90^\circ,135^\circ\}\) defines the polarizer-response matrix \(\mathbf A\in\mathbb R^{4\mathord{\times}3}\) with rows \(0.5[1,\cos 2\theta,\sin 2\theta]\). Passing \(\mathbf s_c(\boldsymbol\omega)\) through \(\mathbf A\), per-channel filtering with \(W_{c,\theta}(\boldsymbol\omega)\), and the pseudoinverse \(\mathbf A^\dagger\in\mathbb R^{3\mathord{\times}4}\) of \(\mathbf A\) gives
\begin{equation}
\begin{aligned}
\mathbf T_c(\boldsymbol\omega)&=\mathbf A^\dagger\operatorname{diag}\!\left(W_{c,0^\circ},W_{c,45^\circ},W_{c,90^\circ},W_{c,135^\circ}\right)\mathbf A\\
&=\frac12\begin{bmatrix}
\tfrac12\sum_\theta W_{c,\theta} & \tfrac12(W_{c,0^\circ}-W_{c,90^\circ}) & \tfrac12(W_{c,45^\circ}-W_{c,135^\circ}) \\
W_{c,0^\circ}-W_{c,90^\circ} & W_{c,0^\circ}+W_{c,90^\circ} & 0 \\
W_{c,45^\circ}-W_{c,135^\circ} & 0 & W_{c,45^\circ}+W_{c,135^\circ}
\end{bmatrix}.
\end{aligned}
\label{eq:supp-color-stokes-transfer-explicit}
\end{equation}
Thus, \cref{eq:supp-color-stokes-transfer-explicit} reduces to this ideal color-independent diagonal form only when all 12 channel responses \(W_{c,\theta}(\boldsymbol\omega)\) coincide. Otherwise, three artifacts arise: nonzero off-diagonal entries leak intensity into polarization, unequal diagonal entries distort the relative gain among Stokes components (biasing DoLP and AoLP), and differences across colors introduce chromatic shifts.

To see how the moment conditions in \cref{sec:supp-cpfa-moments} align these channel responses at low frequencies, we expand the complex exponential in \cref{eq:supp-color-stokes-transfer} around \(\boldsymbol\omega=\mathbf 0\). Under \cref{eq:supp-moment-conditions}, all 12 channels share the same quadratic frequency response,
\begin{equation}
W_{c,\theta}(\boldsymbol\omega)=1-\frac{\bar m_2}{2}\|\boldsymbol\omega\|_2^2+O(\|\boldsymbol\omega\|_2^3),
\label{eq:supp-freq-channel-expansion}
\end{equation}
where \(O(\|\boldsymbol\omega\|_2^3)\) denotes third- and higher-order terms in \(\boldsymbol\omega\), so \cref{eq:supp-color-stokes-transfer-explicit} reduces to
\begin{equation}
\mathbf T_c(\boldsymbol\omega)=\left(1-\frac{\bar m_2}{2}\|\boldsymbol\omega\|_2^2\right)\mathbf I_3+O(\|\boldsymbol\omega\|_2^3),
\label{eq:supp-freq-moment-expansion}
\end{equation}
where \(\mathbf I_3\) is the \(3\mathord{\times}3\) identity matrix. Thus, matching moments through second order achieves this ideal color-independent diagonal form through quadratic frequency terms, eliminating cross-component leakage, diagonal gain imbalance, and inter-color phase shift up to third- and higher-order residuals.

\subsection{Quantitative filter analysis}
\label{sec:supp-filter-evaluation}

\begin{figure}[!ht]
  \centering
  \input{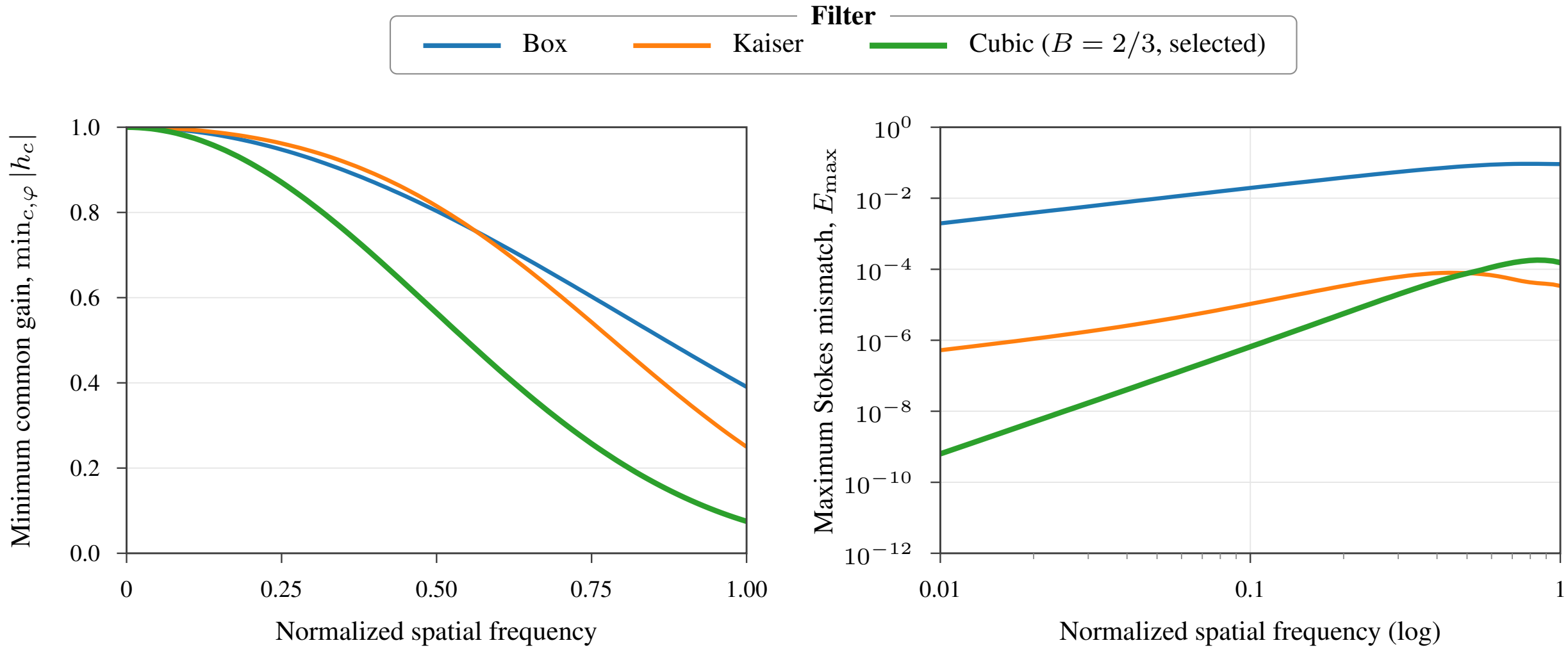}
  \caption{\textbf{Spatial response and Stokes consistency at \(r=16\).} Responses compare Box, Kaiser, and the selected cubic filter with \(B=2/3\) over RGB and spatial directions \(\varphi\), with spatial frequency normalized by the directional output Nyquist frequency. (a) Minimum common gain \(\min_{c,\varphi}|h_c|\), where higher values indicate sharper spatial response and stronger attenuation near the Nyquist limit helps suppress aliasing. (b) Maximum Stokes mismatch \(E_{\max}=\max_{c,\varphi}E_c\) on a log--log scale, where lower values indicate more consistent filtering across Stokes components.}
  \label{fig:supp-downsampling-phase-consistency}
\end{figure}

\paragraph{Evaluation metrics.}
Using the Stokes transfer matrix \(\mathbf T_c(\boldsymbol\omega)\) from \cref{sec:supp-stokes-transfer} and the spatial weights \(w_{c,\theta}[\mathbf n]\), we compare filters along four criteria:
\begin{itemize}[leftmargin=*,nosep]
  \item \textbf{Moment alignment:} in \cref{tab:supp-downsampling-phase-consistency}, we measure center error as the largest weighted-center displacement across all 12 channels in RAW pixels, as illustrated in \cref{fig:supp-downsampling-tile}(b,c), and second-moment spread as the largest range of second moments across channels in squared output pixels. Both vanish when \cref{eq:supp-moment-conditions} holds.
  \item \textbf{Intensity-to-polarization leakage:} in \cref{tab:supp-downsampling-phase-consistency,tab:supp-downsampling-additional-validation}, we measure the spurious polarization amplitude transferred from unpolarized intensity by the Euclidean norm of the two off-diagonal intensity-to-polarization entries in the first column of \(\mathbf T_c(\boldsymbol\omega)\). Baseband leakage is its worst-case value over the output frequency band, capturing leakage from in-band patterns. Folded leakage is the worst-case root-sum-square over aliased CPFA sublattice replicas in the output band, capturing higher-frequency patterns folded in by downsampling.
  \item \textbf{Spatial bandwidth and edge overshoot:} in \cref{tab:supp-downsampling-phase-consistency,tab:supp-downsampling-additional-validation}, we measure worst-case spatial sharpness across all Stokes components via Stokes MTF50, defined as the lowest normalized spatial frequency at which the smallest singular value of \(\mathbf T_c(\boldsymbol\omega)\) drops to half its zero-frequency value. We also report edge overshoot as the largest ringing deviation of \(s_0\) beyond the unit step range on an unpolarized step edge.
  \item \textbf{Frequency response and Stokes consistency:} for \cref{fig:supp-downsampling-phase-consistency}, to measure how closely \(\mathbf T_c(\boldsymbol\omega)\) matches the ideal diagonal form across spatial frequencies, we define the color-wise mean common gain \(h_c(\boldsymbol\omega)=\tfrac14\sum_\theta W_{c,\theta}(\boldsymbol\omega)\) and the spectral-norm mismatch \(E_c(\boldsymbol\omega)=\|\mathbf T_c(\boldsymbol\omega)-h_c(\boldsymbol\omega)\mathbf I_3\|_2\), which captures both off-diagonal leakage and diagonal gain bias. Across color channels \(c\) and spatial directions \(\varphi\) as a function of spatial frequency normalized by the directional output Nyquist frequency, we plot the minimum common gain \(\min_{c,\varphi}|h_c|\) and the maximum Stokes mismatch \(E_{\max}=\max_{c,\varphi}E_c\).
\end{itemize}

\paragraph{Candidate filters.}
To examine how standard resampling kernels behave on a CPFA lattice, we compare three representative filter families at \(r=16\): Box averaging with \(r\mathord{\times}r\) support, a Kaiser-windowed sinc filter~\cite{kaiser1980i0sinh} with output-Nyquist cutoff, shape parameter 6, and \(4r\mathord{\times}4r\) support, and the Mitchell--Netravali cubic family with \(4r\mathord{\times}4r\) support along \(C=(1-B)/2\) at \(B=1/3\), \(B=2/3\), and \(B=1\).

\paragraph{Filter analysis.}
As shown in \cref{tab:supp-downsampling-phase-consistency,fig:supp-downsampling-tile}(b,c), only the cubic family satisfies \cref{eq:supp-moment-conditions} exactly. Box averaging leaves large center displacements and unequal second moments, causing severe baseband and folded leakage, whereas all three cubic settings eliminate low-order moment errors and reduce both leakage metrics by roughly three orders of magnitude. Although Kaiser achieves slightly lower worst-case leakage in \cref{tab:supp-downsampling-phase-consistency}, it introduces much higher edge overshoot. Within the cubic family, increasing \(B\) monotonically lowers leakage and edge overshoot while narrowing the Stokes MTF50 bandwidth.

\begin{table}[htbp]
  \centering
  \footnotesize
  \setlength{\tabcolsep}{3.5pt}
  \renewcommand{\arraystretch}{1.18}
  \caption{\textbf{Filter comparison at downsampling factor \(r=16\).} Support gives the filter footprint in RAW pixels, and metric definitions follow \cref{sec:supp-filter-evaluation}. For cubic filters, \(C=(1-B)/2\) and \(\bar m_2=B/3\).}
  \label{tab:supp-downsampling-phase-consistency}
  \begin{tabular}{@{}lcrrrrrr@{}}
    \toprule
    & & \multicolumn{2}{c}{Moment alignment} & \multicolumn{2}{c}{Leakage} & \multicolumn{2}{c}{Bandwidth \& overshoot} \\
    \cmidrule(lr){3-4}\cmidrule(lr){5-6}\cmidrule(lr){7-8}
    Filter & Support & \makecell{Center error $\downarrow$\\(RAW px)} & \makecell{2nd-mom. spread $\downarrow$\\(output px$^2$)} & \makecell{Baseband $\downarrow$\\(\%)} & \makecell{Folded $\downarrow$\\(\%)} & \makecell{Stokes MTF50 $\uparrow$\\(/ Nyq.)} & \makecell{Overshoot $\downarrow$\\(\%)} \\
    \midrule
    Box & \(r\times r\) & 2.12 & 0.014 & 9.3 & 27 & 0.85 & 0 \\
    Kaiser & \(4r\times 4r\) & 0.0013 & 0.0002 & 0.0080 & 0.016 & 0.78 & 2.7 \\
    \midrule
    Cubic (\(B{=}1/3\)) & \(4r\times 4r\) & 0 & 0 & 0.028 & 0.11 & 0.67 & 1.8 \\
    \rowcolor[gray]{0.94}
    Cubic (\(B{=}2/3\), selected) & \(4r\times 4r\) & 0 & 0 & 0.018 & 0.066 & 0.55 & 0.31 \\
    Cubic (\(B{=}1\)) & \(4r\times 4r\) & 0 & 0 & 0.0092 & 0.040 & 0.45 & 0 \\
    \bottomrule
  \end{tabular}
\end{table}

\Cref{fig:supp-downsampling-phase-consistency} compares the frequency-domain behavior of Box, Kaiser, and the \(B=2/3\) cubic filter. In \cref{fig:supp-downsampling-phase-consistency}(a), the cubic filter attenuates mid- and high-frequency content more strongly than Box and Kaiser, including stronger roll-off near the Nyquist limit that suppresses high-frequency aliasing. In the log--log mismatch plot of \cref{fig:supp-downsampling-phase-consistency}(b), Box and Kaiser decrease only linearly toward zero frequency, whereas the cubic filter drops cubically because it satisfies \cref{eq:supp-moment-conditions}. Consequently, it stays several orders of magnitude below Box across the entire band and well below Kaiser at low frequencies where natural image energy is concentrated, even though Kaiser dips slightly below it near the Nyquist limit.

\FloatBarrier

\Cref{tab:supp-downsampling-additional-validation} examines sensitivity to the downsampling factor \(r\in\{8,16,32\}\) for the \(B=2/3\) cubic filter. Across all three factors, the filter satisfies \cref{eq:supp-moment-conditions} exactly and maintains nearly constant normalized bandwidth and edge overshoot. Increasing \(r\) from 8 to 16 substantially lowers both baseband and folded leakage, while \(r=32\) reduces leakage further at the cost of much lower output resolution.

\begin{table}[!htbp]
  \centering
  \small
  \setlength{\tabcolsep}{5pt}
  \renewcommand{\arraystretch}{1.18}
  \caption{\textbf{Sensitivity of the \(B=2/3\) cubic filter to downsampling factor.} Output sizes correspond to \(2448\mathord{\times}2048\) RAW captures using \((B,C)=(2/3,1/6)\) with \(4r\mathord{\times}4r\) RAW support. All three factors satisfy \cref{eq:supp-moment-conditions} exactly, with zero center error and zero second-moment spread. Metric definitions match \cref{tab:supp-downsampling-phase-consistency}.}
  \label{tab:supp-downsampling-additional-validation}
  \begin{tabular}{@{}rcrrrr@{}}
    \toprule
    & & \multicolumn{2}{c}{Leakage} & \multicolumn{2}{c}{Bandwidth \& overshoot} \\
    \cmidrule(lr){3-4}\cmidrule(lr){5-6}
    \(r\) & \makecell{Output size\\(pixels)} & \makecell{Baseband $\downarrow$\\(\%)} & \makecell{Folded $\downarrow$\\(\%)} & \makecell{Stokes MTF50 $\uparrow$\\(/ Nyq.)} & \makecell{Overshoot $\downarrow$\\(\%)} \\
    \midrule
    8 & \(306\times256\) & 0.39 & 0.38 & 0.55 & 0.41 \\
    \rowcolor[gray]{0.94}
    16 & \(153\times128\) & 0.018 & 0.066 & 0.55 & 0.31 \\
    32 & \(76\times64\) & 0.0011 & 0.012 & 0.55 & 0.30 \\
    \bottomrule
  \end{tabular}
\end{table}

\FloatBarrier

\subsection{Filter and downsampling factor selection}
\label{sec:supp-filter-selection}

Filter selection is inherently a trade-off between polarization fidelity and spatial sharpness. For learning CPFA demosaicing and denoising, we prioritize accurate polarization supervision over maximum sharpness and accept moderate spatial softening rather than allowing spurious polarization, color fringes, or ringing-induced DoLP distortion into the ground-truth targets. On this basis, we choose the Mitchell--Netravali cubic family over Box, which causes severe leakage, and over Kaiser, which leaves nonzero moment residuals and pronounced edge overshoot despite its higher spatial gain in \cref{fig:supp-downsampling-phase-consistency}(a). Within the cubic family, we select \(B=2/3\) as a middle ground: as shown in \cref{tab:supp-downsampling-phase-consistency}, it keeps edge overshoot nearly negligible while avoiding both the heavier blur at \(B=1\) and the stronger overshoot at \(B=1/3\).

For the downsampling factor, we select \(r=16\) to obtain low leakage while retaining enough spatial extent for \(64\mathord{\times}64\) training crops, whereas \(r=32\) in \cref{tab:supp-downsampling-additional-validation} leaves barely enough height for a single vertical crop placement. More generally, the choice of \((B,r)\) scales with sensor resolution. A higher-resolution sensor allows a larger \(r\), which lowers leakage on its own and leaves room for a smaller \(B\) to gain sharpness without excessive overshoot.

\subsection{Real-image validation}
\label{sec:supp-cpfa-real-comparison}

\begin{figure}[!ht]
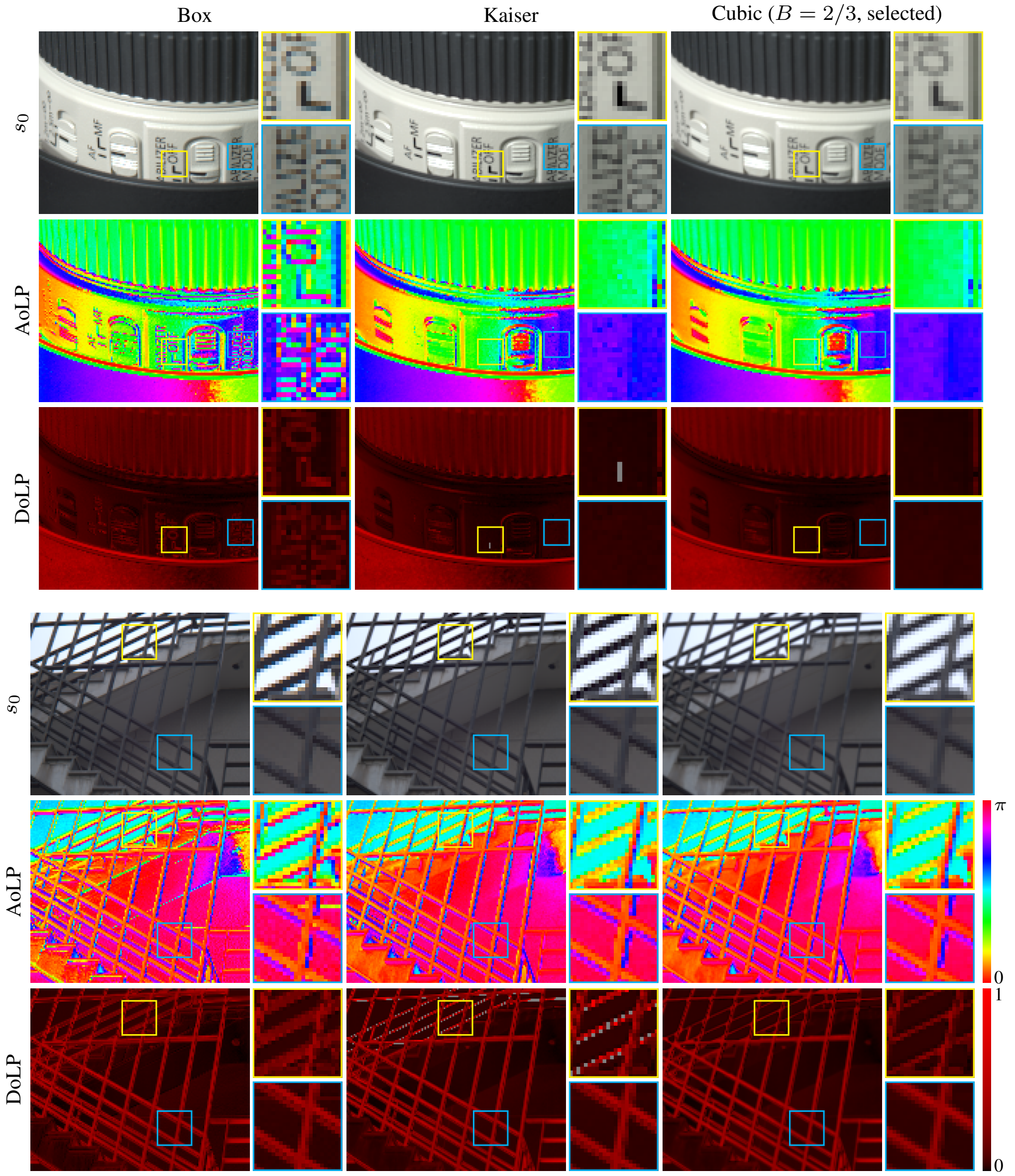

  \centering
  \newlength{\suppFilterUnit}
  \setlength{\suppFilterUnit}{.75mm}
  \input{figures/supp-filter-lens-grid}\par
  \par\vspace{3mm}\nointerlineskip
  \input{figures/downsampling-staircase-grid}\par
  \caption{\textbf{Real-scene filter comparison at \(r=16\).} The Box filter produces color fringes in \(s_0\) and distorts AoLP near intensity edges. The Kaiser filter substantially reduces these artifacts but produces strong edge overshoot, with DoLP locally exceeding \(1.0\), shown in gray in the DoLP panels. The selected cubic filter suppresses both the fringing and AoLP distortion seen with Box and the pronounced edge overshoot seen with Kaiser.}
  \label{fig:supp-downsampling-comparison}
\end{figure}

Real captures confirm the same trade-offs observed in the frequency-domain analysis, as shown in \cref{fig:supp-downsampling-comparison}. We apply each filter at \(r=16\) to the same dark-subtracted, hot-pixel-masked HDR RAW captures on two scenes: a lens scene in the top half and a backlit staircase in the bottom half. In the lens scene, Box introduces color fringes and spurious AoLP and DoLP shifts near intensity boundaries, whereas both Kaiser and the selected cubic filter suppress them. In the staircase crops, Box again produces chromatic artifacts and local AoLP variation, while Kaiser produces ringing that elevates DoLP along the diagonal railings, consistent with its high edge overshoot in \cref{tab:supp-downsampling-phase-consistency}, as negative filter lobes can push individual polarizer channels below zero near sharp edges and drive DoLP above 1. Across both scenes, the selected cubic filter delivers the cleanest color-polarization reconstruction.

\FloatBarrier

\clearpage
\section{Dataset comparison and image quality analysis}
\label{sec:supp-dataset-comparison}

This section details the image quality evaluations summarized in Table~1 of the main paper, along with additional qualitative comparisons and coverage analyses. Following the structure of that table, \cref{sec:supp-dataset-consistency} examines acquisition consistency, \cref{sec:supp-dataset-saturation} analyzes saturation and nonphysical Stokes vectors, and \cref{sec:supp-dataset-scene-coverage} compares polarization distributions and patch diversity.

\subsection{Acquisition consistency}
\label{sec:supp-dataset-consistency}

Existing color-polarization datasets commonly use division-of-time (DoT) acquisition, placing a rotating linear polarizer in front of an RGB camera to capture multiple images at different angles. Although this setup samples the same scene location at every orientation in principle, rotating an external polarizer can introduce spatial and color inconsistencies across captures that corrupt the estimated polarization. In contrast, HDR-Polar1K records all color-polarizer channels simultaneously within each RAW frame using a fixed CPFA camera without moving optical components, resolving the inter-channel spatial offset through the moment-matched downsampling filter in \cref{sec:supp-cpfa-cubic}.

\begin{figure}[!htbp]
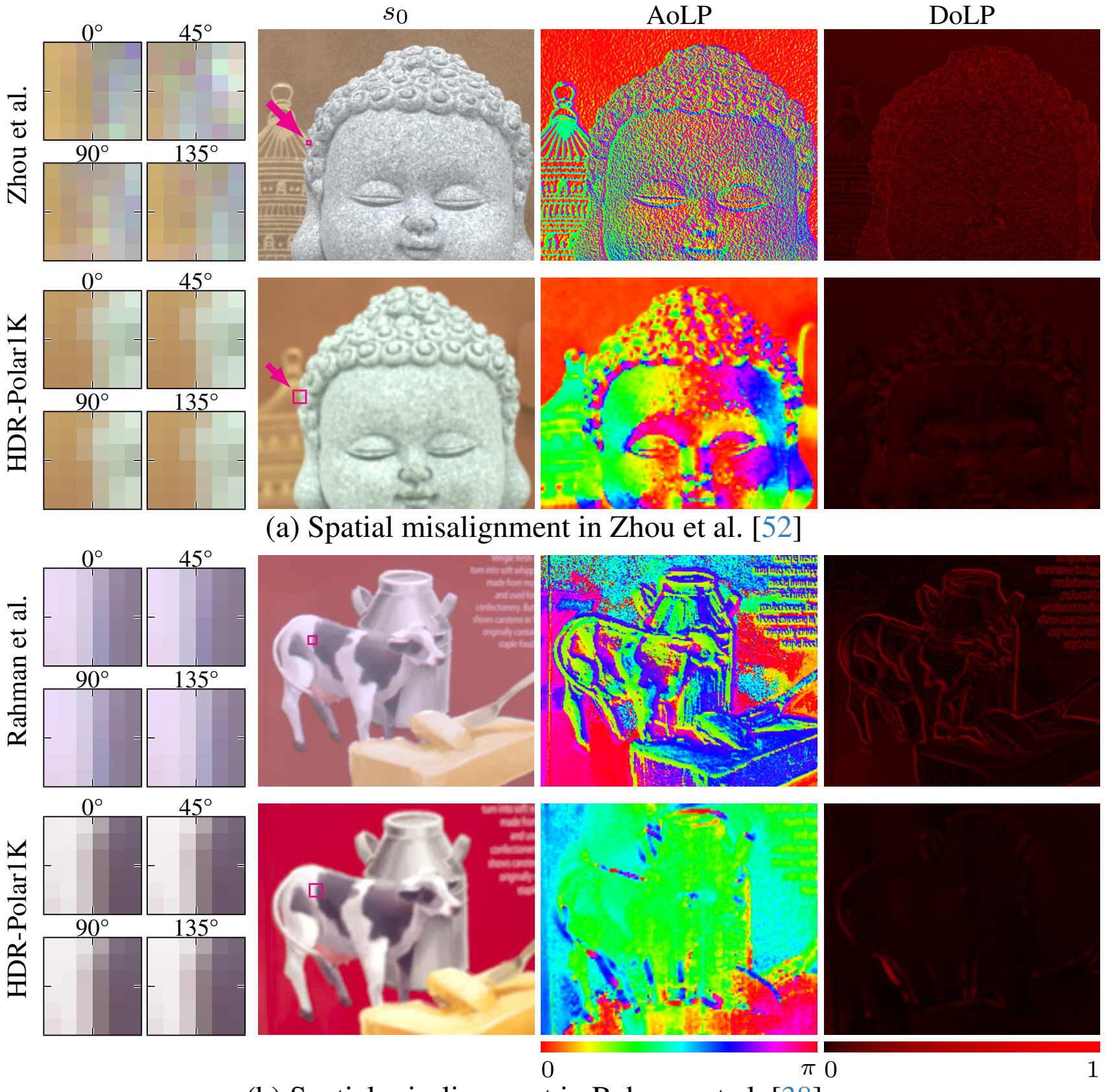

  \centering
  {\PaperSRGBRasters\input{figures/dataset-shift-grid}}\par\vspace{.5mm}
  {\input{figures/dataset-rahman-shift-grid}}
  \caption{\textbf{Spatial misalignment in existing datasets.} Enlarged regions from Zhou et al.~\cite{zhou2025pidsr} (a) and Rahman et al.~\cite{rahman2025polarization} (b) show boundary shifts across polarizer orientations, producing spurious polarization along edges, whereas HDR-Polar1K remains aligned. Panel (b) compares captures of the same type of confectionery tin.}
  \label{fig:supp-dataset-shift-example}
\end{figure}

\paragraph{Spatial misalignment.}
Rotating an external polarizer can displace the image across orientations due to optical refraction or slight mounting tilt, turning intensity gradients near texture boundaries into spurious polarization~\cite{bigue2023production,dumoulin2024impact}. Inspecting enlarged edge crops across the four polarizer orientations for all eight existing datasets reveals visible boundary shifts in seven of them, with only Qiu et al.~\cite{qiu2021linear} showing no noticeable misalignment. \Cref{fig:supp-dataset-shift-example} compares examples from Zhou et al.~\cite{zhou2025pidsr} and Rahman et al.~\cite{rahman2025polarization} with HDR-Polar1K. In both examples, these shifts produce visible AoLP and DoLP artifacts along edges, whereas HDR-Polar1K remains aligned.

\newcommand{\datasetLabelBox}[2][c]{\makebox(0,0)[#1]{#2}}
\newcommand{\datasetIssueHeadings}{%
  \put(57,77.5){\datasetLabelBox{\(s_0\)}}%
  \put(98,77.5){\datasetLabelBox{AoLP}}%
  \put(139,77.5){\datasetLabelBox{DoLP}}}
\newcommand{\datasetIssueBars}{%
  \put(78,3.5){\includegraphics[width=40\unitlength,height=1.5\unitlength]{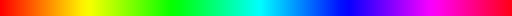}}%
  \put(119,3.5){\includegraphics[width=40\unitlength,height=1.5\unitlength]{figures/hdr-polar1k-dolp-gamma-bar.png}}%
  \put(78,1){\datasetLabelBox[l]{\fontsize{7}{8}\selectfont\(0\)}}%
  \put(118,1){\datasetLabelBox[r]{\fontsize{7}{8}\selectfont\(\pi\)}}%
  \put(119,1){\datasetLabelBox[l]{\fontsize{7}{8}\selectfont\(0\)}}%
  \put(159,1){\datasetLabelBox[r]{\fontsize{7}{8}\selectfont\(1\)}}}

\begin{figure}[!ht]
  \centering
  {\input{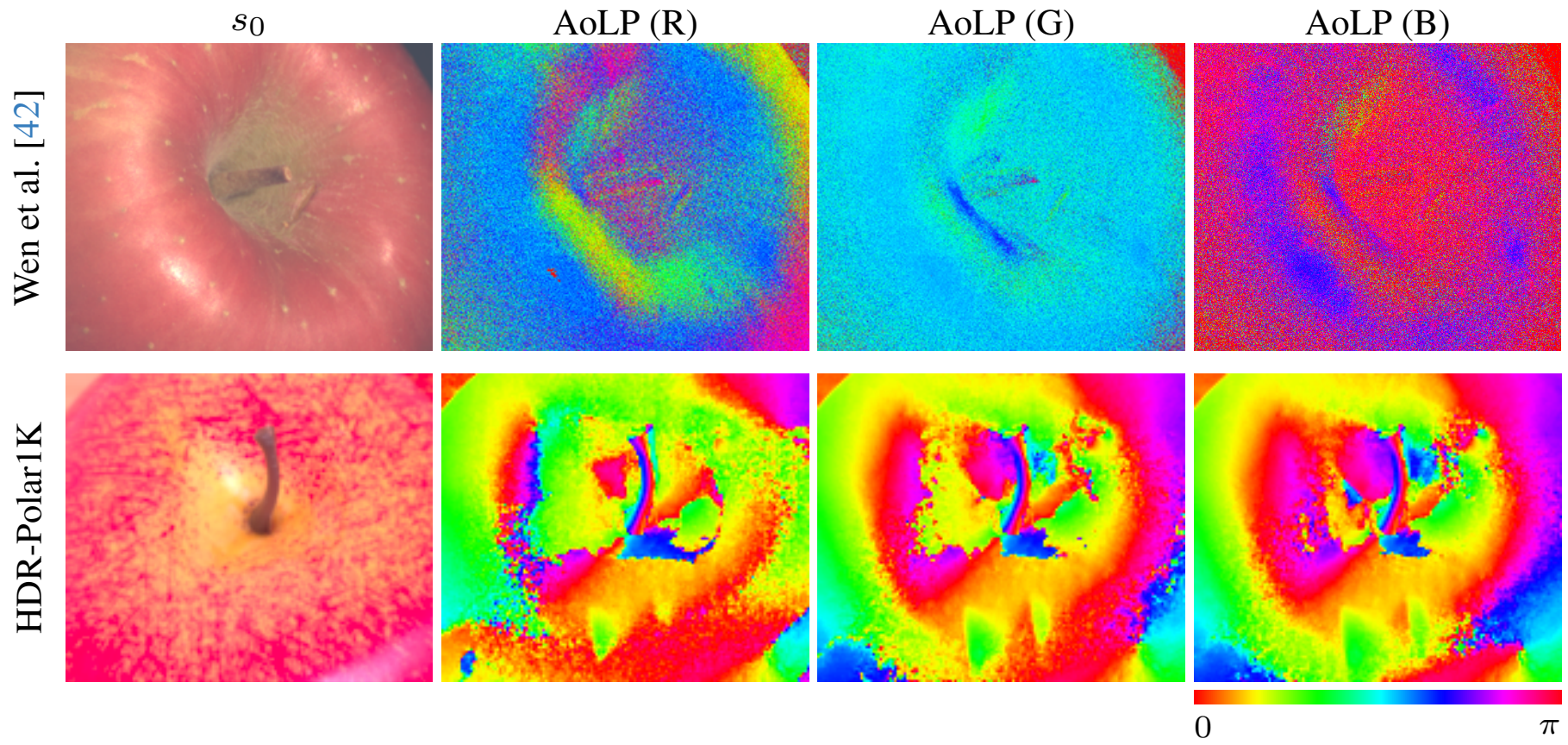}}
  \caption{\textbf{Color inconsistencies in existing datasets.} In Wen et al.~\cite{wen2019cpdnet}, the apple exhibits large discrepancies in AoLP across color channels, whereas HDR-Polar1K yields consistent AoLP estimates across RGB.}
  \label{fig:supp-dataset-color-example}
\end{figure}

\paragraph{Color inconsistencies.}
DoT capture systems can also introduce orientation-dependent color variations that distort Stokes estimates across RGB channels. Among the eight existing datasets, pronounced color inconsistencies appear in Wen et al.~\cite{wen2019cpdnet}, as also reported by Wen et al.~\cite{wen2021sparse}. As shown in \cref{fig:supp-dataset-color-example}, the apple region from Wen et al.~\cite{wen2019cpdnet} exhibits large discrepancies in AoLP across the R, G, and B channels, whereas everyday objects rarely exhibit such large AoLP differences across color channels. In contrast, HDR-Polar1K produces consistent AoLP across color channels.

\FloatBarrier

\subsection{Invalid samples: saturation and nonphysical Stokes vectors}
\label{sec:supp-dataset-saturation}

\paragraph{Saturation.}
All eight existing datasets contain saturated pixels, whereas HDR-Polar1K exhibits zero residual saturation. For existing datasets, we evaluate the released four-orientation images,\footnote{Rahman et al.~\cite{rahman2025polarization} provide captures at multiple noise levels, and we evaluate their low-noise ground-truth subset.} counting an RGB pixel as saturated if any channel reaches its maximum representable integer value. For HDR-Polar1K, we evaluate the 192 individual RAW frames per scene before averaging, dark subtraction, or downsampling, excluding defective pixels as described in \cref{sec:supp-stokes-construction}: a pixel is counted as saturated at 4,094 digital numbers (DN) or above, and a valid RAW site is residually saturated only if it is saturated in all 192 frames.

\begin{figure}[!ht]
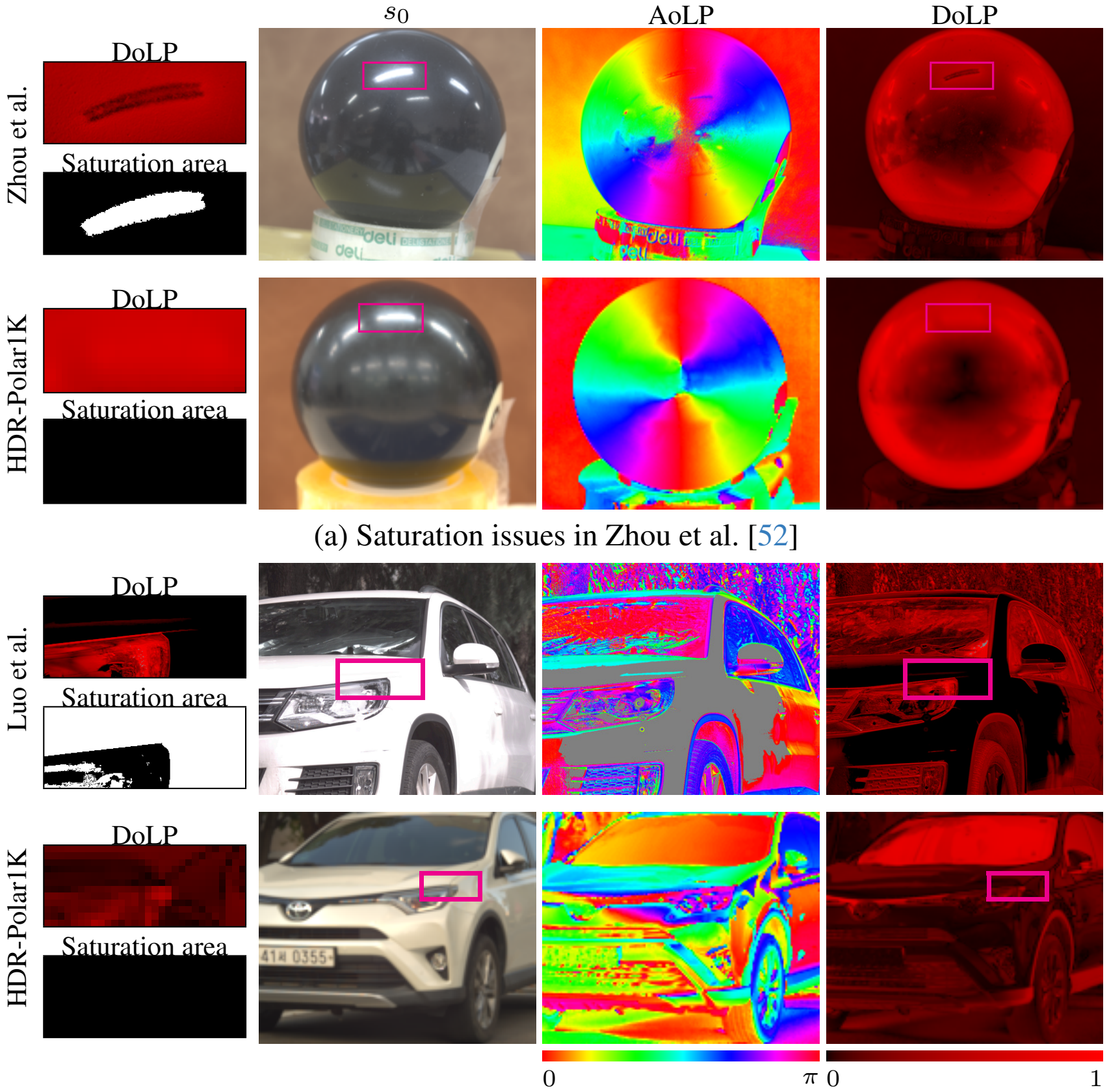

  \centering
  {\input{figures/dataset-saturation-grid}}\par\vspace{.5mm}
  {\input{figures/dataset-opid-saturation-grid}}
  \caption{\textbf{Saturation in existing datasets.} Clipped highlights, shown in white in the masks, cause erroneous DoLP in Zhou et al.~\cite{zhou2025pidsr} (a) and both erroneous DoLP and undefined AoLP, shown in gray, in Luo et al.~\cite{luo2026updem} (b). In HDR-Polar1K, every valid RAW site has at least one unsaturated frame.}
  \label{fig:supp-dataset-saturation-example}
\end{figure}

Intensity clipping corrupts the inter-orientation differences required for Stokes recovery, producing erroneous DoLP and undefined AoLP in highlights, as shown in \cref{fig:supp-dataset-saturation-example}. In contrast, exposure bracketing in HDR-Polar1K ensures at least one unsaturated reading at every valid RAW site, enabling the synthesis of partially saturated inputs while preserving unclipped Stokes targets.

\paragraph{Physical validity.}
Spatial misalignment, color inconsistencies, and intensity saturation can violate the physical Stokes constraints. Seven of the eight existing datasets contain nonphysical vectors, with only Qiu et al.~\cite{qiu2021linear} having none. In HDR-Polar1K, simultaneous fixed-camera capture, exposure bracketing, and low-overshoot moment-matched downsampling eliminate these error sources, and every RGB Stokes vector across all 1,008 scenes satisfies the physical constraints. We compute the nonphysical rate across all scenes and RGB channels using the criterion in \cref{eq:supp-nonphysical-rate} of \cref{sec:supp-hdr-metrics}.
\FloatBarrier

\subsection{Scene and polarization coverage}
\label{sec:supp-dataset-scene-coverage}

\begin{figure}[!htbp]
  \centering
  \includegraphics{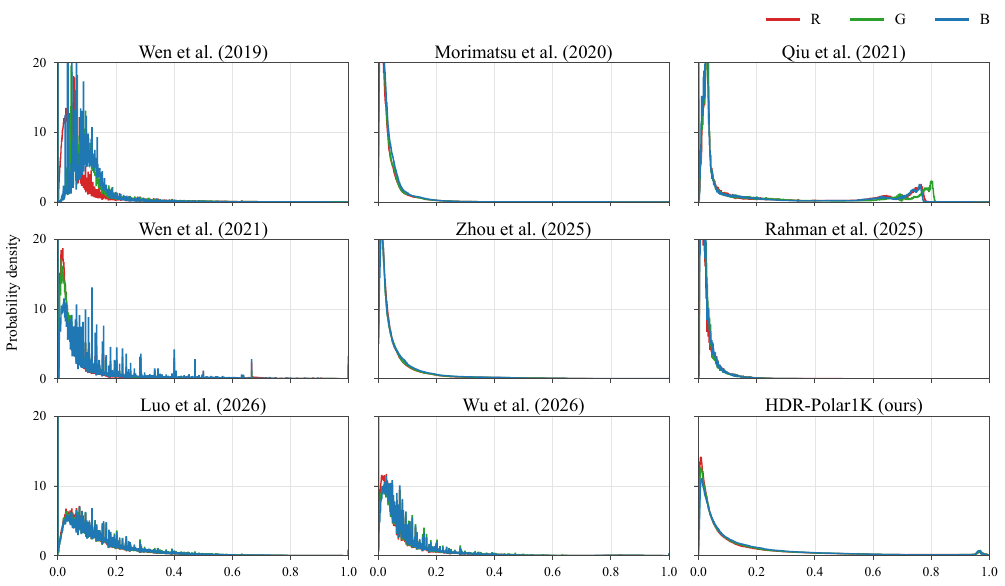}
  \caption{\textbf{DoLP distributions across datasets.} Per-color probability densities over \([0,1]\), grouping values above 1 into the final bin. The vertical axis is truncated at 20.}
  \label{fig:supp-dataset-dolp}
  \vspace{4mm}
  \includegraphics{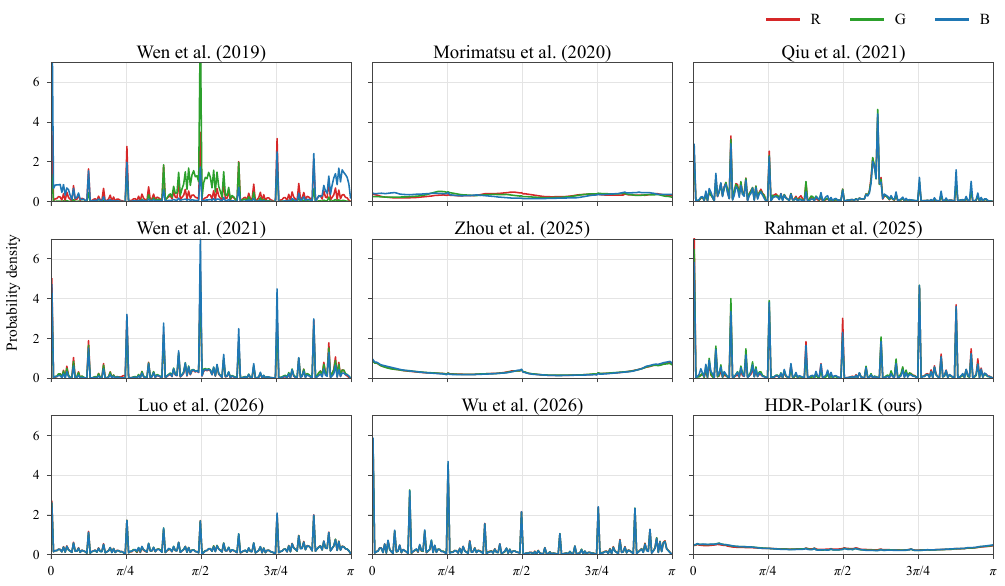}
  \caption{\textbf{AoLP distributions across datasets.} Per-color probability densities over \([0,\pi)\).}
  \label{fig:supp-dataset-aolp}
\end{figure}

\paragraph{Polarization distributions.}
\Cref{fig:supp-dataset-dolp,fig:supp-dataset-aolp} show the per-color DoLP and AoLP distributions across datasets. Several existing datasets exhibit sharp spikes in DoLP and AoLP, and most are heavily concentrated near zero DoLP. In contrast, HDR-Polar1K has a balanced, nearly uniform AoLP distribution across \([0,\pi)\), which helps prevent orientation bias relative to the camera axis, and covers a broader range of DoLP values.

\FloatBarrier

\input{figures/dataset-patch-gallery}

\begin{figure}[!ht]
  \centering
  \includegraphics[width=.91\linewidth]{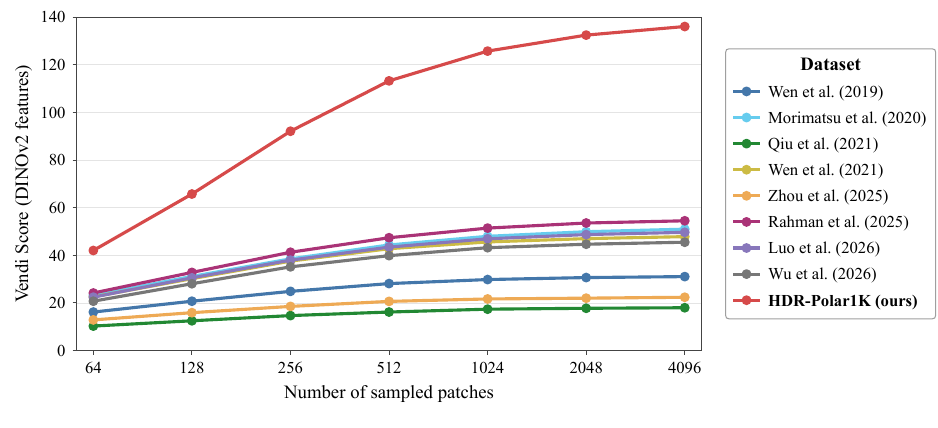}
  \caption{\textbf{Patch appearance diversity versus sample count.} Vendi Score computed from DINOv2 features of \(64\mathord{\times}64\) RGB \(s_0\) patches. Higher scores indicate greater diversity. Each line shows the mean Vendi Score over 30 random draws at each sample count.}
  \label{fig:supp-dataset-patch-vendi}
\end{figure}

\paragraph{Patch appearance diversity.}
HDR-Polar1K also provides substantially higher patch-level visual diversity than all eight existing datasets. \Cref{fig:supp-dataset-patches-1} compares randomly sampled \(64\mathord{\times}64\) RGB \(s_0\) patches extracted at each dataset's native resolution, matching the actual crops seen during patch-based training. Many crops from higher-resolution existing datasets fall on uniform single-surface regions, whereas HDR-Polar1K combines a larger scene count with compact framing to pack a wider variety of textures, colors, and structures into each patch. To quantify this diversity, we compute the Vendi Score~\cite{friedman2023vendi} from cosine similarities between DINOv2 CLS features~\cite{oquab2023dinov2} of uniformly sampled patches. As shown in \cref{fig:supp-dataset-patch-vendi}, HDR-Polar1K consistently outperforms every existing dataset across all sample counts, reaching approximately 136 at 4,096 patches compared with below 55 for all others. Since DINOv2 is designed for standard RGB images, this metric evaluates appearance diversity in \(s_0\) rather than polarization diversity in \(s_1\) and \(s_2\). Nonetheless, it confirms that HDR-Polar1K provides substantially richer visual content for patch-based training.

\FloatBarrier

\clearpage
\section{Sensor calibration and simulation}
\label{sec:supp-raw-simulation}
\label{sec:supp-raw-acquisition}

We calibrate the capture camera's Sony IMX250MYR CPFA sensor for two purposes: constructing coordinate-specific 0~dB dark-signal maps and pixel masks for HDR image construction in \cref{sec:supp-stokes-construction}, and parameterizing a multi-gain sensor simulator for synthesizing degraded training pairs in \cref{sec:supp-raw-training-preprocessing}.

\subsection{Sensor simulation model}
\label{sec:supp-raw-forward-model}

\begin{figure}[!ht]
 \centering
 \input{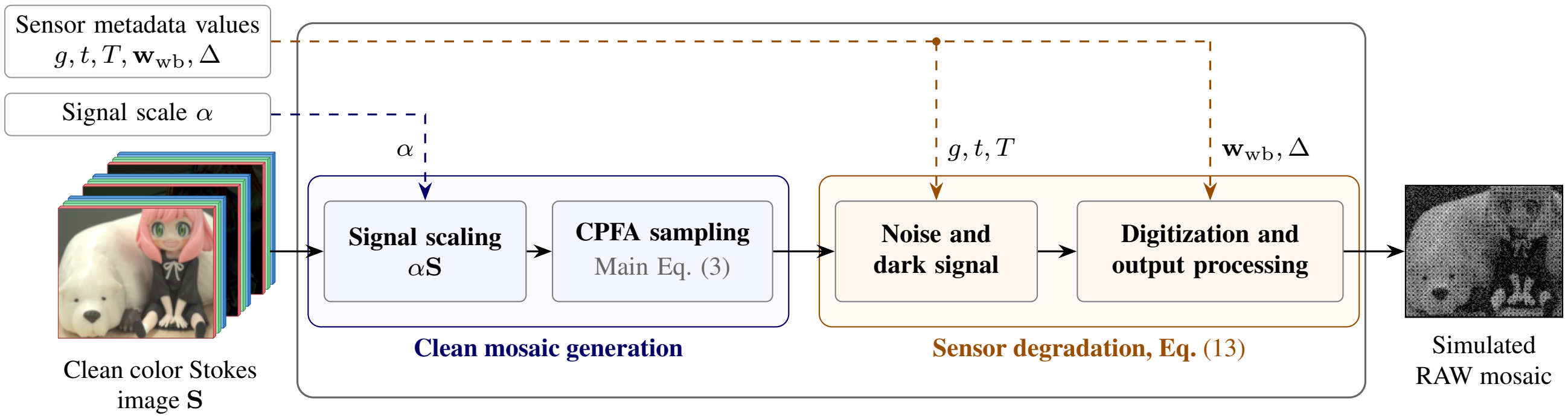}
 \caption{\textbf{Calibrated sensor simulation pipeline.} Starting from a clean color Stokes image \(\mathbf S\), the simulator first generates a clean CPFA mosaic at signal scale \(\alpha\) (left box) and then applies calibrated noise, dark signal, and output processing (right box) using the sensor metadata. Solid arrows trace the signal path, and dashed arrows show the parameter inputs.}
 \label{fig:supp-sensor-simulation-flow}
\end{figure}

To synthesize degraded RAW inputs for training, we build a calibrated sensor simulation model fit to measurements of our sensor, combining photon and dark-current shot noise, Gaussian readout noise, pixel-dependent dark offsets, quantization, and clipping~\cite{wei2022physics,feng2022learnability}. As summarized in \cref{fig:supp-sensor-simulation-flow}, the simulator is conditioned on the camera gain setting \(g\), exposure time \(t\), and sensor temperature \(T\).

\paragraph{Model overview.}
To obtain a clean RAW signal \(x\) at CPFA phase \(q\in\{0,\ldots,15\}\) from \cref{fig:supp-downsampling-tile}, we scale the HDR Stokes image \(\mathbf S\) by a positive signal scale \(\alpha\) as in \cref{sec:supp-exposure-sampling}, sample each pixel's color and polarizer orientation, and multiply by the 12-bit full-scale code (4,095~DN) to match the DN units of the calibrated sensor parameters. Given camera gain setting \(g\), exposure time \(t\), and sensor temperature \(T\), our simulator maps \(x\) to a simulated RAW code \(y\) as
\begin{equation}
 \begin{aligned}
 y &= \operatorname{clip}\!\Bigl(
 Q\!\Bigl(
 \underbrace{b(g,T)}_{\text{dark offset}}
 +\underbrace{\kappa_q(g)n_{\mathrm e}}_{\text{signal and shot noise}}
 +\underbrace{n_{\mathrm{read}}}_{\text{readout noise}}
 \Bigr)\Bigr),\\
 n_{\mathrm e} &\sim \operatorname{Poisson}\!\Bigl(
 \underbrace{\frac{x}{\kappa_q(g)}}_{\text{photoelectrons}}
 +\underbrace{d(T)t}_{\text{dark electrons}}
 \Bigr).
 \end{aligned}
 \label{eq:supp-sensor-model-overview}
\end{equation}
Here, \(b(g,T)\) is the dark offset and \(d(T)t\) denotes the mean dark electrons detailed below, \(n_{\mathrm e}\) is the total electron count combining photoelectrons and dark electrons~\cite{emva2021linear}, \(\kappa_q(g)\) is the phase-level conversion gain (DN/e\(^-\)) at camera gain setting \(g\), and \(n_{\mathrm{read}}\) is the Gaussian readout noise. The operator \(Q\) quantizes the signal to integer DN before clipping to the 12-bit output range. Varying \((g,t,T)\) at fixed \(\alpha\) changes sensor degradation without rescaling \(x\). In the output-processing stage of \cref{fig:supp-sensor-simulation-flow}, optional white-balance gains \(\mathbf w_{\mathrm{wb}}\) scale each color channel with rounding and clipping to full scale, followed by optional bit-depth reduction to 8 or 10 bits (with normalized quantization step \(\Delta\)) and normalization to \([0,1]\).

\paragraph{Dark signal.}
In our simulator, each pixel's pre-digitization mean dark signal combines a temperature-dependent dark offset and an exponential dark-current term:
\begin{equation}
 \mu_{\mathrm{dark}}(g,t,T)
 =\underbrace{b_0(g)+b_1(g)(T-T_0)}_{\text{dark offset } b(g,T)}
 +\kappa_q(g)\underbrace{d_0\exp\!\bigl(\lambda(T-T_0)\bigr)t}_{\text{dark electrons } d(T)t}.
 \label{eq:supp-reference-bl0-dark}
\end{equation}
Here, \(T_0=50\,^{\circ}\mathrm C\) is a reference temperature near the center of our calibration range, the per-pixel parameters \(b_0(g)\) and \(b_1(g)\) denote the dark offset and temperature slope at \(T_0\), and \(d_0\) and \(\lambda\) denote the gain-independent dark-current rate at \(T_0\) and its exponential temperature coefficient.

\paragraph{Independent and correlated readout noise.}
To capture both pixel-wise and row- or column-correlated readout noise, we decompose \(n_{\mathrm{read}}\) at native pixel \(i=(n_y,n_x)\) into pixel-independent, row-pair, and column components:
\begin{equation}
 n_{\mathrm{read},i}
 =\underbrace{\sigma_{\mathrm{ind},i}\epsilon_i}_{\text{pixel-independent}}
 +\underbrace{\sigma_{\mathrm{row},i}\epsilon_{\mathrm{row},\lfloor n_y/2\rfloor}}_{\text{row-pair}}
 +\underbrace{\sigma_{\mathrm{col},i}\epsilon_{\mathrm{col},n_x}}_{\text{column}}.
 \label{eq:supp-dark-noise-components}
\end{equation}
Each pair of native sensor rows shares one row-noise draw, and each column shares one column-noise draw, with pixel-dependent amplitudes. All \(\epsilon\) terms are independent standard normal variables drawn per frame.

We model the total readout variance at pixel \(i\) as
\begin{equation}
 \sigma_{\mathrm{read},i}^2(g,T)
 =\xi_i(g)\,v_q(g)\exp\!\bigl(\eta_q(g)(T-T_0)\bigr),
 \label{eq:supp-readout-temperature-model}
\end{equation}
where \(q\) is the CPFA phase of pixel \(i\), \(\xi_i(g)\) is a per-pixel variance multiplier, and \(v_q(g)\) and \(\eta_q(g)\) are the phase-level reference variance at \(T_0\) and its exponential temperature coefficient. The global row-pair and column reference variances follow analogous exponential temperature laws. To preserve total per-pixel readout variance, we scale them down proportionally whenever their sum exceeds \(\sigma_{\mathrm{read},i}^2\) and assign the remainder to \(\sigma_{\mathrm{ind},i}^2\).

\subsection{Calibration measurements}
\label{sec:supp-calibration-measurements}

We acquire calibration sequences of repeated full-resolution RAW frames at fixed camera settings.

\paragraph{Dark and flat sequences.}
We capture light-blocked dark frames to measure per-pixel dark offsets, exposure- and temperature-dependent dark current, readout noise, and recurring hot pixels. These dark sequences are acquired at 0~dB for per-pixel dark correction and defect exclusion, and across multiple gains for the sensor simulator. We also capture lensless flat frames under uniform diffused LED illumination, using multi-gain sequences to estimate conversion gains and 0~dB sequences under two illumination arrangements to check for photo-response anomalies.

\paragraph{Operating conditions.}
Across these calibration sequences, we vary camera gain \(g\) from 0 to 48~dB and exposure time \(t\) from 21\,\(\mu\)s to 1\,s via camera settings, and we vary sensor temperature \(T\) roughly from 32 to 78\,\(^{\circ}\)C using airflow from an air conditioner while logging the internal sensor temperature for each sequence.

\paragraph{Black level.}
Since dark signals at the 0\% black-level capture setting frequently clip at the lower endpoint, we acquire calibration dark frames with a 10\% black-level pedestal to observe the full dark-signal distribution. We estimate a constant baseline shift of 410 DN from paired 0\% and 10\% dark captures and subtract it to map calibrated offsets back to the 0\% capture setting.

\subsection{Parameter estimation}
\label{sec:supp-parameter-estimation}

\subsubsection{0~dB dark-signal maps and pixel masks}
\label{sec:supp-sensor-calibration}

To construct the HDR-Polar1K ground-truth Stokes images in \cref{sec:supp-stokes-construction}, we calibrate the capture camera at 0~dB directly at each physical pixel coordinate. Specifically, we fix \(g=0\)~dB in \cref{eq:supp-reference-bl0-dark} and fit each pixel's dark offset \(b_0(0)\), temperature slope \(b_1(0)\), dark-current rate \(\kappa_q(0)d_0\) in DN/s, and temperature coefficient \(\lambda\) from unclipped dark-sequence means. During HDR fusion, we evaluate this 0~dB model at each scene's logged temperature, clamping temperatures below the calibrated range.

We also detect and exclude anomalous pixels before HDR fusion. For dark anomalies, we identify pixels with recurring elevated dark signals at fixed coordinates across repeated captures, excluding those whose elevated dark signal cannot be reliably predicted or verified. The remaining candidates are dark-corrected and checked by the post-fusion residual filter. For photo-response anomalies, we compare dark-subtracted flat frames against nearby pixels of the same CPFA phase across two illumination arrangements, finding no reproducible anomalous pixels to exclude.

\subsubsection{Multi-gain simulator parameters}
\label{sec:supp-forward-calibration}

We estimate phase-level conversion gains from flat sequences and per-pixel dark-signal and readout parameters from dark sequences, tracking the same physical pixels across \((g,t,T)\) so their calibrated parameters can be stored jointly.

\paragraph{Conversion gain.}
We estimate the phase-level conversion gain \(\kappa_q(g)\) for each camera gain and CPFA phase via photon-transfer calibration~\cite{emva2021linear,wei2022physics}. For each flat sequence, we aggregate pixel-wise temporal means and variances by CPFA phase to avoid spatial illumination bias, then fit the linear variance-to-mean slope while excluding clipped or nonlinear points, and average the fitted slopes across repeated captures.

\paragraph{Dark signal.}
We fit the dark-signal model of \cref{eq:supp-reference-bl0-dark} across repeated dark sequences of each physical pixel, excluding clipped observations. The dark offset \(b_0(g)\) and temperature gradient \(b_1(g)\) are gain-specific, whereas the dark-current rate \(d_0\geq 0\) and temperature coefficient \(\lambda\) are shared across gains. Hot pixels are included in the same regression and are therefore captured directly by their fitted parameters rather than a separate defect class.

\paragraph{Readout noise.}
We isolate per-pixel readout variance from short-exposure dark frames by subtracting the fitted Poisson dark-current variance and quantization variance~\cite{emva2021linear} from the measured dark temporal variance. We aggregate these residuals over normal pixels per phase and gain to fit the phase-level exponential temperature law in \cref{eq:supp-readout-temperature-model}. We then estimate each pixel's variance multiplier \(\xi_i(g)\), shrinking normal-pixel multipliers to suppress finite-frame noise while retaining elevated multipliers for high-variance pixels. We estimate the row-pair and column noise variances from row- and column-averaged dark residuals after removing per-pixel means and frame-wide offsets. Adjacent rows within each native pair exhibit nearly identical temporal offsets, supporting two-row grouping, whereas the fitted column component is negligible.

\subsection{Calibration results and validation}
\label{sec:supp-calibration-validation}

\begin{figure}[!ht]
 \centering
 \begin{subfigure}[t]{0.32\linewidth}
  \centering
  \includegraphics[width=\linewidth]{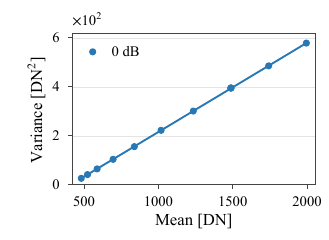}
  \caption{0~dB}
 \end{subfigure}\hfill
 \begin{subfigure}[t]{0.32\linewidth}
  \centering
  \includegraphics[width=\linewidth]{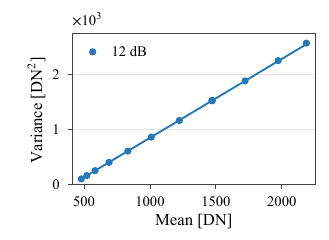}
  \caption{12~dB}
 \end{subfigure}\hfill
 \begin{subfigure}[t]{0.32\linewidth}
  \centering
  \includegraphics[width=\linewidth]{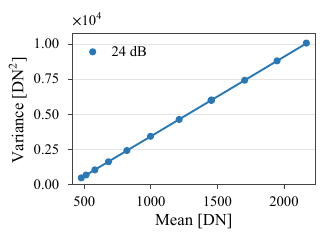}
  \caption{24~dB}
 \end{subfigure}
 \caption{\textbf{Photon-transfer calibration.} Measured RAW means and temporal variances for one CPFA phase, \(q=0\), at (a) 0, (b) 12, and (c) 24~dB. Points and lines show the retained observations and linear fits before dark subtraction, with each panel using its own vertical scale.}
 \label{fig:supp-photon-transfer}
\end{figure}

\begin{figure}[!ht]
 \centering
 \begin{subfigure}[t]{0.48\linewidth}
  \centering
  \includegraphics[width=\linewidth]{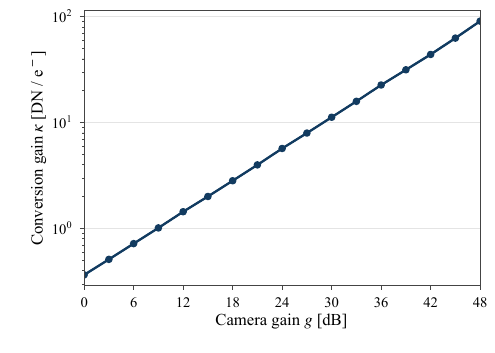}
  \caption{Conversion gain}
  \label{fig:supp-calibrated-conversion-gain}
 \end{subfigure}\hfill
 \begin{subfigure}[t]{0.48\linewidth}
  \centering
  \includegraphics[width=\linewidth]{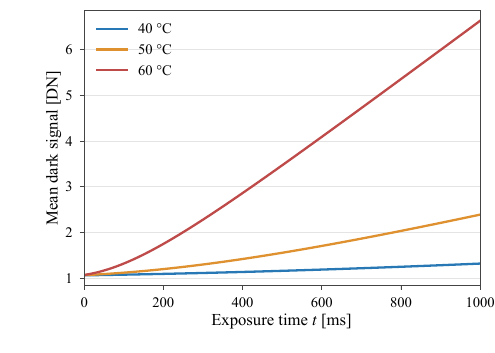}
  \caption{Mean dark signal}
  \label{fig:supp-exposure-dark-bl0}
 \end{subfigure}
 \caption{\textbf{Calibrated conversion gain and predicted dark signal.} (a) Mean conversion gain \(\kappa_q(g)\) over the 16 CPFA phases on a logarithmic vertical axis. (b) Predicted mean dark RAW value at 0~dB gain for \(40\), \(50\), and \(60\,^{\circ}\mathrm C\) over exposures from 21\,\(\mu\)s to 1\,s, averaged over the 16 phases.}
 \label{fig:supp-gain-dark}
\end{figure}

\paragraph{Calibrated parameter behavior.}
The fitted simulator parameters exhibit three consistent trends across operating conditions:
\begin{itemize}[leftmargin=1.5em,itemsep=2pt,topsep=4pt]
 \item \textbf{Conversion gain:} \Cref{fig:supp-photon-transfer} shows representative photon-transfer linear fits at three gains, and Figure~\ref{fig:supp-gain-dark}\subref{fig:supp-calibrated-conversion-gain} confirms that the phase-averaged conversion gains closely follow the expected \(10^{g/20}\) scaling across 0 to 48~dB.
 \item \textbf{Dark signal:} Figure~\ref{fig:supp-gain-dark}\subref{fig:supp-exposure-dark-bl0} plots the predicted mean dark signal at 0~dB across exposure time and temperature, capturing the sharp rise at long exposures and elevated temperatures.
 \item \textbf{Readout noise:} At \(T_0=50\,^{\circ}\mathrm C\), the median per-pixel readout standard deviation spans more than two orders of magnitude across the 48~dB gain range, from approximately 0.8 DN at 0~dB to 7.5 DN at 24~dB and 118 DN at 48~dB.
\end{itemize}

\paragraph{Accuracy against real captures.}
We validate the simulator against held-out real captures. Across nearly all calibration dark sequences (with the 10\% pedestal), predicted temporal variances agree within 5\% of measurements. On standard 0\% black-level dark and flat captures spanning 0--48~dB, 41--53\,\(^{\circ}\)C, and exposures up to 100\,ms, the worst-case phase-averaged mean discrepancy across all 16 CPFA phases stays within 1.2 DN for dark frames and 5.5 DN for flat frames on the 4,095 DN full scale. Clipping rates agree within 3.1 percentage points. Worst-case temporal variance error remains within 5--16\% up to 24~dB, with underestimation reaching 22\% for dark frames at 48~dB. Training with this simulator transfers effectively to real RAW captures, as shown in \cref{sec:supp-reconstruction-full-resolution}.

\paragraph{Scope.}
This calibration comes from a single camera unit. Randomizing per-pixel parameters avoids memorizing that unit's fixed defect map, but unit-to-unit differences in overall gain or noise level are not modeled. In addition, the camera reports a single scalar internal temperature that does not resolve spatial thermal gradients across the sensor array, leaving residual modeling error under extreme combinations of high gain, high temperature, and long exposure.

\subsection{Virtual-sensor sampling and interpolation}
\label{sec:supp-virtual-sensor}

To generate diverse training inputs without memorizing a single camera's defect map, we organize the calibrated simulator parameters into three levels of spatial sharing:
\begin{itemize}[leftmargin=1.5em,itemsep=2pt,topsep=4pt]
 \item \textbf{Global parameters:} Row-pair and column readout reference variances shared across the entire sensor.
 \item \textbf{Phase-level parameters:} Conversion gain \(\kappa_q(g)\) and reference readout variance parameters \(v_q(g),\eta_q(g)\), shared across pixels of the same CPFA phase \(q\).
 \item \textbf{Per-pixel parameters:} Each physical pixel's five-parameter tuple \(\bigl(b_0(g),b_1(g),d_0,\lambda,\xi(g)\bigr)\), capturing individual dark offset, dark current, and readout-variance scaling.
\end{itemize}
Fixing the per-pixel tuples at their physical coordinates during training would cause the network to overfit to one unit's fixed hot-pixel and dark-pattern map. Instead, we keep the global and phase-level parameters fixed and construct a virtual sensor for each training example by randomly resampling each pixel's five-parameter tuple independently from the calibrated empirical distribution of its CPFA phase. This randomization reduces reliance on the fixed defect map of the calibrated camera unit, preserves per-pixel parameter correlations between dark signal and readout variance, and supports arbitrary image resolutions.

\begin{figure}[p]
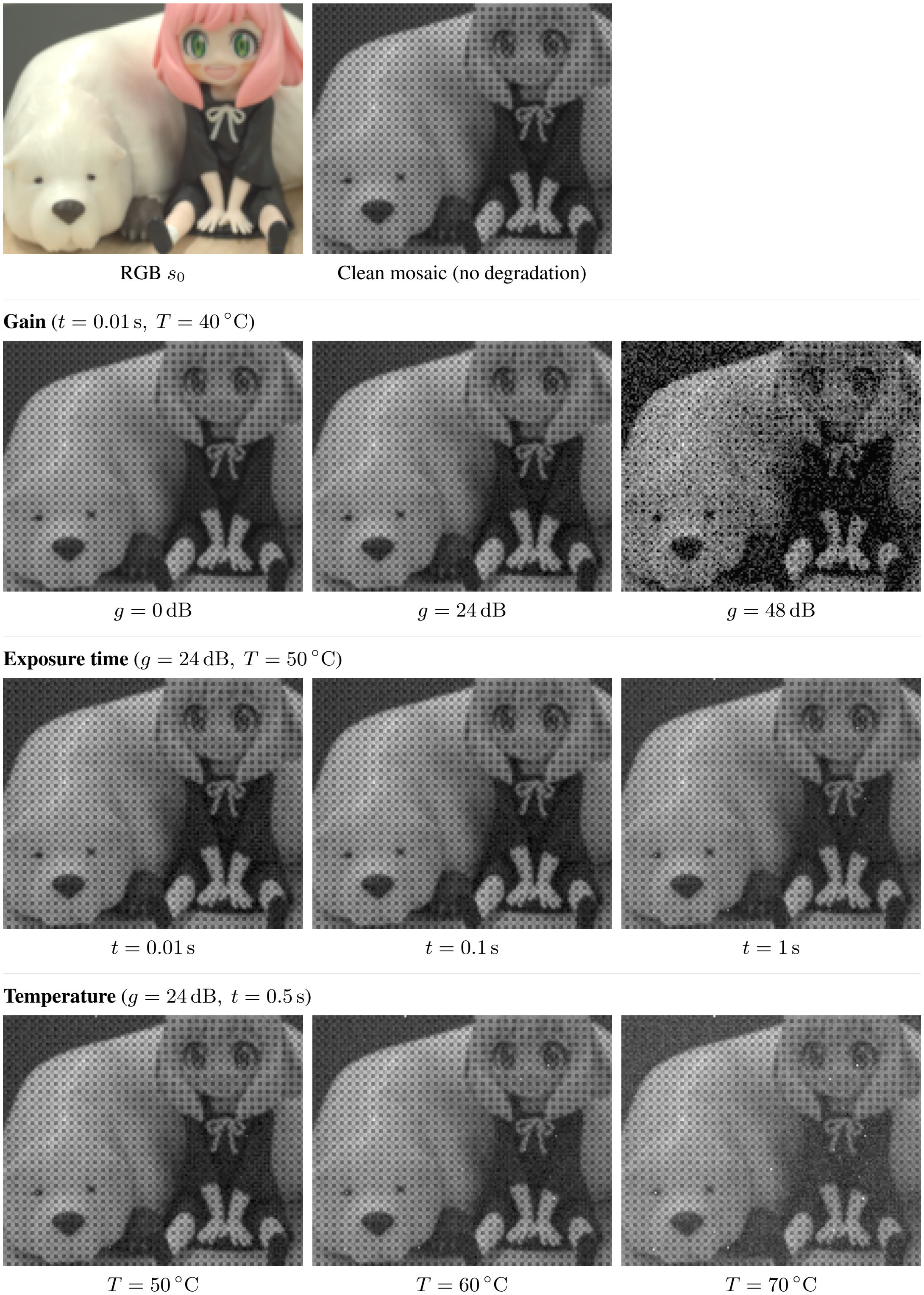

 \centering
 \begingroup
\begin{minipage}{0.82\linewidth}
\centering\fontsize{9}{10}\selectfont
\setlength{\tabcolsep}{0pt}
\renewcommand{\arraystretch}{1}
\edef\sensorpanelwidth{\the\dimexpr(\linewidth-3mm)/3\relax}
\newcommand{\sensorpanel}[1]{\csname ArxivImage:figures/sensor-degradation-panels/#1.png\endcsname{width=\sensorpanelwidth}}
\begin{tabular}{@{}c@{\hspace{1.5mm}}c@{\hspace{1.5mm}}c@{}}
\sensorpanel{reference_s0} & \sensorpanel{ideal_mosaic} & \\\noalign{\vskip.4mm}
RGB \(s_0\) & Clean mosaic (no degradation) & \\
\noalign{\vskip2mm{\color{black!12}\hrule height0.2pt}\vskip2mm}
\multicolumn{3}{@{}l@{}}{\textbf{Gain} (\(t=0.01\,\mathrm{s},\ T=40\,^{\circ}\mathrm{C}\))} \\\noalign{\vskip1mm}
\sensorpanel{gain_0} & \sensorpanel{gain_24} & \sensorpanel{gain_48} \\\noalign{\vskip.4mm}
\(g=0\,\mathrm{dB}\) & \(g=24\,\mathrm{dB}\) & \(g=48\,\mathrm{dB}\) \\
\noalign{\vskip2mm{\color{black!12}\hrule height0.2pt}\vskip2mm}
\multicolumn{3}{@{}l@{}}{\textbf{Exposure time} (\(g=24\,\mathrm{dB},\ T=50\,^{\circ}\mathrm{C}\))} \\\noalign{\vskip1mm}
\sensorpanel{exposure_001} & \sensorpanel{exposure_01} & \sensorpanel{exposure_1} \\\noalign{\vskip.4mm}
\(t=0.01\,\mathrm{s}\) & \(t=0.1\,\mathrm{s}\) & \(t=1\,\mathrm{s}\) \\
\noalign{\vskip2mm{\color{black!12}\hrule height0.2pt}\vskip2mm}
\multicolumn{3}{@{}l@{}}{\textbf{Temperature} (\(g=24\,\mathrm{dB},\ t=0.5\,\mathrm{s}\))} \\\noalign{\vskip1mm}
\sensorpanel{temperature_50} & \sensorpanel{temperature_60} & \sensorpanel{temperature_70} \\\noalign{\vskip.4mm}
\(T=50\,^{\circ}\mathrm{C}\) & \(T=60\,^{\circ}\mathrm{C}\) & \(T=70\,^{\circ}\mathrm{C}\)
\end{tabular}\par
\end{minipage}
\endgroup
 \caption{\textbf{Simulated sensor degradations.} Each degradation row varies one operating parameter while holding the others at the labeled settings. The signal scale \(\alpha\) is fixed across conditions to isolate sensor-model effects. Although simulated degradations are randomized for each training sample, we use a fixed random seed here so that hot pixels appear at the same locations across conditions.}
 \label{fig:supp-sensor-degradation}
\end{figure}

To evaluate arbitrary operating conditions \((g,t,T)\) within the simulator's operating domain of 0--48~dB, 21\,\(\mu\)s--1\,s, and 37--76\,\(^{\circ}\)C, we interpolate gain-dependent parameters between calibrated gain steps, linearly for offsets and multipliers, and in log space for conversion gains and reference variances. Exposure time and temperature are evaluated directly from the continuous physical laws. \Cref{fig:supp-sensor-degradation} visualizes simulated sensor degradations generated by this virtual sensor across gain, exposure time, and temperature.

\FloatBarrier

\section{Experimental setup and Raw2Stokes implementation}
\label{sec:supp-implementation-training}

We detail the common experimental setup and Raw2Stokes implementation used across the quantitative evaluations: training-pair generation, the Raw2Stokes architecture and sensor conditioning, optimization and blind-evaluation protocol, and the four evaluation metrics.

\subsection{Training-pair generation}
\label{sec:supp-raw-training-preprocessing}

We generate each training pair from a clean Stokes image in four steps: we extract a random \(64\mathord{\times}64\) crop, apply random \(90^\circ\)-multiple rotations and reflections with the corresponding Stokes transformations, sample a random signal scale \(\alpha\) as in \cref{sec:supp-exposure-sampling}, and simulate a degraded RAW mosaic under sampled sensor conditions as in \cref{sec:supp-calibrated-raw-generation}.

\subsubsection{Random signal-level scaling}
\label{sec:supp-exposure-sampling}

We randomly scale the brightness of each training crop to cover partial sensor saturation while respecting the redundancy of four-angle polarization sensing. Under an ideal co-located model, any three of the four polarizer intensities suffice to recover the linear Stokes vector, allowing one measurement to saturate without making reconstruction ill-posed~\cite{lapray2020exploiting,zhou2023polarization}.

We first define a nominal brightness scale \(\alpha_0=1/P_{99}(s_0)\), where \(P_{99}(s_0)\) is the 99th percentile of \(s_0\) across all pixels and RGB channels, which serves as a reference brightness scale. To set the upper end, we consider the ideal co-located polarizer intensities for \(0^\circ,45^\circ,90^\circ,\) and \(135^\circ\). At least three of the four intensities remain below the normalized saturation limit of one whenever the second-largest intensity \(I_{(2),c}\) satisfies \(\alpha I_{(2),c}<1\), where
\begin{equation}
  I_{(2),c}=\frac{s_{0,c}+\min\!\left(|s_{1,c}|,|s_{2,c}|\right)}{2}.
  \label{eq:supp-second-analyzer-intensity}
\end{equation}
Taking \(\max_c I_{(2),c}\) at each pixel accounts for the most restrictive color, and we define the polarization-aware upper threshold using the 99th percentile with a 1\% headroom margin:
\begin{equation}
  \alpha_{\mathrm{cap}}=\frac{0.99}{P_{99}\!\left(\max_c I_{(2),c}\right)}.
  \label{eq:supp-exposure-boundary}
\end{equation}
For each crop, we compute these brightness statistics from either the full image or the crop itself with equal probability, balancing global scene brightness with locally exposed patches.

We then sample \(\alpha\) log-uniformly from 4~EV below \(\alpha_0\) to 1~EV above \(\alpha_{\mathrm{cap}}\) to cover both underexposed dark inputs and more severe saturation, clamping the lower bound to \(2\alpha_{\mathrm{cap}}\) whenever \(2^{-4}\alpha_0 > 2\alpha_{\mathrm{cap}}\):
\begin{equation}
  \log_2\alpha\sim\mathcal U\!\left(\min\!\left(\log_2\!\left(2^{-4}\alpha_0\right),\,\log_2\!\left(2\alpha_{\mathrm{cap}}\right)\right),\,\log_2\!\left(2\alpha_{\mathrm{cap}}\right)\right).
  \label{eq:supp-exposure-scaled-target}
\end{equation}
We scale the clean Stokes image to \(\alpha\mathbf s_c\) before CPFA mosaicing and sensor simulation. For external datasets, we apply the same multiplier \(\alpha\) directly to the released polarizer images to form the mosaic, while using their derived Stokes images as reconstruction targets.

\subsubsection{Sensor-condition sampling and training targets}
\label{sec:supp-calibrated-raw-generation}

\paragraph{Sampling sensor conditions.}
Within the simulator's operating domain in \cref{sec:supp-virtual-sensor}, we sample camera gain and temperature uniformly, exposure time log-uniformly, and output bit depth uniformly from 8, 10, and 12 bits. We also sample red and blue WB gains \(\mathbf w_{\mathrm{wb}}=(w_{\mathrm{wb},R},w_{\mathrm{wb},G},w_{\mathrm{wb},B})\) independently and log-uniformly from \([1,2]\), with \(w_{\mathrm{wb},G}=1\). For each crop, we sample a fresh virtual sensor as in \cref{sec:supp-virtual-sensor} and apply the sensor simulator of \cref{sec:supp-raw-forward-model} to the scaled clean mosaic.

\paragraph{Input and target pairs.}
The network takes the normalized simulated RAW mosaic in \([0,1]\) without dark subtraction or WB division, so sensor noise, pixel-dependent dark offsets, hot pixels, quantization, and clipping remain in the input. The reconstruction target is the clean Stokes image scaled by both signal level and WB gains without rounding or clipping, \(\mathbf s_c^\star=w_{\mathrm{wb},c}\alpha\mathbf s_c\).

\subsection{Raw2Stokes architecture}
\label{sec:supp-noise-conditioning}

\begin{figure}[t]
  \centering
  \input{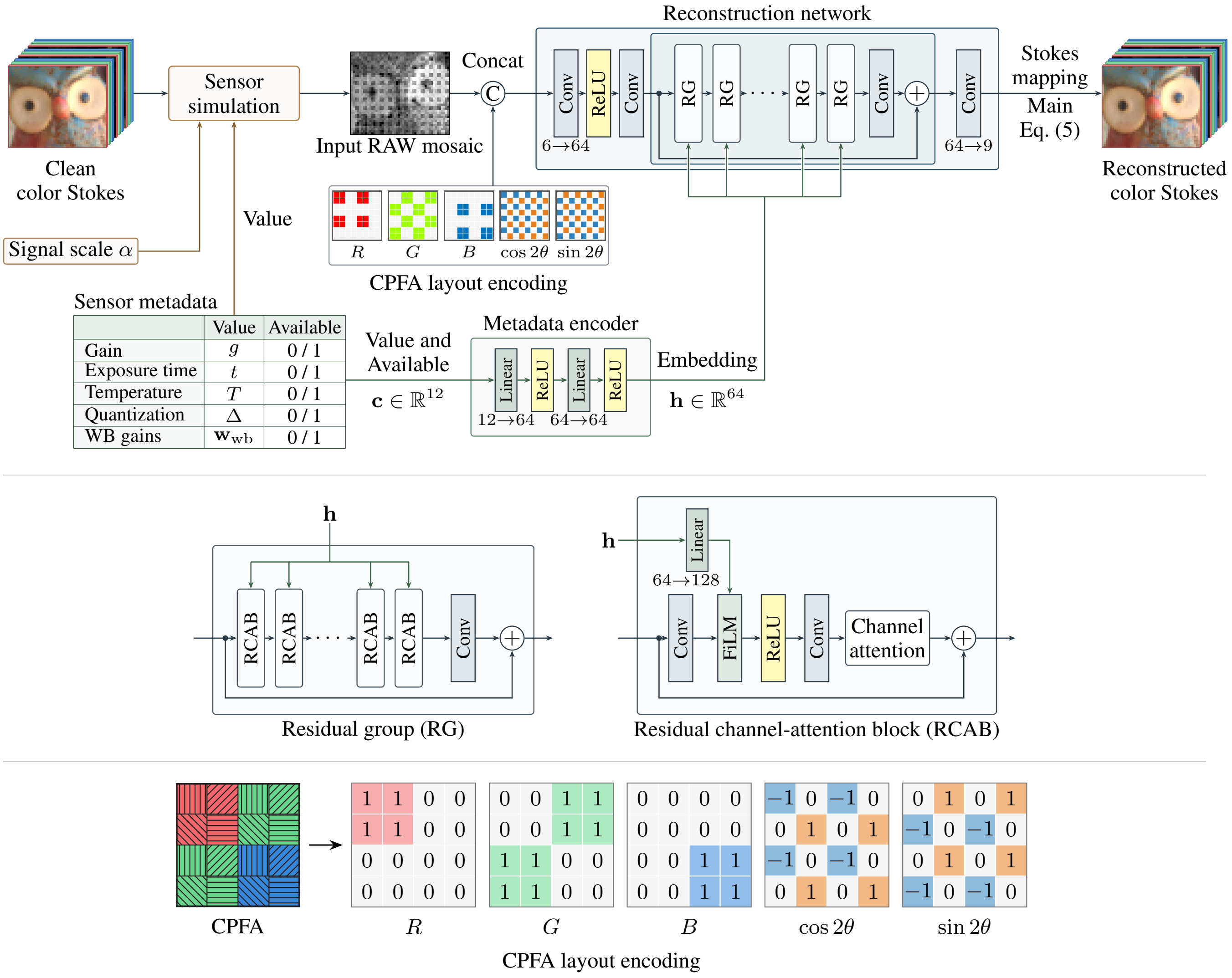}\par
  \vspace{3mm}
  {\color{black!22}\hrule height .35pt}
  \vspace{3mm}
  \begingroup
\definecolor{detailInk}{HTML}{263746}
\definecolor{detailBlue}{HTML}{346689}
\definecolor{detailGreen}{HTML}{477553}
\newcommand{\detailFlowY}{14}
\begin{tikzpicture}[x=0.79mm,y=0.79mm,font=\rmfamily\fontsize{9}{10}\selectfont,
 frame/.style={draw=detailBlue!80!black,line width=0.435pt,rounded corners=1.185pt},
 op/.style={draw=detailInk!85,line width=0.435pt,fill=white,rounded corners=1.185pt,align=center,inner sep=1.58pt},
 plate/.style={draw=detailInk,line width=0.435pt,minimum width=3.476mm,minimum height=12.64mm,inner sep=0pt},
 vertical/.style={rotate=90,inner sep=0pt,font=\rmfamily\fontsize{8}{9}\selectfont},
 sum/.style={circle,draw=detailInk,line width=0.435pt,fill=white,minimum size=3.476mm,inner sep=0pt},
 arrow/.style={draw=detailInk,-{Stealth[length=0.988mm,width=0.75mm]},line width=0.474pt},
 feed/.style={arrow,draw=detailGreen!85!black},
 caption/.style={font=\rmfamily\fontsize{9}{10}\selectfont,inner sep=0pt}]
 \draw[frame,draw=detailInk!85,fill=detailBlue!2] (0,0) rectangle (59,31);
 \foreach \x/\name in {7/bone,14/btwo,29/bnine,36/bten} {
   \node[op,fill=detailBlue!2,minimum width=3.95mm,minimum height=14.22mm] (\name) at (\x,\detailFlowY) {};
   \node[vertical] at (\name.center) {RCAB};
 }
 \node[inner sep=.8pt] (group-latent) at (21.5,37) {$\mathbf h$};
 \draw[draw=detailGreen!85!black,line width=.474pt] (group-latent.south) -- (21.5,28)
   (7,28) -- (36,28);
 \foreach \block in {bone,btwo,bnine,bten}
   \draw[feed] (\block.north |- 0,28) -- (\block.north);
 \node[inner sep=0.316mm] (dots) at (21.5,\detailFlowY) {$\cdots$};
 \node[plate,fill=detailBlue!15] (tail) at (46,\detailFlowY) {};
 \node[vertical] at (tail.center) {Conv};
 \node[sum] (gadd) at (55,\detailFlowY) {$+$};
 \draw[arrow] (-3.5,\detailFlowY) -- (bone.west);
 \foreach \a/\b in {bone/btwo,btwo/dots,dots/bnine,bnine/bten,bten/tail,tail/gadd}
   \draw[arrow] (\a.east) -- (\b.west);
 \draw[arrow] (gadd.east) -- (62.5,\detailFlowY);
 \draw[arrow] (2.3,\detailFlowY) -- (2.3,3) -| (gadd.south);
 \fill[detailInk] (2.3,\detailFlowY) circle[radius=1.145pt];
 \node[caption] at (29.5,-3) {Residual group (RG)};

 \begin{scope}[xshift=7.11mm]
 \draw[frame,draw=detailInk!85,fill=detailBlue!2] (69,0) rectangle (135,40);
 \node[text=black,anchor=east,inner sep=0.79pt] (latent) at (65.5,32) {$\mathbf h$};
 \node[plate,fill=detailGreen!25,minimum width=3.16mm,minimum height=9.085mm] (proj) at (80,32) {};
 \node[vertical,font=\rmfamily\fontsize{7}{8}\selectfont] at (proj.center) {Linear};
 \node[anchor=north,inner sep=0.79pt,font=\rmfamily\fontsize{7}{8}\selectfont] at ([xshift=-1.58mm]proj.south) {$64{\to}128$};
 \draw[feed] (latent.east) -- (proj.west);
 \coordinate (x) at (65.5,\detailFlowY);
 \foreach \xx/\name/\labeltext/\colorname in {77/first/Conv/detailBlue!15,86/mod/FiLM/detailGreen!17,94/relu/ReLU/yellow!32,102/second/Conv/detailBlue!15} {
   \node[plate,fill=\colorname] (\name) at (\xx,\detailFlowY) {};
   \node[vertical] at (\name.center) {\labeltext};
 }
 \node[op,minimum height=7.9mm] (attn) at (115,\detailFlowY) {Channel\\attention};
 \node[sum] (add) at (129,\detailFlowY) {$+$};
 \coordinate (y) at (138.5,\detailFlowY);
 \draw[feed] (proj.east) -- (mod.north |- proj.east) -- (mod.north);
 \foreach \a/\b in {x/first,first/mod,mod/relu,relu/second,second/attn,attn/add,add/y}
   \draw[arrow] (\a.east) -- (\b.west);
 \draw[arrow] (73,\detailFlowY) -- (73,3) -| (add.south);
 \fill[detailInk] (73,\detailFlowY) circle[radius=1.145pt];
 \node[caption] at (102,-3) {Residual channel-attention block (RCAB)};
 \end{scope}
\end{tikzpicture}%
\endgroup\par
  \vspace{3mm}
  {\color{black!22}\hrule height .35pt}
  \vspace{3mm}
  \begingroup
\definecolor{layoutPositive}{RGB}{230,126,34}
\definecolor{layoutNegative}{RGB}{49,130,189}
\begin{tikzpicture}[x=1mm,y=1mm,font=\fontsize{9}{10}\selectfont]
  \node[inner sep=0pt] at (9,9) {\CPFASensor[width=18mm,stripe count=6,stripe width=.9bp,grid width=.9bp,border width=1.45bp,red=cpfaChannelRed,green=cpfaChannelGreen,blue=cpfaChannelBlue]};
  \node at (9,-2.8) {CPFA};
  \draw[-{Stealth},line width=.55pt] (19.3,9) -- (24.2,9);
  \foreach \channel/\labeltext/\activecolor in {0/R/cpfaChannelRed,1/G/cpfaChannelGreen,2/B/cpfaChannelBlue,3/{\cos 2\theta}/layoutPositive,4/{\sin 2\theta}/layoutPositive}{
    \begin{scope}[xshift={25.5mm+20mm*\channel}]
      \foreach \row in {0,...,3}{
        \foreach \col in {0,...,3}{
          \pgfmathtruncatemacro{\colorindex}{floor(\row/2)+floor(\col/2)}
          \pgfmathtruncatemacro{\cellvalue}{\channel<3 ? (\colorindex==\channel) : (\channel==3 ? mod(\row,2)+mod(\col,2)-1 : mod(\col,2)-mod(\row,2))}
          \ifnum\cellvalue=0 \def\shadecolor{black!4}\else
            \ifnum\cellvalue=1 \def\shadecolor{\activecolor!55}\else \def\shadecolor{layoutNegative!55}\fi
          \fi
          \filldraw[fill=\shadecolor,draw=white,line width=.4pt] (4.5*\col,13.5-4.5*\row) rectangle ++(4.5,4.5);
          \node[inner sep=0pt] at (2.25+4.5*\col,15.75-4.5*\row) {$\cellvalue$};
        }
      }
      \draw[black!55,line width=.4pt] (0,0) rectangle (18,18);
      \node at (9,-2.8) {$\labeltext$};
    \end{scope}
  }
  \node[anchor=north,inner sep=0pt] at ([yshift=-1.5mm]current bounding box.south) {CPFA layout encoding};
\end{tikzpicture}
\endgroup
  \caption{\textbf{Raw2Stokes training pipeline and architectural details.} Complementing Figure~6 of the main paper, the top row details how the calibrated sensor simulator from \cref{fig:supp-sensor-simulation-flow} and metadata conditioning integrate with the reconstruction network during training (warm arrows show simulator inputs, green arrows trace FiLM conditioning, and stacks show \(s_0,s_1,s_2\) from front to back). The middle row expands each residual group (RG) into ten FiLM-conditioned residual channel-attention blocks (RCABs). The bottom row details the five-channel CPFA encoding maps for the \(4\times4\) tile.}
  \label{fig:raw2stokes-overview}
\end{figure}

Raw2Stokes builds on an RCAN-based backbone~\cite{zhang2018rcan}, previously used for joint demosaicing and denoising~\cite{xing2021endtoend,tedla2025examining}, with RAW inputs and five-channel CPFA encoding maps inspired by Tedla et al.~\cite{tedla2025examining}, as shown in \cref{fig:raw2stokes-overview}.

\paragraph{Network structure and size.}
The network concatenates the RAW mosaic with the five CPFA encoding maps to form six spatial input channels, processes them with \(3\mathord{\times}3\) convolutions across 64 feature channels and five residual groups of ten residual channel-attention blocks each, and predicts nine output channels. We apply the mapping in Eq.~(5) of the main paper to the output head to produce a physically valid Stokes vector for each color channel at each pixel. Including the FiLM conditioning layers and output head, Raw2Stokes has 4.4 million parameters.

\paragraph{Test-time local pooling.}
During training, each channel-attention block averages its features over the entire \(64\mathord{\times}64\) patch. To keep the pooling region comparable in size when processing larger images, we apply the Test-time Local Converter (TLC)~\cite{chu2022improving} during validation and inference, computing local averages with a stride-one \(96\mathord{\times}96\) window.

\paragraph{Sensor conditioning.}
We condition the network on camera gain \(g\), exposure time \(t\), temperature \(T\), quantization step \(\Delta\), and RGB WB gains \(\mathbf w_{\mathrm{wb}}\). Each metadata value is normalized to a comparable scale and concatenated with its binary availability flag, using one shared flag for the three WB gains, to form a 12-dimensional vector \(\mathbf c\in\mathbb R^{12}\) with unavailable entries set to zero. As illustrated in \cref{fig:raw2stokes-overview}, a two-layer MLP encodes \(\mathbf c\) into a shared 64-dimensional embedding \(\mathbf h\), which each RCAB linearly projects and splits into 64-dimensional channel-wise scale and shift vectors for feature-wise linear modulation (FiLM)~\cite{perez2018film} after the first convolution. These projections are initialized to zero so modulation starts as the identity.

\subsection{Training objective and optimization}
\label{sec:supp-optimization}

\paragraph{Training objective.}
We supervise the network against the scaled Stokes target \(\mathbf s_c^\star\) from \cref{sec:supp-calibrated-raw-generation} over the entire crop without excluding a border, minimizing the weighted Charbonnier loss~\cite{charbonnier1997deterministic}:
\begin{equation}
  \mathcal{L}
  =
  \frac{1}{3|\Omega|(1+2\lambda_{\mathrm{pol}})}
  \sum_{\mathbf u\in\Omega}
  \sum_{c\in\{R,G,B\}}
  \Bigl(
  \Phi\!\left(\widehat s_{0,c}(\mathbf u)-s^\star_{0,c}(\mathbf u)\right)
  +\lambda_{\mathrm{pol}}\sum_{k=1}^{2}
  \Phi\!\left(\widehat s_{k,c}(\mathbf u)-s^\star_{k,c}(\mathbf u)\right)
  \Bigr),
  \label{eq:r2s-loss}
\end{equation}
where \(\Omega\) is the supervised pixel region, \(\Phi(e)=\sqrt{e^2+\epsilon_{\mathcal L}^2}\) is the Charbonnier penalty, and we set \(\epsilon_{\mathcal L}=10^{-3}\) and \(\lambda_{\mathrm{pol}}=5\) to weight the polarization components \((s_1,s_2)\) relative to intensity \(s_0\).

\paragraph{Metadata dropout and blind evaluation.}
During training, we withhold all metadata with probability 0.25. Otherwise, each of gain, exposure time, temperature, quantization step, and WB is independently withheld with probability 0.1. This dropout allows a single trained model to operate with complete, partial, or missing metadata. For blind evaluation in the primary training-data comparison, all availability indicators and encoded values are zero.

\paragraph{Optimization.}
We train Raw2Stokes from scratch using Adam~\cite{kingma2015adam}, with momentum coefficients \((0.9,0.999)\), numerical stability constant \(10^{-8}\), and no weight decay. The batch size is 16. The learning-rate schedule spans 3,000,000 optimizer updates: a linear warmup from \(10^{-6}\) to \(10^{-4}\) over the first 30,000 updates, followed by cosine decay to \(10^{-6}\).

\paragraph{Validation and implementation.}
We split HDR-Polar1K into 858 training, 50 validation, and 100 test scenes, and similarly split each external dataset into training and validation sets. For each training dataset, we select Raw2Stokes weights using the lowest non-blind validation loss on that dataset's validation split under fixed simulated acquisition conditions. We train with FP16 mixed precision by default and use FP32 when the training loss becomes nonfinite.

\subsection{Evaluation protocol and metrics}
\label{sec:supp-hdr-metrics}

We use the same protocol and four metrics across all quantitative experiments in \cref{sec:benchmark,sec:supp-additional-comparisons,sec:supp-model-analyses}. All comparisons evaluate linear Stokes predictions and exclude an eight-pixel border.

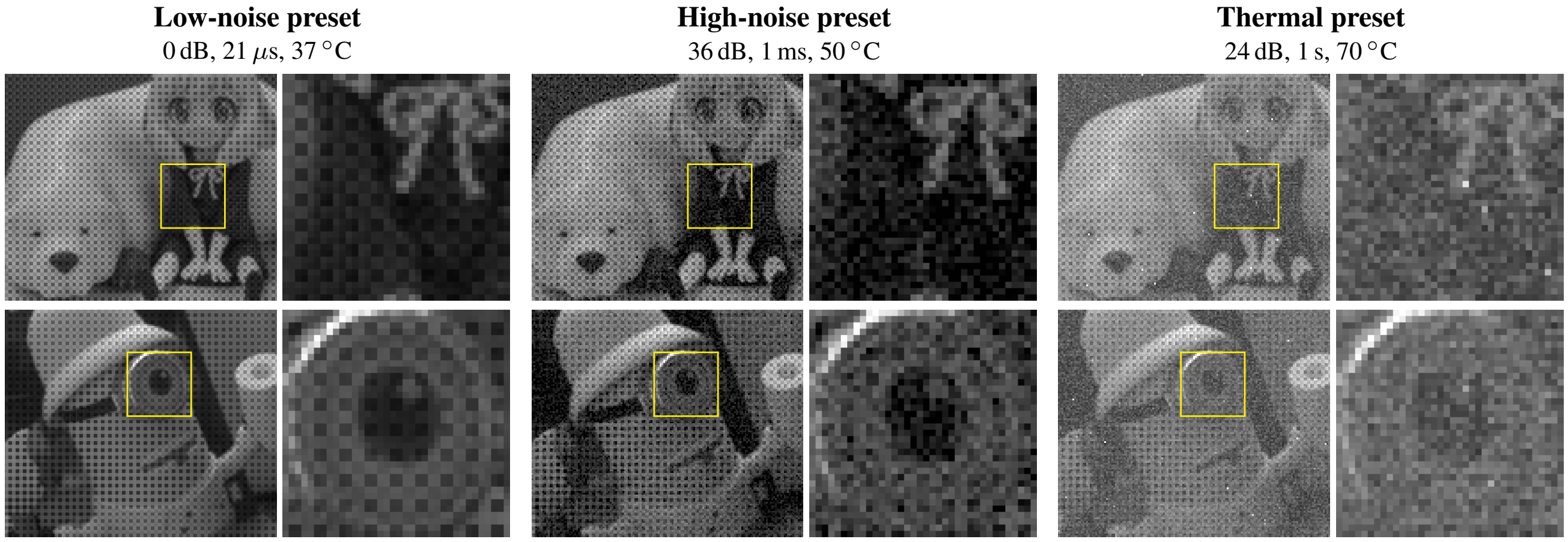
\begin{figure}[!htbp]
  \centering
  \begin{tikzpicture}[x=1mm,y=-1mm,inner sep=0pt,outer sep=0pt]
  \foreach \x/\preset/\titletext/\settings in {
    0/low/{Low-noise preset}/{0\,dB, 21\,$\mu$s, 37\,$^{\circ}$C},
    58/high/{High-noise preset}/{36\,dB, 1\,ms, 50\,$^{\circ}$C},
    116/thermal/{Thermal preset}/{24\,dB, 1\,s, 70\,$^{\circ}$C}}{
    \begin{scope}[shift={(\x,0)}]
      \node[font=\fontsize{9}{10}\selectfont\bfseries] at (27.8,0) {\titletext};
      \node[font=\fontsize{8}{9}\selectfont] at (27.8,3.6) {\settings};
      \foreach \y/\scene/\px/\py in {6/dolls/88/51,32/bird/69/24}{
        \node[anchor=north west] at (0,\y) {\csname ArxivImage:figures/sensor-test-conditions/\scene-\preset.png\endcsname{height=25mm}};
        \draw[yellow,line width=.5pt] ({25*\px/128},{\y+25*\py/128}) rectangle ({25*(\px+36)/128},{\y+25*(\py+36)/128});
        \node[anchor=north west] at (30.6,\y) {\csname ArxivImage:figures/sensor-test-conditions/\scene-\preset-crop.png\endcsname{height=25mm}};
      }
    \end{scope}
  }
\end{tikzpicture}
  \caption{\textbf{Test RAW inputs under the three evaluation presets.} Example simulated RAW inputs from two HDR-Polar1K test scenes under the three fixed sensor presets used in the quantitative evaluations of \cref{sec:benchmark,sec:supp-additional-comparisons,sec:supp-model-analyses}. Compared with the low-noise mosaic, the high-noise preset adds strong readout and shot noise, while the thermal preset raises the dark-current baseline and introduces bright hot pixels.}
  \label{fig:supp-sensor-test-conditions}
\end{figure}

\paragraph{Sensor test conditions.}
Across the experiments, we evaluate under three fixed sensor presets:
\begin{itemize}[leftmargin=1.5em,itemsep=2pt,topsep=4pt]
  \item \textbf{Low noise:} \(g=0\,\mathrm{dB}\), \(t=21\,\mu\mathrm{s}\), \(T=37\,^{\circ}\mathrm{C}\).
  \item \textbf{High noise:} \(g=36\,\mathrm{dB}\), \(t=1\,\mathrm{ms}\), \(T=50\,^{\circ}\mathrm{C}\).
  \item \textbf{Thermal:} \(g=24\,\mathrm{dB}\), \(t=1\,\mathrm{s}\), \(T=70\,^{\circ}\mathrm{C}\).
\end{itemize}
As shown in \cref{fig:supp-sensor-test-conditions}, these presets span distinct degradation conditions: near-clean mosaics under low noise, heavy readout and shot noise under high noise, and an elevated dark-current baseline with hot pixels under thermal conditions. All test inputs are simulated as 12-bit RAW mosaics with unit white-balance gains \(\mathbf w_{\mathrm{wb}}=(1,1,1)\). Unless otherwise noted, each test scene is evaluated at three brightness levels relative to the nominal scale \(\alpha_0\) from \cref{sec:supp-exposure-sampling} (\(-1\), \(0\), and \(+1\)~EV, corresponding to \(\alpha\in\{0.5\alpha_0,\alpha_0,2\alpha_0\}\)) and averaged within each scene before aggregation, while the three-condition comparisons in \cref{sec:supp-degradation-protocol,sec:supp-sensor-information-availability} are evaluated at \(0\)~EV.

\paragraph{Stokes MSE.}
To compare errors across scene brightness levels, we normalize both the simulated Stokes target \(\mathbf s^\star\) and prediction \(\widehat{\mathbf s}\) by the known signal scale \(\alpha\) and by \(P_{99}(s_0)\), the 99th percentile of the unscaled scene's \(s_0\) across all pixels and RGB channels. For each component \(k\in\{0,1,2\}\), we compute the mean squared error over the scoring region \(\Omega_{\mathrm{eval}}\) and RGB channels, including saturated pixels:
\begin{equation}
  E_k
  =\frac{1}{3|\Omega_{\mathrm{eval}}|}
  \sum_{\mathbf u\in\Omega_{\mathrm{eval}}}
  \sum_{c\in\{R,G,B\}}
  \left(\frac{\widehat s_{k,c}(\mathbf u)-s^\star_{k,c}(\mathbf u)}{\alpha\,P_{99}(s_0)}\right)^{\!2}.
  \label{eq:supp-stokes-mse}
\end{equation}
We report Stokes MSE as \((E_0+E_1+E_2)/3\), and show the intensity component \(E_0\) (\(s_0\) MSE) on qualitative comparisons.

\paragraph{DoLP.}
To exclude dark pixels where polarization estimates are ill-conditioned, we restrict DoLP (\(\rho\)) evaluation to the set \(\mathcal D\) of pixel--channel pairs \(i=(\mathbf u,c)\) in \(\Omega_{\mathrm{eval}}\) whose normalized target intensity \(s^\star_{0,i}/(\alpha P_{99}(s_0))\) exceeds \(0.01\). Unconstrained models can occasionally produce extreme out-of-range DoLP values that dominate the average error, so we clip predicted DoLP to \([0,1]\) before computing its mean absolute error:
\begin{equation}
  E_{\mathrm{DoLP}}
  = \frac{1}{|\mathcal D|}\sum_{i\in\mathcal D}
    \left|\operatorname{clip}_{[0,1]}(\widehat\rho_i)-\rho_i\right|.
\end{equation}

\paragraph{AoLP.}
To exclude nearly unpolarized pixels, we restrict AoLP (\(\psi\)) evaluation to the subset \(\mathcal A=\{i\in\mathcal D:\rho_i\geq0.05\}\). With both angles expressed in degrees in \([0^\circ,180^\circ)\) to account for the \(180^\circ\) period, the mean absolute angular error is
\begin{equation}
  E_{\mathrm{AoLP}}
  = \frac{1}{|\mathcal A|}\sum_{i\in\mathcal A}
    \min\!\left(|\widehat\psi_i-\psi_i|,\,180^\circ-|\widehat\psi_i-\psi_i|\right).
\end{equation}

\paragraph{Nonphysical rate.}
We evaluate each RGB channel separately over the entire scoring region using the same criterion as Table~1 of the main paper. We count a predicted linear Stokes vector as physically valid if all three components are finite and satisfy
\begin{equation}
  \widehat s_0\geq 0
  \qquad\text{and}\qquad
  \sqrt{\widehat s_1^2+\widehat s_2^2}\leq(1+\varepsilon)\widehat s_0,
  \label{eq:supp-nonphysical-rate}
\end{equation}
where the relative tolerance \(\varepsilon=10^{-5}\) accounts for floating-point roundoff at the polarization boundary. The nonphysical rate is the percentage of predicted vectors that fail this validity test.

\paragraph{Aggregation.}
Each scene contributes one score per metric. For the main results, we report Stokes MSE as a 5\% trimmed mean across the 100 test scenes, removing the lowest and highest five scene scores for each model. HDR scenes span a wide dynamic range, and some models produce extreme errors on isolated test scenes, so trimming prevents a few outliers from dominating the overall comparison. DoLP error, AoLP error, and nonphysical rate use the arithmetic mean over all 100 scenes. Whenever we report subset or untrimmed averages, we label them explicitly.

\clearpage
\section{Effect of the training dataset}
\label{sec:benchmark}

We compare TCPDNet, PIDSR, and Raw2Stokes trained on each of the nine datasets. All 27 networks receive identical simulated CPFA RAW inputs from the 100 HDR-Polar1K test scenes, with Raw2Stokes evaluated in blind mode.

\subsection{Results across the nine training datasets}
For the main comparison, we evaluate under the low-noise preset defined in \cref{sec:supp-hdr-metrics}. \Cref{tab:supp-training-data-all-scenes} reproduces the four metrics from Figure~7 of the main paper alongside the untrimmed Stokes MSE. Across all three networks, training on HDR-Polar1K yields the lowest Stokes MSE, DoLP error, and AoLP error. The large untrimmed Stokes MSE for Raw2Stokes trained on the Rahman et al. dataset is driven by a single outlier test scene, which is why we also report the trimmed mean in \cref{sec:supp-hdr-metrics} so isolated outliers do not dominate the comparison. Since the physical output constraint bounds polarization relative to intensity rather than capping \(s_0\) itself, such intensity outliers still yield a zero nonphysical rate.
\begin{table}[htbp]
\centering\small
\setlength{\tabcolsep}{4pt}
\caption{\textbf{Effect of training data on reconstruction.} Bold marks the best value within each network (omitted when all rows tie).}
\label{tab:supp-training-data-all-scenes}
\begin{tabular}{@{}llrrrrr@{}}
\toprule
& & \multicolumn{2}{c}{Stokes MSE ($10^{-4}$) $\downarrow$} & & & \\
\cmidrule(lr){3-4}
Network & Training dataset & Trimmed & All scenes & AoLP ($^\circ$) $\downarrow$ & DoLP $\downarrow$ & Nonphysical rate (\%) $\downarrow$ \\
\midrule
TCPDNet & Wen et al. (2019)~\cite{wen2019cpdnet} & 13.0 & 42.5 & 13.5 & 0.053 & 1.57 \\
 & Morimatsu et al.~\cite{morimatsu2020eari} & 12.6 & 42.4 & 13.3 & 0.049 & 0.61 \\
 & Qiu et al.~\cite{qiu2021linear} & 18.8 & 48.5 & 13.7 & 0.051 & 0.42 \\
 & Wen et al. (2021)~\cite{wen2021sparse} & 8.4 & 36.8 & 11.7 & 0.041 & 0.53 \\
 & Zhou et al.~\cite{zhou2025pidsr} & 11.8 & 40.9 & 12.1 & 0.050 & 1.21 \\
 & Rahman et al.~\cite{rahman2025polarization} & 18.5 & 56.8 & 15.0 & 0.080 & 0.58 \\
 & Luo et al.~\cite{luo2026updem} & 10.9 & 39.7 & 11.6 & 0.044 & 0.57 \\
 & Wu et al.~\cite{wu2026deep} & 10.4 & 39.3 & 13.2 & 0.046 & 0.58 \\
 & HDR-Polar1K (ours) & \textbf{2.4} & \textbf{23.0} & \textbf{7.2} & \textbf{0.024} & \textbf{0.20} \\
\midrule
PIDSR & Wen et al. (2019)~\cite{wen2019cpdnet} & 11.6 & 45.5 & 14.6 & 0.059 & 2.26 \\
 & Morimatsu et al.~\cite{morimatsu2020eari} & 14.3 & 43.0 & 12.9 & 0.060 & 3.55 \\
 & Qiu et al.~\cite{qiu2021linear} & 15.3 & 46.9 & 11.5 & 0.065 & 3.32 \\
 & Wen et al. (2021)~\cite{wen2021sparse} & 6.9 & 35.2 & 10.2 & 0.041 & 1.06 \\
 & Zhou et al.~\cite{zhou2025pidsr} & 11.9 & 41.1 & 11.3 & 0.049 & 1.81 \\
 & Rahman et al.~\cite{rahman2025polarization} & 15.7 & 48.4 & 12.5 & 0.067 & 3.79 \\
 & Luo et al.~\cite{luo2026updem} & 13.5 & 41.8 & 11.9 & 0.059 & 2.91 \\
 & Wu et al.~\cite{wu2026deep} & 10.1 & 38.3 & 11.2 & 0.055 & 2.02 \\
 & HDR-Polar1K (ours) & \textbf{2.2} & \textbf{21.4} & \textbf{6.3} & \textbf{0.024} & \textbf{0.61} \\
\midrule
Raw2Stokes (ours) & Wen et al. (2019)~\cite{wen2019cpdnet} & 12.4 & 46.5 & 11.2 & 0.051 & 0 \\
 & Morimatsu et al.~\cite{morimatsu2020eari} & 12.8 & \(4.5\times10^{8}\) & 9.7 & 0.056 & 0 \\
 & Qiu et al.~\cite{qiu2021linear} & 15.9 & 44.9 & 10.3 & 0.048 & 0 \\
 & Wen et al. (2021)~\cite{wen2021sparse} & 5.4 & 34.7 & 7.0 & 0.030 & 0 \\
 & Zhou et al.~\cite{zhou2025pidsr} & 9.1 & 36.5 & 7.6 & 0.036 & 0 \\
 & Rahman et al.~\cite{rahman2025polarization} & 21.7 & \(2.7\times10^{18}\) & 10.0 & 0.063 & 0 \\
 & Luo et al.~\cite{luo2026updem} & 9.0 & 35.7 & 9.0 & 0.038 & 0 \\
 & Wu et al.~\cite{wu2026deep} & 7.2 & 34.1 & 8.2 & 0.034 & 0 \\
 & HDR-Polar1K (ours) & \textbf{1.4} & \textbf{17.6} & \textbf{4.5} & \textbf{0.017} & 0 \\
\bottomrule
\end{tabular}
\end{table}

\FloatBarrier

\clearpage
\subsection{Breakdown by test-scene category}
\label{sec:supp-training-data-breakdowns}

To verify that HDR-Polar1K improves performance across both everyday photography and strongly polarized laboratory scenes, we separate the 82 everyday and 18 controlled test scenes in \cref{tab:supp-training-data-categories}. For every network, training on HDR-Polar1K yields the lowest Stokes MSE, DoLP error, and AoLP error in both categories. For Raw2Stokes trained on Morimatsu et al. and Rahman et al., the large mean Stokes MSEs arise from extreme errors on a few controlled scenes.
\begin{table}[htbp]
\centering\small
\setlength{\tabcolsep}{3pt}
\caption{\textbf{Reconstruction by test-scene category.} Untrimmed means over 82 everyday and 18 controlled scenes. Bold marks the best value within each network.}
\label{tab:supp-training-data-categories}
\begin{tabular}{@{}llrrr@{\hspace{12pt}}rrr@{}}
\toprule
& & \multicolumn{3}{c@{\hspace{12pt}}}{Everyday test scenes (82)} & \multicolumn{3}{c@{}}{Controlled test scenes (18)} \\
\cmidrule(r{12pt}){3-5}\cmidrule(){6-8}
Network & Training dataset & Stokes MSE ($10^{-4}$) $\downarrow$ & AoLP ($^\circ$) $\downarrow$ & DoLP $\downarrow$ & Stokes MSE ($10^{-4}$) $\downarrow$ & AoLP ($^\circ$) $\downarrow$ & DoLP $\downarrow$ \\
\midrule
TCPDNet & Wen et al. (2019)~\cite{wen2019cpdnet} & 42.4 & 14.2 & 0.046 & 43.0 & 10.5 & 0.087 \\
 & Morimatsu et al.~\cite{morimatsu2020eari} & 41.6 & 14.1 & 0.041 & 45.8 & 10.1 & 0.081 \\
 & Qiu et al.~\cite{qiu2021linear} & 51.4 & 14.4 & 0.047 & 35.4 & 10.4 & 0.072 \\
 & Wen et al. (2021)~\cite{wen2021sparse} & 39.3 & 12.4 & 0.037 & 25.6 & 8.7 & 0.060 \\
 & Zhou et al.~\cite{zhou2025pidsr} & 43.4 & 12.8 & 0.047 & 29.6 & 8.6 & 0.067 \\
 & Rahman et al.~\cite{rahman2025polarization} & 45.6 & 15.6 & 0.057 & 107.8 & 12.3 & 0.185 \\
 & Luo et al.~\cite{luo2026updem} & 42.9 & 12.3 & 0.042 & 25.0 & 8.4 & 0.055 \\
 & Wu et al.~\cite{wu2026deep} & 42.4 & 14.0 & 0.043 & 25.7 & 9.2 & 0.058 \\
 & HDR-Polar1K (ours) & \textbf{26.4} & \textbf{7.7} & \textbf{0.023} & \textbf{7.5} & \textbf{5.2} & \textbf{0.030} \\
\midrule
PIDSR & Wen et al. (2019)~\cite{wen2019cpdnet} & 40.5 & 15.1 & 0.052 & 68.4 & 12.7 & 0.091 \\
 & Morimatsu et al.~\cite{morimatsu2020eari} & 40.2 & 13.2 & 0.052 & 55.6 & 11.3 & 0.098 \\
 & Qiu et al.~\cite{qiu2021linear} & 46.4 & 11.7 & 0.059 & 49.4 & 10.6 & 0.095 \\
 & Wen et al. (2021)~\cite{wen2021sparse} & 36.9 & 10.3 & 0.035 & 27.7 & 10.0 & 0.066 \\
 & Zhou et al.~\cite{zhou2025pidsr} & 42.5 & 11.5 & 0.043 & 34.5 & 10.5 & 0.076 \\
 & Rahman et al.~\cite{rahman2025polarization} & 43.1 & 12.6 & 0.056 & 72.6 & 11.8 & 0.114 \\
 & Luo et al.~\cite{luo2026updem} & 41.9 & 12.3 & 0.053 & 41.5 & 10.1 & 0.091 \\
 & Wu et al.~\cite{wu2026deep} & 40.0 & 11.4 & 0.049 & 30.3 & 10.1 & 0.085 \\
 & HDR-Polar1K (ours) & \textbf{24.5} & \textbf{6.6} & \textbf{0.021} & \textbf{7.2} & \textbf{5.3} & \textbf{0.034} \\
\midrule
Raw2Stokes (ours) & Wen et al. (2019)~\cite{wen2019cpdnet} & 42.6 & 11.5 & 0.038 & 64.2 & 9.9 & 0.107 \\
 & Morimatsu et al.~\cite{morimatsu2020eari} & 40.5 & 9.2 & 0.034 & \(2.5\times10^{9}\) & 11.6 & 0.158 \\
 & Qiu et al.~\cite{qiu2021linear} & 42.9 & 10.5 & 0.038 & 53.9 & 9.3 & 0.092 \\
 & Wen et al. (2021)~\cite{wen2021sparse} & 34.7 & 6.9 & 0.023 & 34.8 & 7.4 & 0.059 \\
 & Zhou et al.~\cite{zhou2025pidsr} & 38.9 & 7.8 & 0.028 & 25.6 & 7.0 & 0.069 \\
 & Rahman et al.~\cite{rahman2025polarization} & 43.4 & 9.1 & 0.039 & \(1.5\times10^{19}\) & 14.1 & 0.176 \\
 & Luo et al.~\cite{luo2026updem} & 37.2 & 9.2 & 0.032 & 28.8 & 7.9 & 0.065 \\
 & Wu et al.~\cite{wu2026deep} & 37.0 & 8.3 & 0.029 & 21.2 & 7.7 & 0.056 \\
 & HDR-Polar1K (ours) & \textbf{20.6} & \textbf{4.7} & \textbf{0.016} & \textbf{4.0} & \textbf{3.5} & \textbf{0.023} \\
\bottomrule
\end{tabular}
\end{table}

\clearpage
\subsection{Effect of training set size}
\label{sec:supp-training-set-size}

\begin{figure}[!htbp]
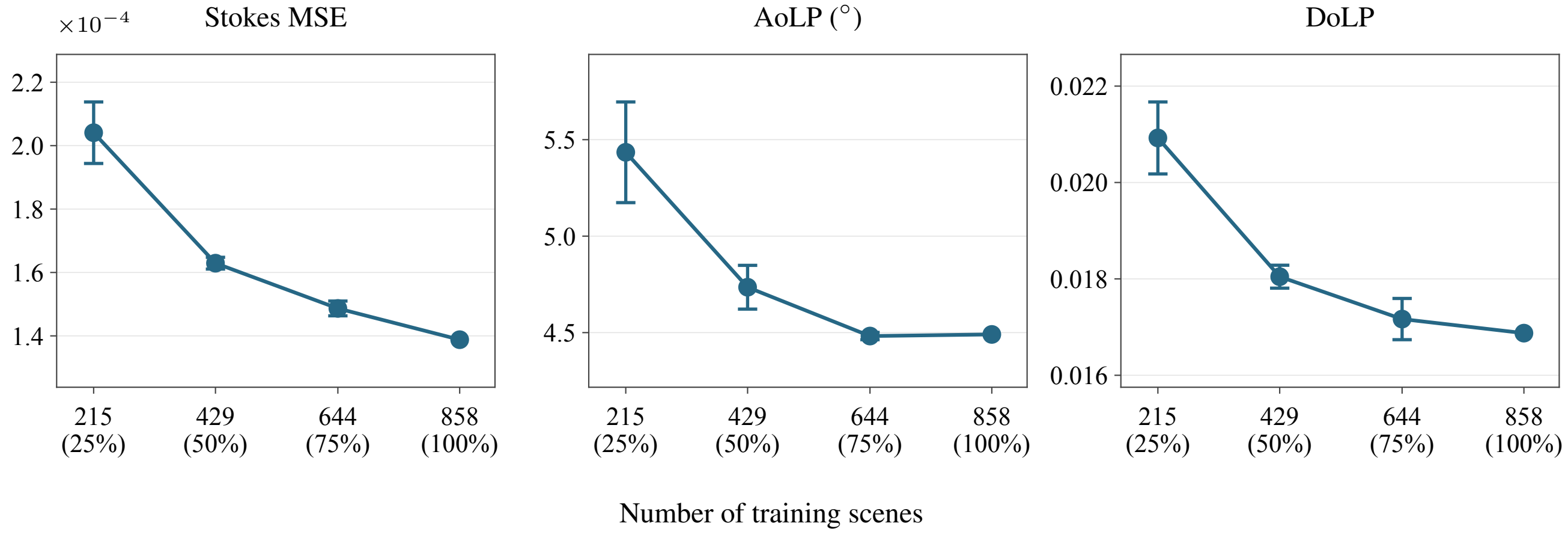

  \centering
  \newcommand{\trainingSizePanel}[3][]{%
    \begin{minipage}[t]{.329\linewidth}
      \noindent\rlap{\begin{minipage}[t]{\linewidth}
        \noindent\hspace*{.15\linewidth}\makebox[.84\linewidth]{\small\strut\raisebox{-1mm}[0pt][0pt]{#2}}\par
        \noindent\csname ArxivImage:figures/supp-training-set-size/#3.pdf\endcsname{width=\linewidth}
      \end{minipage}}%
      \hspace*{.15\linewidth}\raisebox{-1.6mm}[0pt][0pt]{\scriptsize #1}%
    \end{minipage}}
  \trainingSizePanel[\(\times 10^{-4}\)]{Stokes MSE}{mse}\hfill%
  \trainingSizePanel{AoLP (\(^{\circ}\))}{mae_aolp_degrees}\hfill%
  \trainingSizePanel{DoLP}{mae_dolp_clipped}\par
  {\small Number of training scenes\par}
  \caption{\textbf{Effect of training set size.} Scaling the HDR-Polar1K training set from 25\% to 100\% steadily reduces Stokes MSE and DoLP error, while AoLP error changes little beyond 75\%. Points show the mean and \(\pm 1\) standard deviation over three random subsets.}
  \label{fig:supp-training-set-size}
\end{figure}

To assess how dataset scale affects reconstruction accuracy, we compare Raw2Stokes trained on all 858 HDR-Polar1K training scenes with models trained on nested random subsets containing 25\% (215), 50\% (429), or 75\% (644) of those scenes across three random draws, holding validation and test sets fixed. As shown in \cref{fig:supp-training-set-size}, Stokes MSE and DoLP error decrease steadily as the training set grows, confirming the value of collecting a larger training set, while AoLP error changes little beyond 75\% of the training set.

\FloatBarrier
\clearpage
\subsection{Effect of controlled scenes in the training set}
\label{sec:supp-scene-composition}

To isolate the contribution of controlled polarization scenes at a fixed dataset size of 662 scenes, we compare training on all 662 everyday scenes against mixed sets containing 196 controlled scenes plus 466 everyday scenes sampled without replacement across three independent trials, under identical training and validation settings.

\begin{table}[!htbp]
  \centering
  \small
  \setlength{\tabcolsep}{4.5pt}
  \renewcommand{\arraystretch}{0.94}
  \caption{\textbf{Effect of controlled training scenes.} Untrimmed category means under the low-noise preset, with everyday + controlled averaged over three random draws. Bold marks the best value in each column.}
  \label{tab:scene-composition}
  \begin{tabular}{@{}lrrr@{\hspace{12pt}}rrr@{}}
    \toprule
    & \multicolumn{3}{c@{\hspace{12pt}}}{Everyday test scenes (82)} & \multicolumn{3}{c@{}}{Controlled test scenes (18)} \\
    \cmidrule(r{12pt}){2-4}\cmidrule(){5-7}
    Training scenes & Stokes MSE ($10^{-4}$) $\downarrow$ & AoLP ($^\circ$) $\downarrow$ & DoLP $\downarrow$ & Stokes MSE ($10^{-4}$) $\downarrow$ & AoLP ($^\circ$) $\downarrow$ & DoLP $\downarrow$ \\
    \midrule
    Everyday only & 21.62 & \textbf{4.63} & \textbf{0.0156} & 8.75 & 4.83 & 0.0331 \\
    Everyday + controlled & \textbf{21.61} & 4.81 & 0.0159 & \textbf{4.32} & \textbf{3.53} & \textbf{0.0235} \\
    \bottomrule
  \end{tabular}
\end{table}

\begin{figure}[!htbp]
  \centering
  \setlength{\tabcolsep}{2pt}
  \newcommand{\compositionPanel}[1]{\raisebox{-.5\height}{#1}}
  \newcommand{\compositionLegend}[2]{%
    \compositionPanel{\rotatebox{90}{#1}}\,%
    \makebox[.8em][l]{%
      \rlap{\raisebox{\dimexpr.078\linewidth-1ex\relax}{\(\scriptstyle #2\)}}%
      \raisebox{-.078\linewidth}{\(\scriptstyle 0\)}}}
  \begin{tabular}{@{}ccccc@{}}
    & \small Everyday only & \small Everyday + controlled & \small GT & \\[2pt]
    \compositionPanel{\rotatebox{90}{\small \(s_0\)}}
      & \compositionPanel{\suppScoredPanel{\includegraphics[width=.27\linewidth]{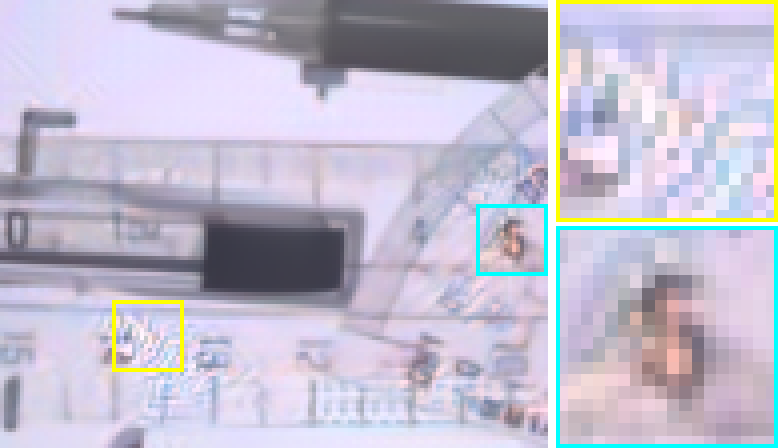}}{0.70437}{18.41}}
      & \compositionPanel{\suppScoredPanel{\includegraphics[width=.27\linewidth]{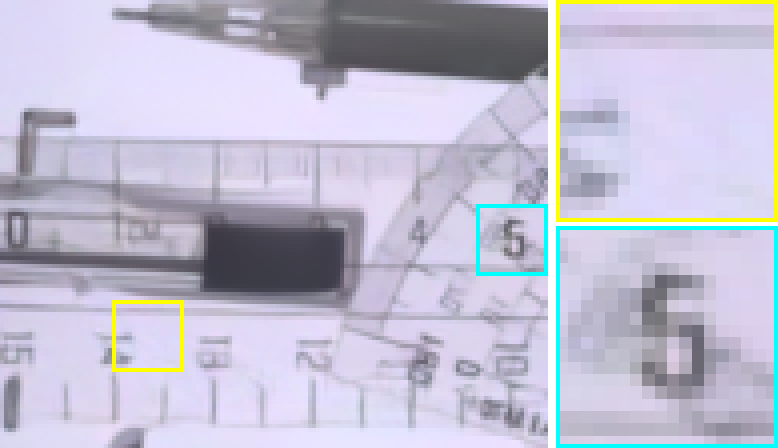}}{0.70437}{7.83}}
      & \compositionPanel{\includegraphics[width=.27\linewidth]{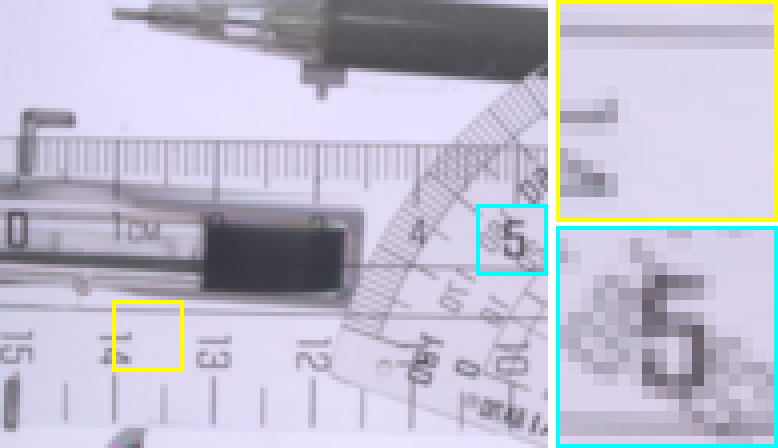}} & \\
    \noalign{\vskip 2pt}
    \compositionPanel{\rotatebox{90}{\small AoLP}}
      & \compositionPanel{\suppScoredPanel{\includegraphics[width=.27\linewidth]{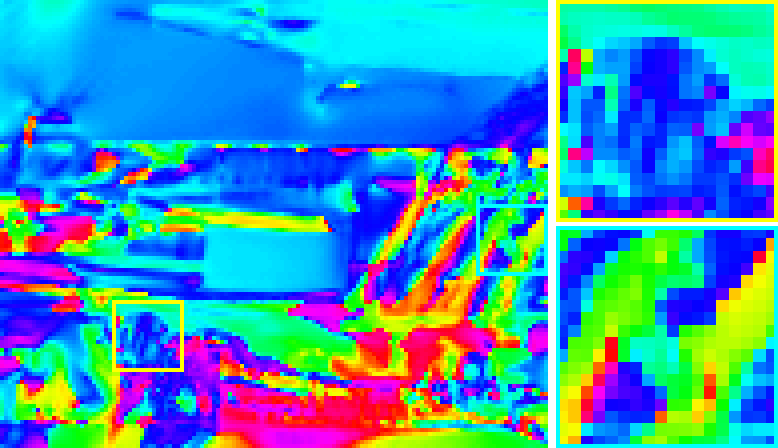}}{0.70437}{13.38^\circ}}
      & \compositionPanel{\suppScoredPanel{\includegraphics[width=.27\linewidth]{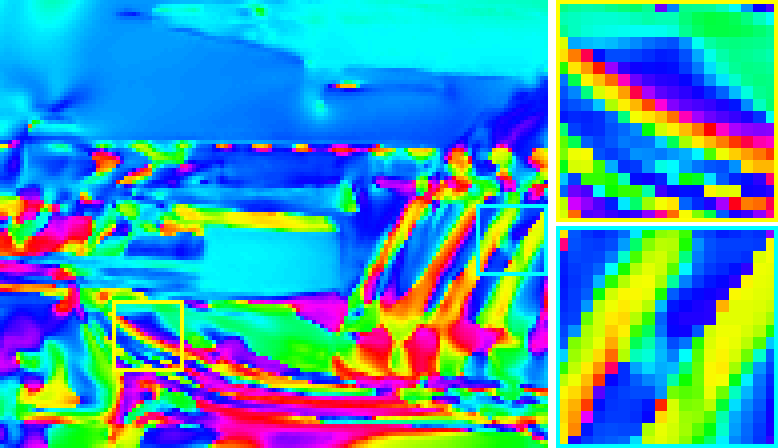}}{0.70437}{8.59^\circ}}
      & \compositionPanel{\includegraphics[width=.27\linewidth]{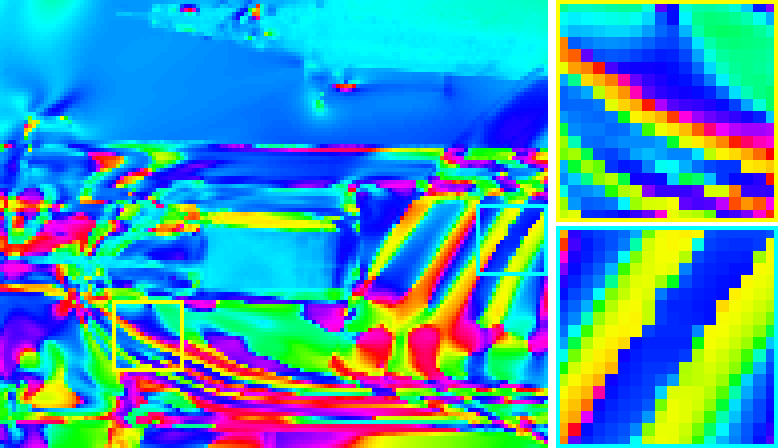}}
      & \compositionLegend{\includegraphics[page=74,width=.156\linewidth,height=.9mm]{figures/training-data-results-panels.pdf}}{\pi} \\
    \noalign{\vskip 2pt}
    \compositionPanel{\rotatebox{90}{\small DoLP}}
      & \compositionPanel{\suppScoredPanel{\includegraphics[width=.27\linewidth]{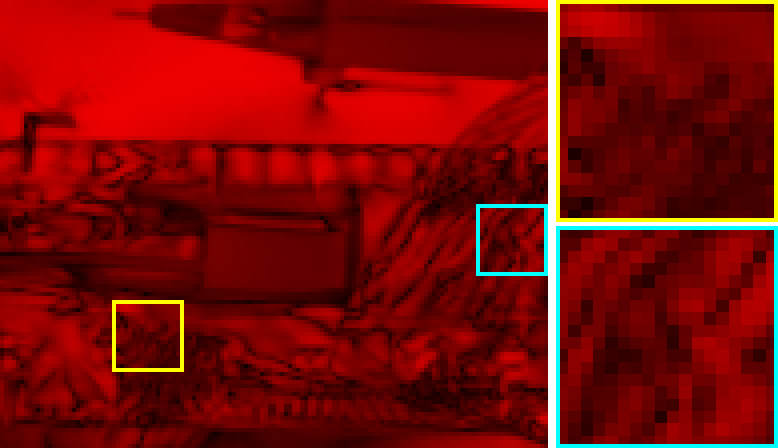}}{0.70437}{0.076}}
      & \compositionPanel{\suppScoredPanel{\includegraphics[width=.27\linewidth]{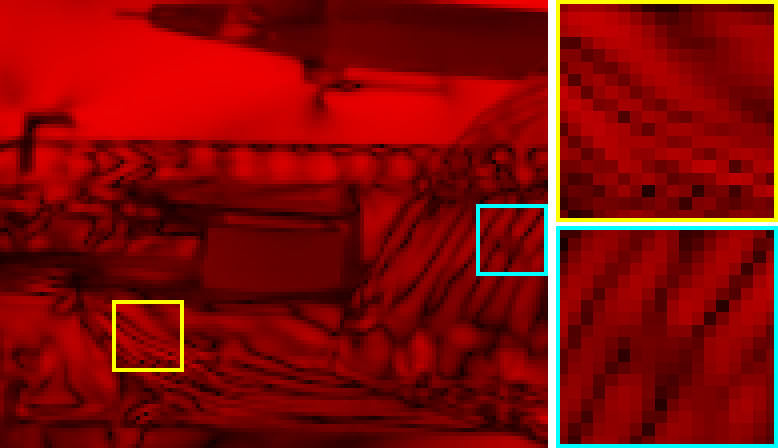}}{0.70437}{0.049}}
      & \compositionPanel{\includegraphics[width=.27\linewidth]{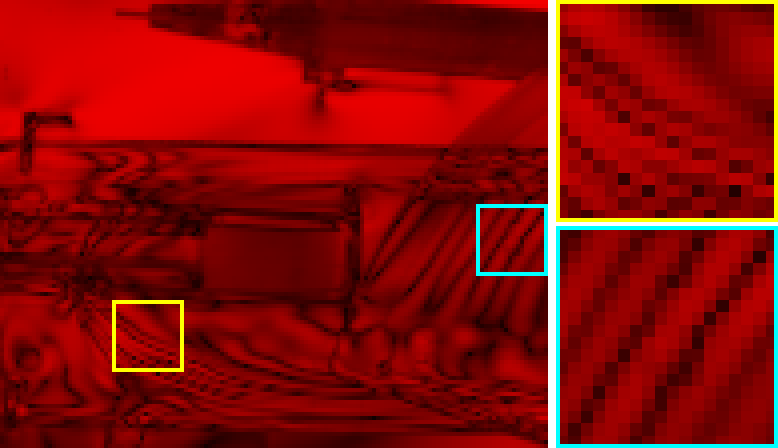}}
      & \compositionLegend{\includegraphics[page=43,width=.156\linewidth,height=.9mm]{figures/training-data-results-panels.pdf}}{1} \\
  \end{tabular}
  \caption{\textbf{Effect of controlled training scenes on photoelastic fringe reconstruction.} Everyday-only training blurs the diagonal stress fringes in AoLP and DoLP, whereas adding 196 controlled scenes at the same 662-scene budget recovers sharper fringes matching the ground truth. Bottom-right numbers report \(s_0\) MSE (\(10^{-4}\)), AoLP error (\(^\circ\)), and DoLP error.}
  \label{fig:supp-composition-photoelastic}
\end{figure}

As shown in \cref{tab:scene-composition}, adding controlled training scenes roughly halves Stokes MSE and substantially lowers both AoLP and DoLP errors on controlled test scenes. On everyday scenes, Stokes MSE is nearly identical with only a minor trade-off in polarization error. In the enlarged photoelastic crops in \cref{fig:supp-composition-photoelastic}, everyday-only training blurs the diagonal stress fringes, whereas mixed training recovers sharper fringes that closely match the ground truth. These results confirm that including controlled scenes in HDR-Polar1K substantially improves reconstruction of strongly polarized structures without degrading everyday scene quality.

\clearpage

\subsection{Effect of training image resolution}
\label{sec:supp-zhou-training-resolution}

Unlike the HDR-Polar1K analyses in \cref{sec:supp-training-set-size,sec:supp-scene-composition}, this experiment uses only the Zhou et al.~\cite{zhou2025pidsr} dataset to examine why models trained on its original images perform differently from the official pretrained PIDSR model evaluated later in \cref{tab:supp-method-conditions}. PIDSR trains its demosaicing stage jointly with \(2\mathord{\times}\) super-resolution by downsampling the original Zhou et al. images to half resolution as demosaicing targets. To isolate the effect of this downsampling, we train a separate matched pair of Raw2Stokes models on the same Zhou et al. scenes at original and half resolution using \(64\mathord{\times}64\) patches, averaging \(2\mathord{\times}2\) blocks separately within each color and polarizer orientation for the half-resolution set before simulating CPFA measurements.

As shown in \cref{tab:supp-zhou-training-resolution}, half-resolution training substantially reduces Stokes MSE and lowers both AoLP and DoLP errors. In \cref{fig:supp-zhou-training-resolution}, it clearly reduces color fringing on the printed text and cleans up the AoLP and DoLP maps, bringing the reconstruction much closer to the official pretrained PIDSR output. Although the official PIDSR model also differs in architecture, joint super-resolution training, noise assumptions, and loss function, this gap indicates that spatial averaging suppresses slight inter-polarizer misalignments and artifacts in the original training images, reinforcing how strongly training-data quality affects reconstruction accuracy.

\newcommand{\zhouResolutionRows}{%
Original & $2448\times2048$ & 9.8 & 9.1 & 0.038 \\%
Half & $1224\times1024$ & \textbf{6.8} & \textbf{7.4} & \textbf{0.033} \\%
}

\begin{table}[!htbp]
  \centering
  \small
  \setlength{\tabcolsep}{5pt}
  \caption{\textbf{Effect of training image resolution.} Bold marks the best value in each column.}
  \label{tab:supp-zhou-training-resolution}
  \begin{tabular}{@{}llrrr@{}}
    \toprule
    Training resolution & Image size & Stokes MSE (\(10^{-4}\)) \(\downarrow\) & AoLP (\(^{\circ}\)) \(\downarrow\) & DoLP \(\downarrow\) \\
    \midrule
    \zhouResolutionRows
    \bottomrule
  \end{tabular}
\end{table}

\begin{figure}[!htbp]
  \centering
  \setlength{\tabcolsep}{2pt}
  \newcommand{\zhouPanel}[1]{\raisebox{-.5\height}{#1}}
  \newcommand{\zhouLegend}[2]{%
    \zhouPanel{\rotatebox{90}{#1}}\,%
    \makebox[.8em][l]{%
      \rlap{\raisebox{\dimexpr.0615\linewidth-1ex\relax}{\(\scriptstyle #2\)}}%
      \raisebox{-.0615\linewidth}{\(\scriptstyle 0\)}}}
  \begin{tabular}{@{}cccccc@{}}
    \noalign{\vskip 9pt}
    & \multicolumn{2}{c}{\small Raw2Stokes (trained on Zhou et al.)} & & & \\
    \cmidrule(lr){2-3}
    \noalign{\vskip -9pt}
    & \small Original resolution & \small Half resolution & \small\shortstack{PIDSR\\(official weights)} & \small GT & \\[2pt]
    \zhouPanel{\rotatebox{90}{\small \(s_0\)}}
      & \zhouPanel{\suppScoredPanel{\includegraphics[width=.214\linewidth]{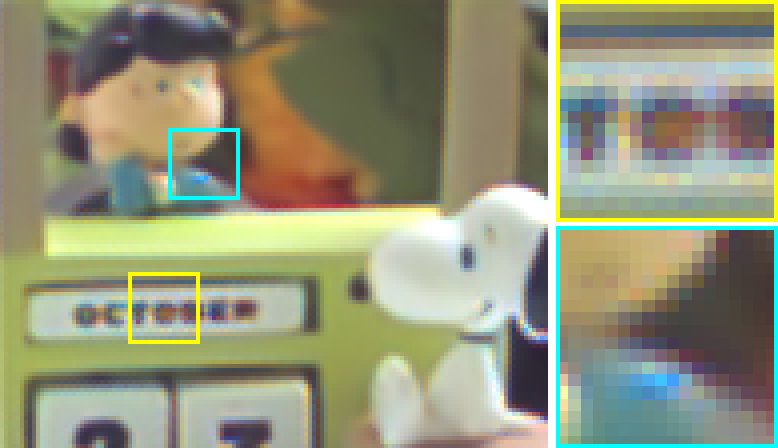}}{0.70437}{6.89}}
      & \zhouPanel{\suppScoredPanel{\includegraphics[width=.214\linewidth]{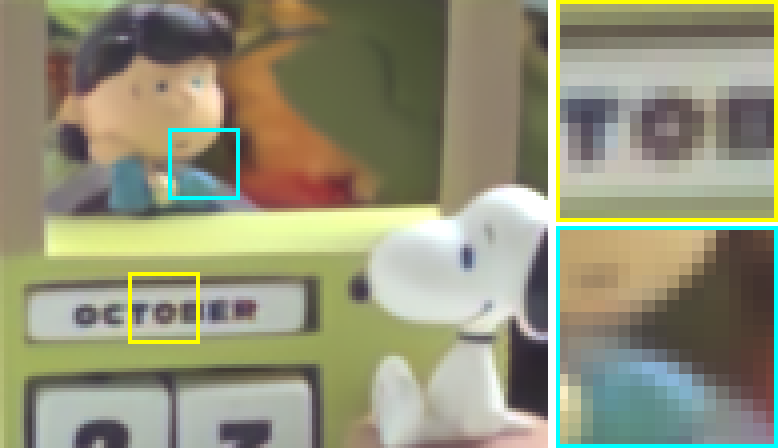}}{0.70437}{2.40}}
      & \zhouPanel{\suppScoredPanel{\includegraphics[width=.214\linewidth]{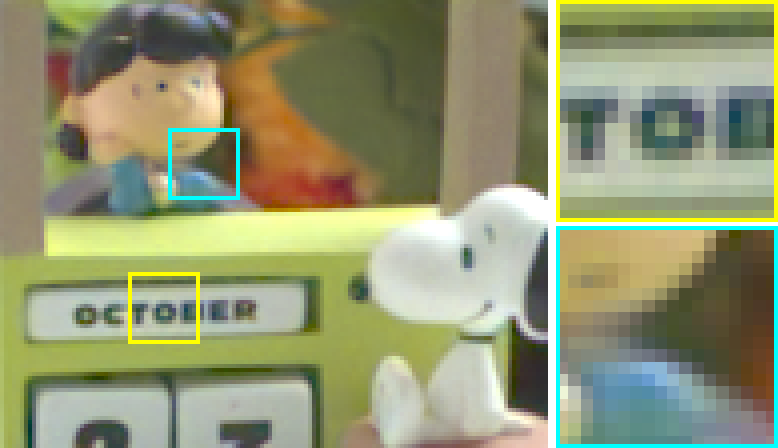}}{0.70437}{1.70}}
      & \zhouPanel{\includegraphics[width=.214\linewidth]{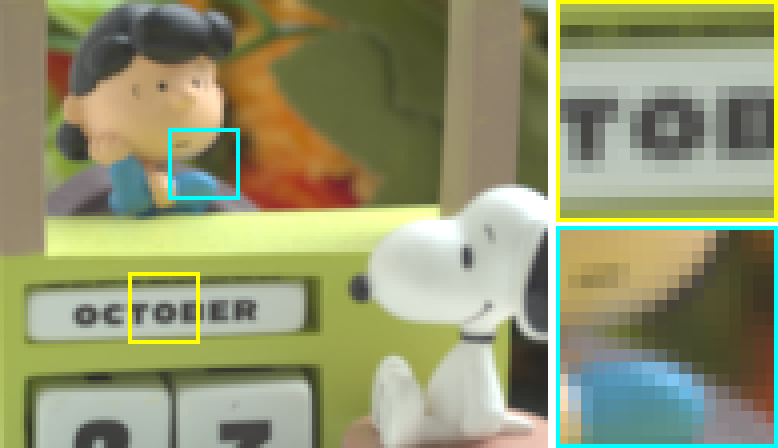}} & \\
    \noalign{\vskip 2pt}
    \zhouPanel{\rotatebox{90}{\small AoLP}}
      & \zhouPanel{\suppScoredPanel{\includegraphics[width=.214\linewidth]{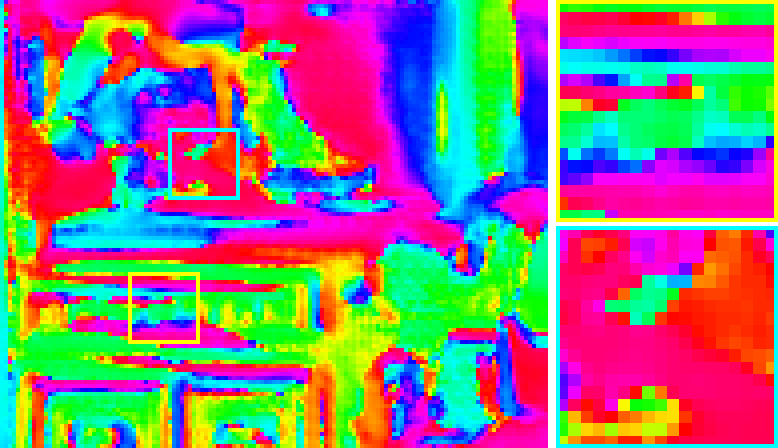}}{0.70437}{15.70^\circ}}
      & \zhouPanel{\suppScoredPanel{\includegraphics[width=.214\linewidth]{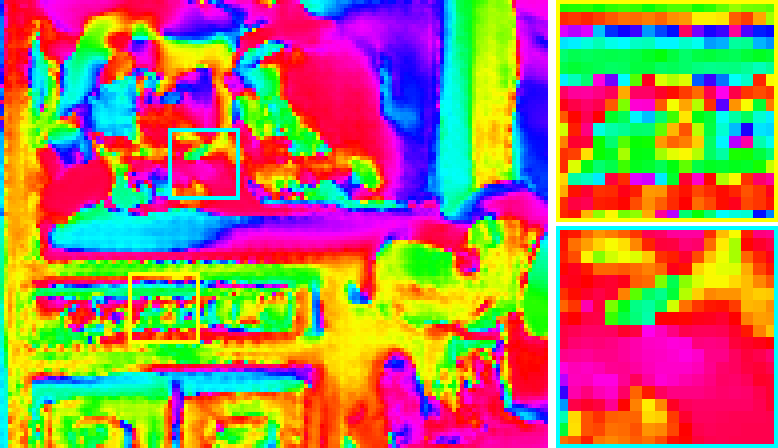}}{0.70437}{10.83^\circ}}
      & \zhouPanel{\suppScoredPanel{\includegraphics[width=.214\linewidth]{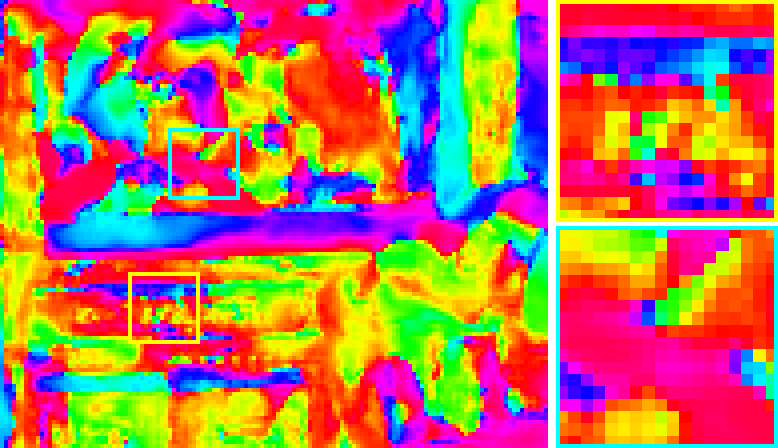}}{0.70437}{9.73^\circ}}
      & \zhouPanel{\includegraphics[width=.214\linewidth]{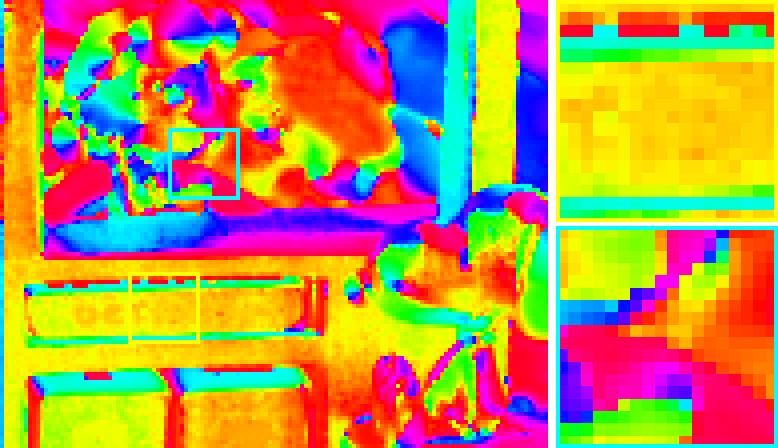}}
      & \zhouLegend{\includegraphics[page=74,width=.123\linewidth,height=.9mm]{figures/training-data-results-panels.pdf}}{\pi} \\
    \noalign{\vskip 2pt}
    \zhouPanel{\rotatebox{90}{\small DoLP}}
      & \zhouPanel{\suppScoredPanel{\includegraphics[width=.214\linewidth]{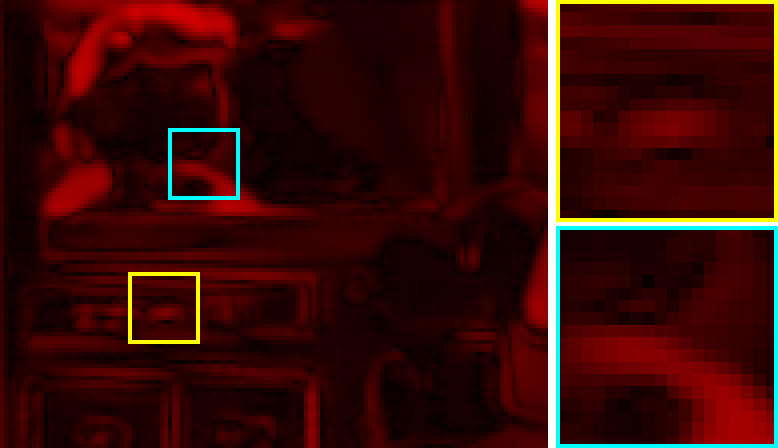}}{0.70437}{0.026}}
      & \zhouPanel{\suppScoredPanel{\includegraphics[width=.214\linewidth]{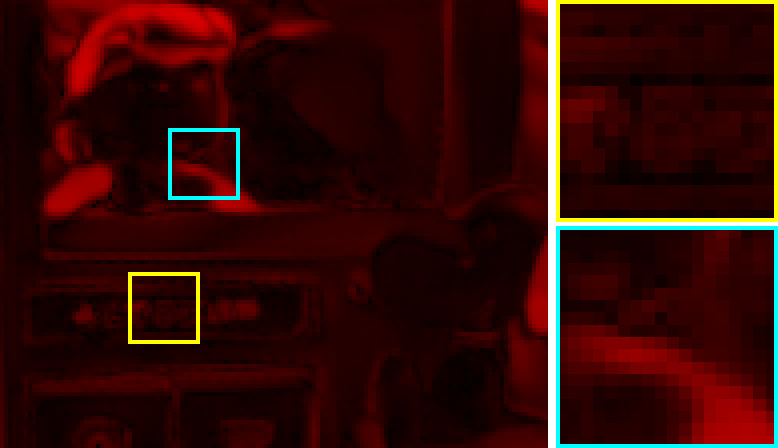}}{0.70437}{0.020}}
      & \zhouPanel{\suppScoredPanel{\includegraphics[width=.214\linewidth]{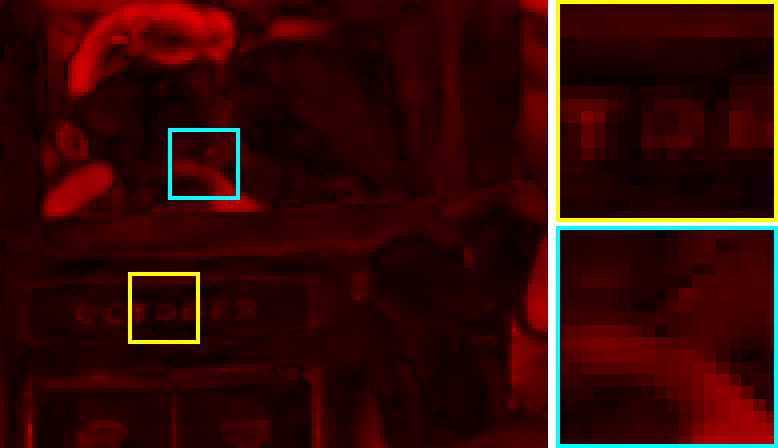}}{0.70437}{0.021}}
      & \zhouPanel{\includegraphics[width=.214\linewidth]{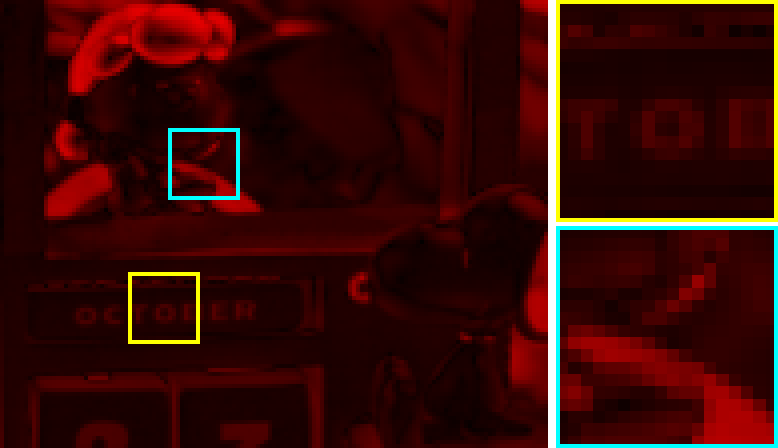}}
      & \zhouLegend{\includegraphics[page=43,width=.123\linewidth,height=.9mm]{figures/training-data-results-panels.pdf}}{1} \\
  \end{tabular}
  \caption{\textbf{Effect of training image resolution.} Comparing Raw2Stokes trained on Zhou et al. images at original versus half (\(2\mathord{\times}2\)-averaged) resolution alongside the official pretrained PIDSR model. Half-resolution training clearly reduces color fringing on the printed lettering and suppresses polarization artifacts, bringing the output much closer to pretrained PIDSR. Bottom-right numbers report \(s_0\) MSE (\(10^{-4}\)), AoLP error (\(^\circ\)), and DoLP error.}
  \label{fig:supp-zhou-training-resolution}
\end{figure}

\FloatBarrier

\subsection{Visual comparison on test scenes}
\label{sec:supp-reconstruction-test-scenes}

\begin{figure}[p]
  \centering
  \input{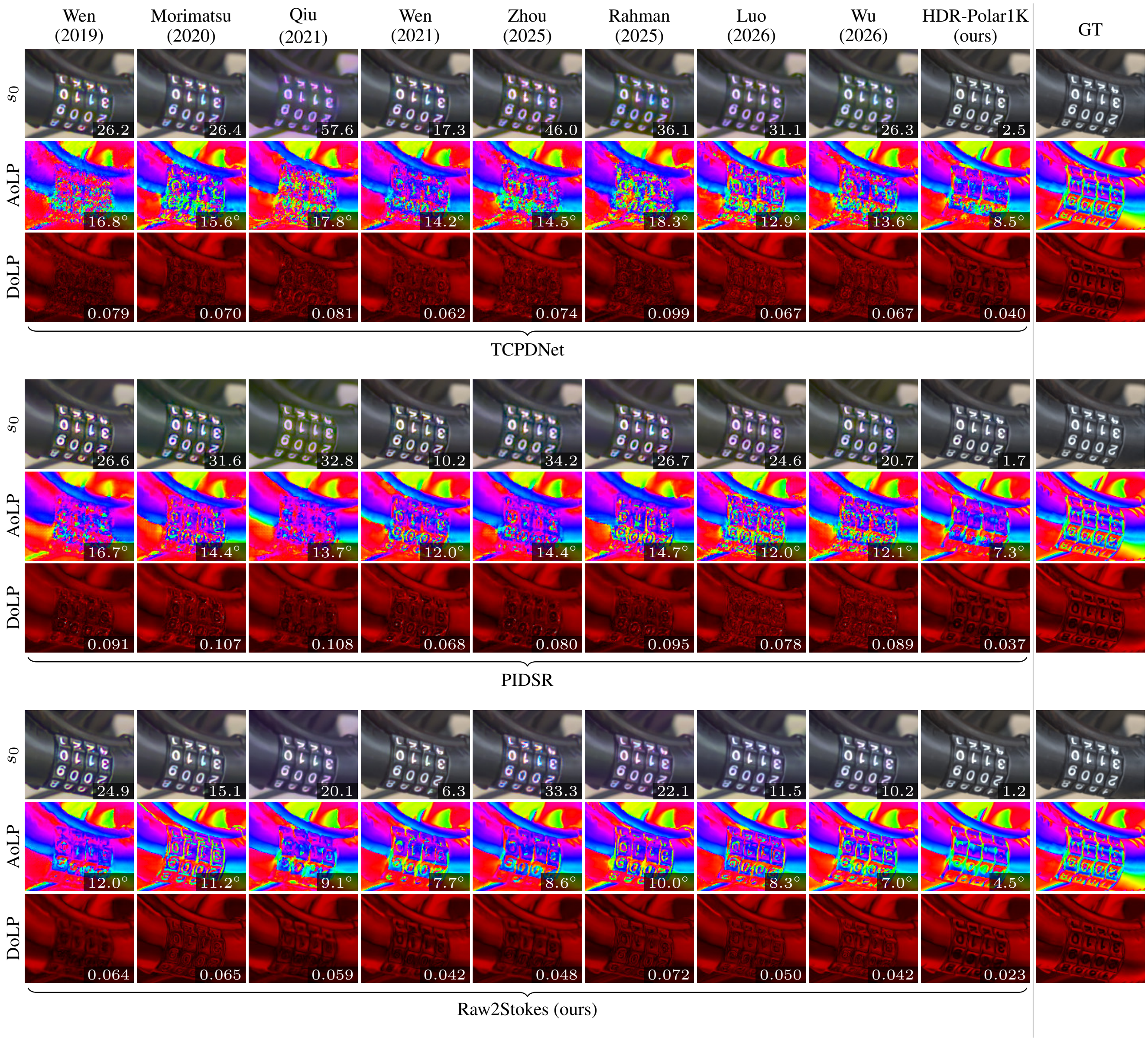}
  \caption{\textbf{Visual comparison across training datasets: everyday scene.} Bottom-right numbers report \(s_0\) MSE (\(10^{-4}\)), AoLP error (\(^\circ\)), and DoLP error.}
  \label{fig:supp-everyday-full}
\end{figure}

\Cref{fig:supp-everyday-full,fig:supp-controlled-full,fig:supp-metal-full,fig:supp-equipment-full,fig:supp-stripes-full,fig:supp-photoelastic-full} compare the nine training datasets across all three networks on six representative HDR-Polar1K test scenes. Each figure uses the same full-frame low-noise RAW input across all reconstructions. Numbers at the bottom right of each image report \(s_0\) MSE in units of \(10^{-4}\), AoLP error in degrees, and DoLP error.

\clearpage
\clearpage

\begin{figure}[p]
  \centering
  \input{figures/supp-controlled-full-grid}
  \caption{\textbf{Visual comparison across training datasets: controlled scene.} Bottom-right numbers report \(s_0\) MSE (\(10^{-4}\)), AoLP error (\(^\circ\)), and DoLP error.}
  \label{fig:supp-controlled-full}
\end{figure}
\clearpage

\begin{figure}[p]
  \centering
  \input{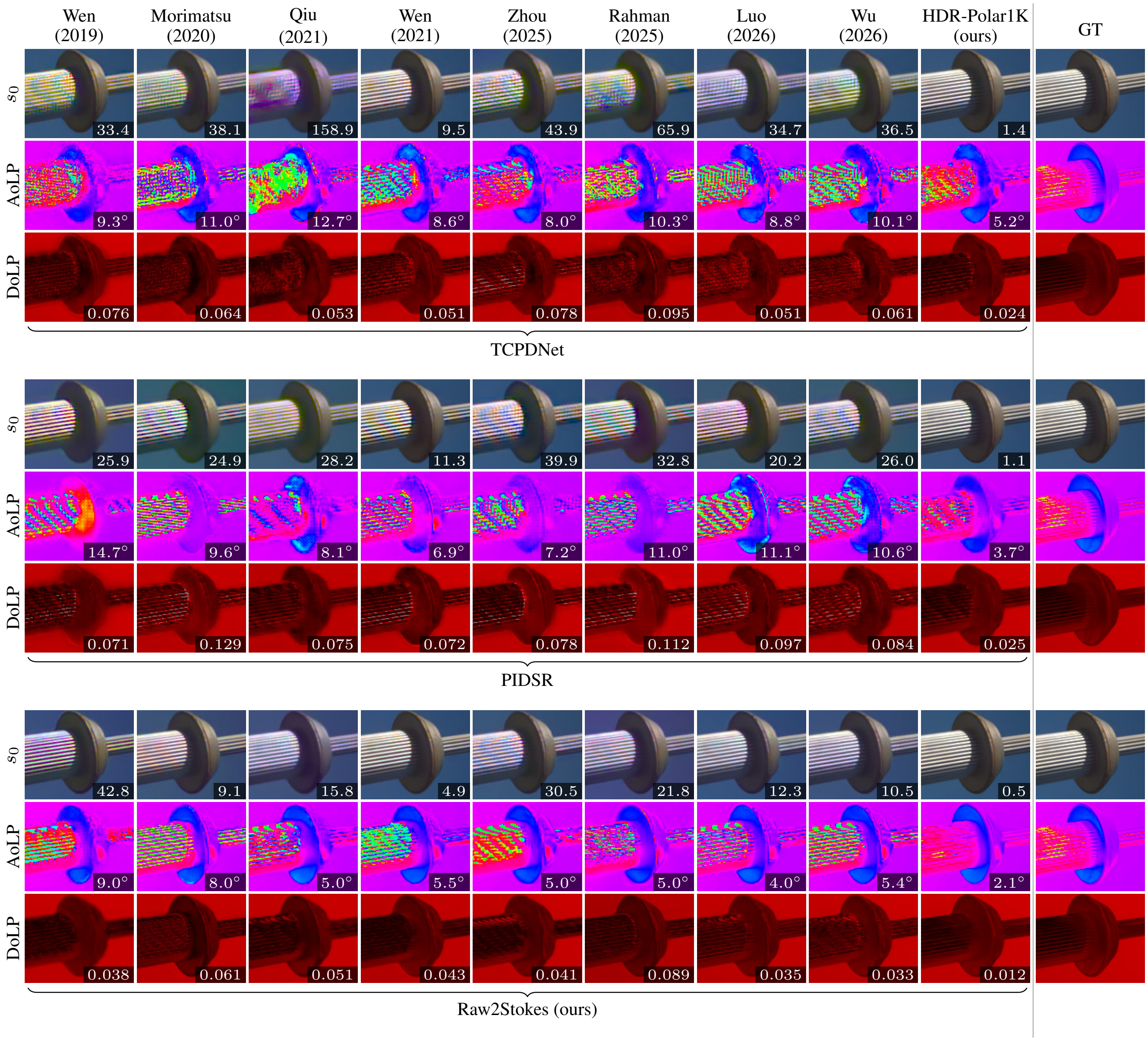}
  \caption{\textbf{Visual comparison across training datasets: metallic surface scene.} Bottom-right numbers report \(s_0\) MSE (\(10^{-4}\)), AoLP error (\(^\circ\)), and DoLP error.}
  \label{fig:supp-metal-full}
\end{figure}
\clearpage

\begin{figure}[p]
  \centering
  \input{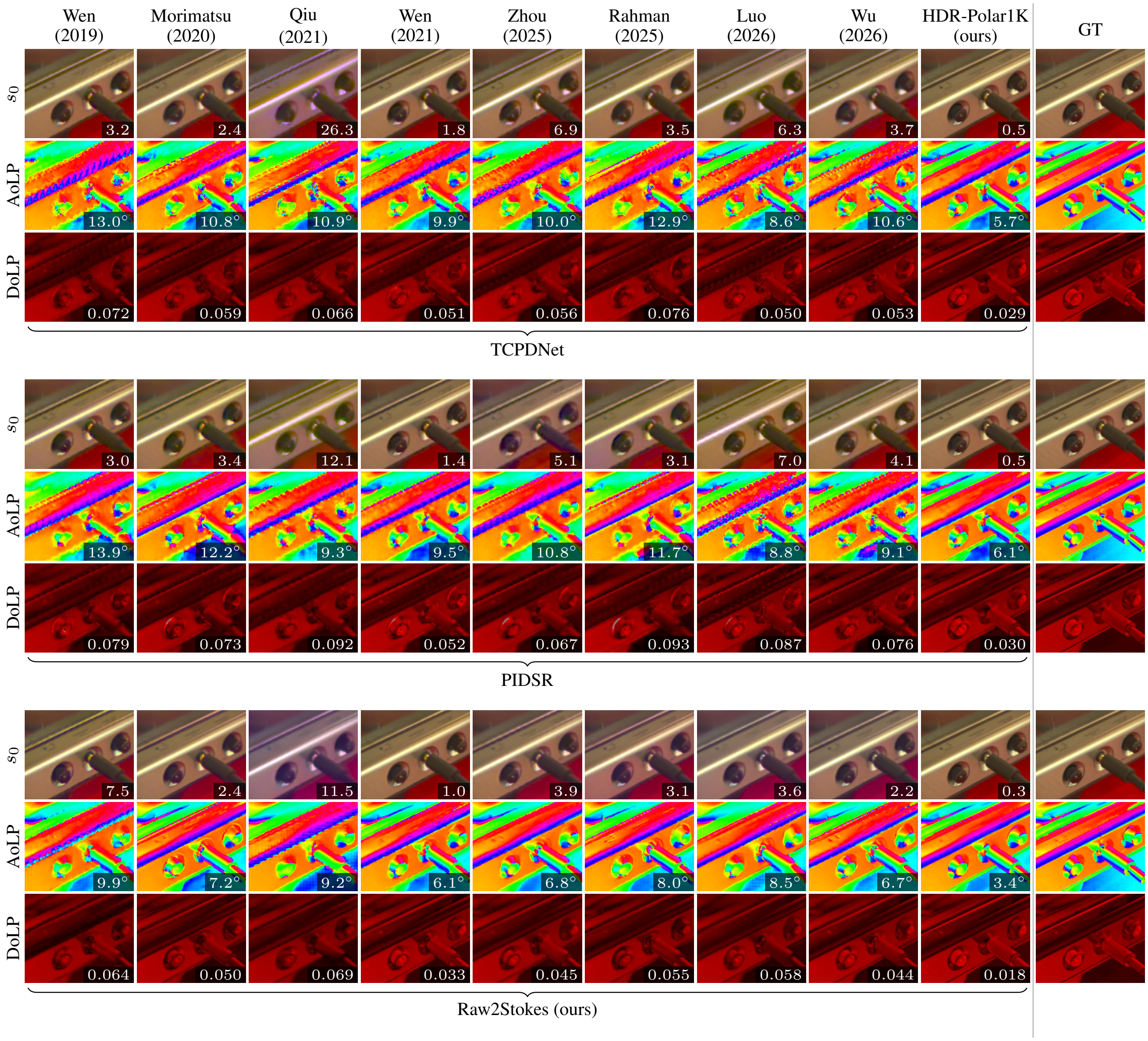}
  \caption{\textbf{Visual comparison across training datasets: laboratory equipment scene.} Bottom-right numbers report \(s_0\) MSE (\(10^{-4}\)), AoLP error (\(^\circ\)), and DoLP error.}
  \label{fig:supp-equipment-full}
\end{figure}
\clearpage

\begin{figure}[p]
  \centering
  \input{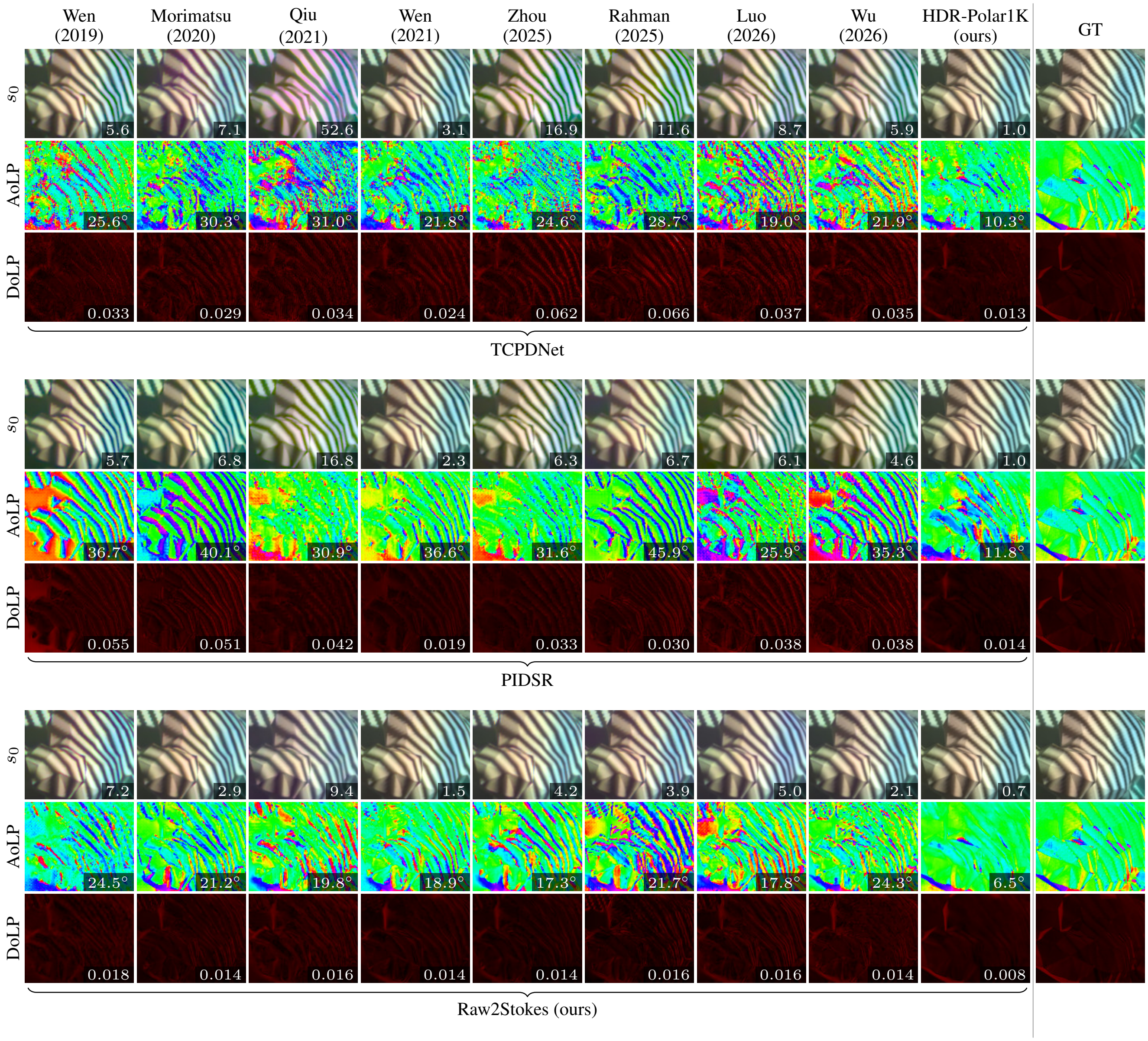}
  \caption{\textbf{Visual comparison across training datasets: high-frequency stripe scene.} Bottom-right numbers report \(s_0\) MSE (\(10^{-4}\)), AoLP error (\(^\circ\)), and DoLP error.}
  \label{fig:supp-stripes-full}
\end{figure}
\clearpage

\begin{figure}[p]
  \centering
  \input{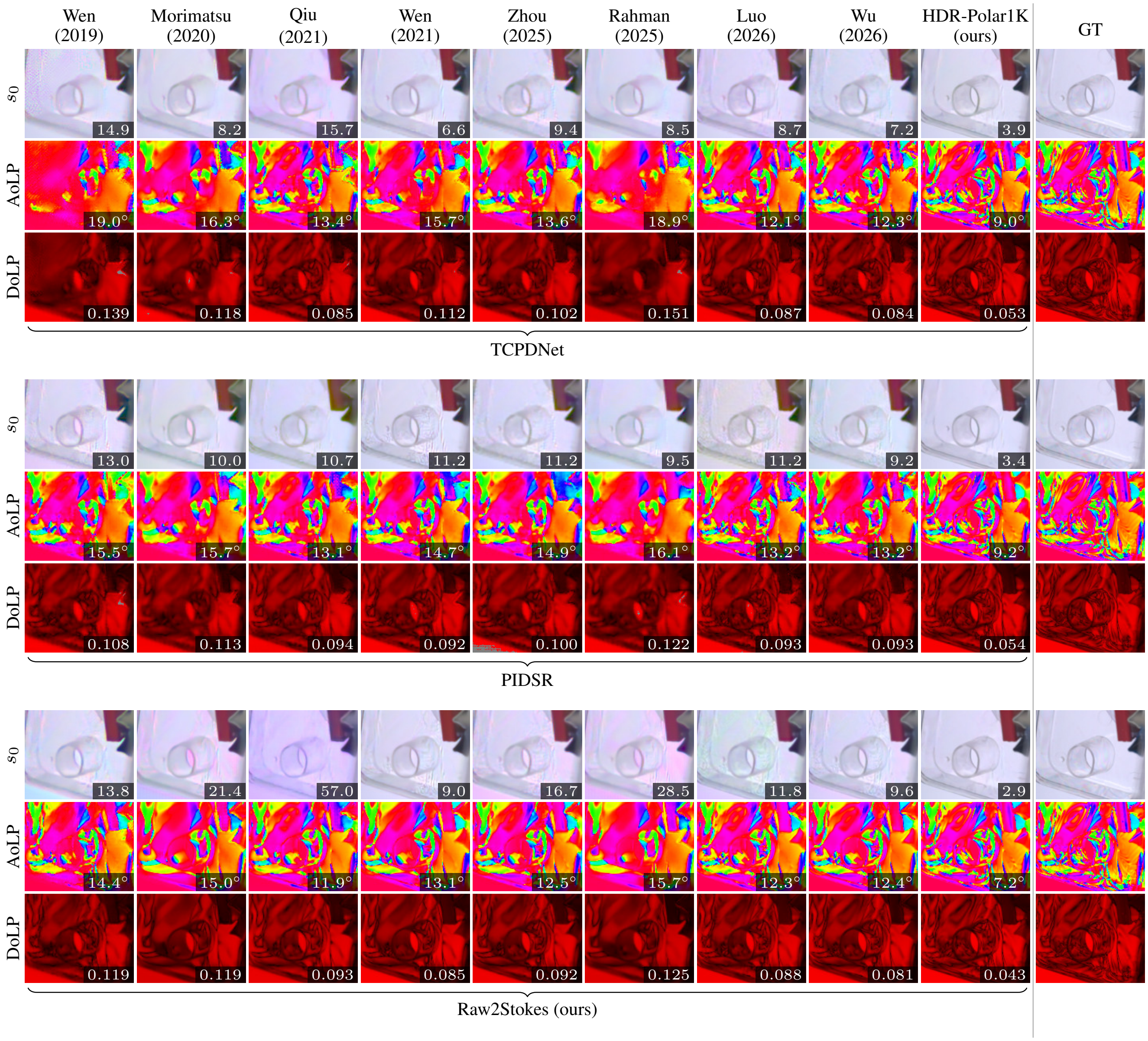}
  \caption{\textbf{Visual comparison across training datasets: photoelastic stress scene.} Bottom-right numbers report \(s_0\) MSE (\(10^{-4}\)), AoLP error (\(^\circ\)), and DoLP error.}
  \label{fig:supp-photoelastic-full}
\end{figure}
\clearpage

\clearpage
\section{Comparison with existing reconstruction methods}
\label{sec:supp-additional-comparisons}

\subsection{Accuracy across sensor conditions}
\label{sec:supp-method-positioning}
\label{sec:supp-degradation-protocol}
\label{sec:supp-primary-restoration-protocol}

\Cref{tab:supp-method-conditions} compares methods that recover color Stokes images from a single CPFA RAW mosaic across the three sensor presets defined in \cref{sec:supp-hdr-metrics}. All methods receive identical simulated RAW mosaics, with Raw2Stokes evaluated without sensor metadata. We group baselines into training-free methods, pretrained methods,\footnote{Since SwinCPD pretrained weights were not released, we trained SwinCPD with its official code on Morimatsu et al.~\cite{morimatsu2020eari}.} and networks trained on HDR-Polar1K using our calibrated sensor simulator under the protocol of \cref{sec:supp-implementation-training}. PIDSR operates at its native demosaicing resolution.

Two consistent trends emerge in \cref{tab:supp-method-conditions}. First, while PIDSR achieves the strongest low-noise accuracy among pretrained baselines, retraining TCPDNet and PIDSR on HDR-Polar1K with our calibrated sensor simulator substantially lowers errors across all conditions, reducing Stokes MSE by more than an order of magnitude under the thermal preset. Second, under identical HDR-Polar1K training data, Raw2Stokes achieves the lowest error across all three sensor conditions and all metrics while guaranteeing a zero nonphysical rate.

\begin{table}[htbp]
  \centering\small
  \setlength{\tabcolsep}{3.2pt}
  \caption{\textbf{Reconstruction across sensor conditions.} Using the three sensor presets of \cref{sec:supp-hdr-metrics}. Bold marks the best value in each column, including ties.}
  \label{tab:supp-method-conditions}
  \begin{tabular}{@{}l@{\hspace{10pt}}rrr@{\hspace{11pt}}rrr@{\hspace{11pt}}rrr@{\hspace{11pt}}rrr@{}}
    \toprule
    & \multicolumn{3}{c@{\hspace{11pt}}}{Stokes MSE (\(10^{-4}\)) \(\downarrow\)} & \multicolumn{3}{c@{\hspace{11pt}}}{AoLP (\(^\circ\)) \(\downarrow\)} & \multicolumn{3}{c@{\hspace{11pt}}}{DoLP \(\downarrow\)} & \multicolumn{3}{c@{}}{Nonphysical rate (\%) \(\downarrow\)} \\
    \cmidrule(r{11pt}){2-4}\cmidrule(r{11pt}){5-7}\cmidrule(r{11pt}){8-10}\cmidrule(){11-13}
    Method & Low & High & Thermal & Low & High & Thermal & Low & High & Thermal & Low & High & Thermal \\
    \midrule
    \multicolumn{13}{@{}l}{\emph{Training-free}} \\
    Bilinear~\cite{maeda2019polanalyser} & 21.0 & 27.6 & 111.2 & 16.4 & 27.9 & 30.4 & 0.074 & 0.210 & 0.132 & 0.42 & 5.86 & 0.05 \\
    IGRI-1 (2020)~\cite{morimatsu2020eari} & 7.8 & 15.7 & 99.6 & 12.6 & 26.7 & 29.6 & 0.053 & 0.209 & 0.127 & 0.56 & 7.16 & 0.02 \\
    Qiu (2021)~\cite{qiu2021linear} & 13.2 & 17.6 & 94.8 & 11.6 & 20.7 & 26.3 & 0.048 & 0.091 & 0.097 & 0.46 & 1.55 & 0.02 \\
    IGRI-2 (2021)~\cite{morimatsu2021igri} & 7.0 & 14.7 & 98.3 & 11.6 & 25.6 & 28.8 & 0.046 & 0.189 & 0.119 & 0.37 & 6.52 & 0.02 \\
    NLCSR (2024)~\cite{luo2024nlcsr} & 14.0 & 15.5 & 89.8 & 10.9 & 18.8 & 24.3 & 0.039 & 0.065 & 0.087 & 0.22 & 1.11 & \textbf{0} \\
    \midrule
    \multicolumn{13}{@{}l}{\emph{Pretrained}} \\
    CPDNet (2019)~\cite{wen2019cpdnet} & 75.5 & 87.1 & 157.5 & 44.9 & 46.0 & 45.5 & 0.101 & 0.105 & 0.111 & 0.03 & 0.08 & \(<0.01\) \\
    JCPD (2021)~\cite{wen2021sparse} & 21.9 & 30.0 & 117.0 & 15.3 & 27.1 & 29.8 & 0.090 & 0.174 & 0.123 & 1.07 & 5.66 & 0.27 \\
    TCPDNet (2022)~\cite{nguyen2022tcpdnet} & 6.9 & 16.2 & 97.3 & 10.8 & 25.0 & 29.3 & 0.038 & 0.127 & 0.098 & 0.52 & 4.00 & \(<0.01\) \\
    PIDSR (2025)~\cite{zhou2025pidsr} & 5.2 & 12.5 & 90.9 & 7.2 & 17.1 & 23.9 & 0.030 & 0.075 & 0.084 & 0.78 & 2.48 & 0.01 \\
    DCPM (2025)~\cite{li2025dcpm} & 5.9 & 14.0 & 92.9 & 10.0 & 24.8 & 28.6 & 0.048 & 0.168 & 0.107 & 2.96 & 6.43 & 0.03 \\
    SwinCPD (2026)~\cite{wang2026swincpd} & 31.4 & 49.0 & 106.8 & 13.8 & 27.6 & 27.7 & 0.096 & 0.215 & 0.093 & 6.68 & 11.16 & 0.21 \\
    PUGDiff (2026)~\cite{li2026pugdiff} & 6.4 & 11.1 & 78.9 & 8.3 & 18.7 & 23.6 & 0.033 & 0.083 & 0.083 & 0.79 & 2.20 & \(<0.01\) \\
    DULR-CPDM (2026)~\cite{wu2026deep} & 5.8 & 12.4 & 91.3 & 8.4 & 21.9 & 25.3 & 0.036 & 0.123 & 0.097 & 0.17 & 1.06 & \(<0.01\) \\
    PolarAPP-SfP (2026)~\cite{luo2026polarapp} & 6.9 & 14.8 & 85.4 & 9.0 & 17.9 & 23.8 & 0.047 & 0.134 & 0.082 & 3.02 & 8.40 & \(<0.01\) \\
    PolarAPP-DfP (2026)~\cite{luo2026polarapp} & 16.0 & 23.9 & 105.9 & 17.0 & 26.6 & 28.7 & 0.108 & 0.217 & 0.123 & 4.37 & 9.83 & 0.18 \\
    CPDDNet (2026)~\cite{zhang2026cpddnet} & 17.1 & 18.9 & 94.4 & 24.4 & 25.8 & 27.3 & 0.121 & 0.140 & 0.109 & 3.45 & 5.01 & \(<0.01\) \\
    UP-Dem (2026)~\cite{luo2026updem} & 6.8 & 13.4 & 108.6 & 9.1 & 17.7 & 23.1 & 0.040 & 0.085 & 0.086 & 0.80 & 2.16 & \(<0.01\) \\
    \midrule
    \multicolumn{13}{@{}l}{\emph{Trained on HDR-Polar1K}} \\
    TCPDNet~\cite{nguyen2022tcpdnet} & 2.1 & 4.5 & 8.7 & 7.2 & 15.0 & 19.8 & 0.024 & 0.043 & 0.063 & 0.25 & 0.26 & 0.03 \\
    PIDSR~\cite{zhou2025pidsr} & 1.8 & 4.0 & 7.4 & 6.2 & 12.3 & 16.2 & 0.023 & 0.039 & 0.054 & 0.74 & 0.68 & 0.14 \\
    \textbf{Raw2Stokes (ours)} & \textbf{1.2} & \textbf{3.3} & \textbf{4.6} & \textbf{4.4} & \textbf{10.7} & \textbf{13.0} & \textbf{0.017} & \textbf{0.034} & \textbf{0.045} & \textbf{0} & \textbf{0} & \textbf{0} \\
    \bottomrule
  \end{tabular}
\end{table}

\FloatBarrier
\clearpage
\subsection{Computational cost}
\label{sec:supp-computational-efficiency}

\Cref{tab:supp-efficiency-resolution} reports parameter counts, multiply--accumulate operations (MACs), latency, and peak GPU memory at two RAW resolutions for untiled batch-size-1 inference in PyTorch eager mode (FP32) on an RTX~3090 (24\,GB). MACs include all diffusion steps where applicable, and latency is measured as the median over five runs after warm-up. At full sensor resolution (\(2448\mathord{\times}2048\)), six of the eleven neural baselines run out of memory during untiled inference. In contrast, Raw2Stokes remains compact and requires the lowest peak GPU memory among all full-resolution-capable models, enabling untiled full-frame inference.

\begin{table}[!htbp]
  \centering
  \small
  \setlength{\tabcolsep}{4pt}
  \renewcommand{\arraystretch}{1.12}
  \caption{\textbf{Computational cost at two RAW resolutions.} OOM: out of memory (24\,GB GPU).}
  \label{tab:supp-efficiency-resolution}
  \begin{tabular}{@{}lrrrrrrr@{}}
    \toprule
    & & \multicolumn{3}{c}{\(256\mathord{\times}256\) RAW} & \multicolumn{3}{c}{\(2448\mathord{\times}2048\) RAW} \\
    \cmidrule(lr){3-5}\cmidrule(l){6-8}
    Method & Parameters & MACs & Latency & Peak memory & MACs & Latency & Peak memory \\
    \midrule
    CPDNet (2019)~\cite{wen2019cpdnet} & 0.5\,M & 34\,G & 6\,ms & 0.9\,GB & 2,622\,G & 0.4\,s & 21.0\,GB \\
    TCPDNet (2022)~\cite{nguyen2022tcpdnet} & 34.5\,M & 181\,G & 33\,ms & 1.0\,GB & 12,559\,G & 1.4\,s & 14.7\,GB \\
    PIDSR (2025)~\cite{zhou2025pidsr} & 7.4\,M & 47\,G & 55\,ms & 1.1\,GB & 3,576\,G & \multicolumn{2}{c}{OOM} \\
    DCPM (2025)~\cite{li2025dcpm} & 143.5\,M & 3,283\,G & 654\,ms & 2.3\,GB & 251,131\,G & \multicolumn{2}{c}{OOM} \\
    SwinCPD (2026)~\cite{wang2026swincpd} & 5.4\,M & 417\,G & 1,526\,ms & 2.1\,GB & 31,872\,G & \multicolumn{2}{c}{OOM} \\
    PUGDiff (2026)~\cite{li2026pugdiff} & 926.9\,M & 2,307\,G & 356\,ms & 7.2\,GB & 314,602\,G & \multicolumn{2}{c}{OOM} \\
    DULR-CPDM (2026)~\cite{wu2026deep} & 34.0\,M & 243\,G & 61\,ms & 2.2\,GB & 18,548\,G & \multicolumn{2}{c}{OOM} \\
    PolarAPP-SfP (2026)~\cite{luo2026polarapp} & 3.8\,M & 33\,G & 31\,ms & 0.9\,GB & 2,542\,G & 1.5\,s & 13.9\,GB \\
    PolarAPP-DfP (2026)~\cite{luo2026polarapp} & 4.4\,M & 33\,G & 31\,ms & 0.8\,GB & 2,542\,G & 1.5\,s & 12.5\,GB \\
    CPDDNet (2026)~\cite{zhang2026cpddnet} & 68.6\,M & 184\,G & 34\,ms & 1.2\,GB & 14,359\,G & 1.4\,s & 11.4\,GB \\
    UP-Dem (2026)~\cite{luo2026updem} & 2.8\,M & 54\,G & 29\,ms & 1.1\,GB & 4,122\,G & \multicolumn{2}{c}{OOM} \\
    \midrule
    \textbf{Raw2Stokes (ours)} & 4.4\,M & 294\,G & 52\,ms & 0.8\,GB & 20,237\,G & 3.3\,s & 9.1\,GB \\
    \bottomrule
  \end{tabular}
\end{table}
\FloatBarrier

\clearpage
\section{Raw2Stokes analyses}
\label{sec:supp-model-analyses}

We analyze four design aspects of Raw2Stokes: the physical output parameterization, training with simulated sensor noise, sensor metadata conditioning, and reconstruction under saturation.

\subsection{Physical output constraint}
\label{sec:supp-physical-output-ablation}

Standard CPFA reconstruction networks typically regress four polarizer intensities~\cite{wen2019cpdnet,nguyen2022tcpdnet,zhou2025pidsr,zhang2026cpddnet} or three Stokes components~\cite{zheng2024scfnet} without constraints. Under low light or heavy noise, unconstrained predictions can violate the physical Stokes bounds. To evaluate how enforcing physical validity affects both validity and reconstruction accuracy, we compare four separately trained Raw2Stokes variants in blind mode:
\begin{itemize}[leftmargin=1.5em,itemsep=2pt,topsep=4pt]
  \item \textbf{Polarizer intensities output:} For each color channel, the network predicts four polarizer intensities \((I_0,I_{45},I_{90},I_{135})\) and converts them via \(\widehat{s}_0=\tfrac12(I_0+I_{45}+I_{90}+I_{135})\), \(\widehat{s}_1=I_0-I_{90}\), and \(\widehat{s}_2=I_{45}-I_{135}\).
  \item \textbf{Direct Stokes output:} The network directly predicts unconstrained Stokes components \((\widehat{s}_0,\widehat{s}_1,\widehat{s}_2)=(z_0,z_1,z_2)\) per color channel.
  \item \textbf{NeSpoF-style constrained output:} Adapting Kim et al.~\cite{kim2023nespof} to HDR linear polarization, we replace the intensity sigmoid with softplus to allow intensities above one while retaining a sigmoid for DoLP and direct angle prediction:
  \begin{equation}
    \widehat{s}_0 = \operatorname{softplus}(z_0), \qquad
    (\widehat{s}_1,\widehat{s}_2) = \widehat{s}_0\,\sigma(z_1)\bigl(\cos(2z_2),\sin(2z_2)\bigr).
    \label{eq:supp-nespof-softplus}
  \end{equation}
\end{itemize}

\begin{figure}[!htbp]
  \centering
  \input{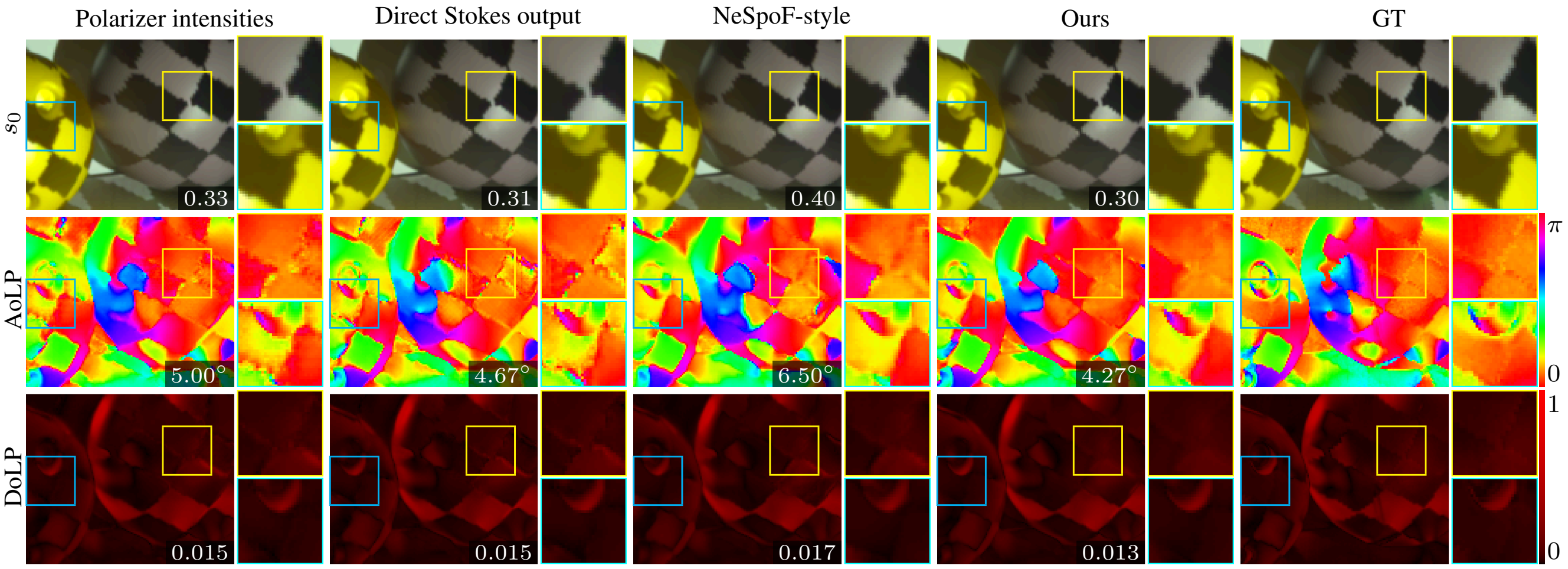}
  \caption{\textbf{Effect of the physical output constraint.} Four output parameterizations on a low-noise test scene. Our constrained mapping reduces polarization errors around fine details. Bottom-right numbers report \(s_0\) MSE (\(10^{-4}\)), AoLP error (\(^\circ\)), and DoLP error.}
  \label{fig:supp-physical-constraint}
\end{figure}

In \cref{tab:physical-output-ablation}, both unconstrained mappings produce nonphysical Stokes vectors despite training on physically valid ground truth. The NeSpoF-style mapping eliminates nonphysical outputs but increases Stokes MSE and AoLP error. Our smooth parameterization guarantees physically valid outputs, achieves the lowest reconstruction errors, and yields cleaner AoLP and DoLP estimates around fine structures in \cref{fig:supp-physical-constraint}.

\begin{table}[!htbp]
  \centering\small
  \caption{\textbf{Physical output constraint.} Bold marks the best value in each column, including ties.}
  \label{tab:physical-output-ablation}
  \newcommand{\AnalyzerMetrics}{1.42 & 4.8 & 0.0173 & 0.31}
\newcommand{\DirectStokesMetrics}{1.43 & 4.7 & 0.0176 & 0.33}
\newcommand{\NeSpoFSoftplusMetrics}{1.77 & 5.1 & 0.0181 & \textbf{0}}
\newcommand{\ProposedMetrics}{\textbf{1.39} & \textbf{4.4} & \textbf{0.0167} & \textbf{0}}

  \setlength{\tabcolsep}{4pt}
  \begin{tabular}{@{}lrrrr@{}}
    \toprule
    Output mapping & Stokes MSE ($10^{-4}$) $\downarrow$ & AoLP ($^\circ$) $\downarrow$ & DoLP $\downarrow$ & Nonphysical rate (\%) $\downarrow$ \\
    \midrule
    Polarizer intensities output & \AnalyzerMetrics \\
    Direct Stokes output & \DirectStokesMetrics \\
    \midrule
    NeSpoF-style & \NeSpoFSoftplusMetrics \\
    Ours & \ProposedMetrics \\
    \bottomrule
  \end{tabular}
\end{table}

\FloatBarrier
\subsection{Simulated sensor noise during training}
\label{sec:supp-training-sensor-noise}

Most existing CPFA demosaicing models are trained on clean or lightly perturbed mosaics, which limits robustness under real high-gain noise. To isolate the impact of calibrated sensor-noise simulation during training, we train a second Raw2Stokes model on the same 858 HDR-Polar1K scenes without the noise simulator, omitting sensor noise, dark signal, and hot pixels while retaining signal scaling, CPFA mosaicing, clipping, and quantization. Each model is selected on the same 50 validation scenes under its matching training setting and evaluated in blind mode on identical low- and high-noise RAW inputs from \cref{sec:supp-hdr-metrics}.

As shown in \cref{tab:supp-training-sensor-noise}, the models trained without and with the noise simulator perform nearly identically under low noise, indicating that training with broad sensor-noise simulation has little cost under low noise. Under high noise, however, training without the noise simulator increases Stokes MSE by more than five times and sharply degrades both AoLP and DoLP accuracy. In the high-noise example of \cref{fig:supp-training-sensor-noise}, the model trained without the noise simulator produces heavy speckle in AoLP and DoLP, whereas the model trained with the noise simulator preserves smooth polarization structure. The same behavior transfers to the full-resolution real captures from Figure~9(a,b) of the main paper in \cref{fig:supp-real-training-noise}. Both models perform comparably on the low-noise daytime scene (0~dB), whereas on the high-gain bird capture (26.4~dB), omitting the noise simulator leaves strong noise in \(s_0\) and heavy speckle in AoLP and DoLP.

\begin{table}[!htbp]
  \centering
  \small
  \caption{\textbf{Effect of training with simulated sensor noise.} Bold marks the best value per noise level.}
  \label{tab:supp-training-sensor-noise}
  \begin{tabular}{@{}llrrr@{}}
    \toprule
    Test noise & Training inputs & Stokes MSE (\(10^{-4}\)) \(\downarrow\) & AoLP (\(^{\circ}\)) \(\downarrow\) & DoLP \(\downarrow\) \\
    \midrule
    Low & Without noise simulator & \textbf{1.34} & 4.52 & \textbf{0.0168} \\
     & With noise simulator & 1.39 & \textbf{4.49} & 0.0169 \\
    \midrule
    High & Without noise simulator & 20.40 & 25.34 & 0.1654 \\
     & With noise simulator & \textbf{3.75} & \textbf{10.92} & \textbf{0.0345} \\
    \bottomrule
  \end{tabular}
\end{table}

\begin{figure}[!htbp]
  \centering
  \setlength{\tabcolsep}{2pt}
  \newcommand{\noisePanel}[1]{\raisebox{-.5\height}{#1}}
  \newcommand{\noiseImage}[1]{\csname ArxivImage:figures/supp-training-sensor-noise/#1.png\endcsname{viewport=8 8 145 120,clip,width=.130\linewidth}}
  \newcommand{\noiseLegend}[2]{%
    \noisePanel{\rotatebox{90}{#1}}\,%
    \makebox[.8em][l]{%
      \rlap{\raisebox{\dimexpr.044\linewidth-1ex\relax}{\(\scriptstyle #2\)}}%
      \raisebox{-.044\linewidth}{\(\scriptstyle 0\)}}}
  \begin{tabular}{@{}ccccccc@{}}
    & \multicolumn{2}{c}{\footnotesize Low-noise input} & \multicolumn{2}{c}{\footnotesize High-noise input} & & \\
    \cmidrule(lr){2-3}\cmidrule(lr){4-5}
    & \footnotesize\shortstack{Without\\noise simulator} & \footnotesize\shortstack{With\\noise simulator} & \footnotesize\shortstack{Without\\noise simulator} & \footnotesize\shortstack{With\\noise simulator} & \footnotesize GT & \\[2pt]
    \noisePanel{\rotatebox{90}{\small \(s_0\)}} & \noisePanel{\suppScoredPanel{\noiseImage{low-clean-s0}}{1}{0.41}} & \noisePanel{\suppScoredPanel{\noiseImage{low-simulated-s0}}{1}{0.29}} & \noisePanel{\suppScoredPanel{\noiseImage{high-clean-s0}}{1}{9.46}} & \noisePanel{\suppScoredPanel{\noiseImage{high-simulated-s0}}{1}{2.46}} & \noisePanel{\noiseImage{gt-s0}} & \\
    \noalign{\vskip 2pt}
    \noisePanel{\rotatebox{90}{\small AoLP}} & \noisePanel{\suppScoredPanel{\noiseImage{low-clean-aolp}}{1}{5.92^\circ}} & \noisePanel{\suppScoredPanel{\noiseImage{low-simulated-aolp}}{1}{5.69^\circ}} & \noisePanel{\suppScoredPanel{\noiseImage{high-clean-aolp}}{1}{32.63^\circ}} & \noisePanel{\suppScoredPanel{\noiseImage{high-simulated-aolp}}{1}{17.87^\circ}} & \noisePanel{\noiseImage{gt-aolp}} & \noiseLegend{\includegraphics[page=74,width=.089\linewidth,height=.9mm]{figures/training-data-results-panels.pdf}}{\pi} \\
    \noalign{\vskip 2pt}
    \noisePanel{\rotatebox{90}{\small DoLP}} & \noisePanel{\suppScoredPanel{\noiseImage{low-clean-dolp}}{1}{0.008}} & \noisePanel{\suppScoredPanel{\noiseImage{low-simulated-dolp}}{1}{0.009}} & \noisePanel{\suppScoredPanel{\noiseImage{high-clean-dolp}}{1}{0.263}} & \noisePanel{\suppScoredPanel{\noiseImage{high-simulated-dolp}}{1}{0.018}} & \noisePanel{\noiseImage{gt-dolp}} & \noiseLegend{\includegraphics[page=43,width=.089\linewidth,height=.9mm]{figures/training-data-results-panels.pdf}}{1} \\
  \end{tabular}
  \caption{\textbf{Effect of training with simulated sensor noise.} Both models behave similarly under low noise, but under high noise the model trained without the noise simulator produces heavy speckle in AoLP and DoLP, whereas training with the noise simulator preserves smooth polarization structure. Bottom-right numbers report \(s_0\) MSE (\(10^{-4}\)), AoLP error (\(^\circ\)), and DoLP error.}
  \label{fig:supp-training-sensor-noise}
  \vspace{3mm}
  \setlength{\tabcolsep}{2pt}
  \newcommand{\realNoiseCrop}[3]{\raisebox{-.5\height}{\csname ArxivImage:figures/supp-real-training-noise/#1-#2-#3.png\endcsname{width=.098\linewidth}}}
  \newcommand{\realNoisePair}[2]{\realNoiseCrop{#1}{A}{#2}\hspace{2pt}\realNoiseCrop{#1}{C}{#2}}
  \newcommand{\realNoiseRow}[2]{\raisebox{-.5\height}{\rotatebox{90}{\small #2}} & \realNoisePair{day-without}{#1} & \realNoisePair{day-with}{#1} & \realNoisePair{bird-without}{#1} & \realNoisePair{bird-with}{#1}}
  \newcommand{\realNoiseLegend}[2]{%
    \raisebox{-.5\height}{\rotatebox{90}{#1}}\,%
    \makebox[.8em][l]{\rlap{\raisebox{\dimexpr.045\linewidth-1ex\relax}{\(\scriptstyle #2\)}}\raisebox{-.045\linewidth}{\(\scriptstyle 0\)}}}
  \begin{tabular}{@{}cc@{\hspace{5pt}}c@{\hspace{3mm}}c@{\hspace{5pt}}cc@{}}
    & \multicolumn{2}{c}{\footnotesize Low-noise scene (0~dB)} & \multicolumn{2}{c}{\footnotesize High-noise scene (26.4~dB)} \\
    \cmidrule(lr){2-3}\cmidrule(lr){4-5}
    & \footnotesize\shortstack{Without\\noise simulator} & \footnotesize\shortstack{With\\noise simulator} & \footnotesize\shortstack{Without\\noise simulator} & \footnotesize\shortstack{With\\noise simulator} \\[2pt]
    \realNoiseRow{S0}{\(s_0\)} \\
    \noalign{\vskip 2pt}
    \realNoiseRow{AoLP}{AoLP} & \realNoiseLegend{\includegraphics[page=74,width=.090\linewidth,height=.9mm]{figures/training-data-results-panels.pdf}}{\pi} \\
    \noalign{\vskip 2pt}
    \realNoiseRow{DoLP}{DoLP} & \realNoiseLegend{\includegraphics[page=43,width=.090\linewidth,height=.9mm]{figures/training-data-results-panels.pdf}}{1}
  \end{tabular}
  \caption{\textbf{Effect of training with simulated sensor noise on real captures.} Reconstructions of the same full-resolution RAW captures from Figure~9(a,b) of the main paper. Both models produce comparable results on the low-noise daytime scene (0~dB), whereas on the high-gain bird capture (26.4~dB) the model trained without the noise simulator shows strong noise in \(s_0\) and heavy speckle in AoLP and DoLP, while training with the noise simulator preserves clean intensity and smooth polarization structure.}
  \label{fig:supp-real-training-noise}
\end{figure}
\clearpage

\subsection{Sensor metadata availability}
\label{sec:supp-sensor-information-availability}

Camera metadata such as gain, exposure time, and temperature are not always recorded or available at inference time. To measure how much conditioning improves reconstruction, we evaluate the same trained Raw2Stokes model in blind and non-blind modes on identical RAW inputs across the three sensor presets of \cref{sec:supp-hdr-metrics}.

\begin{figure}[!htbp]
  \centering
  \input{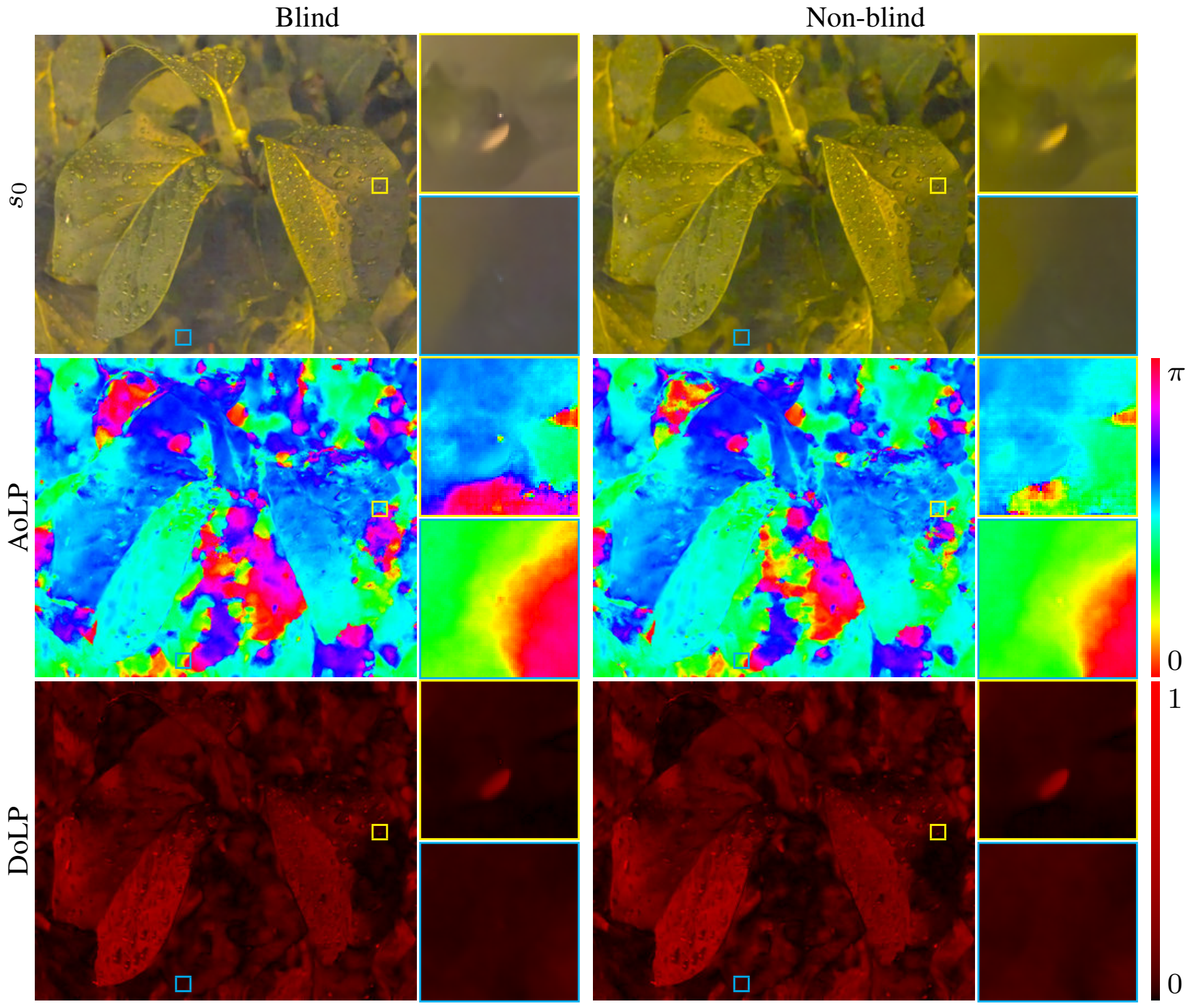}
  \caption{\textbf{Effect of sensor metadata on a real RAW capture.} Real high-gain, elevated-temperature capture (\(g=32.1\,\mathrm{dB}\), \(t=1/20\,\mathrm{s}\), \(T=59\,^{\circ}\mathrm{C}\)). While both modes recover similar scene structure, supplying recorded metadata helps account for sensor degradations under these conditions and yields richer color saturation in the enlarged foliage crops.}
  \label{fig:supp-real-metadata}
\end{figure}

As shown in \cref{tab:conditioning-ablation}, supplying sensor metadata consistently improves every metric across all three conditions, though the magnitude depends strongly on the sensor condition. Under low- and high-noise presets, the improvement in Stokes MSE is modest, since the network can partially infer the spatial noise level directly from the observed RAW mosaic. Under the thermal preset, however, dark-current offsets and hot-pixel patterns depend strongly on exposure time and sensor temperature and are harder to infer from a single frame alone. Providing metadata under this condition substantially reduces Stokes MSE and noticeably lowers both AoLP and DoLP errors. On the full-resolution real foliage capture in \cref{fig:supp-real-metadata}, both modes recover similar scene structure, while supplying recorded metadata helps account for sensor degradations under high gain and elevated temperature, yielding richer color saturation on the leaves.

\begin{table}[!htbp]
  \centering\small
  \caption{\textbf{Sensor metadata availability.} Bold marks the best value per condition.}
  \label{tab:conditioning-ablation}
  \setlength{\tabcolsep}{4pt}
  \begin{tabular}{@{}llrrr@{}}
    \toprule
    Sensor condition & Inference mode & Stokes MSE ($10^{-4}$) $\downarrow$ & AoLP ($^\circ$) $\downarrow$ & DoLP $\downarrow$ \\
    \midrule
    Low noise & Blind & 1.229 & 4.44 & 0.0166 \\
 & Non-blind & \textbf{1.182} & \textbf{4.38} & \textbf{0.0163} \\
\midrule
High noise & Blind & 3.308 & 10.71 & 0.0338 \\
 & Non-blind & \textbf{3.185} & \textbf{10.66} & \textbf{0.0337} \\
\midrule
Thermal & Blind & 4.597 & 13.00 & 0.0448 \\
 & Non-blind & \textbf{3.637} & \textbf{12.59} & \textbf{0.0403}
 \\
    \bottomrule
  \end{tabular}
\end{table}

\clearpage
\subsection{Reconstruction under saturation}

Real scenes frequently clip one or more polarizer channels in bright highlights, whereas non-HDR training datasets cannot supervise recovery above the saturation limit. Under an ideal co-located model, any three unsaturated polarizer orientations suffice to determine a linear Stokes vector~\cite{lapray2020exploiting,zhou2023polarization}. On a CPFA sensor, however, color and polarizer channels are sampled at different pixel locations, so recoverability does not follow a simple pixel-wise rule. Since HDR-Polar1K provides unsaturated HDR Stokes ground truth even when the simulated RAW input is clipped, Raw2Stokes learns to jointly infer missing and saturated measurements from neighboring unsaturated RAW samples and learned image priors.

\begin{figure}[!htbp]
  \centering
  \input{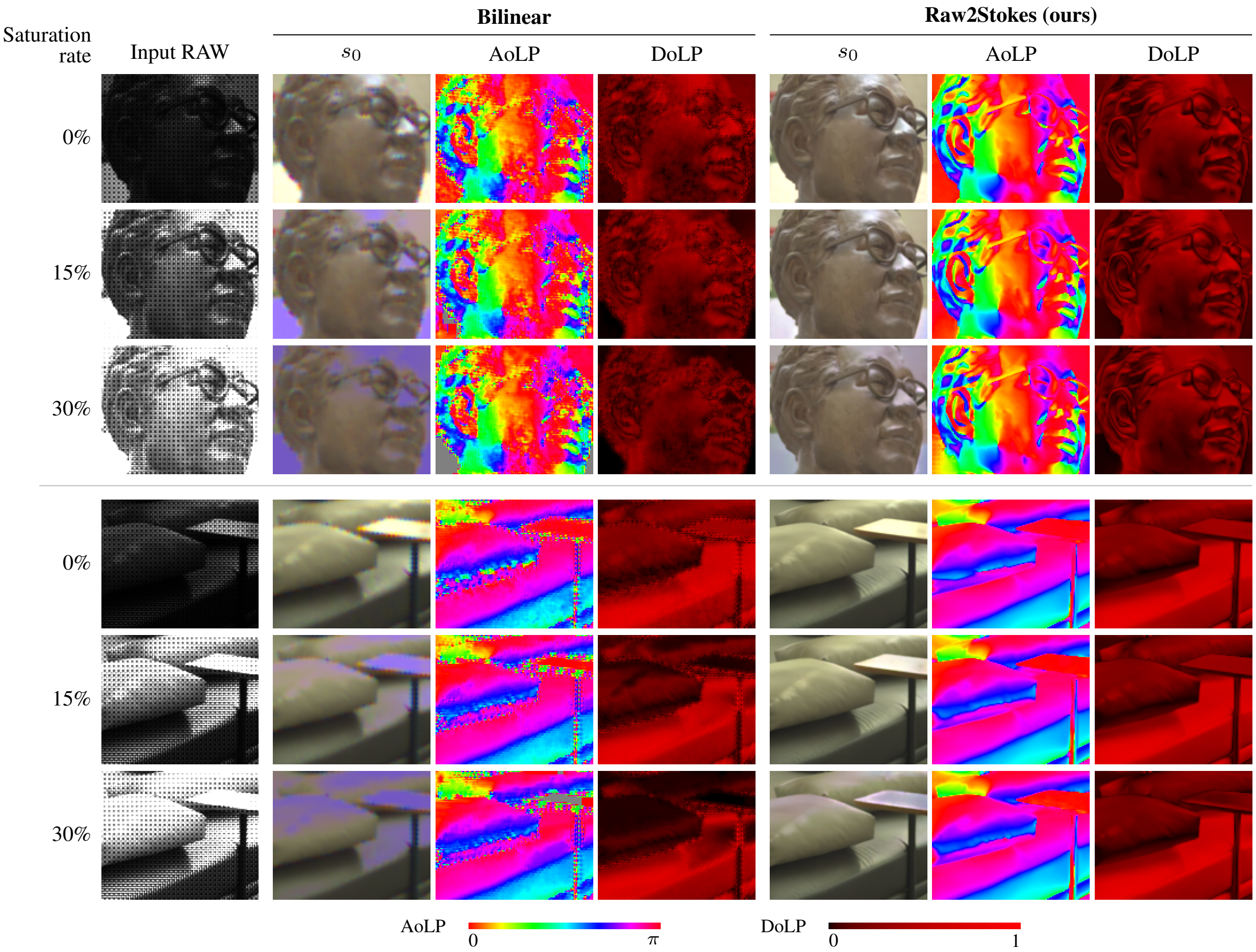}
  \caption{\textbf{Reconstruction under saturation.} Raw2Stokes reconstructs coherent intensity and polarization as scene brightness increases under the low-noise preset to clip 0\%, 15\%, and 30\% of RAW samples, degrading gradually only where large regions saturate across all polarizer angles. Reconstructed \(s_0\) is divided by the applied brightness factor for display.}
  \label{fig:saturation-limits}
\end{figure}

\Cref{fig:saturation-limits} evaluates this behavior under the low-noise preset as scene brightness increases to saturate 0\%, 15\%, and 30\% of RAW samples. While bilinear interpolation loses color and polarization structure in clipped highlights, Raw2Stokes recovers coherent intensity and polarization across heavily saturated regions, degrading gradually only where large areas saturate across all polarizer angles.

\FloatBarrier

\clearpage
\begingroup
\PaperSRGBRasters
\section{Full-resolution real-capture reconstruction}
\label{sec:supp-reconstruction-full-resolution}

Extending the main paper's qualitative evaluation on real captures, we compare the standard training-free Bilinear baseline~\cite{maeda2019polanalyser}, official pretrained PIDSR~\cite{zhou2025pidsr} (the strongest pretrained baseline overall in \cref{tab:supp-method-conditions}), and Raw2Stokes (trained on HDR-Polar1K) on identical full-resolution RAW measurements with matched crop locations. Our camera records 12-bit RAW frames at \(2448\mathord{\times}2048\) pixels, and we supply Raw2Stokes with recorded sensor metadata whenever available. We run untiled full-resolution inference for all methods, including PIDSR, in FP32 on an NVIDIA RTX~6000 Ada (48\,GB) GPU.

\paragraph{Our camera captures.}
\Cref{fig:supp-real-131839,fig:supp-real-bird,fig:supp-real-232956,fig:supp-real-photoelastic-183525} show full-frame reconstructions and matched crops of the daytime, bird, nighttime, and photoelastic examples in Figure~9 of the main paper. \Crefrange{fig:supp-real-173242}{fig:supp-real-controlled-220236} show twelve additional captures, beginning with daytime and indoor scenes, followed by nighttime scenes and a photoelastic example.

\paragraph{RAW images from external datasets.}
\Cref{tab:supp-external-raw} summarizes the external dataset examples. We use the nominal Sony CPFA layout for all five images. For BlackDragon from PISR~\cite{chen2024pisr}, we square the normalized RAW values before reconstruction, consistent with the released AoLP maps. For the other four images, the quantization step is the only sensor metadata provided to Raw2Stokes.

\begin{table}[htbp]
  \centering
  \caption{\textbf{RAW images from external datasets.}}
  \label{tab:supp-external-raw}
  \small
  \setlength{\tabcolsep}{4pt}
  \begin{tabular}{@{}clllc@{}}
    \toprule
    Figure & Dataset & File & RAW resolution & Bit depth \\
    \midrule
    \ref{fig:supp-real-plie-indoor015} & PLIE~\cite{zhou2023lowlight} & \texttt{indoor015\_4000\_*.raw} & \(2448\times2048\) & 8 \\
    \ref{fig:supp-real-potato-tower} & PoTATO~\cite{batista2024potato} & \texttt{exp05\_frame023103\_raw.png} & \(2448\times2048\) & 8 \\
    \ref{fig:supp-real-polargb-01-0007} & PolaRGB~\cite{yao2025polarfree} & \texttt{test/input/01/0007.Bmp} & \(2448\times2048\) & 8 \\
    \ref{fig:supp-real-pisr-blackdragon-025} & PISR~\cite{chen2024pisr} & \texttt{BlackDragon/raw/025.npy} & \(2448\times2048\) & 12 \\
    \ref{fig:supp-real-dir-elephant-072} & Display Inverse Rendering~\cite{choi2025real} & \texttt{072\_440000.png} & \(1224\times1024\) & 8 \\
    \bottomrule
  \end{tabular}
\end{table}

\clearpage

\begin{figure}[p]
  \centering
  \input{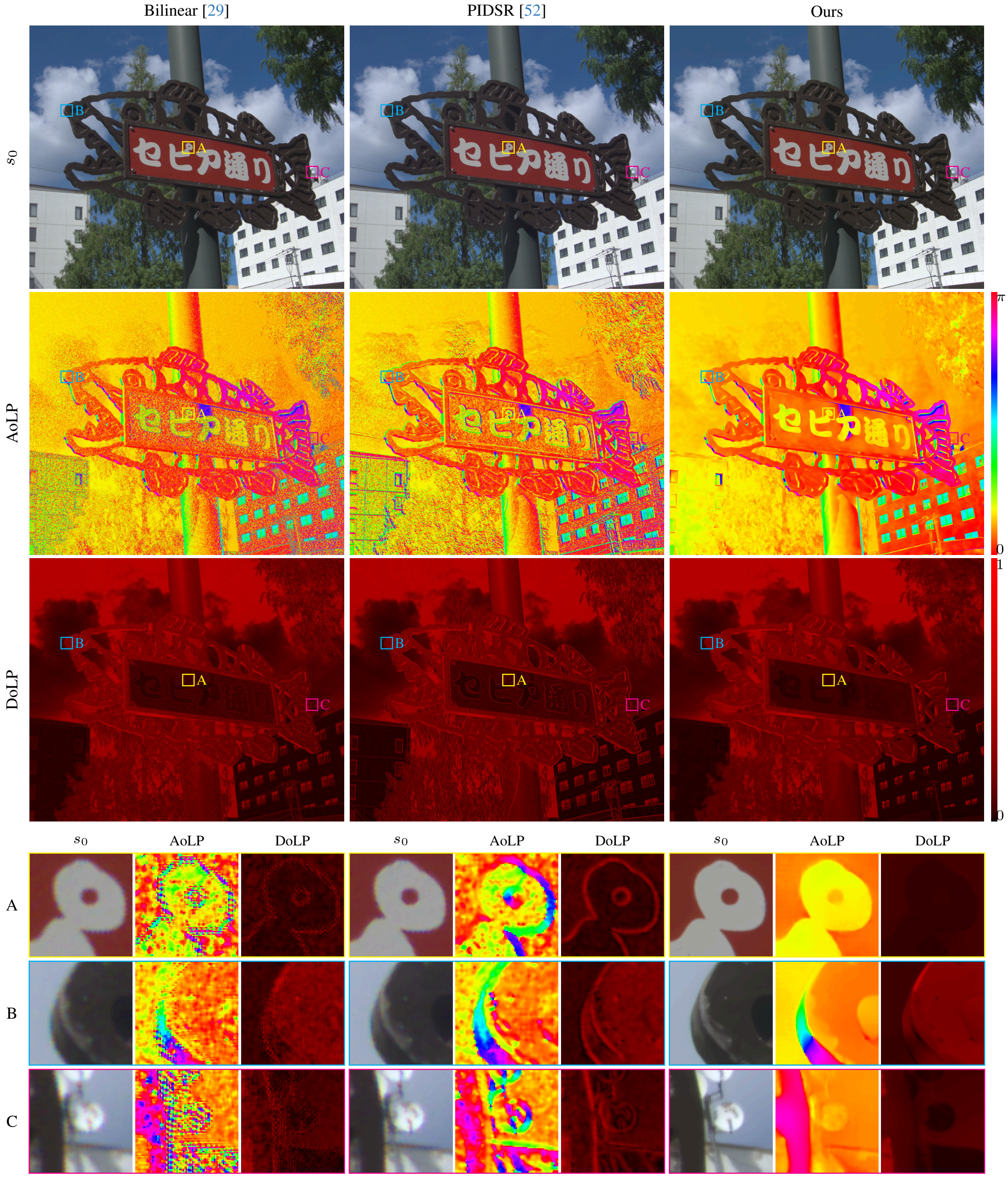}
  \caption{\textbf{RAW-to-Stokes reconstruction of a daytime scene.} \(g=0\,\mathrm{dB}\), \(t=1/1000\,\mathrm{s}\), and \(T=62.3\,^{\circ}\mathrm{C}\).}
  \label{fig:supp-real-131839}
\end{figure}

\clearpage

\begin{figure}[p]
  \centering
  \input{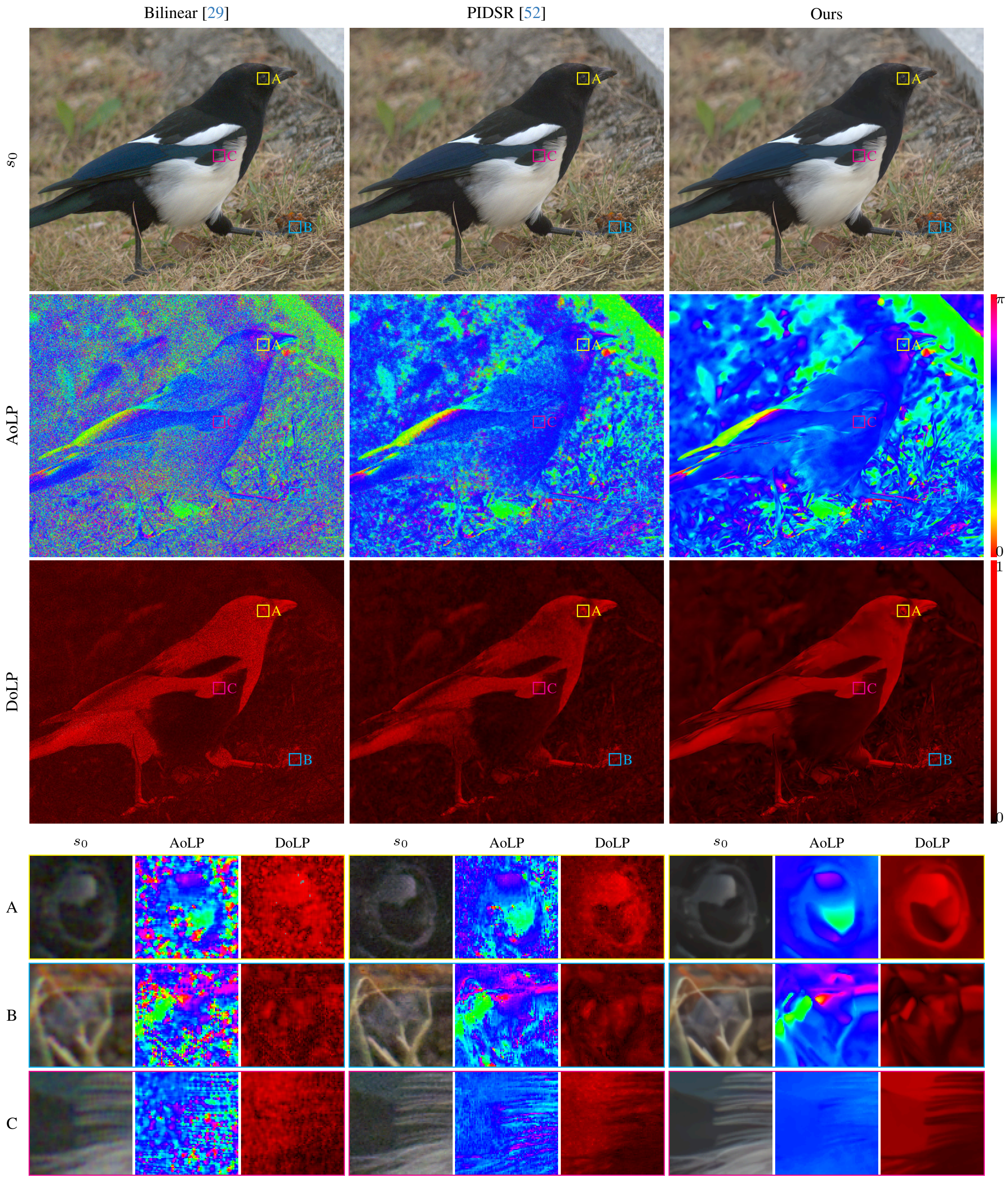}
  \caption{\textbf{RAW-to-Stokes reconstruction of a bird.} \(g=26.4\,\mathrm{dB}\), \(t=1/770\,\mathrm{s}\), and \(T=53.8\,^{\circ}\mathrm{C}\).}
  \label{fig:supp-real-bird}
\end{figure}

\clearpage

\begin{figure}[p]
  \centering
  \input{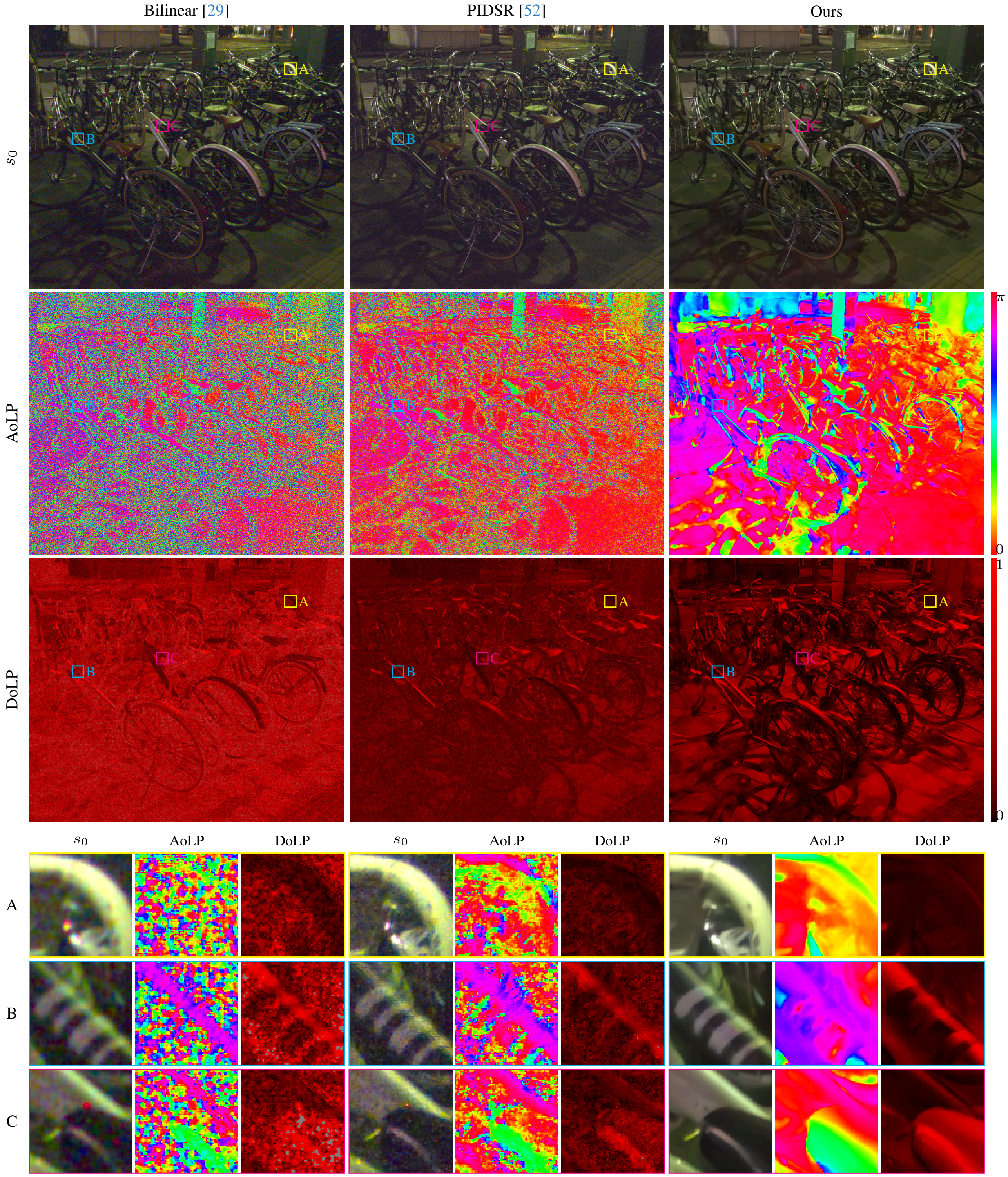}
  \caption{\textbf{RAW-to-Stokes reconstruction of a nighttime scene.} \(g=31\,\mathrm{dB}\), \(t=1/20\,\mathrm{s}\), and \(T=56\,^{\circ}\mathrm{C}\).}
  \label{fig:supp-real-232956}
\end{figure}

\clearpage

\begin{figure}[p]
  \centering
  \input{figures/supp-real-photoelastic-183525-grid}
  \caption{\textbf{RAW-to-Stokes reconstruction of a photoelastic scene.} \(g=12\,\mathrm{dB}\), \(t=1/130\,\mathrm{s}\), and \(T=68.1\,^{\circ}\mathrm{C}\).}
  \label{fig:supp-real-photoelastic-183525}
\end{figure}

\clearpage

\begin{figure}[p]
  \centering
  \input{figures/supp-real-173242-grid}
  \caption{\textbf{RAW-to-Stokes reconstruction of a daytime scene.} \(g=0\,\mathrm{dB}\), \(t=1/1565\,\mathrm{s}\), and \(T=67\,^{\circ}\mathrm{C}\).}
  \label{fig:supp-real-173242}
\end{figure}

\clearpage

\begin{figure}[p]
  \centering
  \input{figures/supp-real-172815-grid}
  \caption{\textbf{RAW-to-Stokes reconstruction of a daytime scene.} \(g=16\,\mathrm{dB}\), \(t=1/140\,\mathrm{s}\), and \(T=50\,^{\circ}\mathrm{C}\).}
  \label{fig:supp-real-172815}
\end{figure}

\clearpage

\begin{figure}[p]
  \centering
  \input{figures/supp-real-150332-grid}
  \caption{\textbf{RAW-to-Stokes reconstruction of a daytime scene.} \(g=24\,\mathrm{dB}\), \(t=1/140\,\mathrm{s}\), and \(T=48\,^{\circ}\mathrm{C}\).}
  \label{fig:supp-real-150332}
\end{figure}

\clearpage

\begin{figure}[p]
  \centering
  \input{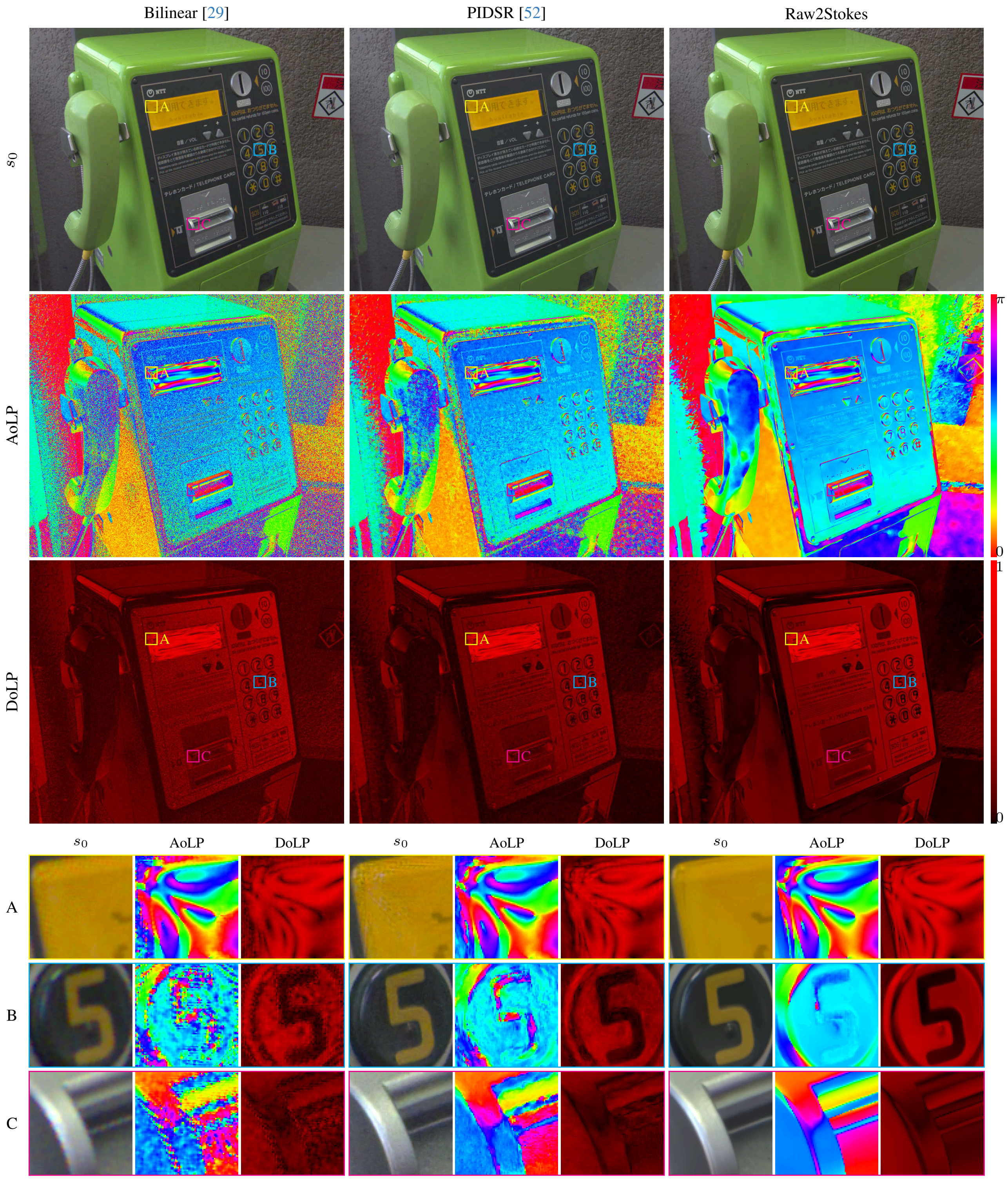}
  \caption{\textbf{RAW-to-Stokes reconstruction of a payphone.} \(g=17.5\,\mathrm{dB}\), \(t=1/39\,\mathrm{s}\), and \(T=73\,^{\circ}\mathrm{C}\).}
  \label{fig:supp-real-184426}
\end{figure}

\clearpage

\begin{figure}[p]
  \centering
  \input{figures/supp-real-175200-grid}
  \caption{\textbf{RAW-to-Stokes reconstruction of an indoor scene.} \(g=10\,\mathrm{dB}\), \(t=1/29\,\mathrm{s}\), and \(T=63\,^{\circ}\mathrm{C}\).}
  \label{fig:supp-real-175200}
\end{figure}

\clearpage

\begin{figure}[p]
  \centering
  \input{figures/supp-real-161258-grid}
  \caption{\textbf{RAW-to-Stokes reconstruction of an indoor scene.} \(g=0\,\mathrm{dB}\), \(t=1/64\,\mathrm{s}\), and \(T=72\,^{\circ}\mathrm{C}\).}
  \label{fig:supp-real-161258}
\end{figure}

\clearpage

\begin{figure}[p]
  \centering
  \input{figures/supp-real-181302-grid}
  \caption{\textbf{RAW-to-Stokes reconstruction of an indoor scene.} \(g=0\,\mathrm{dB}\), \(t=1/62\,\mathrm{s}\), and \(T=70\,^{\circ}\mathrm{C}\).}
  \label{fig:supp-real-181302}
\end{figure}

\clearpage

\begin{figure}[p]
  \centering
  \input{figures/supp-real-231521-grid}
  \caption{\textbf{RAW-to-Stokes reconstruction of a nighttime scene.} \(g=0\,\mathrm{dB}\), \(t=1/39\,\mathrm{s}\), and \(T=51\,^{\circ}\mathrm{C}\).}
  \label{fig:supp-real-231521}
\end{figure}

\clearpage

\begin{figure}[p]
  \centering
  \input{figures/supp-real-185141-grid}
  \caption{\textbf{RAW-to-Stokes reconstruction of a nighttime scene.} \(g=27\,\mathrm{dB}\), \(t=1/25\,\mathrm{s}\), and \(T=55\,^{\circ}\mathrm{C}\).}
  \label{fig:supp-real-185141}
\end{figure}

\clearpage

\begin{figure}[p]
  \centering
  \input{figures/supp-real-190013-grid}
  \caption{\textbf{RAW-to-Stokes reconstruction of a nighttime scene.} \(g=48\,\mathrm{dB}\), \(t=1/26\,\mathrm{s}\), and \(T=59\,^{\circ}\mathrm{C}\).}
  \label{fig:supp-real-190013}
\end{figure}

\clearpage

\begin{figure}[p]
  \centering
  \input{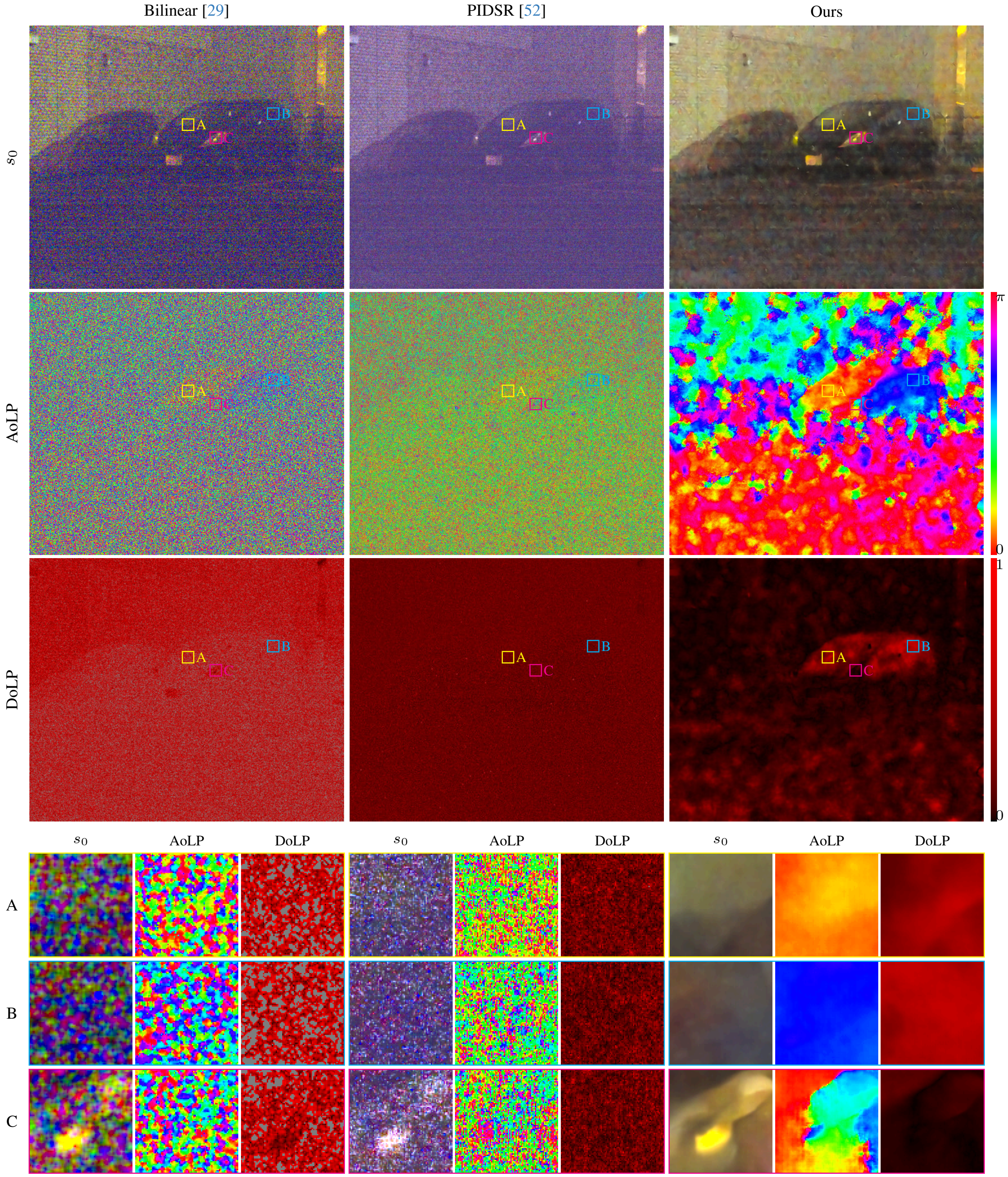}
  \caption{\textbf{RAW-to-Stokes reconstruction of a nighttime scene.} \(g=40\,\mathrm{dB}\), \(t=1/35\,\mathrm{s}\), and \(T=48\,^{\circ}\mathrm{C}\).}
  \label{fig:supp-real-204857}
\end{figure}

\clearpage

\begin{figure}[p]
  \centering
  \input{figures/supp-real-controlled-220236-grid}
  \caption{\textbf{RAW-to-Stokes reconstruction of a photoelastic scene.} \(g=0\,\mathrm{dB}\), \(t=1/102\,\mathrm{s}\), and \(T=48.4\,^{\circ}\mathrm{C}\).}
  \label{fig:supp-real-controlled-220236}
\end{figure}

\clearpage

\begin{figure}[p]
  \centering
  \input{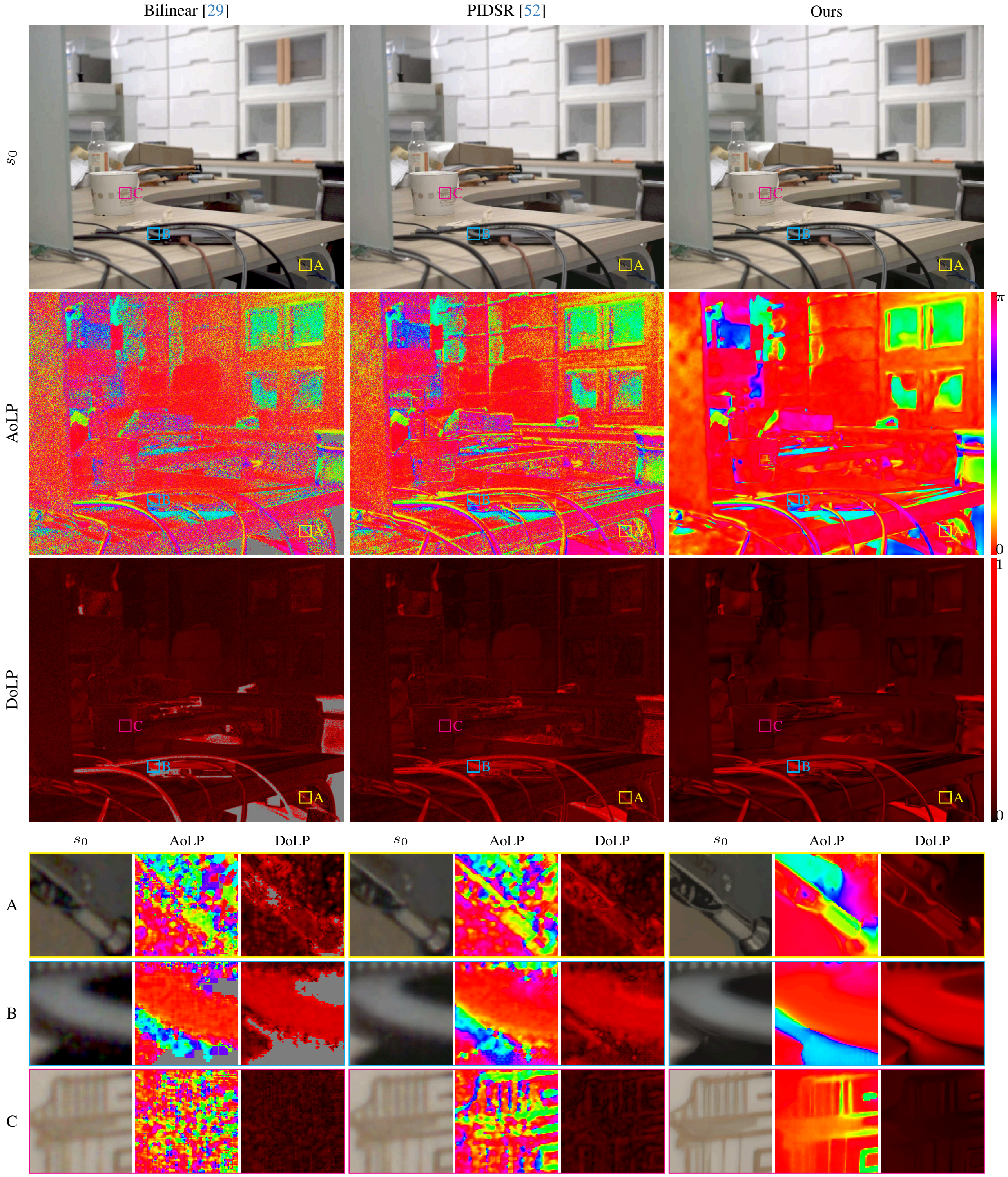}
  \caption{\textbf{RAW-to-Stokes reconstruction of the desk-and-cabinet scene from PLIE~\cite{zhou2023lowlight}.}}
  \label{fig:supp-real-plie-indoor015}
\end{figure}

\clearpage

\begin{figure}[p]
  \centering
  \input{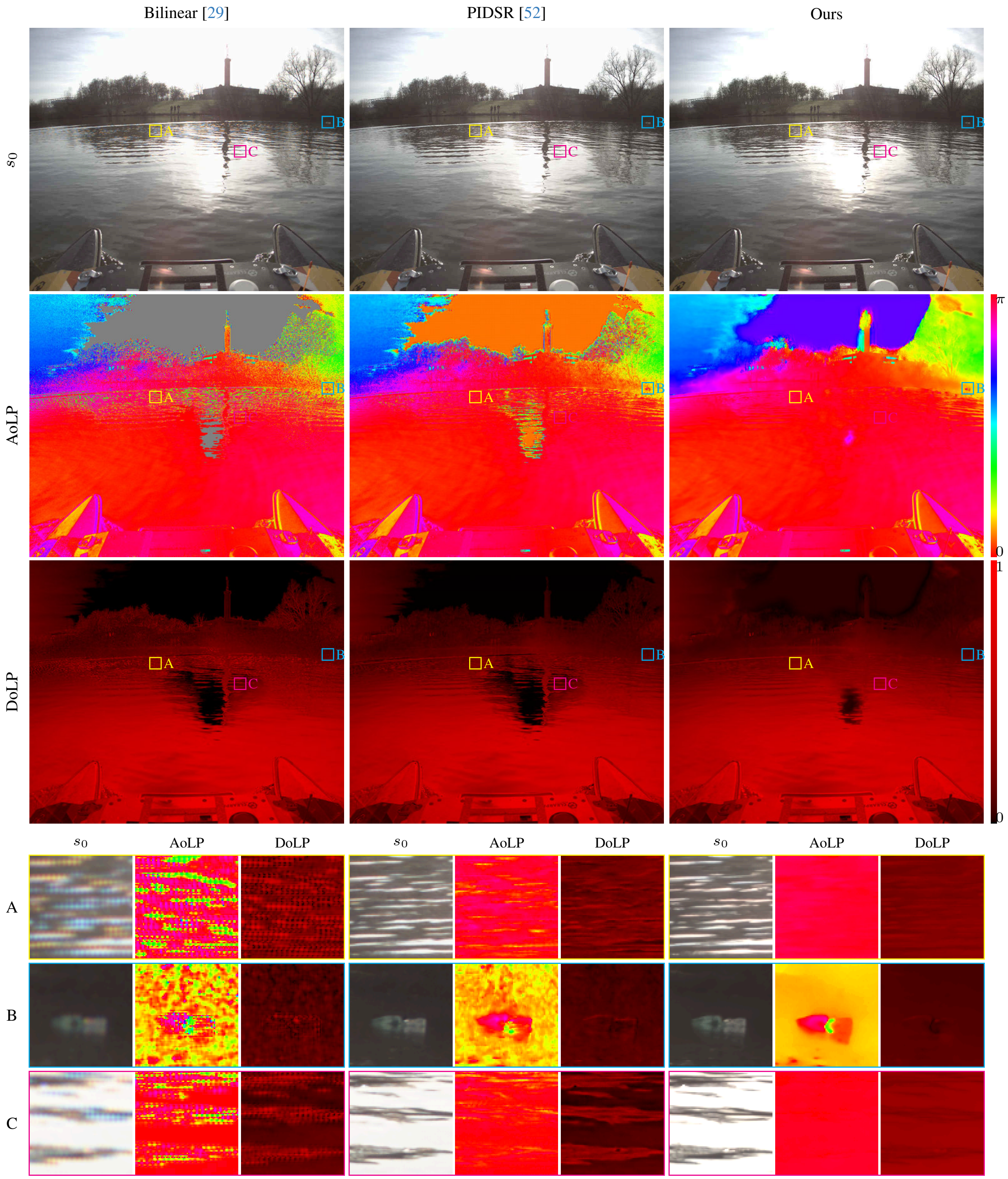}
  \caption{\textbf{RAW-to-Stokes reconstruction of a backlit scene from PoTATO~\cite{batista2024potato}.}}
  \label{fig:supp-real-potato-tower}
\end{figure}

\clearpage

\begin{figure}[p]
  \centering
  \input{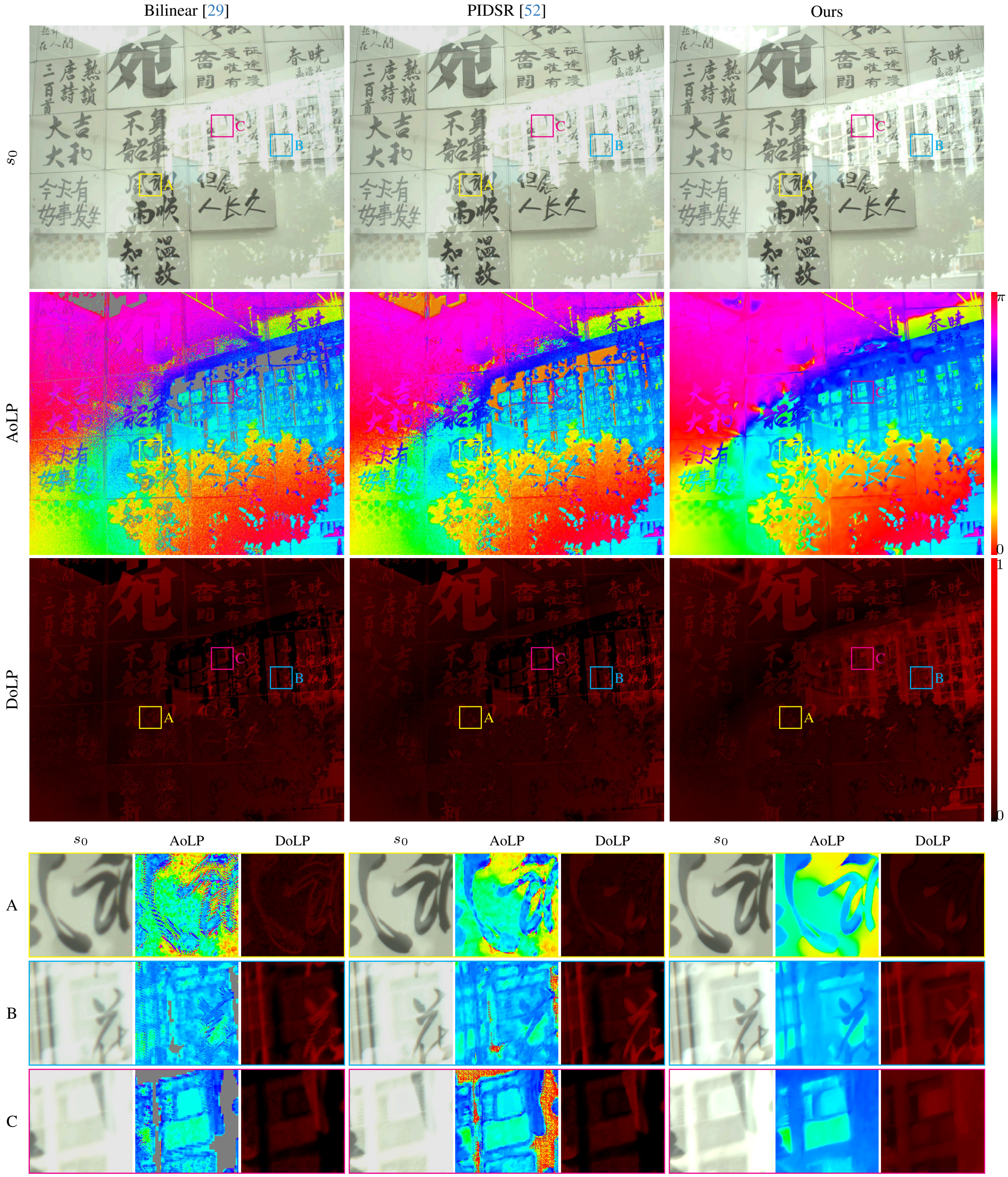}
  \caption{\textbf{RAW-to-Stokes reconstruction of a glass-reflection scene from PolaRGB~\cite{yao2025polarfree}.}}
  \label{fig:supp-real-polargb-01-0007}
\end{figure}

\clearpage

\begin{figure}[p]
  \centering
  \input{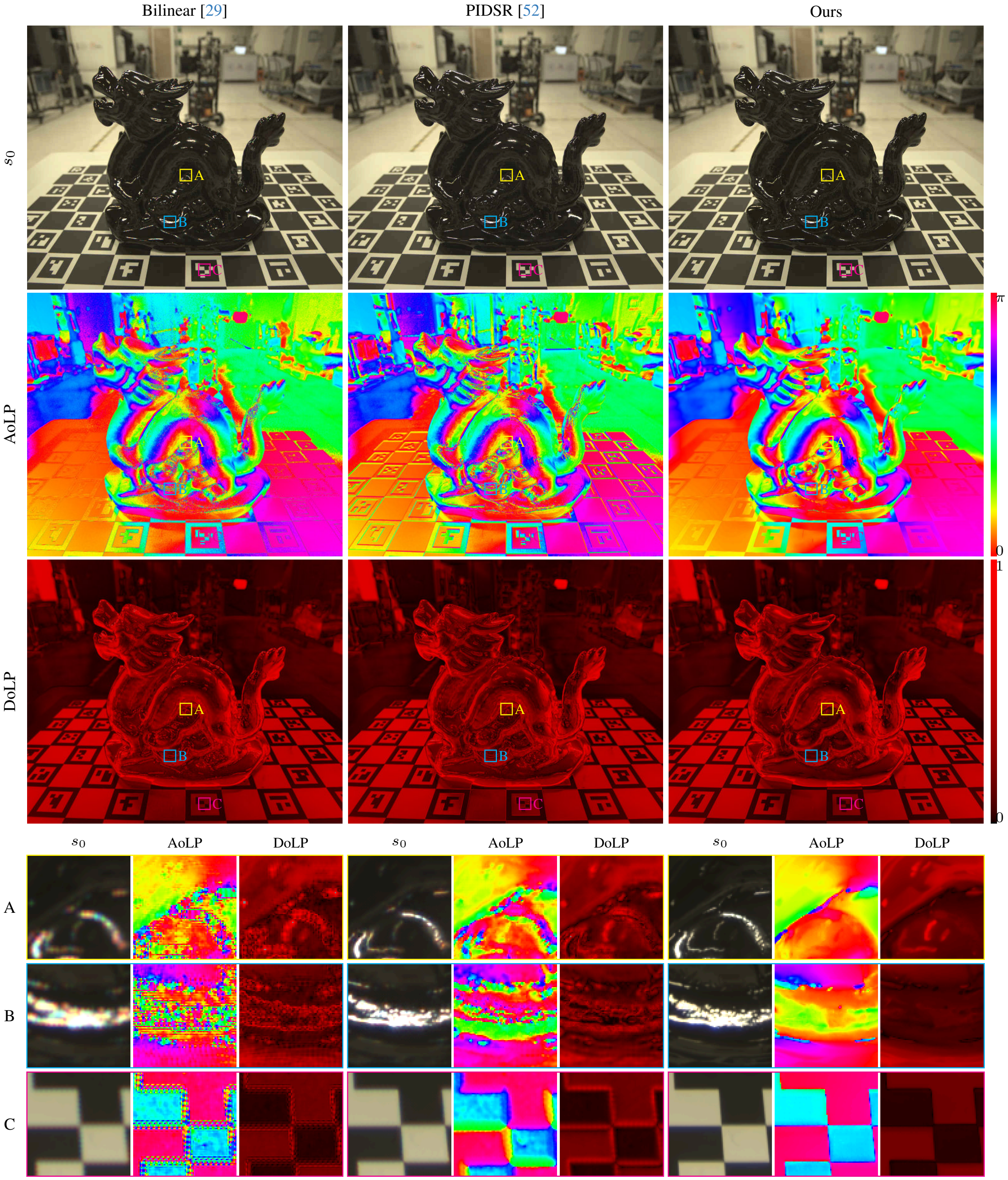}
  \caption{\textbf{RAW-to-Stokes reconstruction of BlackDragon from PISR~\cite{chen2024pisr}.}}
  \label{fig:supp-real-pisr-blackdragon-025}
\end{figure}

\clearpage

\begin{figure}[p]
  \centering
  \input{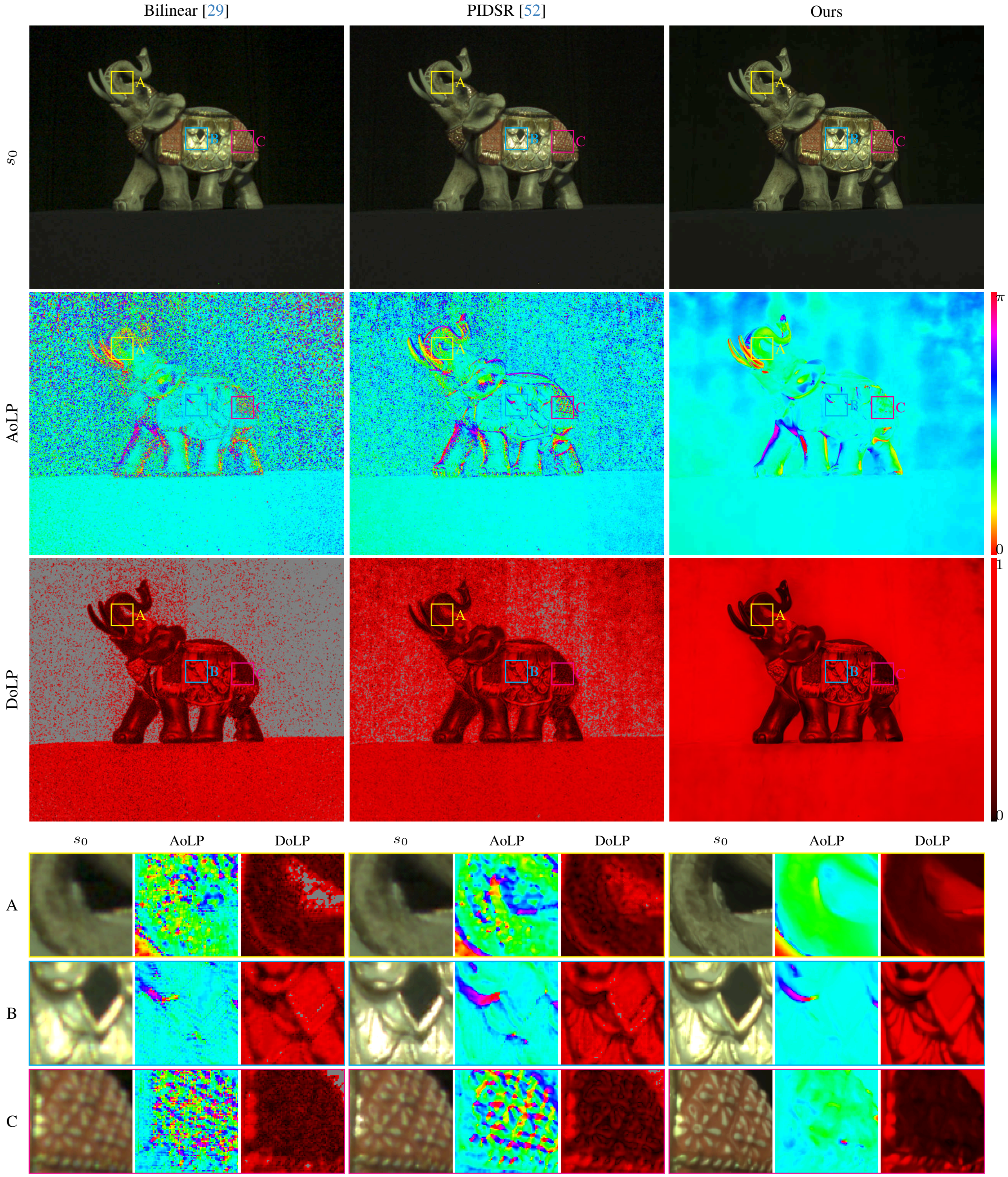}
  \caption{\textbf{RAW-to-Stokes reconstruction of the Elephant scene from the Display Inverse Rendering dataset~\cite{choi2025real}.}}
  \label{fig:supp-real-dir-elephant-072}
\end{figure}

\endgroup

\clearpage
\section{Downstream applications}
\label{sec:supp-downstream-applications}

We demonstrate that Raw2Stokes reconstructions serve as direct inputs for two polarization-based inverse-rendering pipelines, PANDORA~\cite{dave2022pandora} and PDIR~\cite{choi2026snapshot}. In both applications, we hold all downstream calibration, object masks, and optimization or network settings identical across the compared reconstruction methods, processing each RAW frame independently.

\subsection{Multi-view inverse rendering}
\label{sec:supp-downstream-multiview}

\begin{figure}[!htbp]
    \centering
    \input{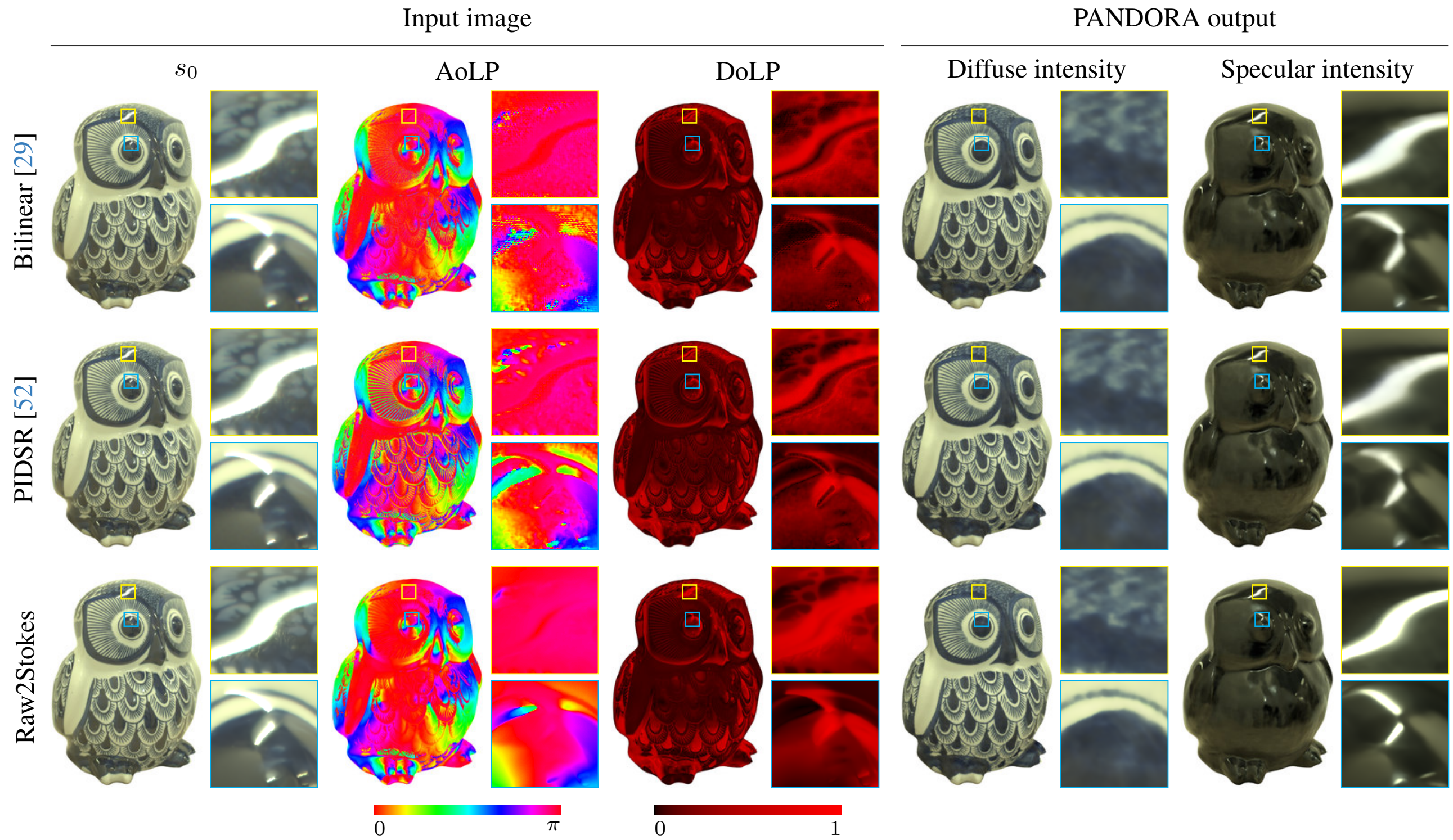}
    \setlength{\abovecaptionskip}{3pt}
    \caption{\textbf{Multi-view inverse rendering with PANDORA on Owl.} Raw2Stokes produces smoother AoLP and DoLP inputs and predicts a higher DoLP in the clipped specular highlight on the head than Bilinear and PIDSR, yielding a slightly tighter specular highlight while overall diffuse and specular decompositions remain visually close.}
    \label{fig:supp-pandora-owl}
\end{figure}

We evaluate PANDORA~\cite{dave2022pandora} on the ceramic Owl scene by reconstructing color Stokes images from the same 35 RAW views using Bilinear~\cite{maeda2019polanalyser}, PIDSR~\cite{zhou2025pidsr}, and Raw2Stokes, then optimizing PANDORA separately on each set under identical camera poses, masks, and optimization settings. As shown in \cref{fig:supp-pandora-owl}, Raw2Stokes produces noticeably smoother AoLP and DoLP inputs than Bilinear and predicts a higher DoLP in the saturated specular highlight on the owl's head, where clipping artificially lowers the Bilinear DoLP estimate. In the resulting PANDORA decompositions, the specular highlight on the head appears slightly tighter and stronger with Raw2Stokes, while the overall diffuse and specular outputs remain visually close for Bilinear and Raw2Stokes.

\FloatBarrier

\subsection{Display-based inverse rendering}
\label{sec:supp-downstream-display}

Display-based snapshot inverse rendering encodes distinct illumination directions across RGB display channels to recover shape and reflectance from a single view. Since the object is illuminated by polarized display light, this setting closely resembles the controlled scenes in HDR-Polar1K. We evaluate PDIR~\cite{choi2026snapshot} on the metallic Case scene by reconstructing the Stokes inputs with Bilinear, PIDSR, and Raw2Stokes and feeding them into the same pretrained PDIR network at \(768\mathord{\times}768\) resolution.

\begin{figure}[!htbp]
  \centering
  \input{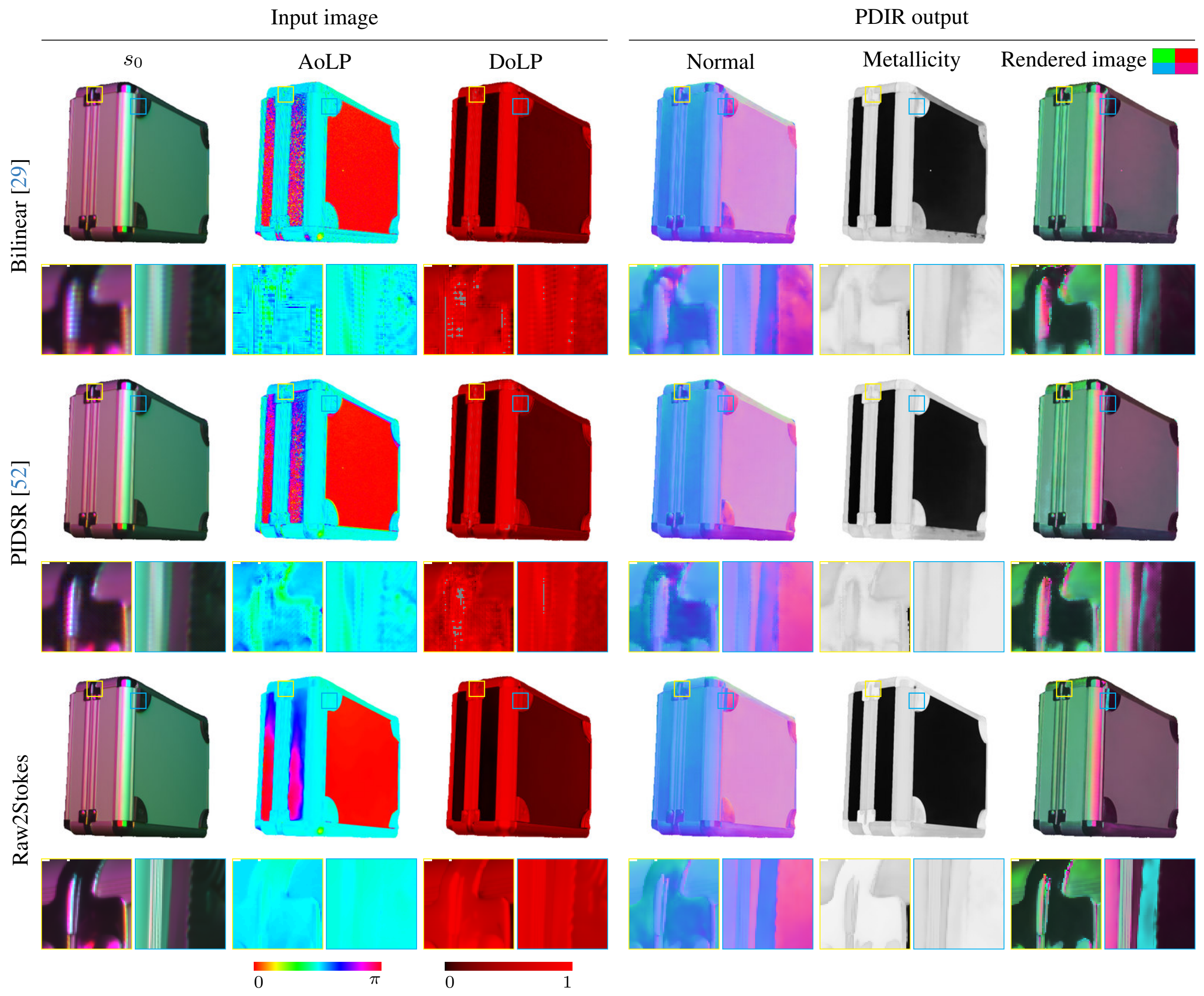}
  \setlength{\abovecaptionskip}{3pt}
  \caption{\textbf{Display-based inverse rendering with PDIR on Case.} On sharp metallic reflections, Bilinear and PIDSR introduce demosaicing artifacts that propagate into PDIR's surface normals, metallicity, and novel-display relighting, whereas Raw2Stokes yields noticeably cleaner inputs and downstream reconstructions with the same pretrained PDIR network.}
  \label{fig:supp-pdir-case}
\end{figure}

As shown in \cref{fig:supp-pdir-case}, this scene exhibits strong metallic reflections and sharp edges, so both Bilinear and PIDSR introduce pronounced demosaicing artifacts in the reconstructed intensity, AoLP, and DoLP maps. These upstream artifacts propagate directly into the downstream predictions. Using Raw2Stokes yields visibly cleaner surface normals, metallicity maps, and relit renderings, even though the pretrained PDIR network was never trained on Raw2Stokes outputs.

\FloatBarrier


{\small
 \begin{thebibliography}{57}
\providecommand{\natexlab}[1]{#1}
\providecommand{\url}[1]{\texttt{#1}}
\expandafter\ifx\csname urlstyle\endcsname\relax
  \providecommand{\doi}[1]{doi: #1}\else
  \providecommand{\doi}{doi: \begingroup \urlstyle{rm}\Url}\fi

\bibitem[Abdelhamed et~al.(2018)Abdelhamed, Lin, and Brown]{abdelhamed2018sidd}
Abdelrahman Abdelhamed, Stephen Lin, and Michael~S. Brown.
\newblock A high-quality denoising dataset for smartphone cameras.
\newblock In \emph{Proceedings of the IEEE Conference on Computer Vision and
  Pattern Recognition (CVPR)}, pages 1692--1700, 2018.

\bibitem[Abubakar et~al.(2018)Abubakar, Zhao, Li, Takruri, Bastaki, and
  Bermak]{abubakar2018bm3d}
Abubakar Abubakar, Xiaojin Zhao, Shiting Li, Maen Takruri, Eesa Bastaki, and
  Amine Bermak.
\newblock A block-matching and 3-d filtering algorithm for {Gaussian} noise in
  {DoFP} polarization images.
\newblock \emph{IEEE Sensors Journal}, 18\penalty0 (18):\penalty0 7429--7435,
  2018.

\bibitem[Baek and Heide(2021)]{baek2021transport}
Seung-Hwan Baek and Felix Heide.
\newblock Polarimetric spatio-temporal light transport probing.
\newblock \emph{ACM Transactions on Graphics}, 40\penalty0 (6):\penalty0 1--18,
  2021.

\bibitem[Bigu{\'e} et~al.(2023)Bigu{\'e}, Foulonneau, and
  Lapray]{bigue2023production}
Laurent Bigu{\'e}, Alban Foulonneau, and Pierre-Jean Lapray.
\newblock Production of high-resolution reference polarization images from real
  world scenes.
\newblock In \emph{Polarization Science and Remote Sensing XI}, page 126900B,
  2023.

\bibitem[Charbonnier et~al.(1997)Charbonnier, Blanc-F{\'e}raud, Aubert, and
  Barlaud]{charbonnier1997deterministic}
Pierre Charbonnier, Laure Blanc-F{\'e}raud, Gilles Aubert, and Michel Barlaud.
\newblock Deterministic edge-preserving regularization in computed imaging.
\newblock \emph{IEEE Transactions on Image Processing}, 6\penalty0
  (2):\penalty0 298--311, 1997.

\bibitem[Choi et~al.(2026)Choi, Moon, Kang, Chung, Kim, Nam, and
  Baek]{choi2026snapshot}
Seokjun Choi, Yunseong Moon, Kaizhang Kang, Hoon-Gyu Chung, Jin-Nyeong Kim,
  Giljoo Nam, and Seung-Hwan Baek.
\newblock Snapshot polarimetric display inverse rendering.
\newblock \emph{ACM Transactions on Graphics}, 45\penalty0 (6):\penalty0 1--13,
  2026.

\bibitem[Cui et~al.(2017)Cui, Gu, Shi, Tan, and Kautz]{cui2017polarimetric}
Zhaopeng Cui, Jinwei Gu, Boxin Shi, Ping Tan, and Jan Kautz.
\newblock Polarimetric multi-view stereo.
\newblock In \emph{Proceedings of the IEEE Conference on Computer Vision and
  Pattern Recognition (CVPR)}, pages 1558--1567, 2017.

\bibitem[Dave et~al.(2025)Dave, Zhang, Young, Raskar, Heidrich, and
  Veeraraghavan]{dave2025nest}
Akshat Dave, Tianyi Zhang, Aaron Young, Ramesh Raskar, Wolfgang Heidrich, and
  Ashok Veeraraghavan.
\newblock {NeST}: Neural stress tensor tomography by leveraging {3D}
  photoelasticity.
\newblock \emph{ACM Transactions on Graphics}, 44\penalty0 (2):\penalty0 1--18,
  2025.

\bibitem[Debevec and Malik(1997)]{debevec1997recovering}
Paul~E. Debevec and Jitendra Malik.
\newblock Recovering high dynamic range radiance maps from photographs.
\newblock In \emph{Proceedings of the 24th Annual Conference on Computer
  Graphics and Interactive Techniques}, pages 369--378, 1997.

\bibitem[Dumoulin et~al.(2024)Dumoulin, Lapray, Foulonneau, and
  Bigu{\'e}]{dumoulin2024impact}
Ronan Dumoulin, Pierre-Jean Lapray, Alban Foulonneau, and Laurent Bigu{\'e}.
\newblock Impact of interchannel misalignment on the performance of demosaicing
  algorithms dedicated to polarization images.
\newblock In \emph{Unconventional Optical Imaging IV}, page 129960V, 2024.

\bibitem[Dumoulin et~al.(2025)Dumoulin, Spote, Lapray, Thomas, and
  Farup]{dumoulin2025lmmse}
Ronan Dumoulin, Alexandra Spote, Pierre-Jean Lapray, Jean-Baptiste Thomas, and
  Ivar Farup.
\newblock Color--polarization filter array image demosaicing: Linear minimum
  mean squared error augmented by anisotropic diffusion.
\newblock \emph{Journal of Electronic Imaging}, 34\penalty0 (3):\penalty0
  033044, 2025.

\bibitem[Foi et~al.(2008)Foi, Trimeche, Katkovnik, and
  Egiazarian]{foi2008poisson}
Alessandro Foi, Mejdi Trimeche, Vladimir Katkovnik, and Karen Egiazarian.
\newblock Practical {Poissonian-Gaussian} noise modeling and fitting for
  single-image raw-data.
\newblock \emph{IEEE Transactions on Image Processing}, 17\penalty0
  (10):\penalty0 1737--1754, 2008.

\bibitem[Gao and Gruev(2011)]{gao2011bilinear}
Shengkui Gao and Viktor Gruev.
\newblock Bilinear and bicubic interpolation methods for division of focal
  plane polarimeters.
\newblock \emph{Optics Express}, 19\penalty0 (27):\penalty0 26161--26173, 2011.

\bibitem[Gharbi et~al.(2016)Gharbi, Chaurasia, Paris, and
  Durand]{gharbi2016deep}
Micha{\"e}l Gharbi, Gaurav Chaurasia, Sylvain Paris, and Fr{\'e}do Durand.
\newblock Deep joint demosaicking and denoising.
\newblock \emph{ACM Transactions on Graphics}, 35\penalty0 (6):\penalty0
  191:1--191:12, 2016.

\bibitem[Han et~al.(2024)Han, Guo, Fukai, Santo, Shi, Okura, Ma, and
  Jia]{han2024nersp}
Yufei Han, Heng Guo, Koki Fukai, Hiroaki Santo, Boxin Shi, Fumio Okura, Zhanyu
  Ma, and Yunpeng Jia.
\newblock {NeRSP}: Neural {3D} reconstruction for reflective objects with
  sparse polarized images.
\newblock In \emph{Proceedings of the IEEE/CVF Conference on Computer Vision
  and Pattern Recognition (CVPR)}, pages 11821--11830, 2024.

\bibitem[Hwang et~al.(2022)Hwang, Jeon, Mu{\~n}oz, Gutierrez, Tong, and
  Kim]{hwang2022sparse}
Inseung Hwang, Daniel~S. Jeon, Adolfo Mu{\~n}oz, Diego Gutierrez, Xin Tong, and
  Min~H. Kim.
\newblock Sparse ellipsometry: Portable acquisition of polarimetric {SVBRDF}
  and shape with unstructured flash photography.
\newblock \emph{ACM Transactions on Graphics}, 41\penalty0 (4):\penalty0
  133:1--133:14, 2022.

\bibitem[Khashabi et~al.(2014)Khashabi, Nowozin, Jancsary, and
  Fitzgibbon]{khashabi2014joint}
Daniel Khashabi, Sebastian Nowozin, Jeremy Jancsary, and Andrew~W. Fitzgibbon.
\newblock Joint demosaicing and denoising via learned nonparametric random
  fields.
\newblock \emph{IEEE Transactions on Image Processing}, 23\penalty0
  (12):\penalty0 4968--4981, 2014.

\bibitem[Lapray(2020)]{lapray2020exploiting}
Pierre-Jean Lapray.
\newblock Exploiting redundancy in color-polarization filter array images for
  dynamic range enhancement.
\newblock \emph{Optics Letters}, 45\penalty0 (19):\penalty0 5530--5533, 2020.

\bibitem[Lei et~al.(2020)Lei, Huang, Zhang, Yan, Sun, and
  Chen]{lei2020polarized}
Chenyang Lei, Xuhua Huang, Mengdi Zhang, Qiong Yan, Wenxiu Sun, and Qifeng
  Chen.
\newblock Polarized reflection removal with perfect alignment in the wild.
\newblock In \emph{Proceedings of the IEEE/CVF Conference on Computer Vision
  and Pattern Recognition (CVPR)}, pages 1750--1758, 2020.

\bibitem[Li et~al.(2024)Li, Ono, Uemori, Mihara, Gatto, Nagahara, and
  Moriuchi]{li2024neisf}
Chenhao Li, Taishi Ono, Takeshi Uemori, Hajime Mihara, Alexander Gatto, Hajime
  Nagahara, and Yusuke Moriuchi.
\newblock {NeISF}: Neural incident {Stokes} field for geometry and material
  estimation.
\newblock In \emph{Proceedings of the IEEE/CVF Conference on Computer Vision
  and Pattern Recognition (CVPR)}, pages 21434--21445, 2024.

\bibitem[Li et~al.(2026)Li, Luo, Zhang, and Yang]{li2026pugdiff}
Chenggong Li, Yidong Luo, Junchao Zhang, and Degui Yang.
\newblock Polarization uncertainty-guided diffusion model for color
  polarization image demosaicking.
\newblock \emph{Proceedings of the AAAI Conference on Artificial Intelligence},
  40\penalty0 (8):\penalty0 6028--6036, 2026.

\bibitem[Li et~al.(2023)Li, Wang, Goudail, Zhao, and Pan]{li2023joint}
Ning Li, Binglu Wang, Fran{\c{c}}ois Goudail, Yongqiang Zhao, and Quan Pan.
\newblock Joint denoising-demosaicking network for long-wave infrared
  division-of-focal-plane polarization images with mixed noise level
  estimation.
\newblock \emph{IEEE Transactions on Image Processing}, 32:\penalty0
  5961--5976, 2023.

\bibitem[Li et~al.(2020)Li, Li, Lin, Guo, Yang, Yue, Li, Li, Cheng, Hu, and
  Liu]{li2020polarimetricdenoising}
Xiaobo Li, Haiyu Li, Yang Lin, Jianhua Guo, Jingyu Yang, Huanjing Yue, Kun Li,
  Chuan Li, Zhenzhou Cheng, Haofeng Hu, and Tiegen Liu.
\newblock Learning-based denoising for polarimetric images.
\newblock \emph{Optics Express}, 28\penalty0 (11):\penalty0 16309--16321, 2020.

\bibitem[Liang et~al.(2022{\natexlab{a}})Liang, Guo, and Liu]{liang2022bm3d}
Jian-An Liang, Ya-fei Guo, and Bin Liu.
\newblock {BM3D}-based denoising method for color polarization filter array.
\newblock \emph{Optics Express}, 30\penalty0 (12):\penalty0 22107--22122,
  2022{\natexlab{a}}.

\bibitem[Liang et~al.(2022{\natexlab{b}})Liang, Wakaki, Nobuhara, and
  Nishino]{liang2022multimodal}
Yupeng Liang, Ryosuke Wakaki, Shohei Nobuhara, and Ko Nishino.
\newblock Multimodal material segmentation.
\newblock In \emph{Proceedings of the IEEE/CVF Conference on Computer Vision
  and Pattern Recognition (CVPR)}, pages 19800--19808, 2022{\natexlab{b}}.

\bibitem[Liu et~al.(2023{\natexlab{a}})Liu, Li, Cheng, Liu, Zhai, and
  Hu]{liu2023pm3dcnn}
Hedong Liu, Xiaobo Li, Zhenzhou Cheng, Tiegen Liu, Jingsheng Zhai, and Haofeng
  Hu.
\newblock Polarization maintaining 3-d convolutional neural network for color
  polarimetric images denoising.
\newblock \emph{IEEE Transactions on Instrumentation and Measurement},
  72:\penalty0 1--9, 2023{\natexlab{a}}.

\bibitem[Liu et~al.(2023{\natexlab{b}})Liu, Li, Cheng, Liu, Zhai, and
  Hu]{liu2023pol2pol}
Hedong Liu, Xiaobo Li, Zhenzhou Cheng, Tiegen Liu, Jingsheng Zhai, and Haofeng
  Hu.
\newblock {Pol2Pol}: Self-supervised polarimetric image denoising.
\newblock \emph{Optics Letters}, 48\penalty0 (18):\penalty0 4821--4824,
  2023{\natexlab{b}}.

\bibitem[Liu et~al.(2023{\natexlab{c}})Liu, Zhang, Sun, Liang, Zeng, and
  Zhang]{liu2023mobilehdr}
Shuaizheng Liu, Xindong Zhang, Lingchen Sun, Zhetong Liang, Hui Zeng, and Lei
  Zhang.
\newblock Joint {HDR} denoising and fusion: A real-world mobile {HDR} image
  dataset.
\newblock In \emph{Proceedings of the IEEE/CVF Conference on Computer Vision
  and Pattern Recognition (CVPR)}, pages 13966--13975, 2023{\natexlab{c}}.

\bibitem[Lu et~al.(2024)Lu, Tian, Su, Luo, Zhang, and Hao]{lu2024hybrid}
Yang Lu, Jiandong Tian, Yiming Su, Yidong Luo, Junchao Zhang, and Chunhui Hao.
\newblock A hybrid polarization image demosaicking algorithm based on
  inter-channel correlation.
\newblock \emph{IEEE Transactions on Computational Imaging}, 10:\penalty0
  1400--1413, 2024.

\bibitem[Luo et~al.(2024)Luo, Zhang, Shao, Tian, and Ma]{luo2024nlcsr}
Yidong Luo, Junchao Zhang, Jianbo Shao, Jiandong Tian, and Jiayi Ma.
\newblock Learning a non-locally regularized convolutional sparse
  representation for joint chromatic and polarimetric demosaicking.
\newblock \emph{IEEE Transactions on Image Processing}, 33:\penalty0
  5029--5044, 2024.

\bibitem[Luo et~al.(2026)Luo, Wu, Li, Wang, Yuan, Yang, and
  Zhang]{luo2026updem}
Yidong Luo, Caiyun Wu, Chenggong Li, Ping Wang, Xin Yuan, Kailun Yang, and
  Junchao Zhang.
\newblock {UP-Dem}: Deep unrolling convolutional sparse coding for color
  polarization image demosaicking.
\newblock \emph{IEEE Transactions on Image Processing}, 35:\penalty0
  10152--10166, 2026.

\bibitem[Maeda(2019)]{maeda2019polanalyser}
Ryota Maeda.
\newblock {Polanalyser}: Polarization image analysis tool.
\newblock \url{https://github.com/elerac/polanalyser}, 2019.
\newblock Software.

\bibitem[Maeda et~al.(2025)Maeda, Moon, and Baek]{maeda2025event}
Ryota Maeda, Yunseong Moon, and Seung-Hwan Baek.
\newblock Event ellipsometer: Event-based {Mueller}-matrix video imaging.
\newblock In \emph{Proceedings of the IEEE/CVF Conference on Computer Vision
  and Pattern Recognition (CVPR)}, pages 21804--21813, 2025.

\bibitem[Maruyama et~al.(2018)Maruyama, Terada, Yamazaki, Uesaka, Nakamura,
  Matoba, Komori, Ohba, Arakawa, Hirasawa, Kondo, Murayama, Akiyama, Oike,
  Sato, and Ezaki]{maruyama2018backilluminated}
Yasushi Maruyama, Takashi Terada, Tomohiro Yamazaki, Yusuke Uesaka, Motoaki
  Nakamura, Yoshihisa Matoba, Kenta Komori, Yoshiyuki Ohba, Shinichi Arakawa,
  Yasutaka Hirasawa, Yuhi Kondo, Jun Murayama, Kentaro Akiyama, Yusuke Oike,
  Shuzo Sato, and Takayuki Ezaki.
\newblock 3.2-{MP} back-illuminated polarization image sensor with
  four-directional air-gap wire grid and 2.5-$\mu$m pixels.
\newblock \emph{IEEE Transactions on Electron Devices}, 65\penalty0
  (6):\penalty0 2544--2551, 2018.

\bibitem[Mei et~al.(2022)Mei, Dong, Dong, Yang, Baek, Heide, Peers, Wei, and
  Yang]{mei2022glass}
Haiyang Mei, Bo Dong, Wen Dong, Jiaxi Yang, Seung-Hwan Baek, Felix Heide,
  Pieter Peers, Xiaopeng Wei, and Xin Yang.
\newblock Glass segmentation using intensity and spectral polarization cues.
\newblock In \emph{Proceedings of the IEEE/CVF Conference on Computer Vision
  and Pattern Recognition (CVPR)}, pages 12622--12631, 2022.

\bibitem[Mitchell and Netravali(1988)]{mitchell1988reconstruction}
Don~P. Mitchell and Arun~N. Netravali.
\newblock Reconstruction filters in computer graphics.
\newblock \emph{ACM SIGGRAPH Computer Graphics}, 22\penalty0 (4):\penalty0
  221--228, 1988.

\bibitem[Morimatsu et~al.(2020)Morimatsu, Monno, Tanaka, and
  Okutomi]{morimatsu2020eari}
Miki Morimatsu, Yusuke Monno, Masayuki Tanaka, and Masatoshi Okutomi.
\newblock Monochrome and color polarization demosaicking using edge-aware
  residual interpolation.
\newblock In \emph{Proceedings of the IEEE International Conference on Image
  Processing (ICIP)}, pages 2571--2575, 2020.

\bibitem[Morimatsu et~al.(2021)Morimatsu, Monno, Tanaka, and
  Okutomi]{morimatsu2021igri}
Miki Morimatsu, Yusuke Monno, Masayuki Tanaka, and Masatoshi Okutomi.
\newblock Monochrome and color polarization demosaicking based on
  intensity-guided residual interpolation.
\newblock \emph{IEEE Sensors Journal}, 21\penalty0 (23):\penalty0 26985--26996,
  2021.

\bibitem[Nguyen et~al.(2022)Nguyen, Tanaka, Monno, and
  Okutomi]{nguyen2022tcpdnet}
Vy Nguyen, Masayuki Tanaka, Yusuke Monno, and Masatoshi Okutomi.
\newblock Two-step color-polarization demosaicking network.
\newblock In \emph{Proceedings of the IEEE International Conference on Image
  Processing (ICIP)}, pages 1011--1015, 2022.

\bibitem[Perez et~al.(2018)Perez, Strub, de~Vries, Dumoulin, and
  Courville]{perez2018film}
Ethan Perez, Florian Strub, Harm de Vries, Vincent Dumoulin, and Aaron
  Courville.
\newblock {FiLM}: Visual reasoning with a general conditioning layer.
\newblock In \emph{Proceedings of the AAAI Conference on Artificial
  Intelligence}, pages 3942--3951, 2018.

\bibitem[Qiu et~al.(2021)Qiu, Fu, Wang, and Heidrich]{qiu2021linear}
Simeng Qiu, Qiang Fu, Congli Wang, and Wolfgang Heidrich.
\newblock Linear polarization demosaicking for monochrome and colour
  polarization focal plane arrays.
\newblock \emph{Computer Graphics Forum}, 40\penalty0 (6):\penalty0 77--89,
  2021.

\bibitem[Rahman et~al.(2025)Rahman, Monno, Tanaka, and
  Okutomi]{rahman2025polarization}
Muhamad Daniel Ariff Bin~Abdul Rahman, Yusuke Monno, Masayuki Tanaka, and
  Masatoshi Okutomi.
\newblock Polarization denoising and demosaicking: Dataset and baseline method.
\newblock In \emph{Proceedings of the IEEE International Conference on Image
  Processing (ICIP)}, pages 2724--2729, 2025.

\bibitem[Ratliff et~al.(2007)Ratliff, Tyo, Boger, Black, Bowers, and
  Fetrow]{ratliff2007dead}
Bradley~M. Ratliff, J.~Scott Tyo, James~K. Boger, Wiley~T. Black, David~L.
  Bowers, and Matthew~P. Fetrow.
\newblock Dead pixel replacement in {LWIR} microgrid polarimeters.
\newblock \emph{Optics Express}, 15\penalty0 (12):\penalty0 7596--7609, 2007.

\bibitem[Su et~al.(2025)Su, Hao, Chen, Lu, and Tian]{su2025universal}
Yiming Su, Chunhui Hao, Wen Chen, Yang Lu, and Jiandong Tian.
\newblock Universal polarization demosaicking using {Newton} polynomial and
  residual interpolation.
\newblock \emph{IEEE Sensors Journal}, 25\penalty0 (7):\penalty0 11508--11520,
  2025.

\bibitem[Tedla et~al.(2025)Tedla, Punnappurath, Zhao, and
  Brown]{tedla2025examining}
SaiKiran Tedla, Abhijith Punnappurath, Luxi Zhao, and Michael~S. Brown.
\newblock Examining joint demosaicing and denoising for single-, quad-, and
  nona-{Bayer} patterns.
\newblock In \emph{Proceedings of the IEEE International Conference on
  Computational Photography (ICCP)}, pages 1--10, 2025.

\bibitem[Wang et~al.(2026)Wang, Zhang, Li, Zheng, Ru, and Xu]{wang2026swincpd}
Tianyi Wang, Xiangyue Zhang, Junlin Li, Yubo Zheng, Jingyu Ru, and Hongli Xu.
\newblock {SwinCPD}: A {Swin Transformer}-based network for color-polarization
  demosaicing.
\newblock \emph{IEEE Sensors Journal}, 26\penalty0 (1):\penalty0 637--651,
  2026.

\bibitem[Wen et~al.(2019)Wen, Zheng, Lu, and Zhao]{wen2019cpdnet}
Sijia Wen, Yinqiang Zheng, Feng Lu, and Qinping Zhao.
\newblock Convolutional demosaicing network for joint chromatic and
  polarimetric imagery.
\newblock \emph{Optics Letters}, 44\penalty0 (22):\penalty0 5646--5649, 2019.

\bibitem[Wen et~al.(2021)Wen, Zheng, and Lu]{wen2021sparse}
Sijia Wen, Yinqiang Zheng, and Feng Lu.
\newblock A sparse representation based joint demosaicing method for
  single-chip polarized color sensor.
\newblock \emph{IEEE Transactions on Image Processing}, 30:\penalty0
  4171--4182, 2021.

\bibitem[Wu et~al.(2026)Wu, Luo, Li, Zhang, and Liang]{wu2026deep}
Caiyun Wu, Yidong Luo, Chenggong Li, Junchao Zhang, and Buge Liang.
\newblock Deep unfolding with learned regularization for color polarization
  demosaicking.
\newblock \emph{Advanced Imaging}, 3\penalty0 (3):\penalty0 031003, 2026.

\bibitem[Yang et~al.(2026)Yang, Liu, Lu, Liu, and Tian]{yang2026zeropie}
Zhenshuo Yang, Zhiyuan Liu, Yang Lu, Jiawei Liu, and Jiandong Tian.
\newblock {ZeroPIE}: Zero-reference polarization-aware low-light enhancement
  via physically derived illumination-invariant priors.
\newblock \emph{IEEE Transactions on Pattern Analysis and Machine
  Intelligence}, 48\penalty0 (11):\penalty0 15216--15233, 2026.

\bibitem[Yao et~al.(2025)Yao, Wang, Tam, Li, Xue, and Gu]{yao2025polarfree}
Mingde Yao, Menglu Wang, King-Man Tam, Lingen Li, Tianfan Xue, and Jinwei Gu.
\newblock {PolarFree}: Polarization-based reflection-free imaging.
\newblock In \emph{Proceedings of the IEEE/CVF Conference on Computer Vision
  and Pattern Recognition (CVPR)}, pages 10890--10899, 2025.

\bibitem[Zhang et~al.(2017)Zhang, Luo, Liang, Zhou, Hui, and
  Chang]{zhang2017pca}
Junchao Zhang, Haibo Luo, Rongguang Liang, Wei Zhou, Bin Hui, and Zheng Chang.
\newblock {PCA}-based denoising method for division of focal plane
  polarimeters.
\newblock \emph{Optics Express}, 25\penalty0 (3):\penalty0 2391--2400, 2017.

\bibitem[Zhang et~al.(2026)Zhang, Monno, Tanaka, and Okutomi]{zhang2026cpddnet}
Qihang Zhang, Yusuke Monno, Masayuki Tanaka, and Masatoshi Okutomi.
\newblock {CPDDNet}: Color-polarization denoising and demosaicking network.
\newblock In \emph{Proceedings of the IEEE International Conference on Image
  Processing (ICIP)}, pages 1--6, 2026.

\bibitem[Zhang et~al.(2018)Zhang, Li, Li, Wang, Zhong, and Fu]{zhang2018rcan}
Yulun Zhang, Kunpeng Li, Kai Li, Lichen Wang, Bineng Zhong, and Yun Fu.
\newblock Image super-resolution using very deep residual channel attention
  networks.
\newblock In \emph{Proceedings of the European Conference on Computer Vision
  (ECCV)}, pages 286--301, 2018.

\bibitem[Zheng et~al.(2024)Zheng, Zhang, Wu, Ji, and Ru]{zheng2024scfnet}
Yubo Zheng, Xiangyue Zhang, Chengdong Wu, Peng Ji, and Jingyu Ru.
\newblock Color polarization imaging demosaicing based on {Stokes} vector
  information complementation and fusion.
\newblock \emph{Optics Express}, 32\penalty0 (25):\penalty0 44049--44066, 2024.

\bibitem[Zhou et~al.(2023)Zhou, Teng, Lyu, Li, Xu, and Shi]{zhou2023lowlight}
Chu Zhou, Minggui Teng, Youwei Lyu, Si Li, Chao Xu, and Boxin Shi.
\newblock Polarization-aware low-light image enhancement.
\newblock \emph{Proceedings of the AAAI Conference on Artificial Intelligence},
  37\penalty0 (3):\penalty0 3742--3750, 2023.

\bibitem[Zhou et~al.(2025)Zhou, Zhou, Lyu, Guo, Ma, Shi, and
  Sato]{zhou2025pidsr}
Shuangfan Zhou, Chu Zhou, Youwei Lyu, Heng Guo, Zhanyu Ma, Boxin Shi, and Imari
  Sato.
\newblock {PIDSR}: Complementary polarized image demosaicing and
  super-resolution.
\newblock In \emph{Proceedings of the IEEE/CVF Conference on Computer Vision
  and Pattern Recognition (CVPR)}, pages 16081--16090, 2025.

\end{thebibliography}

}


\clearpage

{\small\begin{thebibliography}{52}
\providecommand{\natexlab}[1]{#1}
\providecommand{\url}[1]{\texttt{#1}}
\expandafter\ifx\csname urlstyle\endcsname\relax
  \providecommand{\doi}[1]{doi: #1}\else
  \providecommand{\doi}{doi: \begingroup \urlstyle{rm}\Url}\fi

\bibitem[Abdelhamed et~al.(2018)Abdelhamed, Lin, and Brown]{S-abdelhamed2018sidd}
Abdelrahman Abdelhamed, Stephen Lin, and Michael~S. Brown.
\newblock A high-quality denoising dataset for smartphone cameras.
\newblock In \emph{Proceedings of the IEEE Conference on Computer Vision and
  Pattern Recognition (CVPR)}, pages 1692--1700, 2018.

\bibitem[Afifi et~al.(2021)Afifi, Barron, LeGendre, Tsai, and
  Bleibel]{S-afifi2021crosscamera}
Mahmoud Afifi, Jonathan~T. Barron, Chloe LeGendre, Yun-Ta Tsai, and Francois
  Bleibel.
\newblock Cross-camera convolutional color constancy.
\newblock In \emph{Proceedings of the IEEE/CVF International Conference on
  Computer Vision (ICCV)}, 2021.

\bibitem[Baek and Heide(2021)]{S-baek2021transport}
Seung-Hwan Baek and Felix Heide.
\newblock Polarimetric spatio-temporal light transport probing.
\newblock \emph{ACM Transactions on Graphics}, 40\penalty0 (6):\penalty0 1--18,
  2021.

\bibitem[Batista et~al.(2025)Batista, Khazem, Adibi, Hutchinson, and
  Pradalier]{S-batista2024potato}
Luis F.~W. Batista, Salim Khazem, Mehran Adibi, Seth Hutchinson, and Cedric
  Pradalier.
\newblock {PoTATO}: A dataset for analyzing polarimetric traces of afloat trash
  objects.
\newblock In \emph{Computer Vision -- ECCV 2024 Workshops}, pages 190--205.
  Springer Nature Switzerland, 2025.

\bibitem[Bigu{\'e} et~al.(2023)Bigu{\'e}, Foulonneau, and
  Lapray]{S-bigue2023production}
Laurent Bigu{\'e}, Alban Foulonneau, and Pierre-Jean Lapray.
\newblock Production of high-resolution reference polarization images from real
  world scenes.
\newblock In \emph{Polarization Science and Remote Sensing XI}, page 126900B,
  2023.

\bibitem[Charbonnier et~al.(1997)Charbonnier, Blanc-F{\'e}raud, Aubert, and
  Barlaud]{S-charbonnier1997deterministic}
Pierre Charbonnier, Laure Blanc-F{\'e}raud, Gilles Aubert, and Michel Barlaud.
\newblock Deterministic edge-preserving regularization in computed imaging.
\newblock \emph{IEEE Transactions on Image Processing}, 6\penalty0
  (2):\penalty0 298--311, 1997.

\bibitem[Chen et~al.(2024)Chen, He, He, and Zhang]{S-chen2024pisr}
Guangcheng Chen, Yicheng He, Li He, and Hong Zhang.
\newblock {PISR}: Polarimetric neural implicit surface reconstruction for
  textureless and specular objects.
\newblock In \emph{European Conference on Computer Vision (ECCV)}, 2024.

\bibitem[Choi et~al.(2025)Choi, Chung, Jeon, Nam, and Baek]{S-choi2025real}
Seokjun Choi, Hoon-Gyu Chung, Yujin Jeon, Giljoo Nam, and Seung-Hwan Baek.
\newblock A real-world display inverse rendering dataset.
\newblock In \emph{Proceedings of the IEEE/CVF International Conference on
  Computer Vision (ICCV)}, 2025.

\bibitem[Choi et~al.(2026)Choi, Moon, Kang, Chung, Kim, Nam, and
  Baek]{S-choi2026snapshot}
Seokjun Choi, Yunseong Moon, Kaizhang Kang, Hoon-Gyu Chung, Jin-Nyeong Kim,
  Giljoo Nam, and Seung-Hwan Baek.
\newblock Snapshot polarimetric display inverse rendering.
\newblock \emph{ACM Transactions on Graphics}, 45\penalty0 (6):\penalty0 1--13,
  2026.

\bibitem[Chu et~al.(2022)Chu, Chen, Chen, and Lu]{S-chu2022improving}
Xiaojie Chu, Liangyu Chen, Chengpeng Chen, and Xin Lu.
\newblock Improving image restoration by revisiting global information
  aggregation.
\newblock In \emph{Proceedings of the European Conference on Computer Vision
  (ECCV)}, pages 53--71, 2022.

\bibitem[Dave et~al.(2022)Dave, Zhao, and Veeraraghavan]{S-dave2022pandora}
Akshat Dave, Yongyi Zhao, and Ashok Veeraraghavan.
\newblock {PANDORA}: Polarization-aided neural decomposition of radiance.
\newblock In \emph{Proceedings of the European Conference on Computer Vision
  (ECCV)}, 2022.

\bibitem[Dave et~al.(2025)Dave, Zhang, Young, Raskar, Heidrich, and
  Veeraraghavan]{S-dave2025nest}
Akshat Dave, Tianyi Zhang, Aaron Young, Ramesh Raskar, Wolfgang Heidrich, and
  Ashok Veeraraghavan.
\newblock {NeST}: Neural stress tensor tomography by leveraging {3D}
  photoelasticity.
\newblock \emph{ACM Transactions on Graphics}, 44\penalty0 (2):\penalty0 1--18,
  2025.

\bibitem[Debevec and Malik(1997)]{S-debevec1997recovering}
Paul~E. Debevec and Jitendra Malik.
\newblock Recovering high dynamic range radiance maps from photographs.
\newblock In \emph{Proceedings of the 24th Annual Conference on Computer
  Graphics and Interactive Techniques}, pages 369--378, 1997.

\bibitem[Dumoulin et~al.(2024)Dumoulin, Lapray, Foulonneau, and
  Bigu{\'e}]{S-dumoulin2024impact}
Ronan Dumoulin, Pierre-Jean Lapray, Alban Foulonneau, and Laurent Bigu{\'e}.
\newblock Impact of interchannel misalignment on the performance of demosaicing
  algorithms dedicated to polarization images.
\newblock In \emph{Unconventional Optical Imaging IV}, page 129960V, 2024.

\bibitem[{European Machine Vision Association}(2021)]{S-emva2021linear}
{European Machine Vision Association}.
\newblock {EMVA Standard 1288}: Standard for characterization of image sensors
  and cameras, release 4.0 linear.
\newblock Technical report, European Machine Vision Association, 2021.

\bibitem[Feng et~al.(2022)Feng, Wang, Wang, and Huang]{S-feng2022learnability}
Hansen Feng, Lizhi Wang, Yuzhi Wang, and Hua Huang.
\newblock Learnability enhancement for low-light raw denoising: Where paired
  real data meets noise modeling.
\newblock In \emph{Proceedings of the ACM International Conference on
  Multimedia}, pages 1436--1444, 2022.

\bibitem[Friedman and Dieng(2023)]{S-friedman2023vendi}
Dan Friedman and Adji~Bousso Dieng.
\newblock The {Vendi Score}: A diversity evaluation metric for machine
  learning.
\newblock \emph{Transactions on Machine Learning Research}, 2023.

\bibitem[Hwang et~al.(2022)Hwang, Jeon, Mu{\~n}oz, Gutierrez, Tong, and
  Kim]{S-hwang2022sparse}
Inseung Hwang, Daniel~S. Jeon, Adolfo Mu{\~n}oz, Diego Gutierrez, Xin Tong, and
  Min~H. Kim.
\newblock Sparse ellipsometry: Portable acquisition of polarimetric {SVBRDF}
  and shape with unstructured flash photography.
\newblock \emph{ACM Transactions on Graphics}, 41\penalty0 (4):\penalty0
  133:1--133:14, 2022.

\bibitem[Kaiser and Schafer(1980)]{S-kaiser1980i0sinh}
James~F. Kaiser and Ronald~W. Schafer.
\newblock On the use of the {$I_0$}-sinh window for spectrum analysis.
\newblock \emph{IEEE Transactions on Acoustics, Speech, and Signal Processing},
  28\penalty0 (1):\penalty0 105--107, 1980.

\bibitem[Khashabi et~al.(2014)Khashabi, Nowozin, Jancsary, and
  Fitzgibbon]{S-khashabi2014joint}
Daniel Khashabi, Sebastian Nowozin, Jeremy Jancsary, and Andrew~W. Fitzgibbon.
\newblock Joint demosaicing and denoising via learned nonparametric random
  fields.
\newblock \emph{IEEE Transactions on Image Processing}, 23\penalty0
  (12):\penalty0 4968--4981, 2014.

\bibitem[Kim et~al.(2023)Kim, Jin, Cho, and Baek]{S-kim2023nespof}
Youngchan Kim, Wonjoon Jin, Sunghyun Cho, and Seung-Hwan Baek.
\newblock Neural spectro-polarimetric fields.
\newblock In \emph{SIGGRAPH Asia 2023 Conference Papers}, pages 1--11, 2023.

\bibitem[Kingma and Ba(2015)]{S-kingma2015adam}
Diederik~P. Kingma and Jimmy Ba.
\newblock Adam: A method for stochastic optimization.
\newblock In \emph{International Conference on Learning Representations
  (ICLR)}, 2015.

\bibitem[Lapray(2020)]{S-lapray2020exploiting}
Pierre-Jean Lapray.
\newblock Exploiting redundancy in color-polarization filter array images for
  dynamic range enhancement.
\newblock \emph{Optics Letters}, 45\penalty0 (19):\penalty0 5530--5533, 2020.

\bibitem[Li et~al.(2025)Li, Luo, Wu, Zhang, Yang, and Zhao]{S-li2025dcpm}
Chenggong Li, Yidong Luo, Caiyun Wu, Junchao Zhang, Degui Yang, and Dangjun
  Zhao.
\newblock Demosaicking customized diffusion model for snapshot polarization
  imaging.
\newblock \emph{Optics \& Laser Technology}, 188:\penalty0 112868, 2025.

\bibitem[Li et~al.(2026)Li, Luo, Zhang, and Yang]{S-li2026pugdiff}
Chenggong Li, Yidong Luo, Junchao Zhang, and Degui Yang.
\newblock Polarization uncertainty-guided diffusion model for color
  polarization image demosaicking.
\newblock \emph{Proceedings of the AAAI Conference on Artificial Intelligence},
  40\penalty0 (8):\penalty0 6028--6036, 2026.

\bibitem[Luo et~al.(2024)Luo, Zhang, Shao, Tian, and Ma]{S-luo2024nlcsr}
Yidong Luo, Junchao Zhang, Jianbo Shao, Jiandong Tian, and Jiayi Ma.
\newblock Learning a non-locally regularized convolutional sparse
  representation for joint chromatic and polarimetric demosaicking.
\newblock \emph{IEEE Transactions on Image Processing}, 33:\penalty0
  5029--5044, 2024.

\bibitem[Luo et~al.(2026{\natexlab{a}})Luo, Li, Song, Wang, Shi, Zhang, and
  Yuan]{S-luo2026polarapp}
Yidong Luo, Chenggong Li, Yunfeng Song, Ping Wang, Boxin Shi, Junchao Zhang,
  and Xin Yuan.
\newblock {PolarAPP}: Beyond polarization demosaicking for polarimetric
  applications.
\newblock In \emph{Proceedings of the European Conference on Computer Vision
  (ECCV)}, pages 115--133, 2026{\natexlab{a}}.

\bibitem[Luo et~al.(2026{\natexlab{b}})Luo, Wu, Li, Wang, Yuan, Yang, and
  Zhang]{S-luo2026updem}
Yidong Luo, Caiyun Wu, Chenggong Li, Ping Wang, Xin Yuan, Kailun Yang, and
  Junchao Zhang.
\newblock {UP-Dem}: Deep unrolling convolutional sparse coding for color
  polarization image demosaicking.
\newblock \emph{IEEE Transactions on Image Processing}, 35:\penalty0
  10152--10166, 2026{\natexlab{b}}.

\bibitem[Maeda(2019)]{S-maeda2019polanalyser}
Ryota Maeda.
\newblock {Polanalyser}: Polarization image analysis tool.
\newblock \url{https://github.com/elerac/polanalyser}, 2019.
\newblock Software.

\bibitem[Maeda et~al.(2025)Maeda, Moon, and Baek]{S-maeda2025event}
Ryota Maeda, Yunseong Moon, and Seung-Hwan Baek.
\newblock Event ellipsometer: Event-based {Mueller}-matrix video imaging.
\newblock In \emph{Proceedings of the IEEE/CVF Conference on Computer Vision
  and Pattern Recognition (CVPR)}, pages 21804--21813, 2025.

\bibitem[Mitchell and Netravali(1988)]{S-mitchell1988reconstruction}
Don~P. Mitchell and Arun~N. Netravali.
\newblock Reconstruction filters in computer graphics.
\newblock \emph{ACM SIGGRAPH Computer Graphics}, 22\penalty0 (4):\penalty0
  221--228, 1988.

\bibitem[Morimatsu et~al.(2020)Morimatsu, Monno, Tanaka, and
  Okutomi]{S-morimatsu2020eari}
Miki Morimatsu, Yusuke Monno, Masayuki Tanaka, and Masatoshi Okutomi.
\newblock Monochrome and color polarization demosaicking using edge-aware
  residual interpolation.
\newblock In \emph{Proceedings of the IEEE International Conference on Image
  Processing (ICIP)}, pages 2571--2575, 2020.

\bibitem[Morimatsu et~al.(2021)Morimatsu, Monno, Tanaka, and
  Okutomi]{S-morimatsu2021igri}
Miki Morimatsu, Yusuke Monno, Masayuki Tanaka, and Masatoshi Okutomi.
\newblock Monochrome and color polarization demosaicking based on
  intensity-guided residual interpolation.
\newblock \emph{IEEE Sensors Journal}, 21\penalty0 (23):\penalty0 26985--26996,
  2021.

\bibitem[Nguyen et~al.(2022)Nguyen, Tanaka, Monno, and
  Okutomi]{S-nguyen2022tcpdnet}
Vy Nguyen, Masayuki Tanaka, Yusuke Monno, and Masatoshi Okutomi.
\newblock Two-step color-polarization demosaicking network.
\newblock In \emph{Proceedings of the IEEE International Conference on Image
  Processing (ICIP)}, pages 1011--1015, 2022.

\bibitem[Oquab et~al.(2024)Oquab, Darcet, Moutakanni, Vo, Szafraniec, Khalidov,
  Fernandez, Haziza, Massa, El-Nouby, Assran, Ballas, Galuba, Howes, Huang, Li,
  Misra, Rabbat, Sharma, Synnaeve, Xu, J{\'e}gou, Mairal, Labatut, Joulin, and
  Bojanowski]{S-oquab2023dinov2}
Maxime Oquab, Timoth{\'e}e Darcet, Th{\'e}o Moutakanni, Huy~V. Vo, Marc
  Szafraniec, Vasil Khalidov, Pierre Fernandez, Daniel Haziza, Francisco Massa,
  Alaaeldin El-Nouby, Mahmoud Assran, Nicolas Ballas, Wojciech Galuba, Russell
  Howes, Po-Yao Huang, Shang-Wen Li, Ishan Misra, Michael Rabbat, Vasu Sharma,
  Gabriel Synnaeve, Hu Xu, Herv{\'e} J{\'e}gou, Julien Mairal, Patrick Labatut,
  Armand Joulin, and Piotr Bojanowski.
\newblock {DINOv2}: Learning robust visual features without supervision.
\newblock \emph{Transactions on Machine Learning Research}, 2024.

\bibitem[Perez et~al.(2018)Perez, Strub, de~Vries, Dumoulin, and
  Courville]{S-perez2018film}
Ethan Perez, Florian Strub, Harm de Vries, Vincent Dumoulin, and Aaron
  Courville.
\newblock {FiLM}: Visual reasoning with a general conditioning layer.
\newblock In \emph{Proceedings of the AAAI Conference on Artificial
  Intelligence}, pages 3942--3951, 2018.

\bibitem[Qiu et~al.(2021)Qiu, Fu, Wang, and Heidrich]{S-qiu2021linear}
Simeng Qiu, Qiang Fu, Congli Wang, and Wolfgang Heidrich.
\newblock Linear polarization demosaicking for monochrome and colour
  polarization focal plane arrays.
\newblock \emph{Computer Graphics Forum}, 40\penalty0 (6):\penalty0 77--89,
  2021.

\bibitem[Rahman et~al.(2025)Rahman, Monno, Tanaka, and
  Okutomi]{S-rahman2025polarization}
Muhamad Daniel Ariff Bin~Abdul Rahman, Yusuke Monno, Masayuki Tanaka, and
  Masatoshi Okutomi.
\newblock Polarization denoising and demosaicking: Dataset and baseline method.
\newblock In \emph{Proceedings of the IEEE International Conference on Image
  Processing (ICIP)}, pages 2724--2729, 2025.

\bibitem[Tedla et~al.(2025)Tedla, Punnappurath, Zhao, and
  Brown]{S-tedla2025examining}
SaiKiran Tedla, Abhijith Punnappurath, Luxi Zhao, and Michael~S. Brown.
\newblock Examining joint demosaicing and denoising for single-, quad-, and
  nona-{Bayer} patterns.
\newblock In \emph{Proceedings of the IEEE International Conference on
  Computational Photography (ICCP)}, pages 1--10, 2025.

\bibitem[Wang et~al.(2026)Wang, Zhang, Li, Zheng, Ru, and Xu]{S-wang2026swincpd}
Tianyi Wang, Xiangyue Zhang, Junlin Li, Yubo Zheng, Jingyu Ru, and Hongli Xu.
\newblock {SwinCPD}: A {Swin Transformer}-based network for color-polarization
  demosaicing.
\newblock \emph{IEEE Sensors Journal}, 26\penalty0 (1):\penalty0 637--651,
  2026.

\bibitem[Wei et~al.(2022)Wei, Fu, Zheng, and Yang]{S-wei2022physics}
Kaixuan Wei, Ying Fu, Yinqiang Zheng, and Jiaolong Yang.
\newblock Physics-based noise modeling for extreme low-light photography.
\newblock \emph{IEEE Transactions on Pattern Analysis and Machine
  Intelligence}, 44\penalty0 (11):\penalty0 8520--8537, 2022.

\bibitem[Wen et~al.(2019)Wen, Zheng, Lu, and Zhao]{S-wen2019cpdnet}
Sijia Wen, Yinqiang Zheng, Feng Lu, and Qinping Zhao.
\newblock Convolutional demosaicing network for joint chromatic and
  polarimetric imagery.
\newblock \emph{Optics Letters}, 44\penalty0 (22):\penalty0 5646--5649, 2019.

\bibitem[Wen et~al.(2021)Wen, Zheng, and Lu]{S-wen2021sparse}
Sijia Wen, Yinqiang Zheng, and Feng Lu.
\newblock A sparse representation based joint demosaicing method for
  single-chip polarized color sensor.
\newblock \emph{IEEE Transactions on Image Processing}, 30:\penalty0
  4171--4182, 2021.

\bibitem[Wu et~al.(2026)Wu, Luo, Li, Zhang, and Liang]{S-wu2026deep}
Caiyun Wu, Yidong Luo, Chenggong Li, Junchao Zhang, and Buge Liang.
\newblock Deep unfolding with learned regularization for color polarization
  demosaicking.
\newblock \emph{Advanced Imaging}, 3\penalty0 (3):\penalty0 031003, 2026.

\bibitem[Xing and Egiazarian(2021)]{S-xing2021endtoend}
Wenzhu Xing and Karen Egiazarian.
\newblock End-to-end learning for joint image demosaicing, denoising and
  super-resolution.
\newblock In \emph{Proceedings of the IEEE/CVF Conference on Computer Vision
  and Pattern Recognition (CVPR)}, pages 3507--3516, 2021.

\bibitem[Yao et~al.(2025)Yao, Wang, Tam, Li, Xue, and Gu]{S-yao2025polarfree}
Mingde Yao, Menglu Wang, King-Man Tam, Lingen Li, Tianfan Xue, and Jinwei Gu.
\newblock {PolarFree}: Polarization-based reflection-free imaging.
\newblock In \emph{Proceedings of the IEEE/CVF Conference on Computer Vision
  and Pattern Recognition (CVPR)}, pages 10890--10899, 2025.

\bibitem[Zhang et~al.(2026)Zhang, Monno, Tanaka, and Okutomi]{S-zhang2026cpddnet}
Qihang Zhang, Yusuke Monno, Masayuki Tanaka, and Masatoshi Okutomi.
\newblock {CPDDNet}: Color-polarization denoising and demosaicking network.
\newblock In \emph{Proceedings of the IEEE International Conference on Image
  Processing (ICIP)}, pages 1--6, 2026.

\bibitem[Zhang et~al.(2018)Zhang, Li, Li, Wang, Zhong, and Fu]{S-zhang2018rcan}
Yulun Zhang, Kunpeng Li, Kai Li, Lichen Wang, Bineng Zhong, and Yun Fu.
\newblock Image super-resolution using very deep residual channel attention
  networks.
\newblock In \emph{Proceedings of the European Conference on Computer Vision
  (ECCV)}, pages 286--301, 2018.

\bibitem[Zheng et~al.(2024)Zheng, Zhang, Wu, Ji, and Ru]{S-zheng2024scfnet}
Yubo Zheng, Xiangyue Zhang, Chengdong Wu, Peng Ji, and Jingyu Ru.
\newblock Color polarization imaging demosaicing based on {Stokes} vector
  information complementation and fusion.
\newblock \emph{Optics Express}, 32\penalty0 (25):\penalty0 44049--44066, 2024.

\bibitem[Zhou et~al.(2023{\natexlab{a}})Zhou, Han, Teng, Han, Li, Xu, and
  Shi]{S-zhou2023polarization}
Chu Zhou, Yufei Han, Minggui Teng, Jin Han, Si Li, Chao Xu, and Boxin Shi.
\newblock Polarization guided {HDR} reconstruction via pixel-wise
  depolarization.
\newblock \emph{IEEE Transactions on Image Processing}, 32:\penalty0
  1774--1787, 2023{\natexlab{a}}.

\bibitem[Zhou et~al.(2023{\natexlab{b}})Zhou, Teng, Lyu, Li, Xu, and
  Shi]{S-zhou2023lowlight}
Chu Zhou, Minggui Teng, Youwei Lyu, Si Li, Chao Xu, and Boxin Shi.
\newblock Polarization-aware low-light image enhancement.
\newblock \emph{Proceedings of the AAAI Conference on Artificial Intelligence},
  37\penalty0 (3):\penalty0 3742--3750, 2023{\natexlab{b}}.

\bibitem[Zhou et~al.(2025)Zhou, Zhou, Lyu, Guo, Ma, Shi, and
  Sato]{S-zhou2025pidsr}
Shuangfan Zhou, Chu Zhou, Youwei Lyu, Heng Guo, Zhanyu Ma, Boxin Shi, and Imari
  Sato.
\newblock {PIDSR}: Complementary polarized image demosaicing and
  super-resolution.
\newblock In \emph{Proceedings of the IEEE/CVF Conference on Computer Vision
  and Pattern Recognition (CVPR)}, pages 16081--16090, 2025.

\end{thebibliography}
}
\end{document}